\documentclass{dcthesis}

\committee[F. Jon Kull, Ph.D.]{}{}{}{}
\school{Dartmouth College}{Hanover, New Hampshire}
\degree{Doctor of Philosophy}
\field{Computer Science}
\usepackage{microtype}
\usepackage{graphicx}
\usepackage{subfigure}
\usepackage{booktabs} 
\usepackage{hyperref}
\usepackage{amsmath,amssymb,amsthm}
\usepackage{junstyle}
\usepackage{natbib}
\usepackage{bibentry}
\usepackage{algorithm}
\usepackage{algorithmic}
\usepackage{enumerate}
\usepackage{dsfont}
\usepackage{xcolor}
\usepackage{mathrsfs}
\usepackage{amsmath}
\numberwithin{equation}{chapter}
\newcommand{\prop}{\rho}
\renewcommand{\F}{\mathcal{F}}

\title{SCALABLE APPROXIMATE INFERENCE AND SOME APPLICATIONS}
\author{Jun Han}
\date{October 2019}
\field{Computer Science}
\degree{Doctor of Philosophy}
\committee{Lorenzo Torresani, Ph.D.}{Qiang Liu, Ph.D.}{Bo Zhu, Ph.D.}{Martin Renqiang Min, Ph.D.}

\begin{document}
\frontmatter
\newgeometry{left=1.5in,top=1in,bottom=1in,right=1in}
\maketitle
\chapter*{Abstract} 
Approximate inference in probability models is a fundamental task in machine learning. Approximate inference provides powerful tools to Bayesian reasoning, decision making, and Bayesian deep learning. The main goal is to estimate the expectation of interested functions w.r.t. a target distribution. When it comes to high dimensional probability models and large datasets, efficient approximate inference becomes critically important.

There are three main traditional frameworks to perform approximate inference. Firstly, adaptive importance sampling methods (IS) draws samples from the adaptively improved importance proposal and correct the bias with importance weights. IS provides an unbiased estimation but is difficult to adaptively optimize the importance proposal because of the large variance from Monte Carlo estimation of the objective. Secondly, Markov chain Monte Carlo~(MCMC) runs a long Markov chain to approximate the target. MCMC is theoretically sound and asymptotically consistent but is often slow to converge in practice. Thirdly, variational inference~(VI) uses samples from the approximate distribution. VI is practically faster but has been known to lack theoretical consistency guarantees. 
In this thesis, we propose a new framework for approximate inference, which combines the advantages of these three frameworks and overcomes their limitations. Our proposed four algorithms are motivated by the recent computational progress of Stein's method. Our proposed algorithms are applied to continuous and discrete distributions under the setting when the gradient information of the target distribution is available or unavailable. Theoretical analysis is provided to prove the convergence of our proposed algorithms. Our adaptive IS algorithm iteratively improves the importance proposal by functionally decreasing the $\KL$ divergence between the updated proposal and the target. When the gradient of the target is unavailable, our proposed sampling algorithm leverages the gradient of a surrogate model and corrects induced bias with importance weights, which significantly outperforms other gradient-free sampling algorithms. In addition, our theoretical results enable us to perform the goodness-of-fit test on discrete distributions. 

At the end of the thesis, we propose an importance-weighted method to efficiently aggregate local models in distributed learning with one-shot communication. Results on simulated and real datasets indicate the statistical efficiency and wide applicability of our algorithm.

\chapter*{Declaration}
I, Jun Han, confirm that the works presented in this thesis,
except as referenced herein, are done by me and have not been submitted in whole or in part for consideration for any other degree or qualification in this, or any other university. Where the information has been derived
from other sources, I confirm that this has been indicated in the thesis.

\chapter*{Acknowledgments}
\addcontentsline{toc}{section}{Abstract}
First and foremost, I would like to thank my advisors, Prof. Qiang Liu and Prof. Lorenzo Torresani, for their great advice and enormous support through my graduate journey. Prof. Qiang Liu and Prof. Lorenzo Torresani are great mentors. Their strong sense of responsibility, hard working practices, and commitment to perfection have made a profound impact on me. Without their supervision, I cannot finish my Ph.D. in four years after transferring from mathematics program to computer sciences program. Prof. Qiang Liu's commitment to highest standard research consistently motivates me to contribute to best research projects in future. I always remember he has worked very late for a lot of nights to help revise our papers. Prof. Lorenzo Torresani's provides enormous support and very insightful suggestions in my third-year and fourth-year Ph.D. study.

I am fortunate to have a wonderful internship at Disney research with Prof. Stephan Mandt and Dr. Christopher Schroers. We have worked in an interesting project neural video compression, which is published at NeurIPS 2019. I would like to thank inspiring discussions with Dr. Martin Renqiang Min from NEC Labs America, INC. I thank the committee members Prof. Bo Zhu and Dr. Renqiang Min for for their time, comments and suggestions.

I would thank machine learning folks in Prof. Liu's group. We have wonderful Ping Pong time at GDC building. The exercise keeps us healthy and relieves the stress. They also help me a lot in my daily life. I thank some my close friends, Ji Chen, Rui Liu and Hanyu Xue for their helps and encouragements during my Ph.D study.

I can never thank my family enough. I especially thank my
wife, \href{https://sites.google.com/view/shanhuang-homepage/home}{Shan Huang}, for her love, understanding, and support. She is a smart lady who graduated with a mathematics Ph.D. from Department of Mathematics, National University of Singapore. We are lucky to have a lovely daughter, Marina Han who was born in Austin in March 2018. I thank our parents and sisters for their encouragement, support and love.

In the end, I would like to thank Dartmouth graduate fellowship and national science foundation award CRII 1565796 to support my Ph.D. study.

\tableofcontents

\listoffigures

\listoftables

\backmatter

\chapter{Introduction to Approximate Inference\label{chap:intro}}
\section{Background}
Probabilistic models provide a powerful framework to capture complex phenomenons and patterns of the data. In discriminative
supervised learning, given the input variable $\vx$ and response variable $\vy,$ one would define a conditional distribution $p(\vy|\vx,\vthe).$ where $\vthe$ is the parameter of the probability model to be learned. When observations $\CD=\{\vx^i, \vy^i\}_{i=1}^N$ are available,  the task is to learn the parameter $\vthe$. One popular way to learn $\vthe$ in discriminative
supervised learning is to maximize the log likelihood, 
\begin{equation}
\vthe^*=\argmax_{\vthe} \sum_{i=1}^N \log p(\vy^i\mid\vx^i,\vthe).
\end{equation}
Formally, $\vthe^*$ gives the most probable interpretation of the model given the data $\CD.$  

In Bayesian setting, instead of a deterministic variable $\vthe$, $\vthe$ is a random variable. Suppose we have a prior belief distribution on $\vthe$, $p_0(\vthe).$ By Bayesian's rule, the posterior distribution of $\vthe$ is 
\begin{equation}
p(\vthe\mid \CD) = \frac{p(\CD\mid \vthe)p_0(\vthe)}{p(\CD)},    
\end{equation}
where $p(\CD\mid \vthe)=\prod_i p(\vy^i\mid \vx^i, \vthe)$. Here $p(\CD)$ is the normalization constant, 
\begin{equation}
p(\CD)=\int p(\CD\mid \vthe)p_0(\vthe) d\vthe,    
\end{equation}
which is difficult to compute when the dimension of $\vthe$ is high and typically intractable in practice. $P(\CD)$ has wide applications on Bayesian model selections and Bayesian analysis \citep{murphy2012machine}. In this proposal, we will develop an efficient method to effectively estimate $P(\CD).$

To predict the response variable $\vy$ on test data $\vx$ from Bayesian perspective, the task is to compute the predictive probability, 
\begin{equation}
\label{bayes:int}
p(\vy\mid \vx, \CD) = \int p(\vy\mid \vx, \vthe)p(\vthe\mid \CD)d\vthe,    
\end{equation}
where a challenging integral over $\vthe$ needs to be solved. In practice, the integration \eqref{bayes:int} is usually intractable. One efficient way is to draw samples $\{\vthe^i\}_{i=1}^n$ from $p(\vthe\mid \CD),$ and estimate the integration \eqref{bayes:int} using Monte Carlo method,
\begin{equation}
p(\vy\mid \vx, \CD) = \frac1n\sum_{i=1}^n p(\vy\mid \vx, \vthe^i).    
\end{equation}
The key challenging reduces to sample from the posterior distribution $p(\vthe\mid \CD).$ The difficulties come from two parts: the distribution of the data is highly complex when the dimension of the data is high; the dataset $\CD$ is huge in modern machine learning setting. In the next section, we will introduce two main stream algorithms to tackle such approximate inference problem.
\section{Approximate Inference}
In this section, we will introduce two popular algorithms to perform approximate inference, Markov Chain Monte Carlo(MCMC)~\citep{hastings1970monte, metropolis1953equation, hastings1970monte, metropolis1953equation}, and variational inference~\citep{blei2017variational}. MCMC runs a Markov chain to evolve a set of samples to approximate the target distributions. MCMC is theoretically
sound and asymptotically consistent, but is often slow to
converge in practice. Variational inference seeks an approximate distributional family and optimizes the approximate distribution whose sample is easy to draw to match the target distribution under some divergence metrics. Variational inference algorithms are practically faster but has been
known to lack theoretical consistency guarantees. Finally, we introduce a recently proposed approximate inference algorithm, Stein variational gradient descent~(SVGD, \cite{liu2016stein}), which combines advantages of both MCMC and variational inference.

\subsection{Markov Chain Monte Carlo}
Markov chain Monte Carlo (MCMC) methods comprise a class of algorithms for sampling from the distribution of interest. The Metropolis-Hastings (MH) algorithm is the most popular MCMC method~\cite{hastings1970monte, metropolis1953equation}. Let $p(\vthe)$ be the target distribution. MH algorithm proposes a transition distribution $q(\vthe'|\vthe)$ to sample a candidate value $\vthe'$ given the current value $\vthe$ according to the transition distribution $q(\vthe'|\vthe).$ At each step, the Markov Chain moves torwards $\vthe'$ with probability 
\begin{equation}
\A(\vthe, \vthe')=\min(1, \frac{p(\vthe')q(\vthe|\vthe')}{p(\vthe)q(\vthe'|\vthe)}),    
\end{equation}
otherwise it remains to stay at $\vthe.$

\begin{algorithm}[thb] %
\caption{Metropolis-Hasting (MH) Algorithm~\citep{hastings1970monte, metropolis1953equation}}  
\label{alg:mh}
\begin{algorithmic}
\STATE {\bf Input:} initial distribution $p_0(\vthe)$ and proposal distribution $q(\vthe'|\vthe).$ 
\STATE Initialize $\vthe^0$ from $p_0(\vthe)$.
\FOR{$i=0$ to $M-1$}
\STATE -sample $u$ from uniform distribution $\U[0,1]$.
\STATE -sample $\vthe'$ from $q(\vthe'|\vthe)$.
\STATE -if $u< \A(\vthe^i, \vthe')=\min(1, \frac{p(\vthe')q(\vthe^i|\vthe')}{p(\vthe^i)q(\vthe'|\vthe^i)}),$ let $\vthe^{i+1}=\vthe'$; else $\vthe^{i+1}=\vthe^{i}.$
\ENDFOR
\STATE {\bf Remark:} $M$ is number of Markov Chain iteration.
\end{algorithmic}
\end{algorithm}
The transition kernel for MH algorithm is 
\begin{equation}
\label{eq:MH}
K_{\mathrm{MH}}(\vthe^{i+1}|\vthe^{i})=q(\vthe^{i+1}|\vthe^{i})\A(\vthe^i, \vthe')+\delta_{\vthe^{i}}(\vthe^{i+1})r(\vthe^{i})    
\end{equation}
where $\delta_x(y)=1$ iff $x=y$ and $r(\vthe^{i})$ is the term associated with rejection,
\begin{equation*}
r(\vthe^{i})=\int_{\X} q(\vthe'|\vthe)(1-\A(\vthe^i, \vthe'))d\vthe'.    
\end{equation*}
It is straightforward to verify that $K_{\mathrm{MH}} $ satisfies the \emph{detailed balance condition},
\begin{equation*}
p(\vthe^i)K_{\mathrm{MH}}(\vthe^{i+1}|\vthe^{i})=p(\vthe^{i+1})K_{\mathrm{MH}}(\vthe^{i}|\vthe^{i+1}),    
\end{equation*}
which is the sufficient and necessary condition for the Markov chain to converge to the stationary distribution $p(\vthe).$

Most practical MCMC
algorithms, such as Monte Carlo expectation-maximization (MCEM)~\cite{wei1990monte} and Hybrid Monte Carlo~\cite{duane1987hybrid, neal2012bayesian}, can be interpreted as special cases or extensions of the MH algorithm. 

\paragraph{Hybrid Monte Carlo (HMC)~\cite{duane1987hybrid, neal2012bayesian}} HMC introduces a set of auxiliary "momentum" $\vv v$ and defines the extended target density
\begin{equation}
p(\vthe, \vv v)=p(\vthe)\N(\vv v;\vv 0, \I),    
\end{equation}
where $\N(\vv v;\vv 0, \I)$ is the standard Gaussian distribution. Let $\Delta(\vthe)=\partial \log p(\vthe)/\partial \vthe$ and $\epsilon$ be the fixed step size.

\begin{algorithm}[thb] %
\caption{Hamiltonian Monte Carlo (HMC) Algorithm~\citep{duane1987hybrid, neal2012bayesian}}  
\label{alg:hmc}
\begin{algorithmic}
\STATE {\bf Input:} initial distribution $p_0(\vthe).$
\STATE Initialize $\vthe^0$ from  $p_0(\vthe)$ 
\FOR{$i=0$ to $M-1$}
\STATE -sample $u$ from uniform distribution $\U[0,1]$ and $\vv v'\sim \N(\vv v;\vv 0, \I).$
\STATE -Let $\vthe_0=\vthe^i$ and $\vv v_0=\vv v'+\epsilon \Delta(\vthe_0).$
\FOR{$\ell=1,\cdots, L$}
\STATE -$\vthe_{\ell}=\vthe_{\ell-1}+\epsilon \vv v_{\ell-1}.$
\STATE -$\vv v_{\ell}=\vv v_{\ell-1}+\epsilon_{\ell}\Delta(\vthe_{\ell}),$ where $\epsilon_{\ell}=\epsilon$ for $\ell<L$ and $\epsilon_L=\epsilon/2.$

\ENDFOR
\STATE -if $u<\A=\min(1, \frac{p(\vthe_L)}{p(\vthe^i)})\exp(-\frac{1}{2}(\vv v_{L}^\top \vv v_{L})-\vv v'^\top \vv v'),$ then $(\vthe^{i+1}, \vv v^{i+1})=(\vthe_L, \vv v_L);$ \STATE -else $(\vthe^{i+1}, \vv v^{i+1})=(\vthe^i, \vv v').$
\ENDFOR
\STATE {\bf Remark:} $L$ is the number of leap-frog steps.
\end{algorithmic}
\end{algorithm}
When $L=1$ in HMC Algorithm~\ref{alg:hmc}, it reduces to well-known \emph{Langevin} algorithm. Two major drawbacks of MCMC algorithms limit their applications to approximate inference. The first limitation is that it takes a long time for the Markov chains to converge. The second limitation is that it is difficult to measure whether the Markov chains have converged or not. Lastly, widely used MCMC algorithms such as Langevin algorithm and HMC algorithm require the availability of the gradient information of the target distributions, which is impractical in some applications. In some real settings, the gradient of the target distribution is too expensive to calculate or intractable. The major drawbacks of MCMC algorithms motivate us to design better approximate inference algorithms. 

\subsection{Variational Inference\label{cha:def:vi}}
In MCMC methods, our goal is to draw a set of samples $\{\vthe_i\}_{i=1}^n$ to approximate $p(\vthe|\CD)$. Although there is theoretical guarantee that the Markov chains will converge to the target distribution, it is too expensive to draw samples to approximate $p(\vthe|\CD)$ in some applications. In this case, \citet{wainwright2008graphical, blei2017variational} use a variational distribution $q(\vthe)$ from a distribution family $\mathcal{Q}$, which is easy to sample from, to approximate $p(\vthe|\CD).$ To simplify the notation without confusion, we abbreviate $p(\vthe|\CD)$ as $p(\vthe).$ Before introducing variational methods, let us first introduce the divergence between distributions.
\begin{mydef}
The {\bf $f$-divergence} between two probability distributions $q(\vthe)$ and $p(\vthe)$ is
\begin{equation}
\BD_f(q(\vthe)||p(\vthe)) = \E_q[f(\frac{p(\vthe)}{q(\vthe)})-f(1)],    
\end{equation}
where $f: \R_+\rightarrow \R$ is any convex function. 
\end{mydef}

\paragraph{Variational inference by $f$-divergence}: As it is intractable to draw samples from the distribution $p(\vthe)$ of interest, variational inference uses a simpler distribution $q_\ff(\vthe)$, parametrized by $\ff$, to approximate $p(\vthe)$ and draw samples from $q_\ff(\vthe)$ instead to perform approximate inference. The problem is how to ensure $q_\ff(\vthe)$ to approximate the distribution $p(\vthe)$ of interest. In variational inference, we optimize the parameter $\ff$ to 
\begin{equation}
\min_{\phi} \BD_f(q_{\ff}(\vthe)||p(\vthe)) =  \min_{\phi} \E_q[f(\frac{p(\vthe)}{q_{\ff}(\vthe)})-f(1)].
\end{equation}
For details, see~\cite{blei2017variational, wang2018variational}.

\paragraph{Choices of function $f$} One nature choice of $f$ is $f(x)=-\log(x).$ We have the $\KL$ divergence between $q_\ff(\vthe)$ and $p(\vthe),$
\begin{equation}
\label{cha:intro:kldef}
\min_{\phi} \mathcal{L}(\phi) =\min_{\phi} \KL(q_{\ff}(\vthe)||p(\vthe)) =  \min_{\phi} \E_q[\log q_{\ff}(\vthe)-\log p(\vthe)].    
\end{equation}
Another widely used choice of $f$ is $f(x)=t^{\alpha}/(\alpha(\alpha-1))$, $\alpha\in \R/\{0,1\},$ which is called $\alpha$-divergence~\citep{hernandez2016black}. Variational inference algorithms typically converge faster than MCMC algorithms. While one major drawback of variational inference algorithms is that it restrict the approximate distribution from the predefined family $\{q_{\phi}(\vthe)\},$ which gives poor approximation when the predefined distribution family $\{q_{\phi}(\vthe)\}$ deviates from the complex target distribution $p(\vthe).$ In most cases, the expectation \eqref{cha:intro:kldef} doesn't have a closed form, which casts a challenging optimization problem. In practice, to optimize the parameter $\phi$, VI algorithms typically draw a set of samples $\{\vthe_i\}_{i=1}^n$ and do Monte Carlo estimation of the objective \eqref{cha:intro:kldef}, 
\begin{equation}
\label{cha:mc:kldef}
\nabla_{\phi} \mathcal{L}(\phi) \approx \frac1n\sum_{i=1}^n [\nabla_{\phi}\log q_{\phi}(\vthe_i) (\log p(\vthe_i) -\log q_{\phi}(\vthe_i))].
\end{equation}
However, the Monte Carlo estimation \eqref{cha:mc:kldef} has large variance. We will discuss techniques to reduce the variance of such Monte Carlo estimation.

\paragraph{Black-Box Variational Inference} (BBVI, \citet{ranganath2014black}) In some applications, the gradient of the target distribution $p(\vthe)$ w.r.t. $\vthe$ is unavailable. Based on the fact that $  \E_{q_{\phi}(\vthe)} [\nabla_{\phi} \log q_{\phi}(\vthe)] =0, $ we can construct a simple form of control variates to reduce the variance, 
\begin{equation}
\label{cha:intro:sc}
\nabla_{\phi} \mathcal{L}(\phi)\approx \frac1n\sum_{i=1}^n [\nabla_{\phi}\log q_{\phi}(\vthe_i) (\log q_{\phi}(\vthe_i) - \log p(\vthe_i))] + \lambda\frac1n\sum_{i=1}^n  \nabla_{\phi} \log q_{\phi}(\vthe_i)],
\end{equation}
where $\lambda$ is the coefficient and has the optimal form
\begin{equation}
\label{intro:control}
\lambda = -\mathrm{Var}(\nabla_{\phi}\log q_{\phi}(\vthe))^{-1}\mathrm{Cov}[\nabla_{\phi}\log q_{\phi}(\vthe)(\log q_{\phi}(\vthe) - \log p(\vthe), \nabla_{\phi} \log q_{\phi}(\vthe)],    
\end{equation}
and the variance, covariance matrix can be estimated empirically. 
In some cases, \eqref{cha:intro:sc} still has relatively large variance. In order to have a smaller variance, large sample size $\{\vthe_i\}_{i=1}^n$ is required, which might be impractical when the evaluation of the target distribution $\log p(\vthe)$ is expensive. At the end of the thesis, we will adopt a more efficient method to reduce the variance. 

\paragraph{Discussions on implicit and semi-implicit choice of $q_\ff(\vthe)$} To remove the restriction of choosing the surrogate distribution $q_{\phi}(\vthe)$ from the predefined family $\{q_{\phi}(\vthe)\},$ the implicit choice and semi-implicit choice of the distribution $\{q_{\phi}(\vthe)\}$ have recently been proposed \citep{wang2016learning, mescheder2017adversarial, tran2017hierarchical, yin2018semi}. Basically, they construct a powerful variable transform parametrized by a deep neural network as follows,
\begin{equation}
\label{def:imp}
\vthe = f_{\phi}(\epsilon),~q_\ff(\vthe)=q_0(f_{\phi}^{-1}(\vthe))|\frac{\partial f_{\phi}^{-1}(\vthe)}{\partial\vthe}|,~\mathrm{where}~\epsilon\sim q_0(\epsilon),     
\end{equation}
and optimizes a certain divergence between the transformed distribution $q_\ff(\vthe)$ and the target distribution $p(\vthe).$  As long as the transform is expressive enough, the transformed distribution $q_\ff(\vthe)$ can arbitrarily approximate the target distribution $p(\vthe).$ Implicit probability can be applied to applications when the samples from $q_\ff(\vthe)$ is needed. However, as shown in \eqref{def:imp}, it is challenging to calculate the density realization of $q_\ff(\vthe)$ as the inverse of $f_{\phi}$ is typically unavailable and cumbersome to compute the determinant of the Jacobian matrix $\frac{\partial f_{\phi}^{-1}(\vthe)}{\partial\vthe}$, which limits its application. Most importantly, it is also difficult to stably train such a transform using standard optimization methods, which is the main reason limiting its real applications. Semi-implicit $q_\ff(\vthe)$ \citep{yin2018semi} has been proposed to alleviate the problem in implicit model $q_\ff(\vthe)$. But the unstable problem still exists.

\subsection{Stein Variational Gradient Descent}
Stein variational gradient descent (SVGD) \citep{liu2016stein} is a 
nonparametric variational inference algorithm that 
iteratively transports a set of particles 
to approximate a given target distribution by performing a type of functional gradient descent on the KL divergence. We give a quick overview of its main idea in this section. To make notations easy to read, we use the notation $p(\vx)$ to be the target distribution instead of $p(\vthe)$ in the following.

\paragraph{Preliminary Notations} Before introducing SVGD, let us define some notations, which will be used in the whole thesis. We always assume $\vx=[x_1,\cdots, x_d]^\top \in \R^d$.
Given a positive definite kernel $k(\vx,\vx')$, there exists an unique reproducing kernel Hilbert space (RKHS) $\H_0$,
formed by the closure of functions of form $f(\vx) = \sum_{i} a_i k(\vx,\vx_i)$ where $a_i \in \RR$, equipped with inner product
$\la f, ~ g\ra_{\H_0} = \sum_{ij}a_i k(\vx_i, \vx_j) b_j$ for $g(\vx) = \sum_j b_j k(\vx, \vx_j)$. 
Denote by $\H  = \H_0^d = \H_0 \times \cdots \times \H_0$ the vector-valued function space formed by $\vv f = [f_1, \ldots, f_d]^\top$, where $f_i \in \H_0$, $i=1,\ldots, d$, equipped with inner product $\la \vv f, ~ \vv g\ra_{\H}=\sum_{l=1}^d \la f_l, ~ g_l\ra_{\H_0}, $ for $\vv g =[g_1, \ldots, g_d]^\top.$ 
Equivalently, $\H$ is the closure of functions of form  $\vv f(\vx) = \sum_{i} \vv a_i k(\vx,\vx_i)$ where $\vv a_i \in \RR^d $ 
with inner product $\la \vv f, ~ \vv g\ra_{\H} = \sum_{ij}\vv a_i^\top \vv b_j k(\vx_i, \vx_j)$ for $\vv g(\vx) = \sum_{i} \vv b_i k(\vx,\vx_i)$. 
See e.g.,~\citet{berlinet2011reproducing} for more background on RKHS. 

Let $p(\vx)$ be a continuous-valued positive density function on $\R^d$ which we want to approximate with a set of particles 
$\{ \vx_i\}_{i=1}^n$. 
SVGD starts with a set of initial particles $\{\vx_i\}_{i=1}^n$, 
and updates the particles iteratively by 
\begin{align}\label{equ:xxii}
\vx_i  \gets \vx_i +  \epsilon \ff(\vx_i),  ~~~~ \forall i = 1, \ldots, n,  
\end{align}
where $\epsilon$ is a step size, and 
$\ff\colon \RR^d \to \RR^d$ is a velocity field which should be chosen to drive the particle distribution closer to the target.
Assume the distribution of the particles at the current iteration is $q$, 
and $q_{[\epsilon\ff]}$ is the distribution of the updated particles $\vx^\prime = \vx + \epsilon \ff(\vx)$. 
The optimal choice of $\ff$ can be framed into the following optimization problem:  
\begin{align}\label{intro:equ:ff00}
\ff^* =   \argmax_{\ff \in \F}  \bigg\{  -   \frac{d}{d\epsilon} \KL(q_{[\epsilon\ff]} ~|| ~ p) \big |_{\epsilon = 0}  \bigg\}, 
\end{align}
where $\F$ is the set of candidate velocity fields, and  $\ff$ is chosen in $\F$ to maximize the decreasing rate on the KL divergence between the particle distribution and the target. 

In SVGD, $\F$ is chosen to be the unit ball of a vector-valued reproducing kernel Hilbert space (RKHS) $\H = \H_0 \times \cdots \times \H_0$,
where  $\H_0$ is a RKHS formed by scalar-valued functions associated with a positive definite kernel $k(\vx,\vx')$, that is, 
$\F = \{\ff \in \H \colon ||\ff||_{\H}\leq  1 \}.$
This choice of $\F$ allows us to 
consider velocity fields in infinite dimensional function spaces while still obtaining computationally tractable solution. 

A key step towards solving \eqref{intro:equ:ff00} is to observe that the objective function 
in \eqref{intro:equ:ff00} is a simple linear functional of $\ff$ that connects to Stein operator \citep{oates2017control, gorham2015measuring, liu2016stein, gorham2017measuring, chen2018stein}, 
\begin{align}\label{equ:klstein00}
&~ - \frac{d}{d\epsilon} \KL(q_{[\epsilon\ff]} ~|| ~ p) \big |_{\epsilon = 0}  = \E_{x\sim q}[\steinpxtransp  \ff(\vx)], \\[.5\baselineskip]
&\!\!\!\!\!\!\!\text{with}~~~ \steinpxtransp \ff(\vx)  = \nabla_\vx \log p(\vx) ^\top \ff (\vx)+ \nabla_{\vx}^\top\ff(\vx),  
\end{align}
where $\steinpx$ is a linear operator 
called \emph{Stein operator} and is formally viewed as a column vector similar to the gradient operator $\nabla_{\vx}$. 
The Stein operator $\steinpx$ is connected to Stein's identity which shows that the RHS of \eqref{equ:klstein00} is zero if $p = q$: 
\begin{align}\label{equ:steinid}
\E_{\vx\sim p}[\steinpxtransp \ff(\vx)]  = 0. 
\end{align}
This corresponds to $\frac{d}{d\epsilon} \KL(q_{[\epsilon\ff]} ~|| ~ p) \big |_{\epsilon = 0} = 0$ since there is no way to further decease the KL divergence when $p=q$. 
Eq. \eqref{equ:steinid} is a simple result of integration by parts assuming the value of $p(\vx)\ff(\vx)$ vanishes on the boundary of the integration domain.   

Therefore, the optimization in \eqref{intro:equ:ff00} reduces to 
\begin{align}
\label{solvksd}
\!\!\!\!\D_{\F}(q || p) \overset{def}{=} 
\max_{\ff \in \F} \left\{ \E_{\vx \sim q} \left [\steinpxtransp \ff(\vx)\right] \right\}, 
\end{align}
where $\mathbb{D}_\F(q ~||~ p)$ is the kernelized Stein discrepancy (KSD) defined in \citet{liu2016kernelized,  chwialkowski2016kernel}. 

Observing that \eqref{solvksd} 
is ``simple'' in that it is a linear functional optimization on a 
unit ball of a Hilbert space, \citet{liu2016stein} showed that \eqref{intro:equ:ff00} has a simple closed-form solution: 
\begin{align}\label{svgdoptimal} 
\ff^*(\vx') \propto  \E_{\vx\sim q}[\steinpx k(\vx, \vx')], 
\end{align} 
where $\steinpx$ is applied to variable $\vx$, and 
\begin{equation}
\label{ksddef}
\mathbb{D}^2_\F(q\mid\mid p) = \E_{\vx, \vx'\sim q}[\kappa_p (\bd{x},  \bd{x}')],
\end{equation}
where $\kappa_p (\bd{x},  \bd{x}') := (\stein_p')^\top(\steinpx  k(\vx, \vx'))$ and $\newsteinpx$ denotes the Stein operator applied on variable $\vx'$. 
Here $\kappa_p(\bd x, \bd x')$ 
can be calculated explicitly in Theorem~3.6 of \citet{liu2016kernelized}, 
\begin{equation}
\label{intro:ksd}
\begin{aligned}
\kappa_{p} (\bd{x},  \bd{y}) = & \bd{s}_{p}(\bd{x})^\top k(\bd{x},\bd{y})\bd{s}_{p}(\bd{y}) +\bd{s}_{p}(\bd{x})^\top \nabla_{\bd{y}}k(\bd{x},\bd{y}) \\
&+\bd{s}_{p}(\bd{y})^\top \nabla_{\bd{x}} k(\bd{x},\bd{y})+\nabla_{\bd{x}}\cdot(\nabla_{\bd{y}}k(\bd{x}, \bd{y})).
\end{aligned}
\end{equation}

The Stein variational gradient direction $\ff^*$ provides a theoretically optimal direction that drives the particles towards the target $p(\vx)$ as fast as possible. In practice, SVGD approximates $q(\vx)$ using a set of particles, $\{\vx_i\}_{i=1}^n$ iteratively updated by 
\begin{align}
\label{update11}
\!\!\vx_i \gets \vx_i  + \frac{\epsilon}{n} \Delta \vx_i, \text{~where~} \Delta \vx_i =\sum_{j=1}^n [\nabla \log p(\vx_j)k(\vx_j, \vx_i) +  \nabla_{\vx_j}k(\vx_j,\vx_i)], 
\end{align}
where $k(\vx,\vx')$ is any positive definite kernel; the term with the gradient $\nabla\log p(\vx)$ drives the particles to the high probability regions of $p(\vx)$,
and the term with $\nabla k(\vx, \vx')$ acts as a repulsive force to keep the particles away from each other to quantify the uncertainty. 

\section{Thesis Outline and Contributions}
In this section, we will first provide the outline flow and dependence of different chapters in the thesis. Then we will briefly summarize the contributions of our thesis in each chapter.
\begin{figure}[htb]
   \centering
   \includegraphics[width=.95\textwidth]{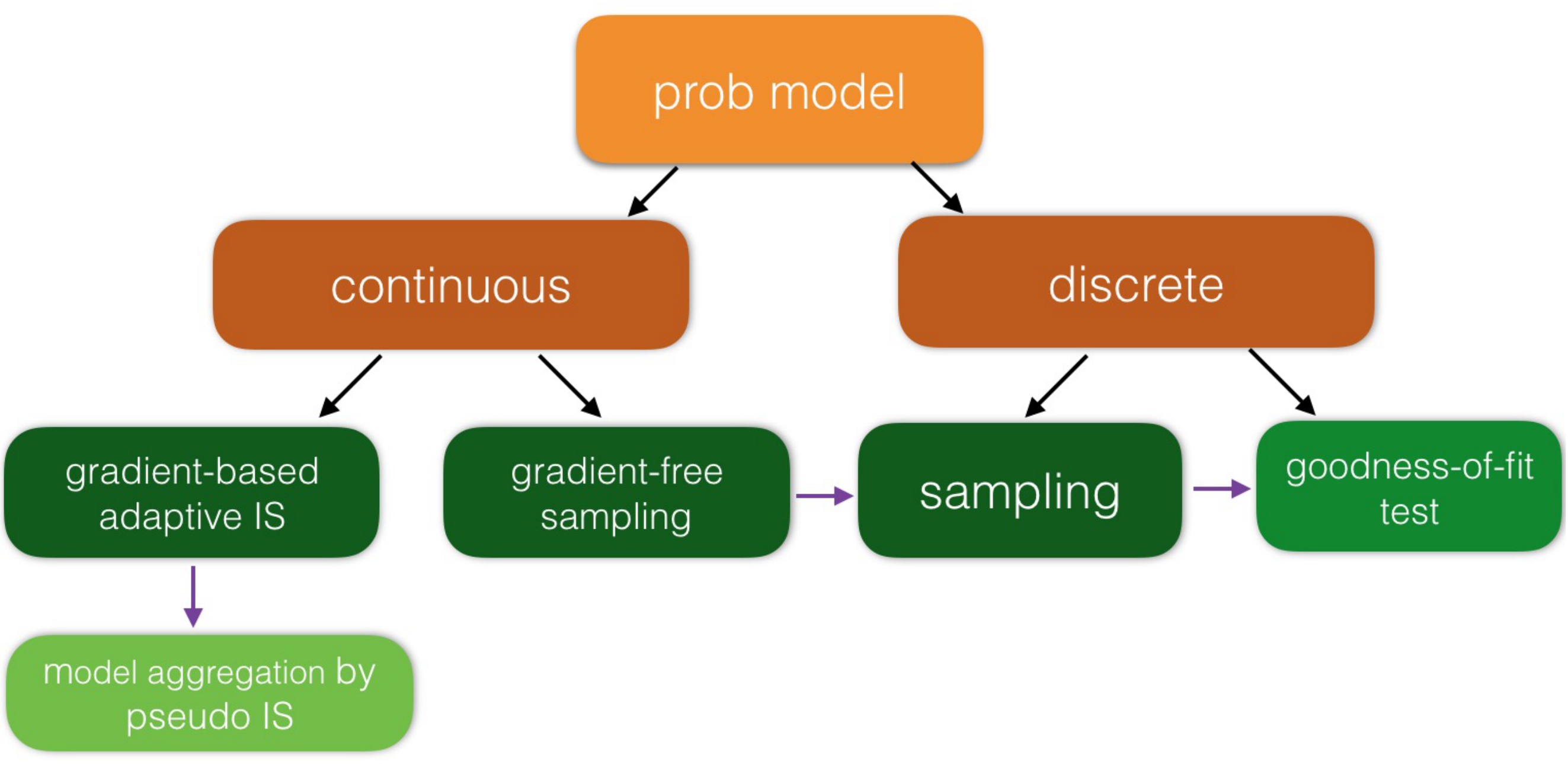}
  \caption[The flow and dependence of the thesis]{The flow and dependence of the thesis. A gradient-based adaptive importance sampling on continuous-valued distribution is proposed in Chapter~\ref{chap:is}. A gradient-free sampling on continuous-valued distribution is proposed in Chapter~\ref{chap:gf}. A sampling algorithm on discrete-valued distribution is proposed in Chapter~\ref{chap:disc}, which is motivated from gradient-free sampling method. A goodness-of-fit test algorithm is proposed in Chapter~\ref{chap:gof}, which is motivated by results of Chapter~\ref{chap:disc}. Distributed model aggregation algorithm by pseudo importance sampling is proposed in Chapter~\ref{chap:boot}, which is closely related to the importance sampling trick used in Chapter~\ref{chap:is}. \label{thesis:flow}}
\end{figure} 

\paragraph{Thesis Flow and Dependence} Firstly, we propose a gradient-based adaptive importance sampling on continuous-valued distribution  in Chapter~\ref{chap:is}. Secondly, we propose a gradient-free sampling on continuous-valued distribution in Chapter~\ref{chap:gf}. Thirdly, we propose a sampling algorithm on discrete-valued distribution in Chapter~\ref{chap:disc}, which is motivated from gradient-free sampling method. Fourthly, based on results of Chapter~\ref{chap:disc}, we propose a goodness-of-fit test algorithm in Chapter~\ref{chap:gof}. Finally, we propose importance-weighted method to distributed model aggregation, which is motivated by a form of by importance sampling and is termed as pseudo importance sampling in Chapter~\ref{chap:boot}.

\paragraph{Thesis Contributions} The contributions of the thesis can be summarized in three main parts. Firstly, we propose three new approximate inference algorithms which can be applied to continuous-valued distributions. Secondly, we propose one sampling algorithm on discrete-valued distributions and one goodness-of-fit testing algorithm, which test whether a set of data $\{\vz_i\}_{i=1}^n$ is the candidate distribution, on discrete-valued distributions. Finally, we present one effective method, which is motivated from the widely-used tool in approximate inference, to efficiently aggregate distributed models in the one-shot communication setting. Theoretical analysis is provided to analyze the convergence or other properties of our algorithms. 
Extensive experiments are conducted to demonstrate the effectiveness and wide applicability of all our proposed algorithms.

In the following, we first emphasize our contributions of approximate inference algorithms on continuous-valued distributions. Specifically, we propose a nonparametric adaptive importance sampling algorithm by decoupling the iteratively updated particles of SVGD $\{\vx_i^\ell\}_{i=1}^n$ into two sets: \emph{leader particles} $\vx_A^\ell = \{\vx_i^\ell  \colon  i\in A\}$ and \emph{follower particles}
$\vx_B^\ell = \{\vx_i^\ell  \colon  i\in B\}$, with $B =  \{1,\ldots, n\} \setminus A$. The leader particles is applied to construct the transform and the follower particles are updated by the constructed transform, $\vv x_{i}^{\ell+1} \gets \vv x_i^\ell + \epsilon \ff_{\ell+1}(\vv x_i^\ell), ~ \forall i \in A\cup B,$ where $\ff_{\ell+1}(\cdot)$ is constructed by only  
using particles in set $A,$
\begin{align}\notag
 \!\!\!\! \!\!\!\!
\ff_{\ell+1}(\cdot) =  \frac{1}{|A|}\sum_{j\in A} [\nabla \log p(\vv x_j^\ell) k(\vv x_j^\ell,  \cdot) + \nabla_{\vv x_j^\ell} k(\vv x_j^\ell, \cdot)].
\end{align}

With such a transform, the distribution of the updated particles in $\vx_B$ satisfies \begin{align}\label{intro:equ:qt}
q_{\ell} =  (\T_{\ell} \circ \cdots \circ \T_{1})\sharp q_0, \quad \ell=1, \ldots, K,
\end{align}
where the importance proposal $q_\ell$ forms increasingly better approximation of the target $p$ as $\ell$ increases. Conditional on $\vx_A^\ell,$ particles in $\vx_B^\ell$ are i.i.d. and hence can provide an unbiased estimation of the integral $\E_{\vx\sim q_\ell}[\frac{p(\vx)}{q_\ell(\vx)}h(\vx)]$ for any function $h(\vx).$ Our importance proposal is not restricted to the predefined distributional family as traditional adaptive importance sampling methods do. The $\KL$ divergence between the updated proposal $q_\ell$ and the target distribution is also maximally decreased in a functional space, which inherits from the theory of SVGD~\citep{liu2017stein}. We apply our proposed algorithm to evaluate the normalization constant of various probability models including restricted Boltzmann machine and deep generative model to demonstrate the effectiveness of our proposed algorithm, where the original SVGD cannot be applied in such tasks. We propose a novel sampling algorithm for continuous-valued target distribution when the gradient information of the target distribution is unavailable. 
iteratively updated by 
$\vx_i \gets \vx_i  + \frac{\epsilon}{n} \Delta \vx_i$, where 
\begin{align}
\label{intro:gf:upd}
 \Delta \vx_i \propto
\sum_{j=1}^n 
\! w(\vx_j) \big[\nabla \log \rho(\vx_j) k( \vx_j, \vx_i) + \nabla_{\vx_j} k(\vx_j, \vx_i) \big], 
\end{align}
which replaces the true gradient $\nabla \log p(
\vx)$ with a surrogate gradient $\nabla\log \rho(\vx)$ of an arbitrary auxiliary distribution $\rho(\vx)$, and then uses an importance weight $w(\vx_j):=\rho(\vx_j)/p(\vx_j)$ to correct the bias introduced by the surrogate $\rho(\vx)$.
Perhaps surprisingly, 
we show that the new update can be derived as a standard SVGD update by using an importance weighted kernel $w(\vx)w(\vx')k(\vx,\vx')$, and hence immediately inherits the theoretical proprieties of SVGD; for example, particles updated by \eqref{intro:gf:upd} can be viewed as a gradient flow of KL divergence similar to the original SVGD \citep{liu2017stein}. Empirical experiments demonstrate that our proposed gradient-free SVGD significantly outperforms gradient-free Markov Chain Monte Carlo sampling baselines on various probability models with intractable normalization constant and unavailable gradient information of the target distribution.

We propose a gradient-free black-box importance sampling algorithm, which equips any given set of particles $\{\vx_i\}_{i=1}^n$ with a set of importance weights $\{u_i\}_{i=1}^n$ such that
\begin{equation}
\label{isproblem}
\sum_{i=1}^n u_i h(\bd{x}_i) \approx \E_{\vx\sim p} [h(\bd{x})], 
\end{equation}
for general test function $h(\bd{x}).$ To achieve this goal, we will leverage our result from gradient-free KSD defined as follows,
\begin{equation}
\label{intro:bbis:ksd}
\wt{\mathbb{S}}(q, p) =\E_{\bd{x},\bd{y}\sim q}[w(\bd{x})\kappa_{\rho} (\bd{x},  \bd{y})w(\bd{y})]\ge 0,
\end{equation}
where $\kappa_{\rho} (\bd{x},  \bd{y})$ can be evaluated by the formula~\eqref{intro:ksd} and does not require the gradient of the target distribution $p(\vx).$ \eqref{intro:bbis:ksd} provides a metric to measure the closeness between $q(\vx)$ and $p(\vx)$ when the samples $\{\vx_i\}_{i=1}^n$ and the evaluation of the target $p(\vx)$ are available. Motivated from~\citep{liu2016black}, we propose a gradient-free black-box importance sampling algorithm by optimizing a set of importance weights $\{u_i\}_{i=1}^n$ for any given set of particles $\{\vx_i\}_{i=1}^n$ through the following quadratic optmization
\begin{align}
\label{intro:gfbbis:pro}
\begin{split}
\hat {\vv u} =  \argmin_{\vv u}\bigg\{ \vv u^\top \wt{\bd{K}}_p \vv u,  ~~  s.t.~~ \sum_{i=1}^n u_i = 1, ~~~ u_i \geq 0\bigg\},
\end{split}
\end{align}
where $\wt{\bd{K}}_p = \{ w(\vx_i)\kappa_{\rho}(\bd{x}_i, \bd{x}_j)w(\vx_j) \}_{i,j=1}^n$ and $\vv u = \{ u_i \}_{i=1}^n.$ For more details of the idea and the approximation error, please refer to Chapter~\ref{chap:gf}.

In the second part of the thesis, we propose two approximate inference algorithms on discrete-valued distributions. We propose a new algorithm to sample from the discrete-valued distributions. Our proposed algorithm is based on the fact that the discrete-valued distributions can be bijectively mapped to the piecewise continuous-valued distributions. Since the piecewise continuous-valued distributions are non-differentiable, gradient-based sampling algorithms cannot be applied in this setting. Our proposed sample-efficient GF-SVGD is a natural choice. To construct effective surrogate distributions $\rho(\vx)$ in GF-SVGD, we propose a simple transformation, the inverse of dimension-wise Gaussian c.d.f. $F(\vx)$ (its p.d.f. $p_0(\vx)$, $F'(\vx)=p_0(\vx)$), to transform the piecewise continuous-valued distributions to a simple form of continuous distributions. With such a straightforward transform, the effective surrogate distribution $\rho(\vx)$ in GF-SVGD is natural to construct.
The detail of our sampling algorithm is provided in Chapter~\ref{chap:disc}. Empirical experiments on large-scale discrete graphical models demonstrate the effectiveness of our proposed algorithm. 

As a direct application, we propose a principled ensemble method to train the binarized neural networks~(BNN). We train an ensemble of $n$ neural networks (NN) with the same architecture ($n\ge 2$). Let $\vw_i^b$ be the binary weight of model $i$, for $i=1,\cdots, n$, and $p_*(\vw_i^b;D)$ be the target probability model with softmax layer as last layer given the data $D$. Learning the target probability model is framed as drawing $n$ samples $\{\vw_i^b\}_{i=1}^n$ to approximate the posterior distribution $p_*(\vw^b; D)$. We apply multi-dimensional transform $F$ to transform the original discrete-valued target to the target distribution of real-valued $\vw\in\R^d$. Let $p_0(w)$ be the base function, which is the product of the p.d.f. of the standard Gaussian distribution over the dimension $d.$ Based on the derivation in Section 3, the distribution of $\vw$ has the form $p_c(\vw;D)\propto p_*(\sign(\vw);D)p_0(\vw)$ with weight $\vw$ and the $\sign$ function is applied to each dimension of $\vw$. To backpropagate the gradient to the non-differentiable target, we construct a surrogate probability model $\rho(\vw;D)$ which approximates $\sign(\vw)$ in the transformed target $\wt{p}(\sigma(\vw);D)p_0(\vw)$ by $\sigma(\vx)$ and relax the binary activation function $\{-1, 1\}$ by $\sigma$, where $\sigma$ is defined as $\sigma(\vx)=\frac{2}{1+\exp(-\vx)}-1.$ Here $\wt{p}(\sigma(\vw);D)$ is a differentiable approximation of $p_*(\sign(\vw);D).$  Then we apply GF-SVGD to update $\{\vw_i\}$ to approximate the transformed target distribution of $p_c(\vw;D)$ of $\vw$ as follows, 
$\vw_i \leftarrow \vw_i+\frac{\epsilon_{i}}{\Omega}\Delta \vw_i$, $\forall i=1,\cdots, n,$
\begin{equation}\label{bnn:update}
 \Delta \vw_i \!\! \leftarrow \!\! \! \sum_{j=1}^n \! \gamma_j [\nabla_{\vw}\log \rho(\vw_j;\!D_i)k(\vw_j\!,\!\vw_i)
             +\!\nabla_{\vw_j} k(\vw_j\!,\!\vw_i)]   
\end{equation}
where $D_i$ is batch data $i$ and $\mu_j =\rho(\vw_j; D_i)/p_c(\vw_j;D_i)$, $H(t) \overset{\mathrm{def}}{=}\sum_{j=1}^n \mathbb{I}(\mu_j\ge t)/n$, $\gamma_j= (H(\vw_j))^{-1}$ and $\Omega=\sum_{j=1}^n \gamma_j$. Empirical results on CIFAR-10 dataset shows that our method, which is applied to a popular network architecture, AlexNet, outperforms various baselines of ensemble learning BNN such as Adaboost and bagging. 

We propose a new goodness-of-fit testing method on discrete distributions, which evaluates whether a set of data $\{\vz_i\}_{i=1}^n$ match the proposed distribution $p_*(\vz)$. Our algorithm is motivated from the goodness-of-fit test method for continuous-valued distributions~\citep{liu2016kernelized}.
To leverage the gradient-free KSD to perform the goodness-of-fit test, we first transform the data $\{\vz_i\}_{i=1}^n$ and the candidate distribution $p_*(\vz)$ to the corresponding continuous-valued data and distributions using the transformation constructed in discrete distributional sampling aforementioned. Our method performs better and more robust than maximum mean discrepancy and discrete KSD methods under different setting on various discrete models. 

At the end of the thesis, we leverage some powerful tools from approximate inference to perform some applications on distributed model aggregation. In distributed, or privacy-preserving learning, a large dataset $\{\vx_i\}_{i=1}^N$ is distributed in $d$ local machines. We consider the setting where the data are evenly partitioned in each local machine to ease the notation, which can be easily generalized to uneven partition. We learn a $\{p(\boldsymbol{x}|\boldsymbol{\hat{\theta}}_k)\}_{k=1}^d$ in each local machine, where $\boldsymbol{\hat{\theta}}_k$ is the parameter of the probabilistic model. Our goal is to combine local models into a single model $p(\boldsymbol{x}|\boldsymbol{\hat{\theta}})$ that gives efficient statistical estimation. We focuses on a \emph{one-shot} approach for distributed learning, in which the learned local models $\{p(\boldsymbol{x}|\boldsymbol{\hat{\theta}}_k)\}_{k=1}^d$ are sent to a fusion center to form a single model that integrates all the information in the local models. A simple method is to linearly average the parameters of the local models, $\boldsymbol{\hat{\theta}}=\frac{1}{d}\sum_{j=1}^d \boldsymbol{\hat{\theta}}_j$, which tends to degenerate in practical scenarios for models with non-convex log-likelihood or non-identifiable parameters (such as latent variable models and neural models), and is not applicable at all for models with non-additive parameters (e.g., when the parameters have discrete or categorical values, the number of parameters in local models are different, or the parameter dimensions of the local models are different). Instead of linearly averaging the parameters, it is more meaningful way to geometrically average these local models in distribution space. To find such a geometrical mean model, our goal now is to find a model $p(\boldsymbol{x}|\boldsymbol{\hat{\theta}})$ such that the sum of the $\KL$ divergence between $p(\boldsymbol{x}|\boldsymbol{\hat{\theta}})$ and $p(\boldsymbol{x}\mid \boldsymbol{\hat{\theta}}_k)$ is minimized, $\boldsymbol{\hat{\theta}}=\argmax_{\vthe} \sum_{j=1}^d \KL(p(\boldsymbol{x}\mid \boldsymbol{\hat{\theta}}_k)|| p(\vx|\vthe)),$ which is called the \emph{KL-averaging} framework. To minimize this objective, it is equivalent to solve the following optimization problem, $\boldsymbol{\hat{\theta}}=\argmax_{\vthe}\sum_{k=1}^d\int p(\boldsymbol{x}\mid \boldsymbol{\hat{\theta}}_k)\log
p(\boldsymbol{x}\mid \boldsymbol{\theta})d\boldsymbol{x},$ where the integration cannot be evaluated in most cases. It casts a challenging optimization problem. To solve such an optimization, one more practical strategy is to generate bootstrap samples $\{\vv{\widetilde{x}}_j^k\}_{j=1}^n$ from each local model $p(\boldsymbol{x}\mid \boldsymbol{\hat{\theta}}_k)$, where $n$ is the number of the bootstrapped samples drawn from each local model $p(\boldsymbol{x}\mid \boldsymbol{\hat{\theta}}_k)$, and use the typical Monte Carlo method to estimate the integration~\citet{liu2014distributed}, $$\bd{\hat{\vthe}}_{\KL} =\argmax_{\vthe}\sum_{k=1}^d\frac1n\sum_{j=1}^n \log
p(\vv{\widetilde{x}}_j^k \mid \boldsymbol{\theta}).$$ Typical gradient descent method can be applied to solve this optimization to obtain a joint model. Unfortunately, the bootstrap procedure introduces additional noise and can significantly deteriorate the performance of the learned joint model. We prove that the mean square error (MSE) between $\bd{\hat{\vthe}}_{\KL}$ and the ground truth $\vthe^*$ has rate $O(N^{-1}+(nd)^{-1}).$ In order to has error rate $O(N^{-1})$, the total number of bootstrapped samples $nd$ should be proportional to $N,$ which is undesirable.
To reduce the induced variance, we introduce two variance-reduced techniques to more efficiently combine the local models, including a weighted M-estimator
that is both statistically efficient and practically powerful. The weighted M-estimator method is motivated from  \citet{henmi2007importance} to reduce the asymptotic variance in importance sampling, 
\begin{equation*}
\label{KLweigthed}
{\boldsymbol{\hat\theta}}_{\KL-W} =
\argmax_{\boldsymbol{\theta}\in\Theta} \bigg\{ \widetilde{\eta}(\boldsymbol{\theta})  \equiv \sum_{k=1}^d\frac1n\sum_{j=1}^n
\frac{p(\boldsymbol{\widetilde{x}}_j^k|\boldsymbol{\hat{\theta}}_k)}{p(\boldsymbol{\widetilde{x}}_j^k| \boldsymbol{\widetilde{\theta}}_k)}\log p(\boldsymbol{\widetilde{x}}_j^k|\boldsymbol{\theta}) \bigg\},
\end{equation*}
which can be viewed as a form of importance sampling as $\{\boldsymbol{\widetilde{x}}_j^k\}_{j=1}^n$ is more likely drawn from $p(\boldsymbol{x}\mid \boldsymbol{\hat{\theta}}_k).$
In Chapter~\ref{chap:boot}, we prove that the MSE between ${\boldsymbol{\hat\theta}}_{\KL-W}$ and the ground truth $\vthe^*$ has smaller rate than that of the naive estimator $\bd{\hat{\vthe}}_{\KL}.$ Typically, ${\boldsymbol{\hat\theta}}_{\KL-W}$ has MSE rate $O(N^{-1}+d^{-1}n^{-2}).$ The experimental results on simulated data have verified the correctness of our theoretical analysis. The experimental results on real data demonstrate the wide applicability of our proposed methods.
\chapter{Adaptive Importance Sampling\label{chap:is}}

Probabilistic modeling provides a fundamental framework for reasoning under uncertainty and modeling complex relations in machine learning. A critical challenge, however, is to develop efficient computational techniques for
approximating complex distributions. Specifically,
given a complex distribution $p(\bd x)$,
often known only up to a normalization constant,
we are interested in estimating integral quantities $\E_{p}[f(\vx)]$  for test functions $f(\vx).$
Popular approximation algorithms
include particle-based methods, such as
Monte Carlo, which construct a set of independent particles $\{\vv x_i\}_{i=1}^n$
whose empirical averaging $\frac{1}{n}\sum_{i=1}^n f(\vv x_i)$
forms unbiased estimates of $\E_p[f(\vx)]$. However, in real world applications, it is typically intractable to directly draw samples from $p(\bd x)$. Markov Chain Monte Carlo (MCMC), which has been introduced in previous chapter, is introduced to draw a set of samples~$\{\vv x_i\}_{i=1}^n$ to approximate the target $p(\bd x)$. In practice, it is difficult to examine when the Markov chains will converge to the stationary distribution $p(\bd x)$ and the samples $\{\vv x_i\}_{i=1}^n$ provides a good approximation of the target distribution $p(\bd x).$ Therefore, MCMC usually gives a biased estimation of the integral $\E_{p}[f(\vx)]$. Variational inference
which approximates $p$ with a simpler surrogate distribution $q(\vx)$
by minimizing a certain divergence between the target $p(\bd x)$ and the surrogate distribution $q(\vx)$ within a predefined parametric family of distributions.
Modern variational inference methods have found successful applications in
highly complex learning systems \citep[e.g., ][]{hoffman2013stochastic, kingma2013auto}. However, variational inference critically depends on the choice of parametric families. When the target distribution is not from the predefined parametric family of distributions, variational inference algorithms will definitely give a bias estimation of the integral $\E_{p}[f(\vx)].$ In practice, it is impossible to ensure the complex target distributions are within the predefined distribution family.

Stein variational gradient descent (SVGD, \citep{liu2016stein}) is an alternative
framework that integrates both the advantages of particle-based methods and variational inference algorithms. It starts with a set of initial particles $\{\vv x_i^0\}_{i=1}^n$, and
iteratively updates the particles using adaptively constructed {deterministic variable transforms:} $\bd{x}_i^{\ell} \gets \T_{\ell}(\bd{x}_i^{\ell-1}), ~~~~\forall i= 1,\ldots, n,$
where $\T_\ell$ is a variable transformation at the $\ell$-th iteration that maps current particles to new ones,
constructed adaptively at each iteration based on the most recent particles $\{\bd{x}_i^{\ell-1}\}_{i=1}^n$
that guarantee to push the particles ``closer'' to the target distribution $p$, in the sense that the KL divergence between the distribution of the particles and the target distribution $p$ can be iteratively decreased. More details on the construction of $\T_\ell$ can be found in this chapter. In practice, SVGD stops the iteration in finite iteration and use $\{\bd{x}_i^{\ell}\}_{i=1}^n$ to estimate the integral $\E_{p}[f(\vx)].$ However, the distribution $q_{\ell}(\vx)$ of the final particles $\{\bd{x}_i^{\ell}\}_{i=1}^n$ are different from $p(\vx).$ Therefore, SVGD cannot give a unbiased estimation of the integral $\E_{p}[f(\vx)].$ 

To address the problem of the bias estimations in MCMC, variational inference and SVGD, we introduce a family of algorithms in this chapter, importance sampling, which can give unbiased estimation of the integral $\E_{p}[f(\vx)].$ Importance sampling is a simple yet widely used technique in machine learning~\citep{bishop2006pattern}, deep learning~(importance weighted autoencoders\citep{burda2015importance}, etc.) and reinforcement learning~(proximal policy optimization algorithm\citep{schulman2017proximal}, etc.). Basically, importance sampling estimates the following expectation of the function $f(\vx)$ w.r.t. probability model $p(\vx)$ with a different distribution $q(\vx)$, which is easy to sample, and corrects the induced bias with importance weights,
\begin{equation}
\label{def:imp}
\E_p[f(\vx)] = \E_q[\frac{p(\vx)}{q(\vx)}f(\vx)]\approx \sum_{i=1}^n w(\vx_i)f(\vx_i)/(\sum_{i=1}^n w(\vx_i)),    
\end{equation}
where i.i.d. sample $\{\vx_i\}$ is drawn from $q$ and the weight is defined as $w(\vx_i)=p(\vx_i)/q(\vx_i).$ Importance sampling \eqref{def:imp} gives a unbiased estimation of the integral $\E_{p}[f(\vx)].$ However, in practice, when the surrogate distribution $q(\vx)$ is different from the target distribution $p(\vx)$, the importance weights $\{w(\vx_i)\}$ in \eqref{def:imp} usually have large variance. When the dimension of the input $\vx$ is high, it is typically challenging to construct the surrogate distribution $q(\vx)$ to ensure the variance of $\{w(\vx_i)\}$ is small. This will give a poor estimation for the expectation~\eqref{def:imp}. A family of adaptive importance sampling algorithms have been proposed to adaptively improve the approximation of the surrogate distribution $q(\vx)$ to the target distribution $p(\vx).$  

In the following, we will first discuss existing adaptive parametric importance sampling algorithms. Then we propose our main algorithm in this chapter, a novel non-parametric importance sampling algorithm. Finally, we will introduce a stochastic version of a widely used robust importance sampling algorithm, annealed importance sampling, when the posterior distribution (the target distribution) is defined over a large amount of data. 


\section{Parametric Adaptive Importance Sampling}
In order to improve the approximation of the surrogate distribution $q(\vx)$ to the target distribution $p(\vx),$ it is straightforward to come up with using a parametric form of the surrogate distribution $q(\vx)$ and optimizing $q(\vx)$ within the distribution family to find the best $q(\vx)$ to fit the target $p(\vx).$ In practice, the parametric distribution family $\{q_{\phi}(\vx)\}$ is typically chosen as exponential family or Gaussian mixture family \citep{cappe2008adaptive, ryu2014adaptive,  cotter2015parallel}. \citet{ryu2014adaptive} optimizes $q_{\phi}(\vx)$ within the exponential family by minimizing the variance of the estimation~\eqref{def:imp}, 
\begin{equation}
\label{imp:opt}
\mathcal{L}(\phi)=\mathrm{Var}(\frac{p(\vx)}{q_{\phi}(\vx)}f(\vx)) = \int_{\vx} \frac{p^2(\vx)}{q_{\phi}(\vx)}f^2(\vx)d\vx-\mathrm{Constant}.     
\end{equation}
The optimal $q_{\phi}(\vx)$ is proportional to $p(\vx)|f(\vx)|,$ which induces zero variance. But it is intractable to draw samples from such $q_{\phi}(\vx)\sim p(\vx)|f(\vx)|.$ In practice, we optimize \eqref{imp:opt} by using the gradient descent,
\begin{equation}
\label{imp:opt:est}
\nabla_{\phi} \mathcal{L}(\phi) = - \E_{q_{\phi}}[\frac{p^2(\vx)}{q_{\phi}^2(\vx)}f^2(\vx) \nabla_{\phi} \log q_{\phi}(\vx)].       
\end{equation}
The optimization objective in \eqref{imp:opt} itself has large variance and it is challenging to optimize such an objective to ensure $q_{\phi}(\vx)$ to approximate the target distribution $p(\vx)$ in high dimensional setting. In order to reduce the variance from the Monte Carlo estimation of \eqref{imp:opt:est}, one simple way is to introduce the score function method, $\E_{\vx \sim q_{\phi}(\vx)}[\nabla_{\phi} \log q_{\phi}(\vx)]=0,$

\begin{equation}
\label{imp:opt:cont}
\nabla_{\phi} \mathcal{L}(\phi) = - \E_{q_{\phi}}[\frac{p^2(\vx)}{q_{\phi}^2(\vx)}f^2(\vx)  \nabla_{\phi} \log q_{\phi}(\vx)]+\lambda\E_{q_{\phi}}[\nabla_{\phi} \log q_{\phi}(\vx)],   \end{equation}
where the optimal $\lambda$ has closed form,
\begin{equation}
\label{imp::opt:coeff}
\lambda = \mathrm{Var}(\nabla_{\phi}\log q_{\phi}(\vx))^{-1}\mathrm{Cov}[\frac{p^2(\vx)}{q_{\phi}^2(\vx)}f^2(\vx)\nabla_{\phi} \log q_{\phi}(\vx), \nabla_{\phi} \log q_{\phi}(\vx)],    
\end{equation}
and can be empirically estimated by samples $\{\vx_i\}_{i=1}^n$ from $q_{\phi}(\vx).$

In addition, the parametric assumptions restrict the choice of the proposal distributions and may give poor results when the assumption is inconsistent with
the target distribution $p(\vx)$. These limitations motivate us to develop more effective adaptive importance sampling algorithms.

\section{Non-Parametric Adaptive Importance Sampling}

In this section, we introduce a novel non-parametric adaptive importance sampling algorithm, which is motivated from SVGD~\citep{liu2016stein}. Before introducing our adaptive importance sampling algorithm, let us review SVGD~\citep{liu2016stein} from a slightly different perspective, the optimal variable transform viewpoint.

\begin{figure}[h]
\begin{centering}
 \includegraphics[width=0.99\textwidth]{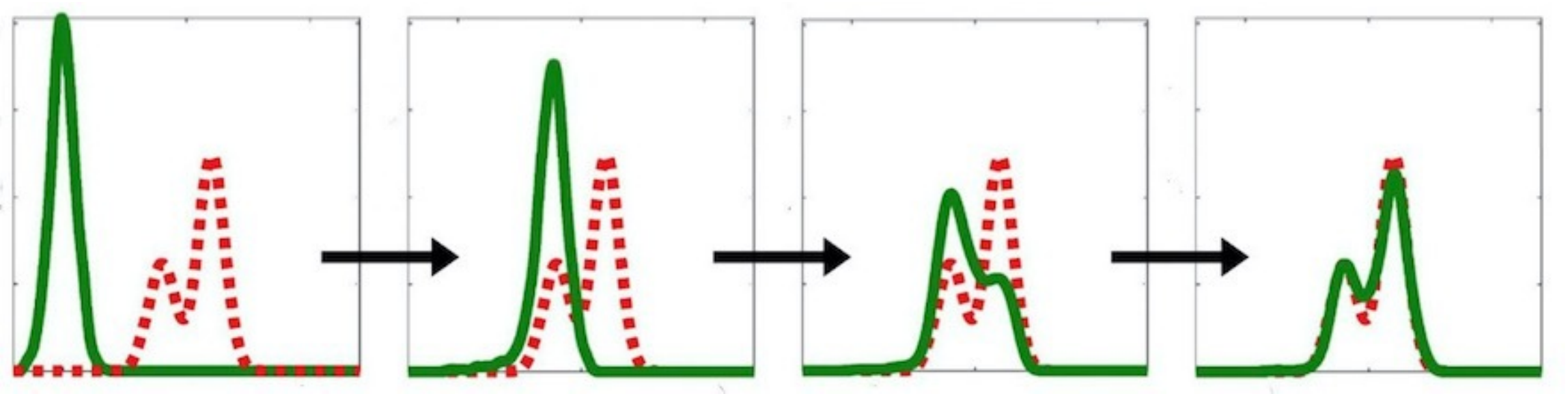} \\
 {$\bd{\red{p}}$: target distribution; ~~~~~~ $\bd{\color{green!50!black}{q_\ell}}$: approximate distribution.}
 \caption[Evolution of the transformed density functions in SVGD]{Evolution of the transformed density functions in SVGD. The target distribution in red dash line is an GMM with two modals. The initial distribution $q_0$ is transported by the transformation constructed by SVGD. \label{def:tran}}
\end{centering}
\end{figure}

SVGD starts with a set of initial particles
$\{\vv x_i^0\}_{i=1}^n$, and
iteratively updates the particles using adaptively constructed {deterministic variable transforms:} 
\begin{equation}
\bd{x}_i^{\ell} \gets \T_{\ell}(\bd{x}_i^{\ell-1}), ~~~~\forall i= 1,\ldots, n,    
\end{equation}
where $\T_\ell$ is a variable transformation at the $\ell$-th iteration that updates the current particles to new ones. The transform is constructed adaptively at each iteration based on the most recent particles $\{\bd{x}_i^{\ell-1}\}_{i=1}^n$
that guarantee to push the particles ``closer'' to the target distribution $p$, in the sense that the KL divergence between the distribution of the particles and the target distribution $p(\vx)$ can be iteratively decreased. Let us see one example in Fig.~\ref{def:tran}. The density functions of updated particles are getting closer and closer to the target distribution(red).

In the view of measure transport, SVGD iteratively transports the initial probability mass of the particles to the target distribution. SVGD constructs a path of distributions that bridges the initial distribution $q_0$ to the target distribution $p$ as follows,
\begin{align}\label{equ:qt1}
q_{\ell} =  (\T_{\ell} \circ \cdots \circ \T_{1})\sharp q_0, \quad \ell=1, \ldots, K.
\end{align}
where $\T \sharp q$ denotes the push-forward measure of $q$ through the transform $\T$, that is the distribution of $\vv z = \T(\vv x)$ when $\vv x\sim q$.

The story, however, is complicated
by the fact that the transform $\T_\ell$ is practically constructed on the fly \emph{depending} on the recent particles $\{\vx_i^{\ell-1}\}_{i=1}^n$,
which introduces complex dependency between the particles at the next iteration,
whose theoretical understanding requires mathematical tools in interacting particle systems \citep[e.g.,][]{braun1977vlasov, spohn2012large, del2013mean} and propagation of chaos \citep[e.g.,][]{sznitman1991topics}.
As a result, $\{\vv x_i^{\ell}\}_{i=1}^n$ can not be viewed as i.i.d. samples from $q_\ell$.

This makes it
difficult to analyze the results of SVGD and quantify their bias and variance. In this paper,
we propose a simple modification of SVGD
that ``decouples'' the particle interaction
and returns particles i.i.d. drawn from $q_\ell$;
we also develop a method to iteratively keep track of the importance weights of these particles,
which makes it possible to give consistent, or unbiased estimators within finite number of iterations of SVGD. 

Our method integrates 
 SVGD with importance sampling (IS) and combines their advantages:
it leverages the SVGD dynamics to obtain high quality proposals $q_\ell$ for IS
and also turns SVGD into a standard IS algorithm, inheriting the interpretability and theoretical properties of IS.
Another advantage of our proposed method is that it
provides an SVGD-based approach for estimating intractable normalization constants,
an inference problem that the original SVGD does not offer to solve.

The proposals $q_\ell$ in our method, however, are obtained by recursive variable transforms constructed in a nonparametric fashion
and become more complex as more transforms $\T_\ell$ are applied. In fact, one can view $q_\ell$ as the result of pushing $q_0$ through a neural network with $\ell$-layers,
constructed in a non-parametric, layer-by-layer fashion,
which provides a much more flexible distribution family than typical parametric families such as mixtures or exponential families.

There has been a collection of recent works,
 \citep[such as][]{rezende2015variational, kingma2016improved, marzouk2016introduction, spantini2017inference},
that approximate the target distributions
with complex proposals obtained by iterative variable transforms in a similar way to our proposals $q_\ell$ in \eqref{equ:qt1}.
The key difference, however, is that these methods explicitly parameterize the transforms $\T_\ell$
and optimize the parameters by back-propagation,
while our method, by leveraging the nonparametric nature of SVGD,
constructs the transforms $\T_\ell$ sequentially in closed forms,
requiring no back-propagation.
We introduce the basic idea of Stein variational gradient descent (SVGD) and Stein discrepancy. 
The readers are referred to \citet{liu2016stein} and \cite{liu2016kernelized} for more detailed introduction.

\subsection{Stein Discrepancy as Gradient of KL Divergence}
Let $p(\vx)$ be a density function on $\R^d$ which we want to approximate. We assume that we know $p(\vx)$ only up to a normalization constant, that is,
\begin{equation}
\label{barp}
p(\vx) = \frac{1}{Z} \bar p(\vx), ~~~~~ Z = \int \bar p(\vx) d\vx,
\end{equation}
where we assume we can only calculate $\bar p(\vx)$ and $Z$ is a normalization constant (known as the partition function) that is intractable to calculate exactly.
We assume that $\log p(\vx)$ is differentiable w.r.t. $\vx$, and we have access to $\nabla \log p(\vx) = \nabla \log \bar p(\vx )$ which does not depend on $Z$.

The main idea of SVGD is to use a set of sequential deterministic transforms
to iteratively push a set of particles $\{\vx_i\}_{i=1}^n$ towards the target distribution:
 \begin{align}
 \label{update}
 \begin{split}
 & \vx_i \leftarrow  \T(\vx_i), ~~~~ \quad \forall i=1, 2,\cdots, n \\
& \T(\vx) =  \bd{x} +\epsilon \bd{\phi}(\bd{x}),
\end{split}
 \end{align}
 where we choose the transform $\T$ to be an additive perturbation by a velocity field  $\ff$,
 with a magnitude controlled by a step size
 $\epsilon$ that is assumed to be small. 

The key question is the choice of the velocity field $\ff $; this is done by choosing $\ff$ to maximally decrease the $\KL$ divergence between the distribution of particles and the target distribution. Assume the current particles are drawn from $q$, and $\T\sharp q$ is the distribution of the updated particles, that is, $\T\sharp q$ is the distribution of
$\bd{\vx}'=\T(\vx) = \bd{\vx}+\epsilon\bd{\phi}(\bd{\vx})$ when $\vx \sim q$.
The optimal $\ff$ should solve the following functional optimization:
 \begin{align}
 \label{vgddecrease}
 \begin{split}
\mathbb{D}(q ~||~ p)  \overset{def}{=}  \max_{\bd{\phi} \in \mathcal \F  \colon  ||\ff ||_{\F}\leq 1 } \bigg\{ - \frac{d}{d\epsilon}\KL( \T\sharp q \mid\mid p)~ \big|_{\epsilon=0}  \bigg\},
\end{split}
 \end{align}
where $\mathcal{F}$ is a vector-valued normed function space that contains the set of candidate velocity fields $\ff$. 

The maximum negative gradient value $\mathbb{D}(q ~||~ p)$ in \eqref{vgddecrease}
provides a discrepancy measure between two distributions $q$ and $p$ and is known as \emph{Stein discrepancy} \citep{gorham2015measuring, liu2016kernelized, chwialkowski2016kernel}: if $\mathcal{F}$ is taken to be large enough, we have $\mathbb{D}(q ~||~ p) = 0$ iff there exists no transform to further improve the KL divergence between $p$ and $q$, namely $p = q$.

It is necessary to use an infinite dimensional function space $\mathcal{F}$ to
obtain good transforms,
which then casts a challenging functional optimization problem.
Fortunately, it turns out that a simple closed form solution can be obtained by taking
$\mathcal{F}$ to be an RKHS $\mathcal{H}=\mathcal{H}_0\times\cdots \mathcal{H}_0$,
where $\H_0$ is a RKHS of scalar-valued functions, associated with a positive definite kernel $k(x,x')$.
In this case, \citet{liu2016kernelized} showed that the optimal solution  of \eqref{vgddecrease} is $\ff^* / ||\ff^* ||_\H$, where
 \begin{equation}
 \label{is:transf}
 \bd{\phi}^*(\cdot)
 =\mathbb{E}_{\bd{x}\sim{q}}[\nabla_{\bd{x}} \log p(\bd{x})k(\bd{x},\cdot)+\nabla_{\bd{x}} k(\bd{x},\cdot)].
 \end{equation}
In addition, the corresponding Stein discrepancy, known as kernelized Stein discrepancy (KSD) \citep{liu2016kernelized, chwialkowski2016kernel, gretton2009fast, oates2016control}, can be shown to have the following closed form
\begin{align}
\label{equ:sdefine}
\mathbb{D}(q ~||~ p) = ||\ff^* ||_\H =  \big(\mathbb{E}_{x,x'\sim q}[\kappa_p(\vx,\vx')]\big)^{1/2},
\end{align}
where $\kappa_p(x,x')$ is a positive definite kernel defined by
 \begin{align}
\kappa_p & (\bd{x},  \bd{x}') 
 = \bd{s}_p(\bd{x})^\top k(\bd{x},\bd{x}')\bd{s}_p(\bd{x}') + \bd{s}_p(\bd{x})^\top \nabla_{\bd{x}'}k(\bd{x},\bd{x}') \notag\\
& +\bd{s}_p(\bd{x}')^\top \nabla_{\bd{x}} k(\bd{x},\bd{x}')+\nabla_{\bd{x}}\cdot(\nabla_{\bd{x}'}k(\bd{x}, \bd{x}')). \label{ksdkernel}
\end{align}
where $\bd s_p(\vx) \overset{def}{=} \nabla \log p(\vx)$. 
We refer to \citet{liu2016kernelized} 
for the  derivation of \eqref{ksdkernel}, and further treatment of KSD in  
\citet{chwialkowski2016kernel, oates2016control, gorham2017measuring}.

\subsection{Complex Dependence of Particles in SVGD}
In order to apply the derived optimal transform in the practical SVGD algorithm, we approximate the expectation $\E_{\vx\sim q}[\cdot]$ in \eqref{is:transf} using  the empirical averaging of the current particles,
that is, given particles $\{\vx_i^\ell\}_{i=1}^n$ at the $\ell$-th iteration, we construct the following velocity field:
\begin{align}\label{equ:phit}
\!\!\! \!  \ff_{\ell+1}(\cdot ) =\frac{1}{n} \sum_{j=1}^n [\nabla \log p(\bd{x}_j^{\ell})k(\bd{x}_j^{\ell},\cdot)+\nabla_{\bd{x}_j^{\ell}} k(\bd{x}_j^{\ell},\cdot)].
\end{align}
The SVGD update at the $\ell$-th iteration is then given by
%
\begin{align}\label{equ:svgdup2}
\begin{split}
& \vx_i^{\ell+1}  \gets \T_{\ell+1}(\vx_{i}^{\ell}), \\
& \T_{\ell+1}(\vx) =\vx + \epsilon \ff_{\ell+1}(\vx).
\end{split}
\end{align}
Here transform $\T_{\ell+1}$ is adaptively constructed based on the most recent particles $\{\vx_i^\ell\}_{i=1}^n$.
 Assume the initial particles $\{\vx_i^0\}_{i=1}^n$ are i.i.d. drawn from some distribution $q_0$, then the pushforward maps of $\T_\ell$ define a sequence of distributions that bridges between $q_0$ and $p$:
\begin{align}\label{equ:qt}
q_{\ell} =  (\T_{\ell} \circ \cdots \circ \T_{1})\sharp q_0, \quad \ell=1, \ldots, K,
\end{align}
where $q_\ell$ forms increasingly better approximation of the target $p$ as $\ell$ increases.
Because $\{\T_\ell\}$ are nonlinear transforms,
$q_\ell$ can represent highly complex distributions even when the original $q_0$ is simple.
In fact, one can view $q_\ell$ as
a deep residual network \citep{he2016deep} constructed layer-by-layer in a fast, nonparametric fashion.

However,
because the transform $\T_{\ell}$ depends on the previous particles $\{\vx_{i}^{\ell-1}\}_{i=1}^n$ as shown in \eqref{equ:phit},
the particles $\{\vx_i^{\ell}\}_{i=1}^n$, after the zero-th iteration, depend on each other in a complex fashion,
and do not, in fact, straightforwardly follow distribution $q_{\ell}$ in \eqref{equ:qt}.
Principled approaches for analyzing such interacting particle systems can be found in~\citet[e.g.,][]{braun1977vlasov, spohn2012large, del2013mean, sznitman1991topics}. The goal of this work, however, is to provide a simple method to ``decouple'' the SVGD dynamics, transforming it into a standard 
importance sampling method that is amendable to easier analysis and interpretability, and also applicable to more general inference tasks such as estimating partition function of unnormalized distribution where SVGD cannot be applied.



\subsection{Stein Variational Adaptive Importance Sampling}
In this section, we introduce our main Stein variational importance sampling~(SteinIS) algorithm. Our idea is simple.
We initialize the particles $\{\vx_i^0\}_{i=1}^n$ by i.i.d. draws from an initial distribution $q_0$ and partition them into two sets,
including a set of \emph{leader particles} $\vx_A^\ell = \{\vx_i^\ell  \colon  i\in A\}$
and
 \emph{follower particles}
$\vx_B^\ell = \{\vx_i^\ell  \colon  i\in B\}$, with $B =  \{1,\ldots, n\} \setminus A$,
where the leader particles $\vx_A^\ell$ are responsible for constructing the transforms, using the standard SVGD update \eqref{equ:svgdup2},
while the follower particles $\vx_B^\ell$ simply follow the transform maps constructed by $\vx_A^\ell$ and do not contribute to the construction of the transforms. 
In this way, the follower particles $\vx_B^{\ell}$ are independent conditional on the leader particles $\vx_A^\ell$. 

\begin{figure}[H]
\centering
\includegraphics[width=.8\linewidth]{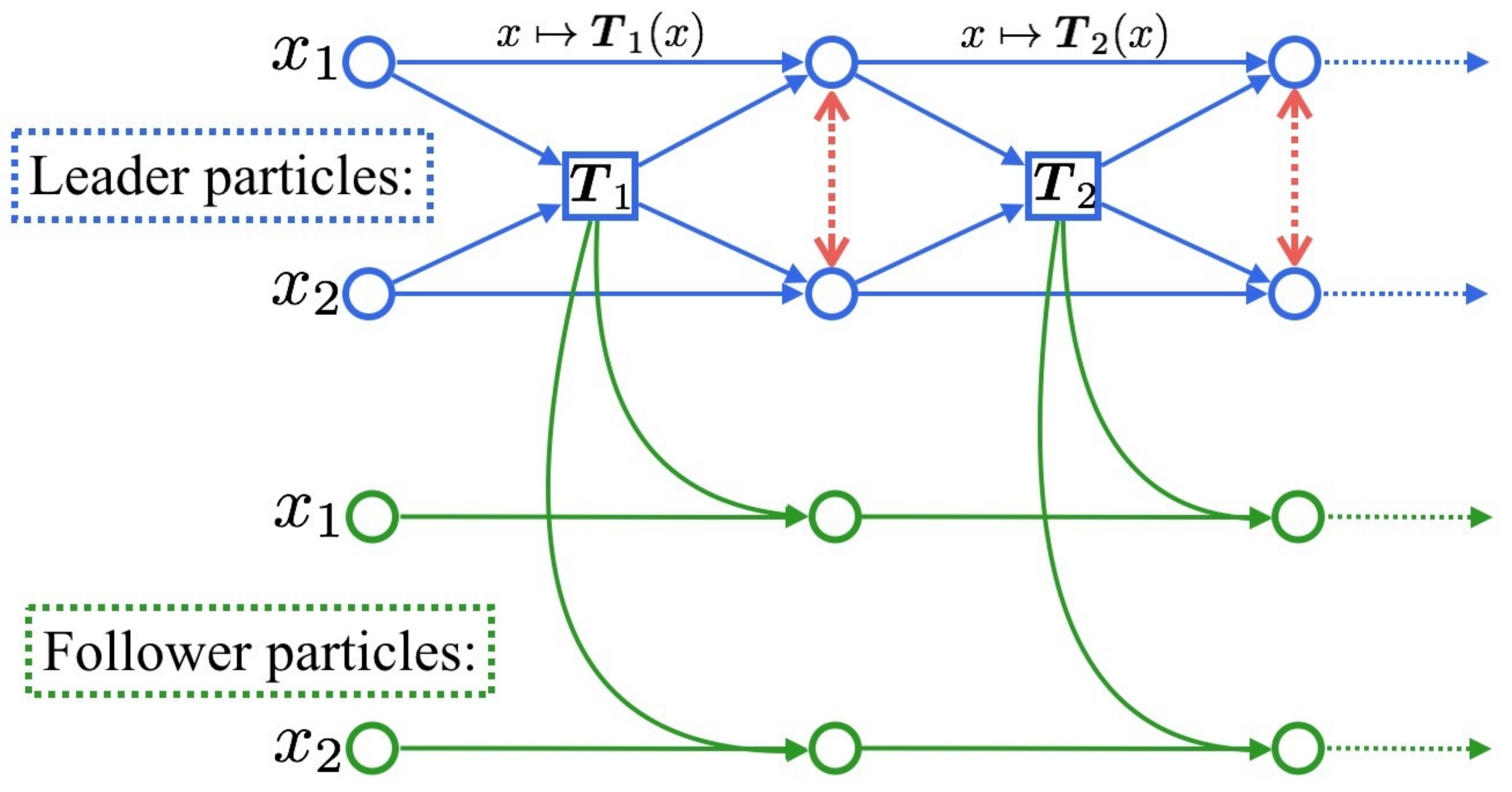}
\caption[Illustration of the decoupled particles and their update in SteinIS]{Illustrating the decoupled particles and their update in SteinIS. The leader particles set $\vx_A^\ell$ is used to construct the transform $T_{\ell}$ and the follower particles  $\vx_B^{\ell}$ is updated by the constructed transform. The leader particles $\vx_A^\ell$ are interactive and dependent on each other. The follower particles $\vx_B^\ell$ can be viewed as i.i.d. draws from $q_\ell$, given fixed leader particles $\vx_A^\ell$. \label{svgd:transp}}
\end{figure}

Conceptually, we can think that we first construct all the maps $\T_\ell$ by evolving the leader particles $\vx_A^\ell$,
and then push the follower particles through $\T_\ell$ in order to draw exact, i.i.d. samples from $q_\ell$ in \eqref{equ:qt}.
Note that this is under the assumption the leader particles $\vx_A^\ell$
has been observed and fixed, which is necessary because the transform $\T_\ell$ and
distribution $q_\ell$ depend on $\vx_A^\ell$.

In practice, however, we can simultaneously update both the leader and follower particles, by a simple modification of the original SVGD \eqref{equ:svgdup2},
\begin{align}\notag
& \vv x_{i}^{\ell+1} \gets \vv x_i^\ell + \epsilon \ff_{\ell+1}(\vv x_i^\ell), ~ \forall i \in A\cup B, \\
&\ff_{\ell+1}(\cdot) =  \frac{1}{|A|}\sum_{j\in A} [\nabla \log p(\vv x_j^\ell) k(\vv x_j^\ell,  \cdot) + \nabla_{\vv x_j^\ell} k(\vv x_j^\ell, \cdot)].
\end{align}
where the only difference is that we restrict the empirical averaging in \eqref{equ:phit} to the set of 
the leader particles $\vx_A^\ell$. The whole procedure is summarized in Algorithm~\ref{VarIS:algo}.
The relationship between the particles in set $A$ and $B$ can be more easily understood in Figure~\ref{svgd:transp}.
\begin{algorithm}[tb] %
\caption{Stein Variational Importance Sampling}  \label{VarIS:algo}
\begin{algorithmic}
\STATE {\bf Goal}: Obtain i.i.d. importance sample $\{\vx_i^K, ~ w_i^K\}$ for $p$.
\STATE {Initialize} $\vx_{A}^0$ and $\vx_B^0$ by i.i.d. draws from $q_0$.
\STATE {Calculate} $\{ q_0(\vx_i^0) \}, \forall i\in B.$
\FOR{iteration $\ell =0, \ldots, K-1$}
\STATE
1. Construct the map using the leader particles $\vx_A^\ell$
\begin{align}\notag
 \!\!\!\! \!\!\!\!
\ff_{\ell+1}(\cdot) =  \frac{1}{|A|}\sum_{j\in A} [\nabla \log p(\vv x_j^\ell) k(\vv x_j^\ell,  \cdot) + \nabla_{\vv x_j^\ell} k(\vv x_j^\ell, \cdot)].
\end{align}
2. Update both the leader and follower particles 
$$
\vv x_{i}^{\ell+1} \gets \vv x_i^\ell + \epsilon \ff_{\ell+1}(\vv x_i^\ell), ~~~~ \forall i \in A\cup B.
$$
3. Update the density values (for $i\in B$) by
\begin{equation*}
q_{\ell+1}(\vx_{i}^{\ell+1}) =  q_{\ell}(\vx_{i}^{\ell}) \cdot  | \mathrm{det}(I ~+~ \epsilon \nabla_{\bd{x}} \bd{\ff}_{\ell+1}(\bd{x}^{\ell}_i))|^{-1}
\end{equation*}
\ENDFOR
\STATE {Calcuate} $w_i^{K} = p(\vv x_i^{K}) / q_K(\vv x_i^{K}), \forall i\in B.$
\STATE {\bf Outputs}: i.i.d. importance sample $\{\vx_i^K, ~ w_i^K\}$ for $i\in B.$
\end{algorithmic}
\end{algorithm}

\paragraph{Calculating the Importance Weights}
Because $q_\ell$ is still different from $p$ when we only apply finite number of iterations $\ell$, which introduces deterministic biases if we directly use $\vx_B^\ell$ to approximate $p$.

We address this problem by further turning the algorithm into an importance sampling algorithm with importance proposal $q_\ell$. Specifically, we calculate the importance weights of the particles $
\{\vx_i^\ell\}$:
\begin{align}\label{equ:wt}
w_i^\ell= \frac{\bar p(\vx_i^{\ell})}{q_\ell(\vx_i^\ell)},
\end{align}
where $\bar p$ is the unnormalized density of $p$, that is, $p(\vx) = \bar p(\vx)/Z$ as in \eqref{barp}. 
In addition, the importance weights in \eqref{equ:wt} can be calculated based on the following formula:
\begin{equation}
\label{density}
q_\ell(\bd{x}^\ell)=q_0(\bd{x}^0)\prod_{\jmath=1}^\ell |\mathrm{det}(\nabla_{\bd{x}} \bd{T}_\jmath(\bd{x}^{\jmath-1}))|^{-1},
\end{equation}
where $\bd{T}_\ell$ is defined in~\eqref{equ:svgdup2} and we assume that
 the step size $\epsilon$ is small enough so that each $\T_\ell$ is an one-to-one map.
 As shown in Algorithm~\ref{VarIS:algo} (step 3), 
 \eqref{density} can be calculated recursively as we update the particles. 

With the importance weights calculated, we turn SVGD into a standard importance sampling algorithm. 
For example, we can now estimate expectations of form $\E_p f$ by 
$$
 \hat \E_{p}[f] = \frac{\sum_{i\in B} w_i^\ell f(\vx_i^\ell)}{\sum_{i\in B} w_i^\ell}, 
$$
which provides a consistent estimator of $\E_{p} f$ when we use finite number $\ell$ of transformations.  
Here we use the self normalized weights because
$\bar p(\vx)$ is unnormalized.
Further, the sum of the unnormalized weights provides an unbiased estimation for the normalization constant $Z$:
$$
\hat Z = \frac{1}{|B|} \sum_{i\in B} w_i^\ell,
$$
which satisfies the unbiasedness property $\E[\hat Z] = Z$. Note that the original SVGD does not provide a method for estimating normalization constants,
although, as a side result of this work, Section 4 will discuss another method for estimating $Z$ that is more directly motivated by SVGD.

We now analyze the time complexity of our algorithm. 
Let $\alpha(d)$ be the cost of computing $\bd{s}_p(\bd{x})$ and 
$\beta(d)$ be the cost of evaluating kernel $k(\vx, \vx')$ and its gradient $\nabla k(\vx, \vx')$. Typically, both $\alpha(d)$ and $\beta(d)$ grow linearly with the dimension $d. $ In most cases, $\alpha(d)$ is much larger than $\beta(d)$. 
The complexity of the original SVGD with $|A|$ particles is $O(|A|\alpha(d)+|A|^2\beta(d))$, 
and the complexity of Algorithm~\ref{VarIS:algo} is $O(|A|\alpha(d)+|A|^2\beta(d)+|B||A|\beta(d)+|B| d^3 ),$   
where the $O(|B|d^3)$ complexity comes from calculating the determinant of the Jacobian matrix, 
which is expensive when dimension $d$ is high, but is the cost to pay for having a consistent importance sampling estimator in finite iterations
and for being able to estimate the normalization constant $Z$.
Also, by calculating the effective sample size based on the importance weights, 
we can assess the accuracy of the estimator, and early stop the algorithm when a confidence threshold is reached. 

One way to speed up our algorithm in empirical experiments is to parallelize the computation of Jacobian matrices for all follower particles in GPU. It is possible, however, to develop efficient approximation for the determinants by leveraging the special structure of the Jacobean matrix; note that 
\begin{align}
&\nabla_{\bd{y}} \bd{T}(\bd{y}) = I  + \epsilon A,  \notag \\
&A = \frac{1}{n}\sum_{j=1}^n [\nabla_{\vv x} \log p(\vv x_j)^\top \nabla_{\vv y} k(\vv x_j, \vv y) +
\nabla_{\vv x} \nabla_{\vv y} k(\vv x_j, \vv y)]. \notag
 \end{align}
 Therefore, $\nabla_{\bd{y}} \bd{T}(\bd{y})$ 
 is close to the identity matrix $I$ when the step size is small.
This allows us to use Taylor expansion for approximation:

\begin{pro}
\label{detapprox}
Assume $\epsilon < 1/\rho(A)$, where $\rho(A)$ is the spectral radius of $A$, that is,
 $\rho(A) =\max_j |\lambda_j(A)|$ and $\{\lambda_j\}$ are the eigenvalues of $A$. We have
\begin{equation}
\label{Jacobapprox}
\mathrm{det}(I +\epsilon A) =\prod_{k=1}^d (1+\epsilon a_{kk})+ O(\epsilon^2),
\end{equation}
where $\{a_{kk}\}$ are the diagonal elements of $A$.
\end{pro}

\begin{proof}
Use the Taylor expansion of $\mathrm{det}(I +\epsilon A)$. 
Note that $\mathrm{det}(I +\epsilon A)=\exp(\mathrm{trace}(\log (I +\epsilon A))),$
and $\log (I +\epsilon A)=1 + \epsilon A + O(\epsilon^2).$ 
and $\log (I +\epsilon A)=\sum_{n=1}^\infty \frac{(-1)^{n-1}}{n}\epsilon^nA^n,$
where $A^n= A^{n-1}A.
$ T $\mathrm{det}(I +\epsilon A) =\prod_k^d (1+\epsilon a_{kk})+ O(\epsilon^2).$
\end{proof}
Therefore, one can approximate the determinant with approximation error $O(\epsilon^2)$ using linear time $O(d)$ w.r.t. the dimension.
Often the step size is decreasing with iterations,
and a way to trade-off the accuracy with computational cost is to 
use the exact calculation in the beginning when the step size is large,
and switch to the approximation when the step size is small.

The idea of constructing a path of distributions $\{q_\ell\}$ to bridge
the target distribution $p$ with a simpler distribution $q_0$
invites connection to ideas such as annealed importance sampling (AIS) \citep{neal2001annealed}
and path sampling (PS) \citep{gelman1998simulating}. These methods typically construct an annealing path using geometric averaging of the initial and target densities instead of variable transforms,
which does not build in a notion of variational optimization as the SVGD path.
In addition, it is often intractable to directly sample distributions on the geometry averaging path, and hence  AIS and PS  need additional mechanisms in order to construct proper estimators.

\paragraph{Monotone Decreasing of KL divergence} 
One nice property of algorithm~\ref{VarIS:algo} is that the KL divergence between the iterative distribution $q_\ell$ and $p$ is monotonically decreasing. This property can be more easily understood by considering our iterative system in continuous evolution time as shown in \citet{liu2017stein}. 
Take the step size $\epsilon$ of the transformation defined in \eqref{update} to be infinitesimal, 
and define the continuos time $t = \epsilon \ell$. Then the evolution equation of random variable $\vx^t$ is governed by the following nonlinear partial differential equation~(PDE), 
\begin{equation}
\label{part}
\frac{d\bd{x}^t}{dt}=\mathbb{E}_{\bd{x}\sim{q_t}}[\bd{s}_p(\bd{x})k(\bd{x},\bd{x}^t)+\nabla_{\bd{x}} k(\bd{x},\bd{x}^t)],
\end{equation}
where $t$ is the current evolution time and $q_t$ is the density function of $\vx^t.$ The current evolution time $t= \epsilon \ell$ when $\epsilon$ is small and $\ell$ is the current iteration. We have the following proposition (see also \citet{liu2017stein}): 
\begin{pro}
\label{pro2}
Suppose random variable $\bd{x}^t$ is governed by PDE \eqref{part}, then its density $q_t$ is characterized by
\begin{equation}
\label{diffode}
\frac{\partial q_t}{\partial t}=-\mathrm{div}(q_t\mathbb{E}_{\bd{x}\sim{q_t}}[\bd{s}_p(\bd{x}) k(\bd{x},\bd{x}^t)+\nabla_{\bd{x}} k(\bd{x},\bd{x}^t)]),
\end{equation}
where $\mathrm{div}(\bd{f})=\trace(\nabla \vv f) = \sum_{i=0}^d \partial f_i(\bd{x})/\partial x_i$, and  $\bd{f}=[f_1,\ldots, f_d]^\top.$ 
\end{pro}
The proof of proposition~\ref{pro2} is similar to the proofs of proposition 1.1 in~\citet{jourdain1998propagation} and lemma 1 in \citet{dai2019opaque}. 
Proposition~\ref{pro2} characterizes the evolution of the density function $q_t(\bd{x}^t)$ when the random variable $\bd{x}^t$ is evolved by ~\eqref{part}. The continuous system captured by~\eqref{part} and ~\eqref{diffode} is a type of Vlasov process which has wide applications in physics, biology and many other areas~\citep[e.g.,][]{braun1977vlasov}.
As a consequence of proposition~\ref{pro2}, one can show the following nice property:
\begin{equation}
\label{klksd}
\frac{d\mathrm{KL}(q_t\mid\mid p)}{dt}=-\mathbb{D}(q_t ~||~ p)^2<0,
\end{equation}
which is proved by theorem 4.4 in \citet{liu2017stein}. 
Equation ~\eqref{klksd} indicates that the KL divergence between the iterative distribution $q_t$ and $p$ is monotonically decreasing with a rate of $\mathbb{D}(q_t~||~ p)^2$. 

\section{A Path Integration Method}
\begin{algorithm}[tb]
\caption{SVGD with Path Integration for estimating $\KL(q_0 ~||~p)$ and $\log Z$}
\label{ksdalgo}
\begin{algorithmic}[1]
\STATE {\bfseries Input:} Target distribution $p(x) = \bar p(x)/Z$; an initial distribution $q_0$.
\STATE {\bfseries Goal:} Estimating $\KL(q_0\mid\mid p)$ and the normalization constant $\log Z.$
\STATE{ Initialize $\hat K =0.$ Initialize particles $\{\vx_i^0\}_{i=1}^n\sim q_0(\bd{x}).$}
\STATE{Compute $\hat{\mathbb{E}}_{q_0}[\log(q_0(\vx)/\overline{p}(\vx))]$ via sampling from $q_0.$}
\WHILE{iteration $\ell$}
\STATE 
$$
\hat K \gets \hat K + \epsilon  \hat{\mathbb{D}}({q}_\ell ~||~ p)^2,
$$
$$\vx_i^{\ell+1} \gets \vx_i^\ell + \bd{\phi}_{\ell +1}(\vx_i^\ell), $$
where $\hat{\mathbb{D}}({q}_\ell ~||~ p)$ is defined in \eqref{equ:hats}.
\ENDWHILE
\STATE {Estimate $\KL(q_0 ~||~  p)$ by $\hat K$ and $\log Z$ by $\hat{\mathbb{D}} - \hat{\mathbb{E}}_{q_0}[\log(q_0(\vx)/\overline{p}(\vx))].$}
\end{algorithmic}
\end{algorithm}

We mentioned that the original SVGD
does not have the ability to estimate the partition function. 
Section 3 addressed this problem by 
turning SVGD into a standard importance sampling algorithm in Section 3.
Here we introduce another method
for estimating KL divergence and normalization constants that is more directly motivated by the original SVGD,
by leveraging the fact that the Stein discrepancy is a type of gradient of KL divergence.
This method does not need to estimate the importance weights but has to run SVGD to converge to diminish the Stein discrepancy between intermediate distribution $q_\ell$ and $p$. In addition, this method does not perform as well as Algorithm 1 as we find empirically.
Nevertheless, we find this idea is conceptually interesting and useful to discuss it.

Recalling equation {\eqref{vgddecrease}} in Section 2.1, we know that
if we perform transform $\T(\vx) = \vx +  \epsilon \ff^*(\vx)$ with $\ff^*$ defined in \eqref{is:transf},
the corresponding decrease of KL divergence would be
\begin{align}
\begin{split}
 \KL(q ~||~ p)  - \KL(\T\sharp q  ~||~ p)
&  \approx \epsilon \cdot || \ff^* ||_\H \cdot \S(q ~|| ~ p)  \\
&  \approx \epsilon \cdot \S(q ~|| ~ p)^2,
\end{split}
\label{def:kldecrease}
\end{align}
where we used the fact that $\S(q~||~p) = ||\ff^*||_\H$, shown in \eqref{equ:sdefine}.
Applying this recursively on $q_\ell$ in \eqref{def:kldecrease}, we get
$$
 \KL(q_0 ~||~ p) - \KL(q_{\ell+1} ~||~ p)  \approx \sum_{\jmath = 0}^\ell
\epsilon \cdot \S(q_{\jmath} ~|| ~ p)^2.
$$
Assuming $\KL(q_{\ell} ~||~ p) \to 0$ when $\ell \to \infty$,
we get
\begin{equation}
\label{klestimate}
\KL(q_0 ~||~ p)  \approx \sum_{\ell = 0}^\infty
\epsilon \cdot \S(q_{\ell} ~|| ~ p)^2.
\end{equation}

By \eqref{equ:sdefine}, the square of the KSD can be empirically estimated via V-statistics, which is given as
\begin{align}\label{equ:hats}
\hat \S(q_{\ell} ~|| ~ p)^2 = \frac{1}{n^2}\sum_{i=1}^n\sum_{j=1}^n \kappa(\vx_i^\ell, \vx_j^\ell).
\end{align}
Overall, equation~\eqref{klestimate} and \eqref{equ:hats} give an estimator of the KL divergence between $q_0$ and $p = \bar p(\bd{x})/Z.$
This can be transformed into an estimator of the log normalization constant $\log Z$ of $p$, by noting that
\begin{equation}
\log Z=\KL(q_0\mid\mid p)-\mathbb{E}_{q_0}[\log(q_0(\vx)/\overline{p}(\vx))],
\end{equation}
where the second term can be estimated by drawing a lot of samples to diminish its variance since the samples from $q_0$ is easy to draw. The whole procedure is summarized in Algorithm~\ref{ksdalgo}.

\section{Empirical Experiments of SteinIS}
\begin{figure}[t]
\centering
\begin{tabular}{cc}
\includegraphics[height=0.3\textwidth]{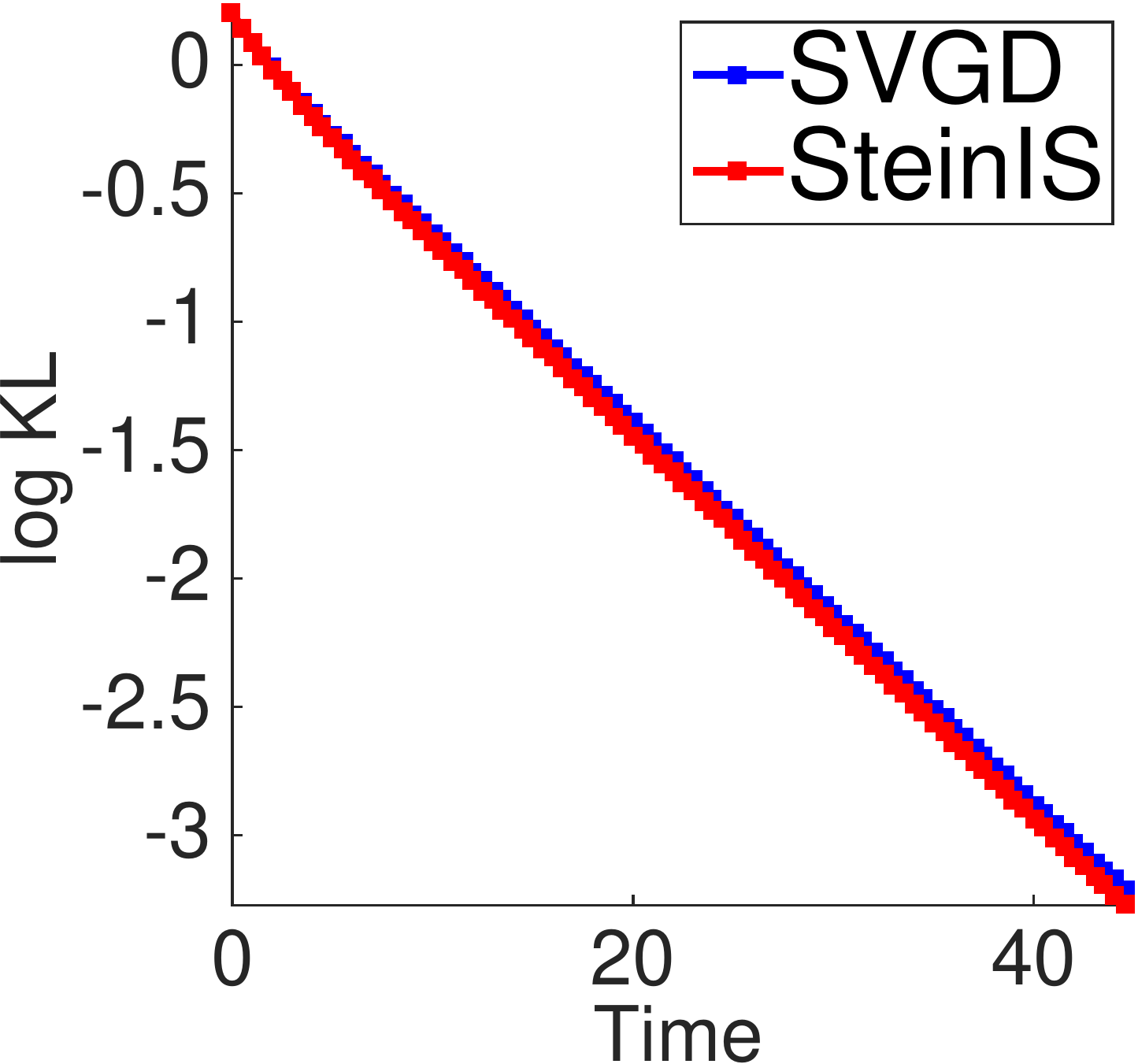} &
\includegraphics[height=0.3\textwidth]{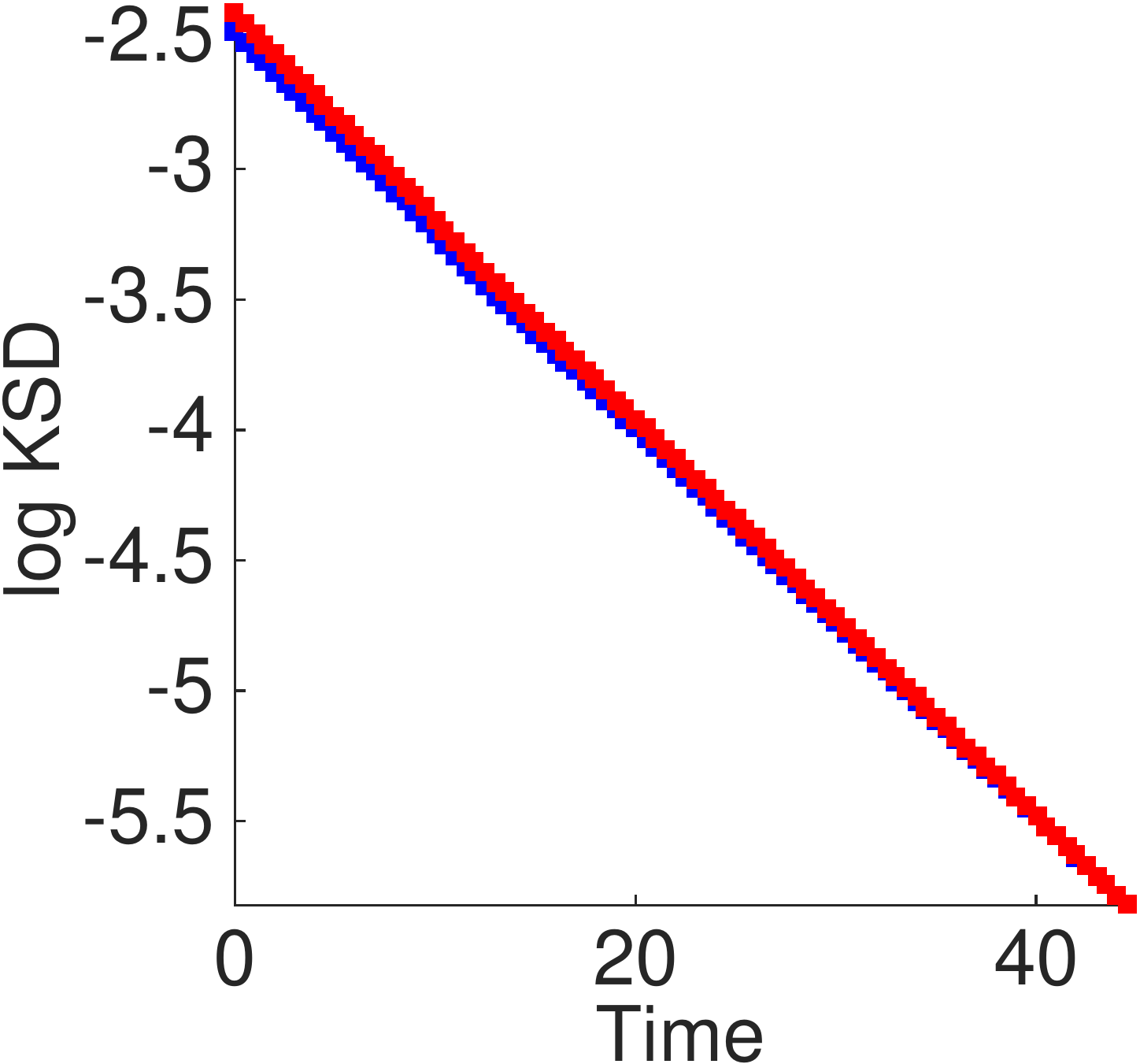} \\
{\small (a)  KL} &
{\small (b)  KSD}
\end{tabular}
\caption[Verification of the monotone deceasing of $\KL$ divergence in SteinIS on GMM with 10 mixture components]{Verifying the monotone deceasing of $\KL$ divergence in SteinIS on GMM with 10 mixture components. $d=1.$ In SVGD, 500 particles are evolved. In SteinIS, $|A|=200$ and $|B|=500$. For SVGD and SteinIS, all particles are drawn from the same Gaussian distribution $q_0(\bd{x}).$ \label{fig:KL}}
\end{figure}

We study the empirical performance of our proposed algorithms on both simulated and real world datasets. 
We start with toy examples to numerically investigate some theoretical properties of our algorithms, 
and compare it with traditional adaptive IS on non-Gaussian, multi-modal distributions. We also employ our algorithm to estimate the partition function of Gaussian-Bernoulli Restricted Boltzmann Machine(RBM), a graphical model widely used in deep learning~\citep{welling2004exponential, hinton2006reducing}, and to evaluate the log likelihood of decoder models in variational autoencoder~\citep{kingma2013auto}.

We summarize some hyperparameters used in our experiments. We use RBF kernel $k(\bd{x}, \bd{x}')=\exp(-\|\bd{x}-\bd{x}'\|^2/h),$ where $h$ is the bandwidth. In most experiments, we let $h {=} \mathrm{med^2}/(2\log(|A|+1))$, where $\mathrm{med}$ is the median of the pairwise distance of the current leader particles $\vx_A^\ell$, 
and $|A|$ is the number of leader particles.  The step sizes in our algorithms are chosen to be $\epsilon =\alpha/(1+\ell)^\beta,$ where $\alpha$ and $\beta$ are hyperparameters chosen from a validation set to achieve best performance. When $\epsilon\le 0.1$, we use first-order approximation to calculate the determinants of Jacobian matrices as illustrated in proposition~\ref{detapprox}.

In what follows, we use ``{AIS}'' to refer to the annealing importance sampling with Langevin dynamics as its Markov transitions, and use ``{HAIS}'' to denote the annealing importance sampling whose Markov transition is Hamiltonian Monte Carlo (HMC). We use "transitions" to denote the number of intermediate distributions constructed in the paths of both SteinIS and AIS. 
A transition of HAIS may include $L$ leapfrog steps, as implemented by ~\citet{wu2016quantitative}. 

\paragraph{Verification of Monotone Decreasing of KL Divergence in SteinIS} We start with testing our methods on simple 2 dimensional Gaussian mixture models with 10 randomly generated mixture components. The dimension of $\vx$ in $p(\vx)$ is one. In SVGD, 500 particles are evolved. In SteinIS, the leader particle set size $|A|=200$ and the follower particle set size $|B|=500$. For SVGD and SteinIS, all particles are drawn from the same Gaussian distribution $q_0(\bd{x}).$ First, we numerically investigate the convergence of KL divergence between the particle distribution $q_t$ (in continuous time) and $p.$ Sufficient particles are drawn and infinitesimal step $\epsilon$ is taken to closely simulate the continuous time system, as defined by \eqref{part}, \eqref{diffode} and \eqref{klksd}. Figure~\ref{fig:KL}(a)-(b) show that the KL divergence $\KL(q_t, p)$, as well as the squared Stein discrepancy $\mathbb{D}(q_t, p)^2$, seem to decay exponentially in both SteinIS and the original SVGD. 
 This suggests that the quality of our importance proposal $q_t$ improves quickly as we apply sufficient transformations. However, it is still an open question to establish the exponential decay theoretically; see \citet{liu2017stein} for a related discussion. 

\begin{figure*}[t]
\centering
\begin{tabular}{cccc}
\includegraphics[height=0.19\textwidth]{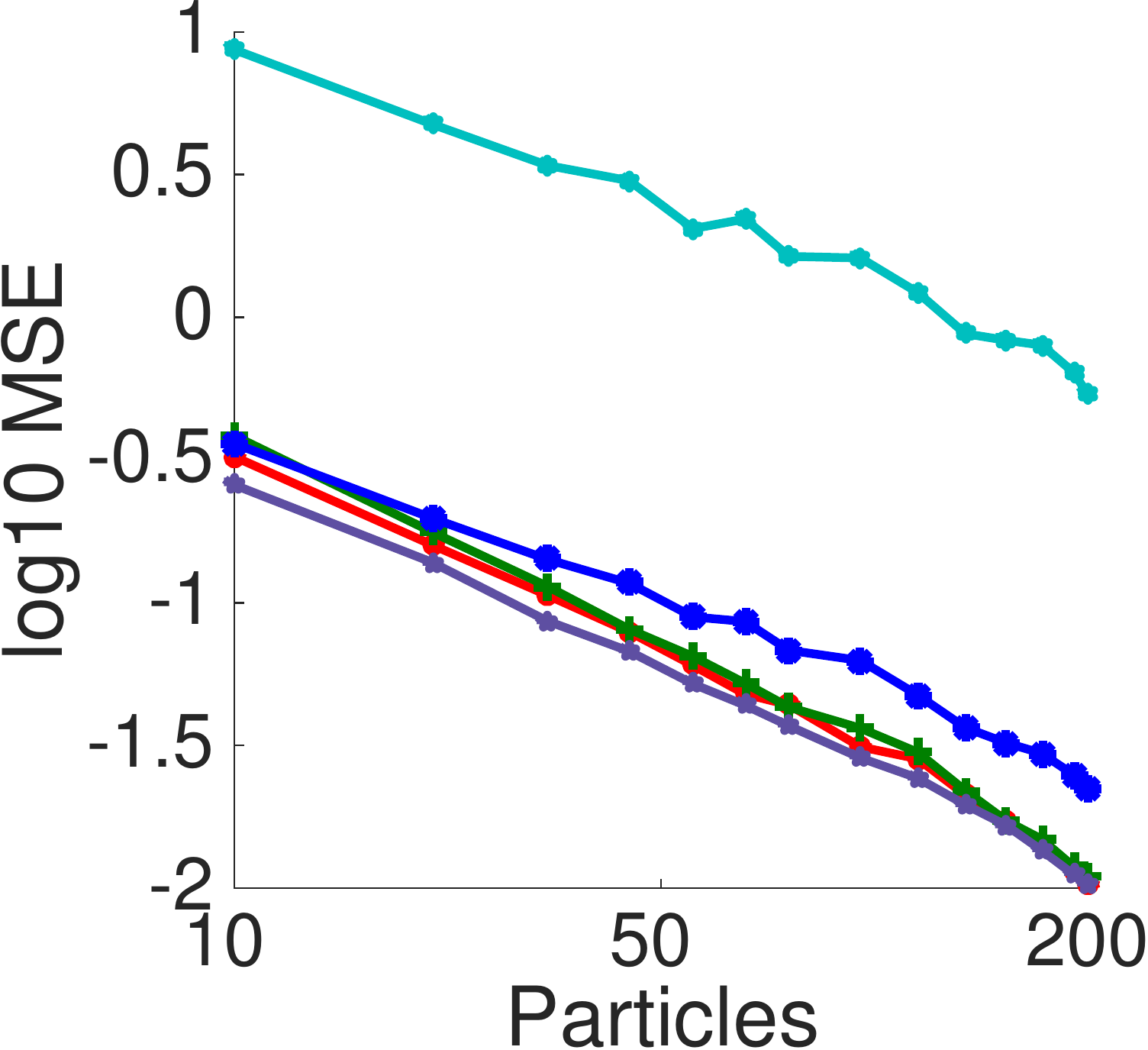} &
\includegraphics[height=0.19\textwidth]{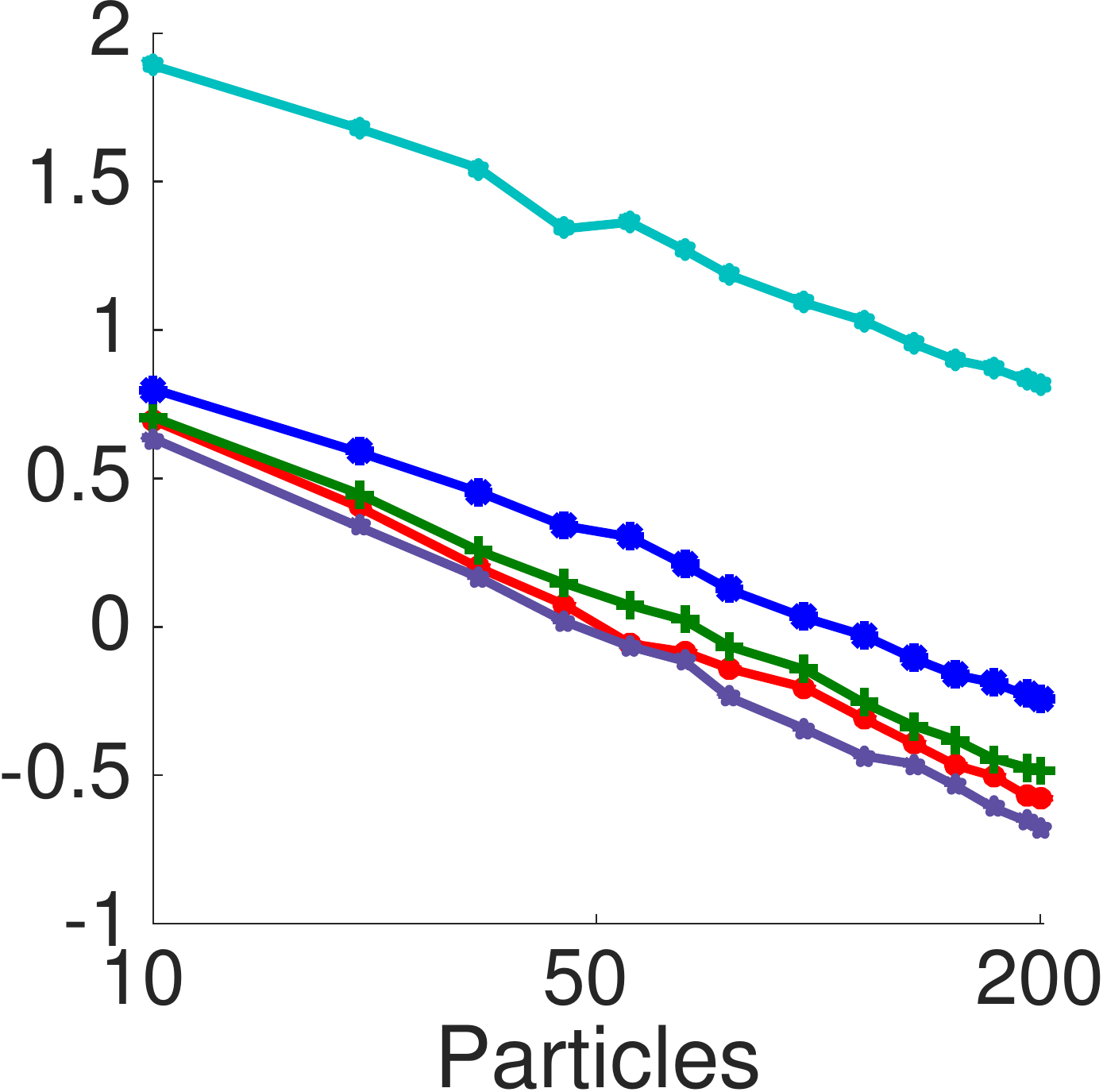} &
\includegraphics[height=0.19\textwidth]{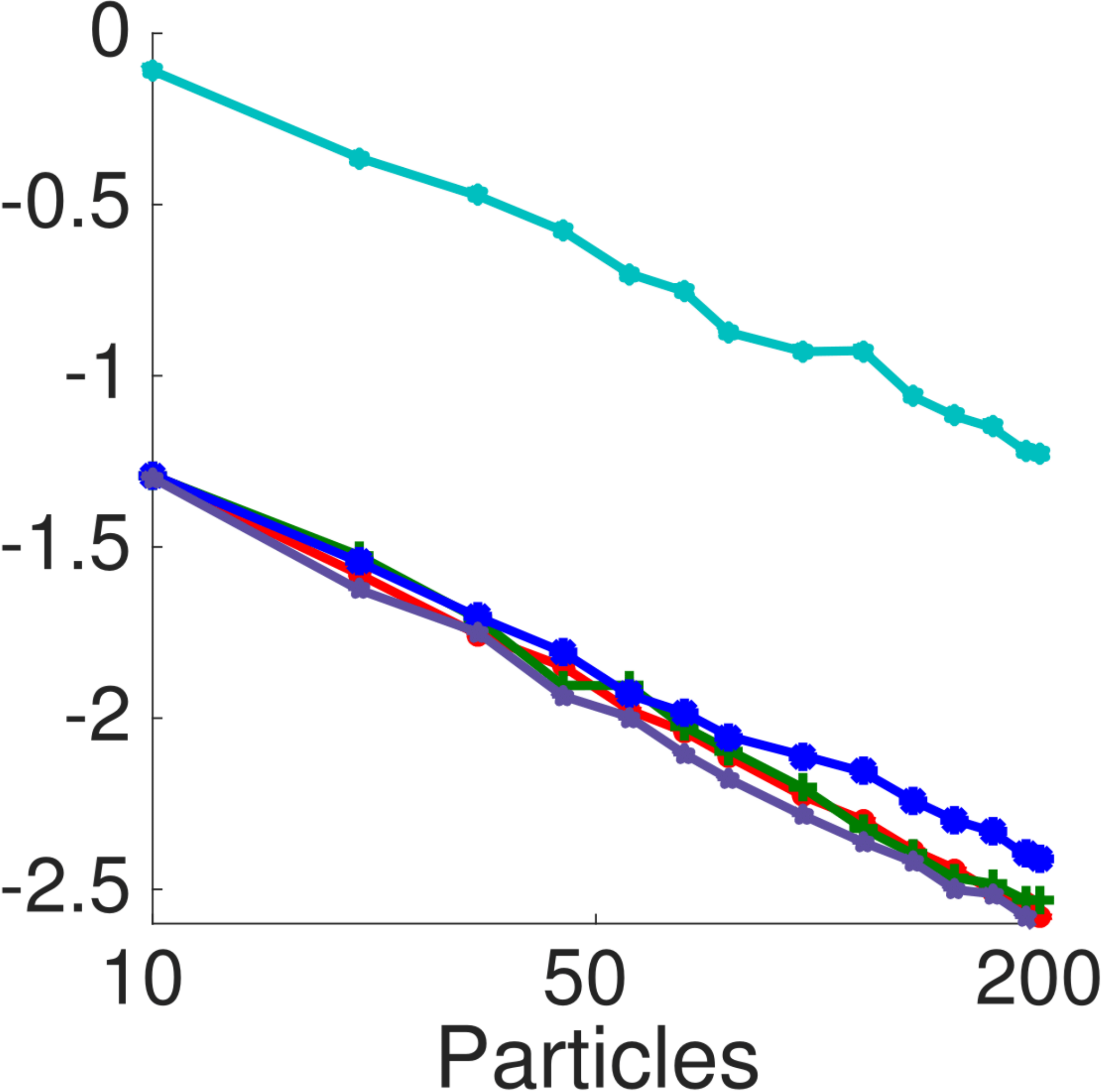} &
\includegraphics[height=0.19\textwidth]{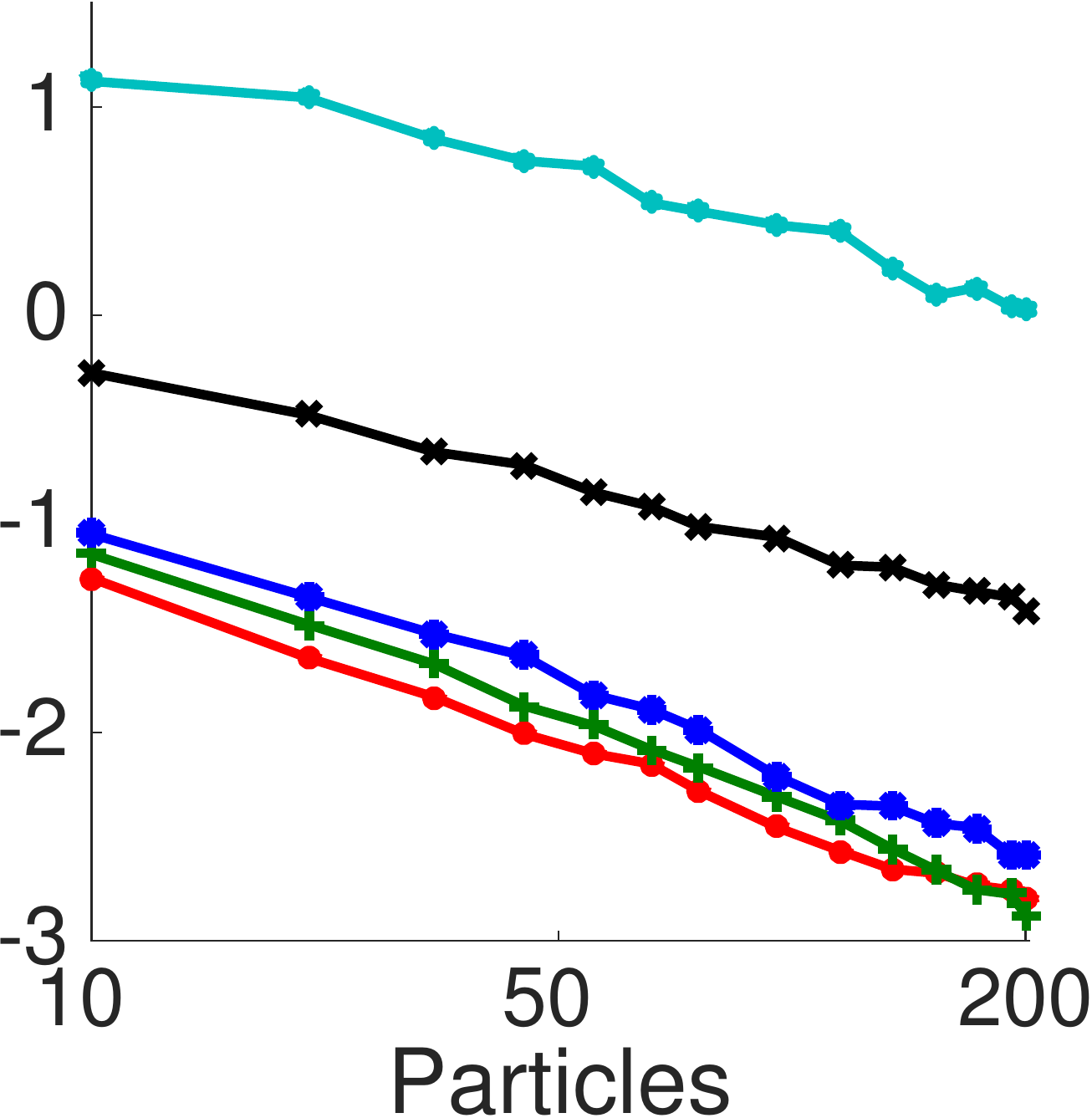}
\raisebox{2em}{ \includegraphics[height=0.1\textwidth]{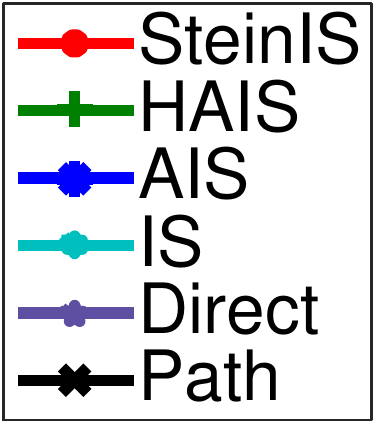}}\\
{\small (a)  $\mathbb{E}[x]$} &
{\small (b)  $\mathbb{E}[x^2]$ } &
{\small (c) $\mathbb{E}[\cos(w x + b)]$} &
{\small (d) Partition Function}\\
\end{tabular}
\caption[Verification of the convergence rate of SteinIS w.r.t. the sample size on 2D GMM with 10 randomly generated mixture components. ]{Verification of the convergence rate of SteinIS w.r.t. the sample size on 2D GMM with 10 randomly generated mixture components. $d=2.$ (a)-(c) shows mean square error(MSE) for estimating $\mathbb{E}_p[h(x)],$ where $h(\bd{x})=x_j,~x_j^2,~\cos(wx_j+b),$ for $j=1,2.$ 800 transitions are used in SteinIS, HAIS and AIS. L=1 in HAIS. The size of leader particles $|A|$ is fixed as 100 and let the size of follower particles $|B|$ vary in SteinIS. The initial proposal $q_0(\bd{x})$ is Gaussian. "Direct" means that samples are directly drawn from $p(\bd{x})$ and is not applicable in (d). "IS" means we directly draw samples from $q_0.$ "Path" denotes the proposed algorithm~\ref{ksdalgo} and is only applicable to estimate (d). The MSE is averaged on each coordinate over 500 independent experiments for SteinIS, HAIS, AIS and Direct, and over 2000 independent experiments for IS. SVGD has similar resluts as our SteinIS on (a), (b), (c) and is not provided in this figure for clarity. SVGD cannot be applied to task (d). The logarithm base is 10. \label{fig:varyparticle}}
\end{figure*}

 \paragraph{Verification of Convergence Property of SteinIS}
We also empirically verify the convergence  of our SteinIS 
as the follower particle size $|B|$ increases (as the leader particle size $|A|$ is fixed) in Fig.~\ref{fig:varyparticle}. We apply SteinIS to estimate $\mathbb{E}_p[h(x)],$ where $h(\bd{x})=x_j,~x_j^2 ~\textit{or}~\cos(wx_j+b)$ with $w \sim \normal(0,1)$ and $b \sim \mathrm{Uniform}([0,1])$ for $j=1,2$, and the partition function (which is trivially $1$ in this case). In Fig.~\ref{fig:varyparticle}(a)-(c) shows mean square error(MSE) for estimating $\mathbb{E}_p[h(x)],$ where $h(\bd{x})=x_j,~x_j^2,~\cos(wx_j+b)$ with $w\sim \normal(0,1)$ and $b\in \mathrm{Uniform}([0,1])$ for $j=1,2$, and the normalization constant (which is $1$ in this case). From Fig.~\ref{fig:varyparticle}, we can see that the mean square error(MSE) of our algorithms follow the typical convergence rate of IS, which is $O(1/\sqrt{|B|}),$ where $|B|$ is the number of samples for performing IS. Figure~\ref{fig:varyparticle} indicates that SteinIS can achieve almost the same performance as the exact Monte Carlo~(which directly draws samples from the target $p(\vx)$), 
indicating the proposal $q_\ell(\vx)$ closely matches the target $p(\vx)$. 

We used 800 transitions in SteinIS, HAIS and AIS, and take $L=1$ in HAIS. We fixed the size of the leader particles $|A|$ to be $100$ and vary the size of follower particles $|B|$ in SteinIS. The initial proposal $q_0$ is the standard Gaussian. "Direct" means that samples are directly drawn from $p$ and is not applicable in (d). "IS" means we directly draw samples from $q_0$ and apply standard importance sampling. "Path" denotes  path integration method in Algorithm~\ref{ksdalgo} and is only applicable to estimate the partition function in (d). The MSE is averaged on each coordinate over 500 independent experiments for SteinIS, HAIS, AIS and Direct, and over 2000 independent experiments for IS. SVGD has similar results (not shown for clarity) as our SteinIS on (a), (b), (c), but can not be applied to estimate the partition function in task (d). 
The logarithm base is 10.

\begin{figure}[t]
\centering
\scalebox{.99}{
\setlength{\tabcolsep}{0em}
\begin{tabular}{cccc}
\hspace{-1.cm} \includegraphics[height=0.255\textwidth]{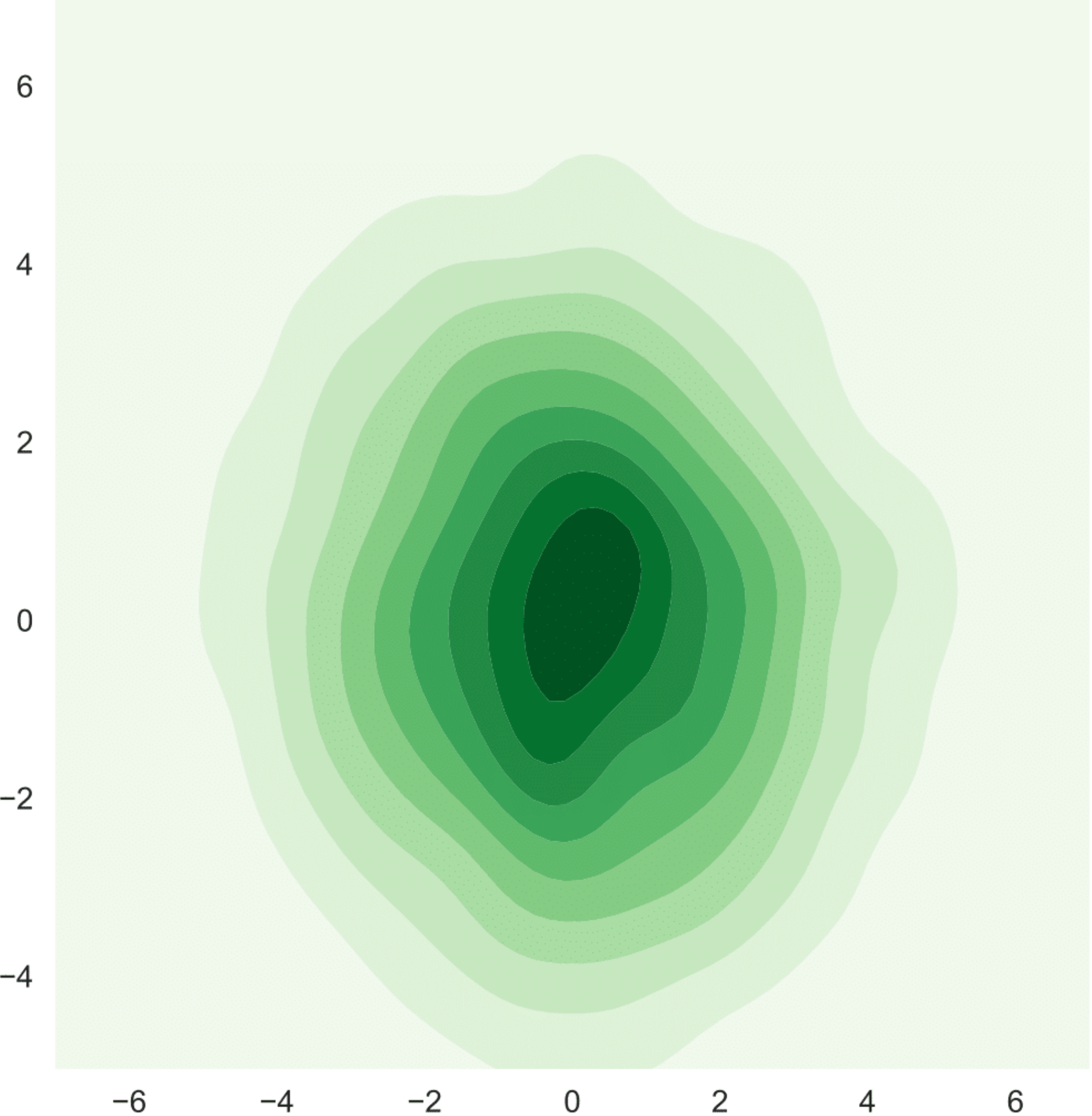}&
\hspace{-1.cm}
\includegraphics[height=0.255\textwidth]{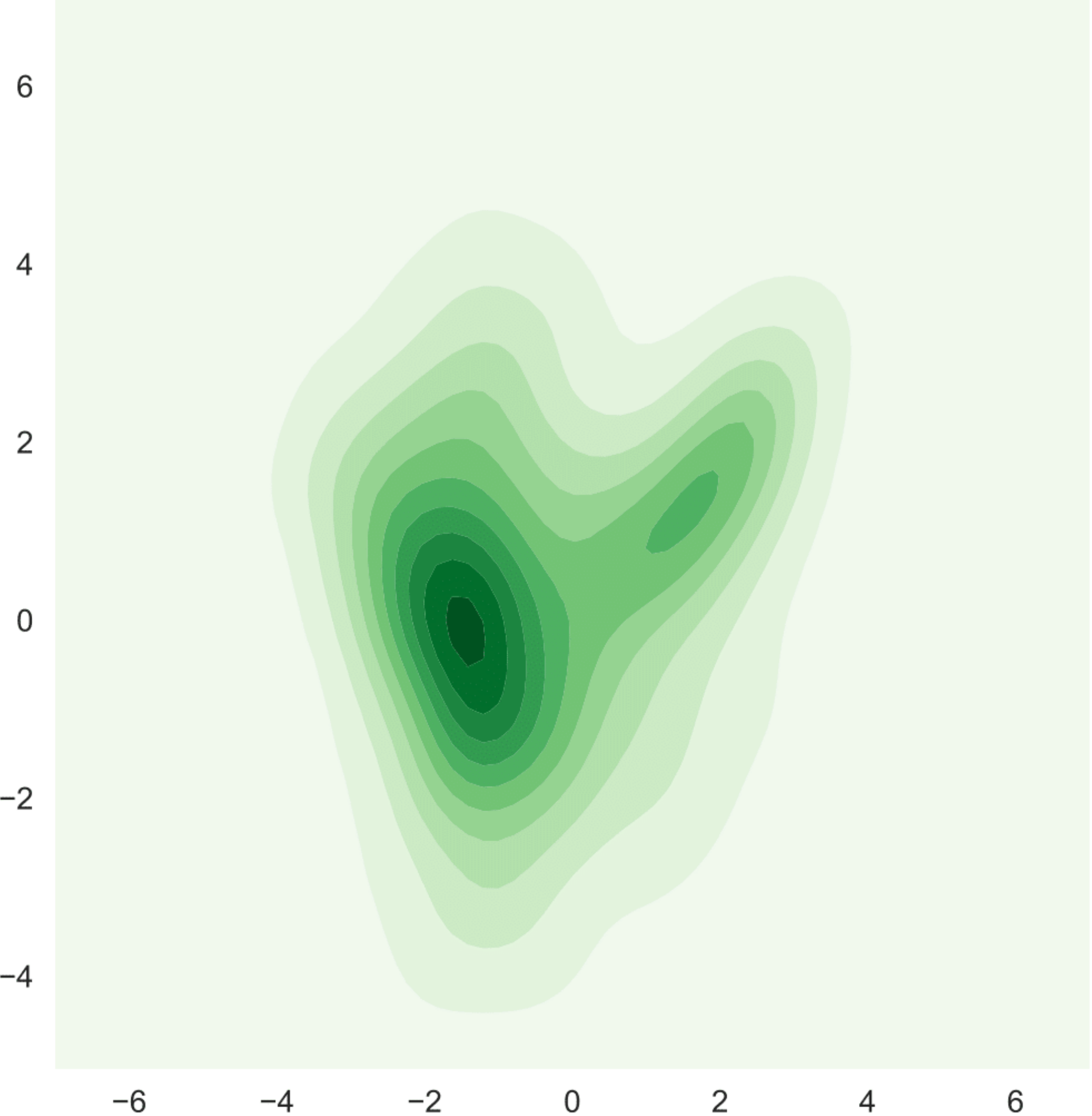} &
\hspace{-1.cm}
\includegraphics[height=0.255\textwidth]{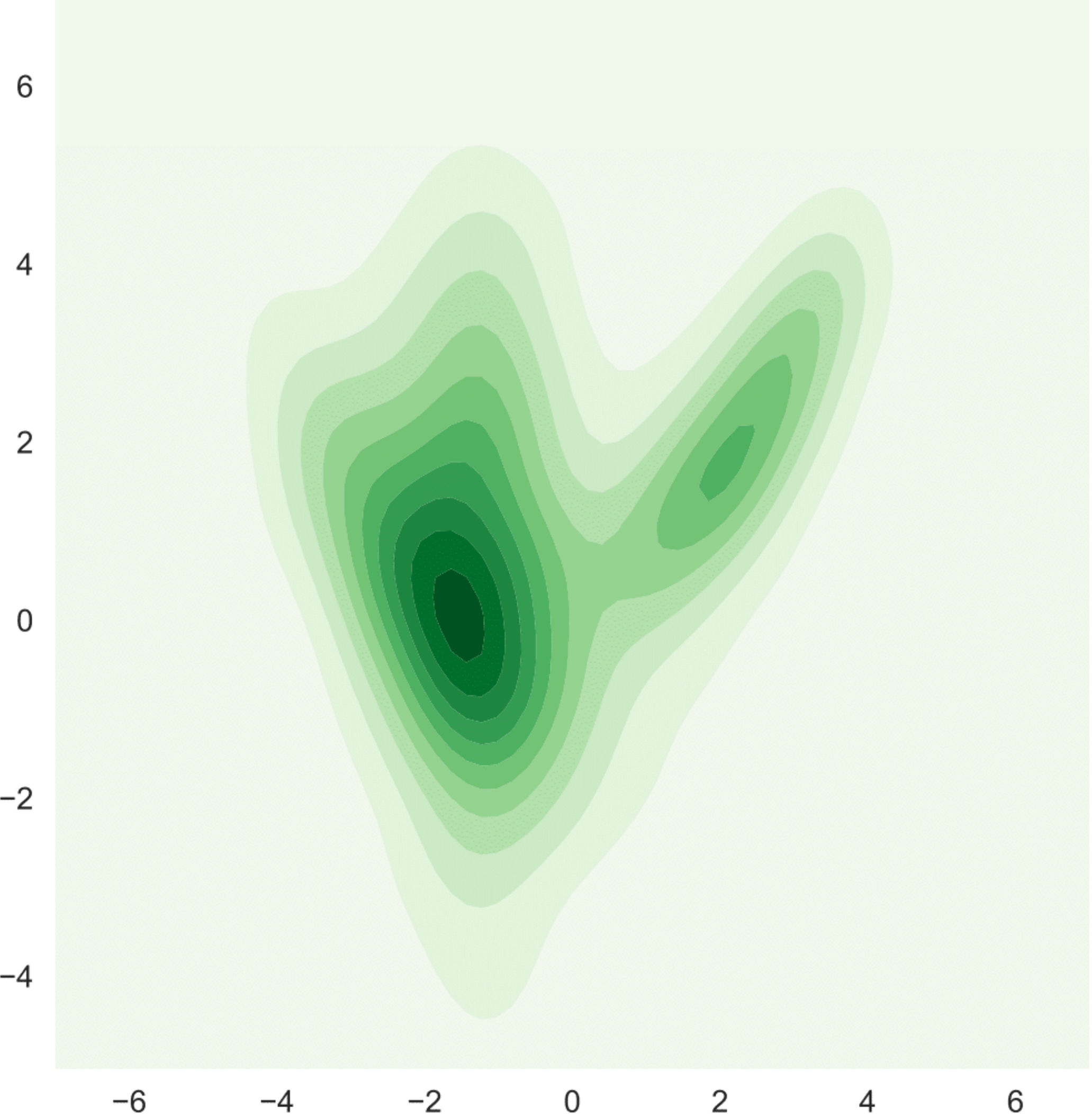} &
\hspace{-1.cm}
\includegraphics[height=0.255\textwidth]{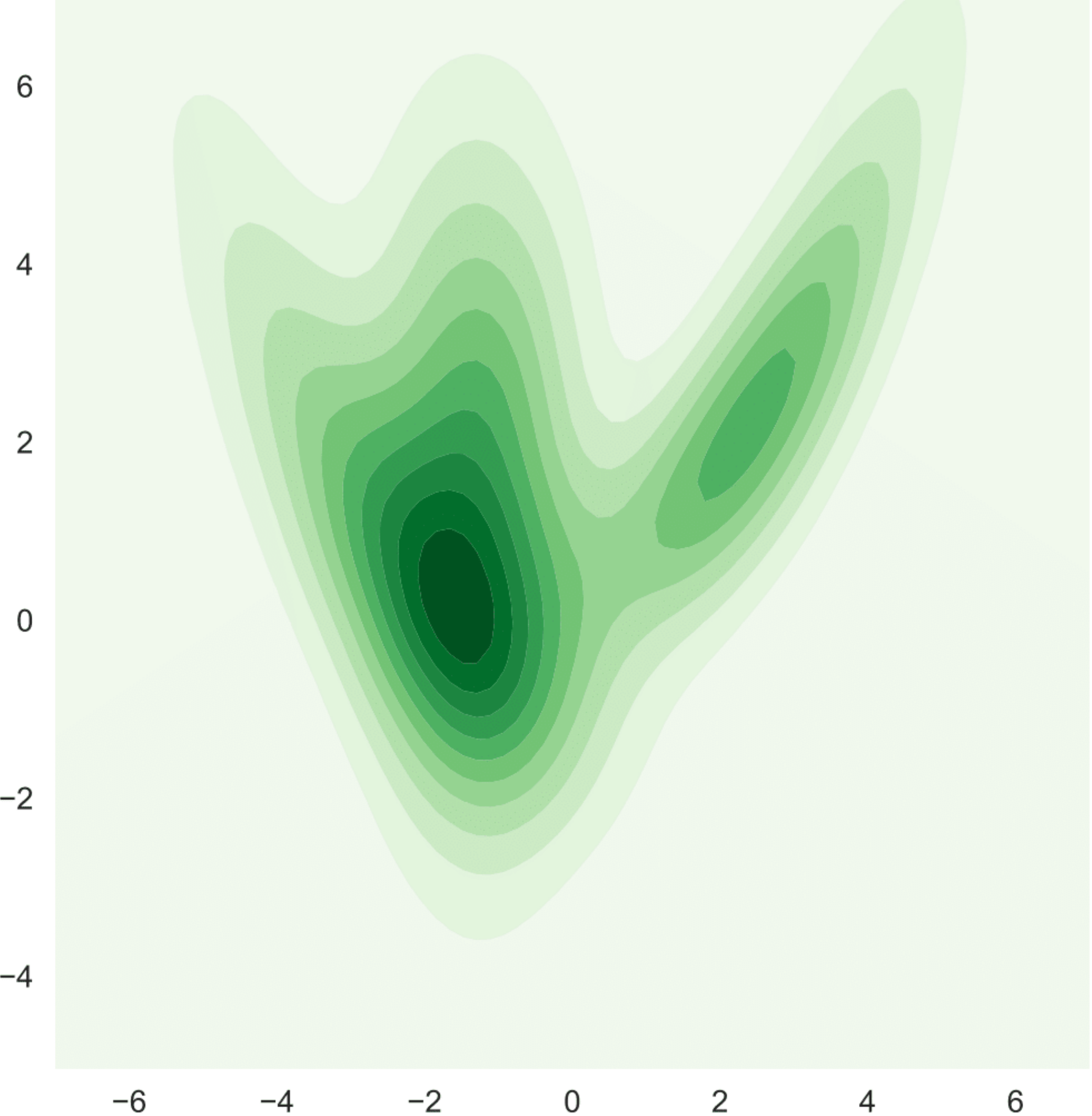} \\
\hspace{-1.cm} {\small (a) SteinIS, $\ell=0$} & {\small (b) SteinIS,  $\ell=50$} & {\small (c)  SteinIS, $\ell=200$} & {\small (d) SteinIS,  $\ell=2000$} \\
\hspace{-1.cm} \includegraphics[height=0.245\textwidth]{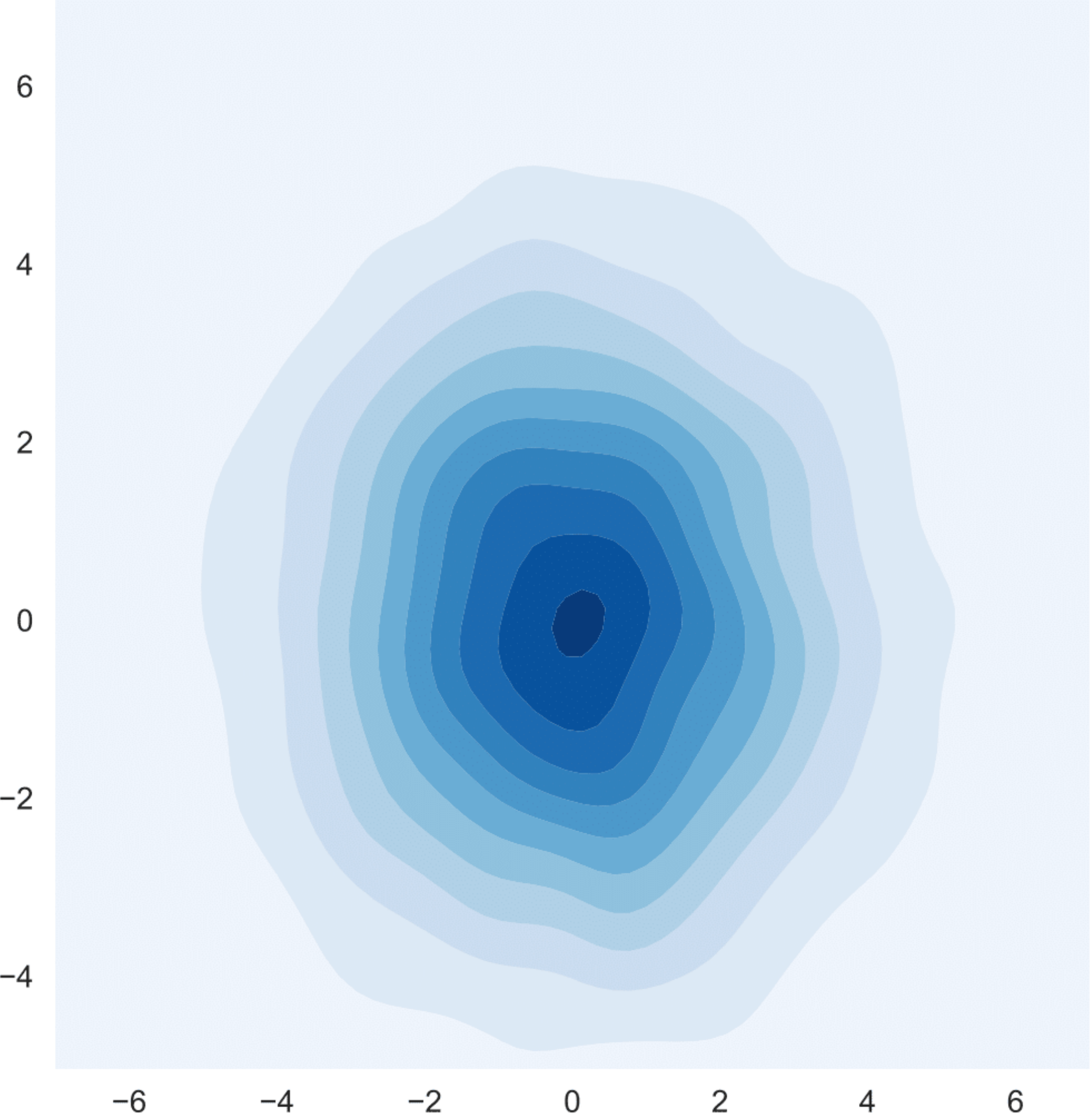} &
\hspace{-1.cm}
\includegraphics[height=0.255\textwidth]{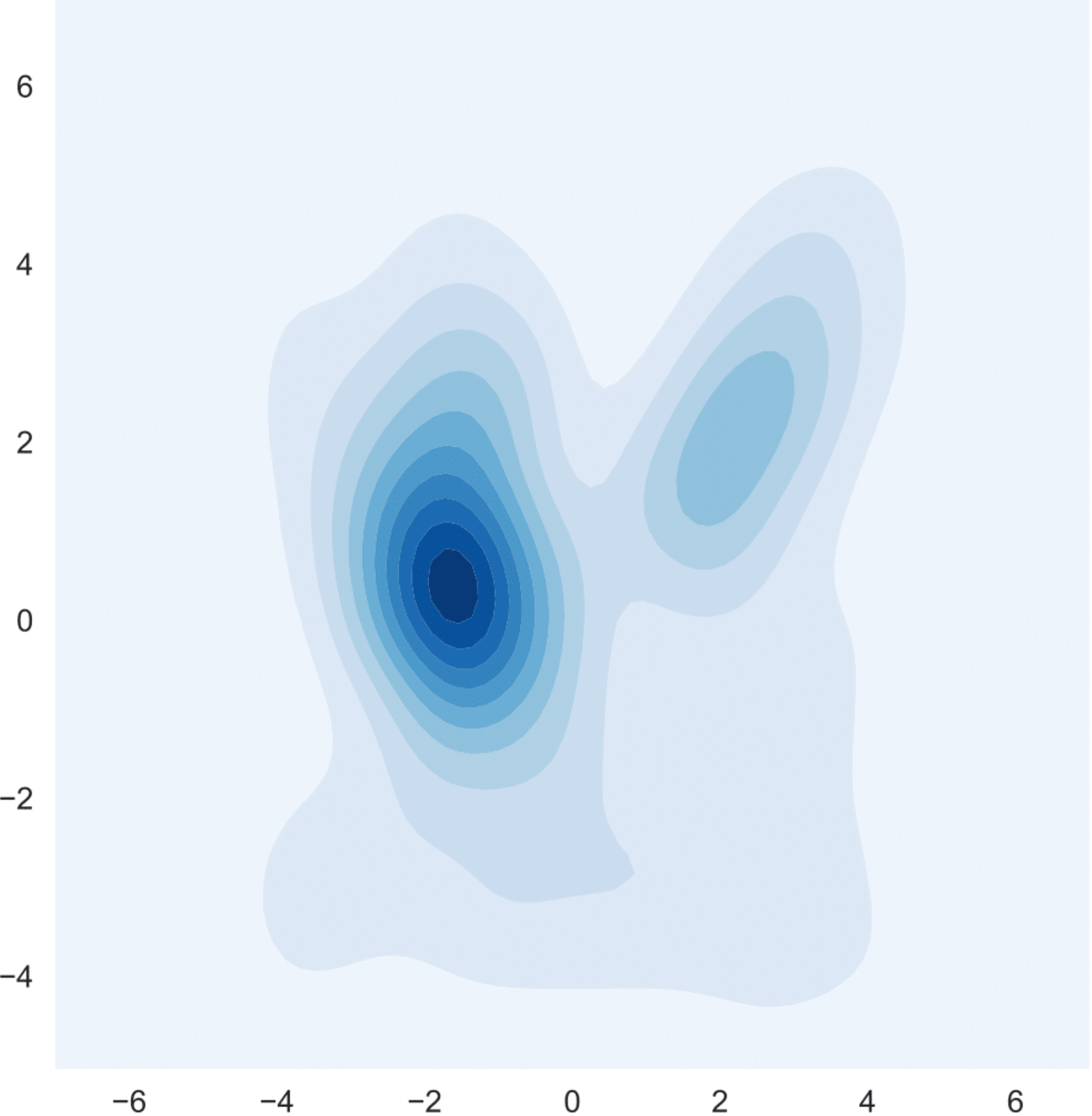} &
\hspace{-1.cm}
\includegraphics[height=0.255\textwidth]{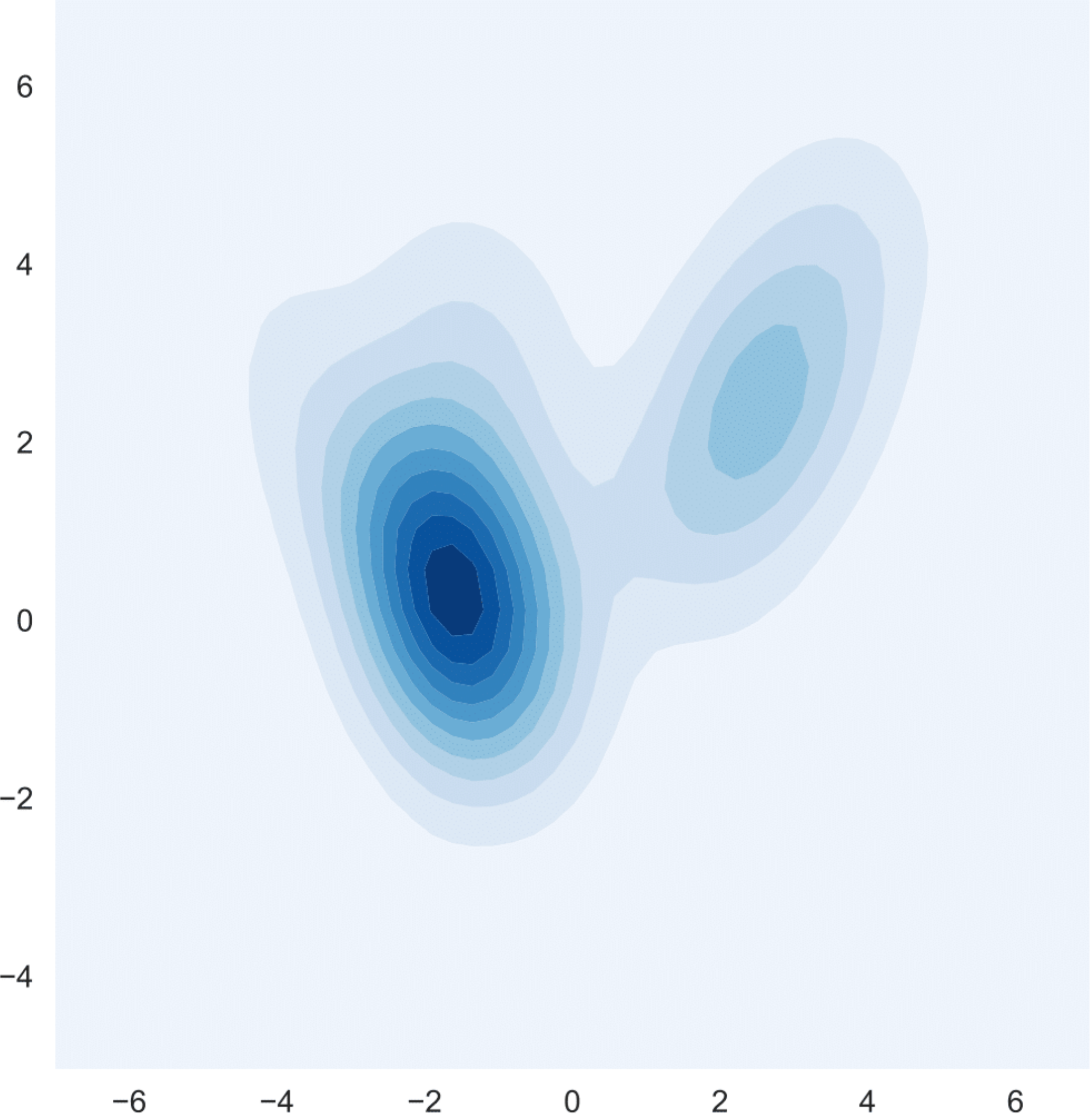} &
\hspace{-1.cm}
\includegraphics[height=0.255\textwidth]{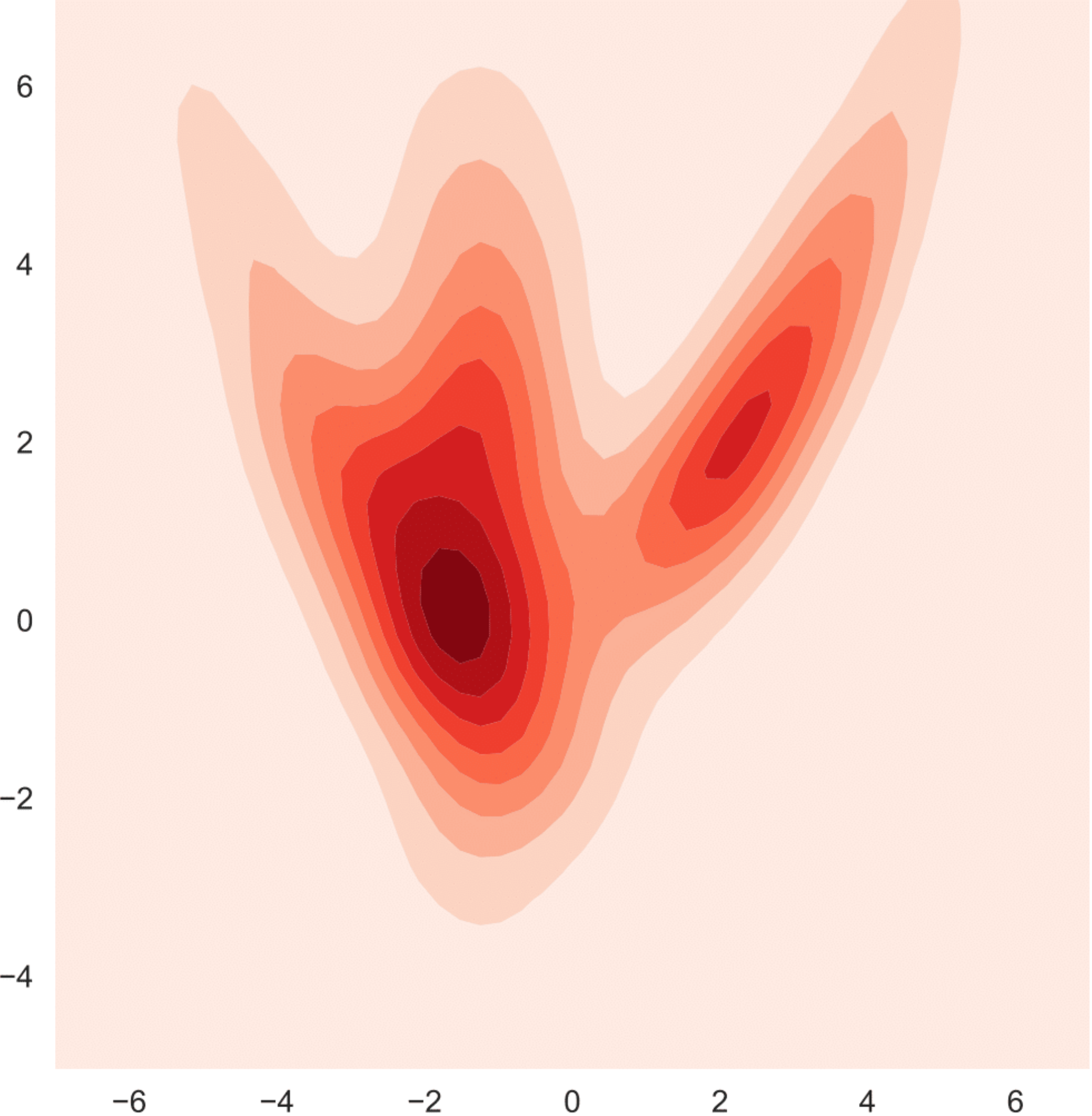} \\
\hspace{-1.cm} {\small (e)Adap IS, $\ell=0$} & {\small (f)Adap IS, $\ell=1000$} & {\small (g)Adap IS, $\ell=10000$} & {\small (h)  Exact}\\
\end{tabular}
}
\caption[Evolution of the contour of density functions for SteinIS and Adaptive IS]{Evolution of the contour of density functions for SteinIS and Adaptive IS. The top line shows the contours of the evolved density functions in SteinIS, i.e., (a, b, c, d). (e, f, g) are the evolved contours of the traditional adaptive IS. (h) is the contour of the target density $p(\bd{x})$. The number of mixture components for adaptive IS is 200 and the number of leader particles for approximating the map in SteinIS is 200. \label{fig:adap}}
\end{figure}
\subsection{Comparison between SteinIS and Adaptive IS}
In the following, we compare SteinIS with traditional adaptive IS~\citep{ryu2014adaptive} on a probability model $p(\bd{x})$, obtained by applying nonlinear transform on a three-component Gaussian mixture model. 

Specifically, let $ \widetilde{q}$ be a 2D Gaussian mixture model, and 
$\bd T$ is a nonlinear transform defined by $\bd{T}(\bd{z})=[a_1z_1+b_1, a_2 z_1^2+a_3z_2+b_2]^{\top}$, where $\bd{z}=[z_1, z_2]^{\top}.$ We define the target $p$ to be the distribution of $\vx  = \bd T (\bd z)$ when $\bd z \sim \widetilde q$. 

The contour of the target density $p$ we constructed is shown in Figure~\ref{fig:adap}(h). 
We test our SteinIS and visualize in Figure~\ref{fig:adap}(a)-(d) the density of the evolved distribution $q_\ell$ using kernel density estimation, by drawing a large number of follower particles at iteration equaling $0$, $50$, $200$, $1000$ respectively. 
We compare our method with 
the adaptive IS by \citep{ryu2014adaptive} 
using a proposal family formed by Gaussian mixture with $200$ components. 
The densities of the proposals obtained by adaptive IS at different iterations
are shown in Figure~\ref{fig:adap}(e)-(g)  at iteration equaling $0$, $1000$, $10000$ respectively. The number of the mixture components for adaptive IS is $200$ and the number of leader particles for approximating the map in SteinIS is also $200$. 

We can see that the evolved proposals of SteinIS converge to the target density $p(\bd{x})$ and approximately match $p(\bd{x})$ at 2000 iterations, but the optimal proposal of adaptive IS with 200 mixture components (at the convergence) can not fit $p(\bd{x})$ well, as indicated by Figure~\ref{fig:adap}(g). This is because the Gaussian mixture proposal family (even with upto 200 components) can not closely approximate the non-Gaussian target distribution we constructed. 
We should remark that SteinIS can be applied to refine 
the optimal proposal given by adaptive IS to get better importance proposal by implementing a set of successive transforms on the given IS proposal. 

Qualitatively, we find that the KL divergence (calculated via kernel density estimation) between our evolved  proposal $q_\ell$ and $p$ decreases to  $\leq 0.003$ after 2000 iterations, while the KL divergence between the optimal adaptive IS proposal  and the target $p$ can be only decreased to $0.42$ even after sufficient optimization. 
\begin{figure}[h]
\centering
\begin{tabular}{cc}
\includegraphics[width=0.3\textwidth]{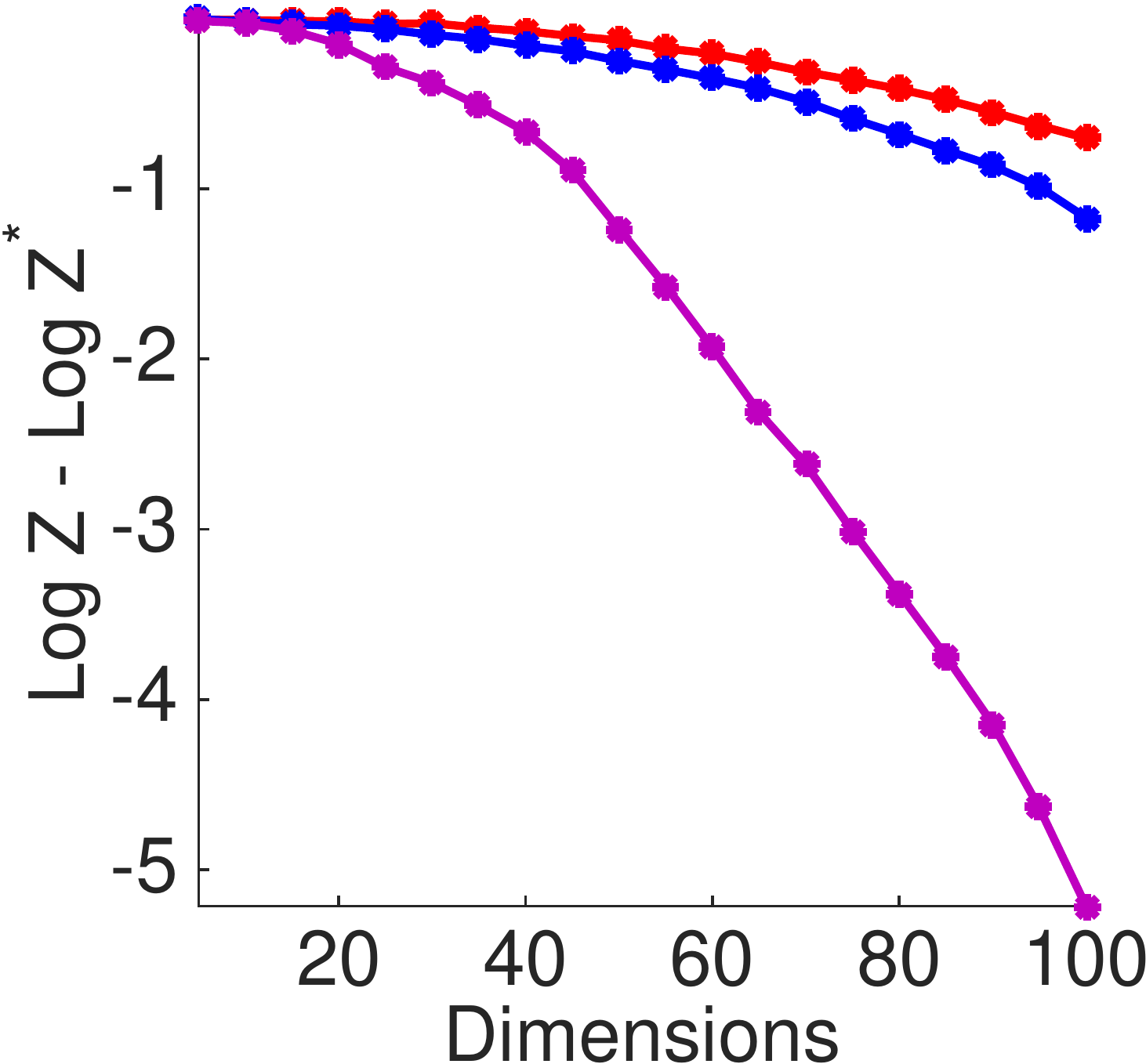} &
\includegraphics[width=0.3\textwidth]{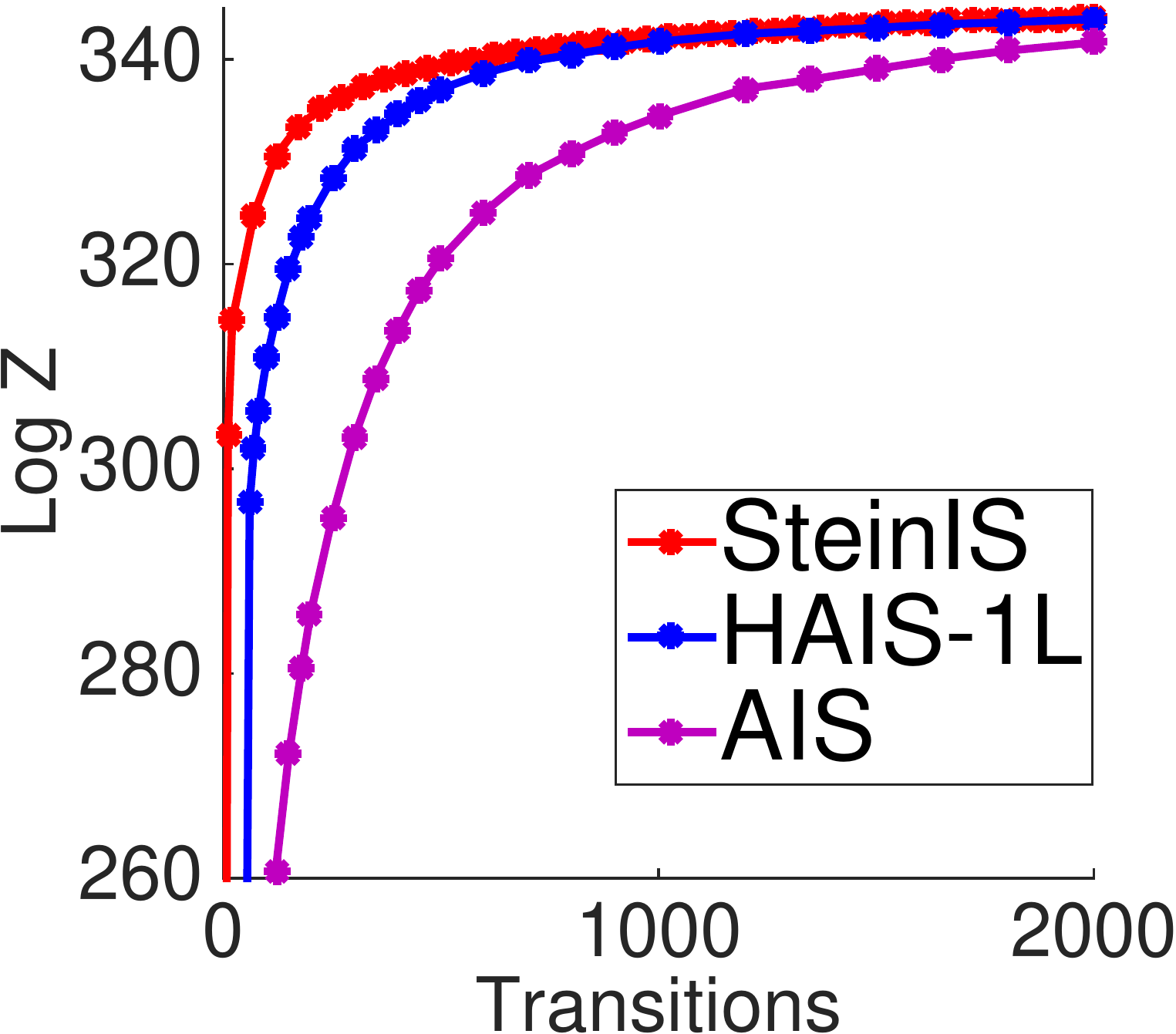} \\
{\small (a)  Vary dimensions} &
{\small (b)  100 dimensions}
\end{tabular}
\caption[Empirical experiments on Gauss-Bernoulli RBM with $d'=10$ hidden variables.]{Gauss-Bernoulli RBM with $d'=10$. The initial distribution $q_0(\bd{x})$ for SteinIS, HAIS and AIS is multivariate Gaussian. We let $|A|=100$ in SteinIS and use 100 samples for implementing IS in SteinIS, HAIS and AIS. In (a), we use 1500 transitions for HAIS, SteinIS and AIS. "HAIS-1L" means $L=1$ in each Markov transition of HAIS. SVGD is not applicable. $\log Z^*$ denotes the logarithm of the exact normalizing constant. All experiments are averaged over 500 independent trails.
\label{fig:GauBerlli}}
\end{figure}

\subsection{Gauss-Bernoulli Restricted Boltzmann Machine}
We apply our method to estimate the partition function of Gauss-Bernoulli Restricted Boltzmann Machine (RBM), which is a multi-modal, hidden variable graphical model. Effective estimation of the partition function is a fundamental task on the application of probabilistic graphical model~\citep{liu2015estimating}. It consists of a continuous observable variable $\bd{x}\in \mathbb{R}^d$ and a binary hidden variable $\bd{h}\in \{\pm 1\}^{d'},$ with a joint probability density function of form 
\begin{equation}
p(\bd{x},\bd{h})= \frac{1}{Z} \exp(\bd{x}^\mathrm{T} B\bd{h}+b^\mathrm{T}\bd{x}+c^\mathrm{T}\bd{h}-\frac{1}{2}\| \bd{x}\|_2^2),
\end{equation}
where $p(\bd{x})=\frac{1}{Z}\sum_{\bd{h}} p(\bd{x},\bd{h})$ and $Z$ is the normalization constant. 
By marginalzing the hidden variable $h$, we can show that $p(\bd{x})$ is 
\begin{equation*}
\begin{aligned}
p(\bd{x})=\frac{1}{Z}\exp(b^\mathrm{T}\bd{x}-\frac12\| \bd{x}\|_2^2)\prod_{i=1}^{d'} [\exp(\varphi_i)+ \exp(-\varphi_i)],
\end{aligned}
\end{equation*}
where $\varphi = B^\mathrm{T}\bd{x} + c$, and its score function $\bd{s}_p$ is easily derived as
$$
\bd{s}_p(\bd{x})  = \nabla_{\bd x} \log p(\bd x) = b - \bd{x} + B\frac{\exp(2\varphi)-1}{\exp(2\varphi)+1}.
$$

In our experiments, we simulate a true model $p(\bd{x})$ by drawing $b$ and $c$ from the standard Gaussian distribution and select $B$ uniformly random from $\{0.5, -0.5\}$ with probability 0.5. The dimension of the latent variable $\bd{h}$ is 10 so that the probability model $p(\bd{x})$ is the mixture of  $2^{10}$ multivariate Gaussian distribution. The exact normalization constant $Z$ can be feasibly calculated using the brute-force algorithm in this case. The initial distribution $q_0(\bd{x})$ for all the methods  is a same multivariate Gaussian. We let $|A|=100$ in SteinIS and use $(B=)$100 importance samples in SteinIS, HAIS and AIS. In (a), we use 1500 transitions for HAIS, SteinIS and AIS. "HAIS-1L" means we use $L=1$ leapfrog in each Markov transition of HAIS. $\log Z^*$ denotes the logarithm of the exact normalizing constant. All experiments are averaged over 500 independent trails. 
Figure~\ref{fig:GauBerlli}(a) and Figure~\ref{fig:GauBerlli}(b) shows the performance of SteinIS on Gauss-Bernoulli RBM 
when we vary the dimensions of the observed variables 
and the number of transitions in SteinIS, respectively.  
 We can see that SteinIS converges slightly faster than HAIS which uses one leapfrog step in each of its Markov transition. Even with the same number of Markov transitions, AIS with Langevin dynamics converges much slower than both SteinIS and HAIS. 
The better performance of HAIS comparing to AIS was also observed by ~\citet{sohl2012hamiltonian} when they first proposed Hamiltonian annealed importance sampling. 

\begin{figure}[t]
\centering
\begin{tabular}{cc}
\includegraphics[width=0.3\textwidth]{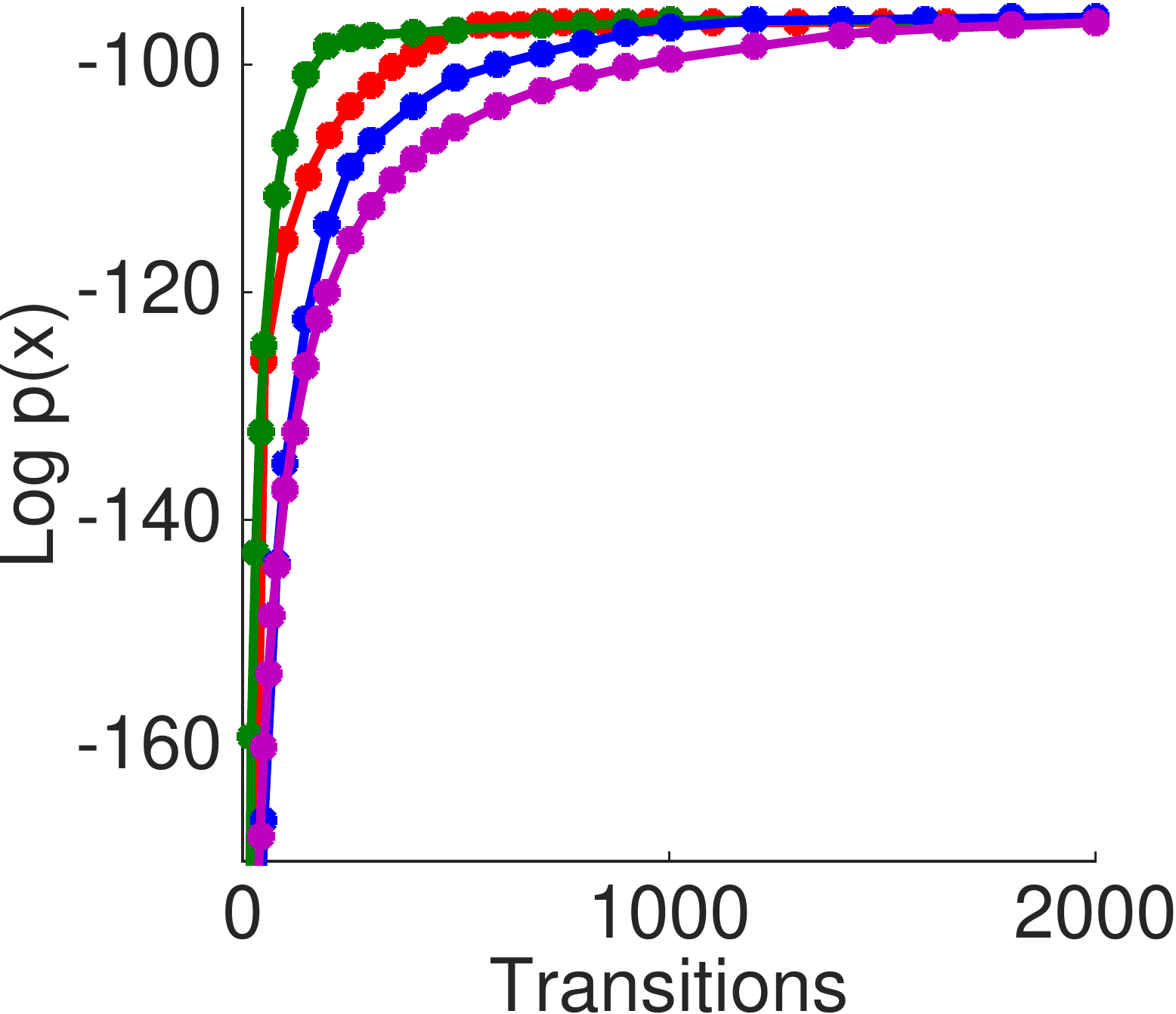}&
\includegraphics[width=0.3\textwidth]{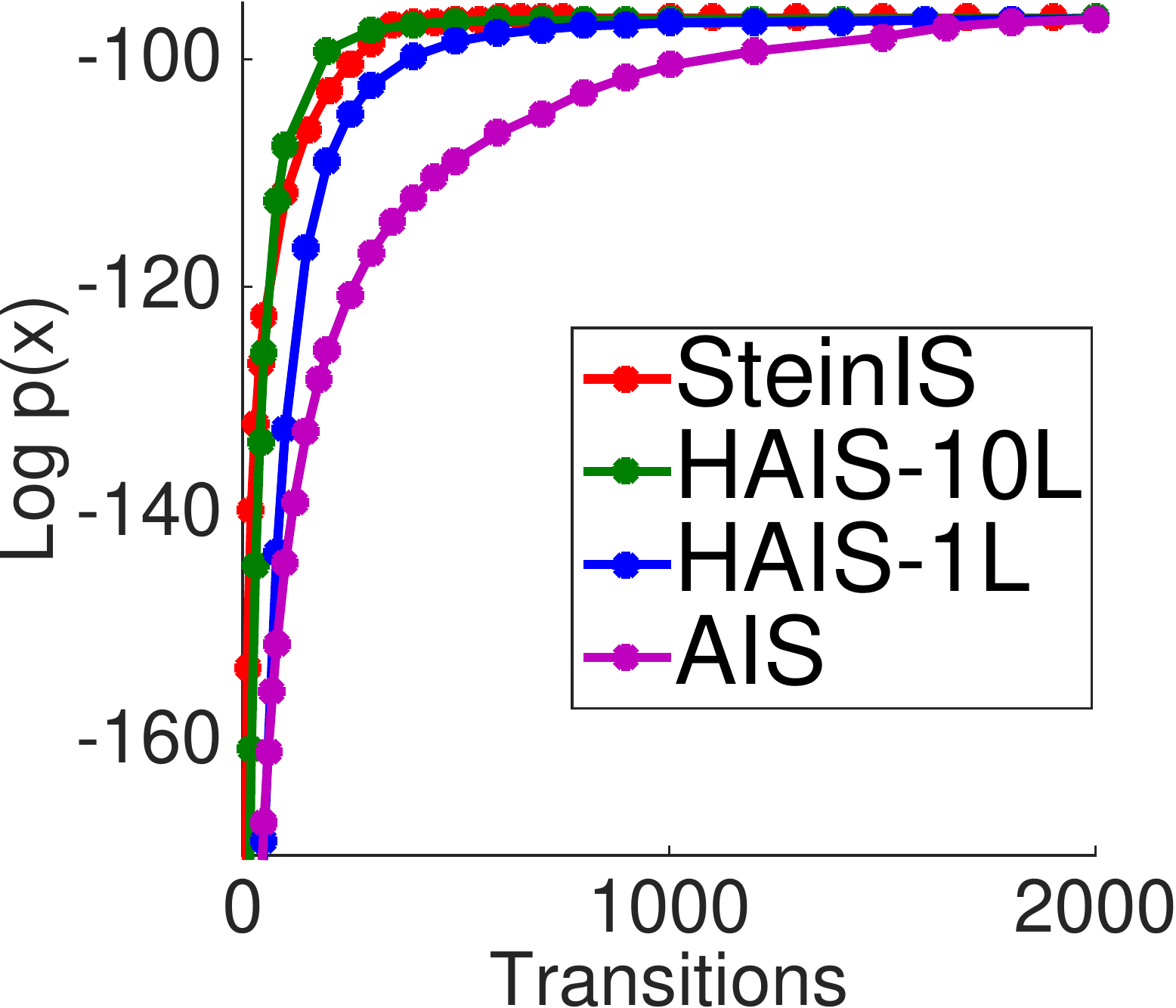} \\
{\small (a)  20 hidden variables}&
{\small (b)  50 hidden variables}
\end{tabular}
\caption[Evaluation of the testing log-likelihood $\log p(x)$ for the deep generative model on MNIST dataset.]{Evaluation of the testing log-likelihood $\log p(x)$ for the decoder-based generative models generative model on MNIST dataset. The initial distribution $q_0(\bd{z})$ for SteinIS, HAIS and AIS is multivariate Gaussian. We let $|A|=60$ in SteinIS and use 60 samples for each image to implement IS in SteinIS, HAIS and AIS. "HAIS-10L" and "HAIS-1L" denote $L=10$ and $L=1$ in each Markov transition of HAIS respectively. The log-likelihood is calculated over 1000 images randomly chosen from MNIST. The dimensions in (a) and (b) are the dimensions of latent space $\bd{z}$ in our setting. \label{figdecode}}
\end{figure}

\subsection{Deep Generative Models}
Finally, we implement our SteinIS to evaluate the $\log$-likelihoods of the decoder models in variational autoencoder (VAE) \citep{kingma2013auto}. VAE is a directed probabilistic graphical model. The decoder-based generative model is defined by a joint distribution over a set of latent random variables $\bd{z}$ and the observed variables $\bd{x}: p(\bd{x}, \bd{z})=p(\bd{x}\mid \bd{z}) p(\bd{z}).$ We use the same network structure as that in \citet{kingma2013auto}. The prior $p(\bd{z})$ is chosen to be a multivariate Gaussian distribution. The log-likelihood is defined as $p(\bd{x})=\int p(\bd{x}\mid\bd{z})p(\bd{z})d\bd{z},$ where $p(\bd{x}\mid \bd{z})$ is the Bernoulli MLP as the decoder model given in \citet{kingma2013auto}.
 In our experiment, we use a two-layer network for $p(\bd{x}\mid\bd{z}),$ whose parameters are estimated using a standard VAE based on the MNIST training set.  
 For a given observed test image $\vx$, we use our method to sample the posterior distribution $p(\bd z | \vx)  = \frac{1}{p(\vx)}  p(\bd{x} | \bd{z}) p(\bd{z})$, and estimate the partition function $p(x)$, which is the testing likelihood of image $\vx$. 

Figure~\ref{figdecode} also indicates that our SteinIS converges slightly faster than HAIS-1L which uses one leapfrog step in each of its Markov transitions, denoted by HAIS-1L. The initial distribution $q_0$ used in SteinIS, HAIS and AIS is a same multivariate Gaussian. 
We let $|A|=60$ in SteinIS and use 60 samples for each image to implement IS in HAIS and AIS. "HAIS-10L" and "HAIS-1L" denote using $L=10$ and $L=1$ in each Markov transition of HAIS, respectively. The log-likelihood $\log p(x)$ is averaged over 1000 images randomly chosen from MNIST. Figure (a) and (b) show the results when using $20$ and $50$ hidden variables, respectively. Note that the dimension of the observable variable $\vx$ is fixed, and is the size of the MNIS images. Meanwhile, the running time of SteinIS and HAIS-1L is also comparable as provided by Table~\ref{timetable}. Although HAIS-10L, which use 10 leapfrog steps in each of its Markov transition, converges faster than our SteinIS, it takes much more time than our SteinIS in our implementation since the leapfrog steps in the Markov transitions of HAIS are sequential and can not be parallelized. See Table~\ref{tab:tab1}.  
Compared with HAIS and AIS, our SteinIS has another advantage: if we want to increase the transitions from 1000 to 2000 for better accuracy, SteinIS can build on the result from 1000 transitions and just need to run another 1000 iterations, while HAIS cannot take advantage of the result from 1000 transitions and have to independently run another 2000 transitions. 

\begin{table}[tbh]
\label{tab:tab1}
\centering
\caption[Running time comparison of different methods on quantitative evaluation of deep generative model on MNIST dataset]{Running time (in seconds) on  MNIST dataset using 1000 transitions in all methods to test the running time.}
\begin{tabular}{|c|c|c|c|}
  \hline
  Dimensions of $\bd{z}$ & 10  & 20  & 50  \\ \hline
  SteinIS & 224.40 & 226.17 & 261.76 \\ \hline
  HAIS-10L & 600.15 & 691.86 & 755.44 \\ \hline
  HAIS-1L & 157.76 & 223.30 & 256.23 \\ \hline
  AIS & 146.75 & 206.89 & 230.14 \\
  \hline
\end{tabular}
\label{timetable}
\end{table}

\section{Stochastic Annealed Importance Sampling}
In the main part of this chapter, we propose an adaptive importance sampling algorithm, which has interesting connections with annealed importance sampling (AIS). Extensive experimental results show that AIS has competitive performance as our SteinIS and is a robust importance sampling algorithm. In practice, AIS has been widely used in various scenario such as Bayesian model selection and quantitative analysis of deep generative models. However, in Bayesian inference, it is intractable to AIS to estimate the normalization constant of the posterior when the dataset is very large as AIS often needs to run a long chain. In this section, we propose a new algorithm to estimate the normalization constant in such setting.

Suppose we are interested in estimating the normalization constant of the posterior distribution in Bayesian inference. Let the data set be $X=\{\vx_i\}_{i=1}^N,$ where $N$ is the number of data and is assumed to be huge. The posterior distribution is proportional to,
\begin{equation}
p(\vthe\mid X)\propto \prod_{i=1}^N p(\vx_i\mid \vthe) p_0(\vthe)   
\end{equation}
where $p_0(\vthe)$ is the prior distribution. Suppose we are interested in estimating the expectation of some interested function $f(\vthe)$, i.e., $\int f(\vthe) p(\vthe| X)d\vthe.$ AIS can provide an unbiased estimation but is impractical when the dataset size $N$ is large since AIS typically requires a long iteration of Markov chains and it is cumbersome to calculate the posterior $p(\vthe\mid X)$ at each iteration. To alleviate the computation cost at each iteration, we randomly sample a subset $\{\vx_i\}_{i=1}^M$ from the whole dataset $X$ and provides an unbiased estimation of the posterior $p(\vthe\mid X)$ as follows,
\begin{equation}
p(\vthe\mid X)\approx \hat{p}(\vthe\mid X),~\mathrm{where}~\hat{p}(\vthe\mid X) = \exp(\frac{N}{M}\sum_{i=1}^M\log p(\vx_i|\vthe)).     
\end{equation}
Let $q(\vthe)$ be any initial distribution where AIS starts from. As AIS will bridge a distribution path between $p(\vthe\mid X)$ and $q(\vthe).$ In stochastic version of AIS, we can define the intermediate distribution as $\hat{p}_j(\vthe)\propto \hat{p}(\theta\mid X)^{\alpha_j} q(\vthe)^{1-\alpha_j},$
where $0=\alpha_T<\alpha_{T-1}<\cdots<\alpha_1<\alpha_{0}=1$, $\{\alpha_j\}$ is a set of temperatures. $\hat{p}_j$ is a stochastic approximation of $p_j(\vthe)\propto p(\theta\mid X)^{\alpha_j} q(\vthe)^{1-\alpha_j}.$ The intermediate distribution $\{p_j\}$ can be chosen in arbitrary way as long as $p_j$ and $p_{j-1}$ are close to each other to calculate IS weight.  

Let $T_{j}(\vz|\vz_{j})$ is implemented by Metropolis-Hastings Algorithm~\citep{ metropolis1953equation, hastings1970monte} with the stationary distribution $\hat{p}_j$. The whole procedure of Stochastic AIS can be illustrated as follows. Initialize the importance weight $w^i=1$, for $i=1,\cdots, K$.
\begin{enumerate}
\item sample $\vthe_{T-1}^i$ from $q(\vthe)$, $\hat{p}_T=q(\vthe)$, draw batched data $\{\vx_i\}_{i=1}^M$ to estimate $\hat{p}_{T-1}$,  calculate $w^i = w^i\frac{\hat{p}_{T-1}(\vthe_{T-1}^i)}{\hat{p}_{T}(\vthe_{T-1}^i)}$ 
\item sample $\vthe_{T-2}^i$ using $T_{T-1}(\vthe|\vthe_{T-1}^i),$ draw batched data $\{\vx_i\}_{i=1}^M$ to estimate $\hat{p}_{T-2}$,  calculate $w^i = w^i\frac{\hat{p}_{T-2}(\theta_{T-2}^i)}{\hat{p}_{T-1}(\vthe_{T-2}^i)}$ 
\item Continuous the same procedure up to $\vthe_{2}^i$
\item sample $\vthe_{1}^i$ using $T_{2}(\vthe|\vthe_{2}^i), $ draw batched data $\{\vx_i\}_{i=1}^M$ to estimate $\hat{p}_{0}$,  calculate $w^i = w^i\frac{\hat{p}_{1}(\vthe_{1}^i)}{\hat{p}_{2}(\vthe_{1}^i)}$ 
\item sample $\vthe_{0}^i$ using $T_{1}(\vthe|\vthe_{1}^i).$ calculate $w^i = w^i\frac{\hat{p}_{0}(\vthe_{0}^i)}{\hat{p}_{1}(\vthe_{0}^i)}$. 
\end{enumerate}
Hence we can apply normalized $\{w^i\}$ to do Bayesian Inference
\begin{equation}
\label{ais:est}
\int f(\vthe) p(\vthe| X)d\vthe \approx \sum_{i=1}^K w^i f(\vthe_0^i)  /(\sum_{i=1}^K w^i). 
\end{equation} 
In the following, we will demonstrate that the stochastic AIS provides an unbiased estimation of $\int f(\vthe) p(\vthe| X)d\vthe.$
From the procedure of stochastic AIS, the target distribution is 
\begin{align}
p(\vthe_0, \vthe_1, \cdots, \vthe_{T-1})& \propto \hat{p}_0(\vthe_0)\frac{\hat{p}_1(\vthe_0)}{\hat{p}_1(\vthe_0)}\widetilde{T}_1(\vthe_0, \vthe_1)\cdots\frac{\hat{p}_{T-1}(\vthe_{T-2})}{\hat{p}_{T-1}(\vthe_{T-2})}\widetilde{T}_{T-1}(\vthe_{T-2}, \vthe_{T-1})\\
& = \frac{\hat{p}_0(\vthe_0)}{\hat{p}_1(\vthe_0)}T_1(\vthe_1, \vthe_0)\cdots\frac{\hat{p}_{T-2}(\vthe_{T-2})}{\hat{p}_{T-1}(\vthe_{T-2})}T_{T-1}(\vthe_{T-1}, \vthe_{T-2})\hat{p}_{T-1}(\vthe_{T-1}). \notag
\end{align}
The importance proposal $q$ is 
\begin{align}
q(\vthe_0, \vthe_1, \cdots, \vthe_{T-1})=\hat{p}_T(\vthe_{T-1})T_{T-1}(\vthe_{T-1}, \vthe_{T-2})\cdots T_1(\vthe_1, \vthe_0).
\end{align}
Therefore, the importance weight $w^i$ is 
\begin{equation}
\label{def:sais}
w^i=w(\vthe_0^i, \vthe_1^i, \cdots, \vthe_{T-1}^i)= \frac{\hat{p}_0(\vthe_0^i)}{\hat{p}_1(\vthe_0^i)} \frac{\hat{p}_1(\vthe_1^i)}{p_2(\vthe_1^i)}\cdots  \frac{\hat{p}_{T-1}(\vthe_{T-1}^i)}{\hat{p}_T(\vthe_{T-1}^i)},    
\end{equation}
where $\hat{p}_T = q$ is the prior where $\{\vthe_{T-1}^i\}$ are drawn from. \eqref{def:sais} is a stochastic approximation of the following importance weight in AIS,
\begin{equation}
w(\vthe_0, \vthe_1, \cdots, \vthe_{T-1}^i)= \frac{p_0(\vthe_0^i)}{p_1(\vthe_0^i)} \frac{p_1(\vthe_1^i)}{p_2(\vthe_1^i)}\cdots\frac{p_{T-1}(\vthe_{T-1}^i)}{p_T(\vthe_{T-1}^i)}.  
\end{equation}
As $T$ is large, \eqref{ais:est} can give a good approximation of the expectation $\int f(\vthe) p(\vthe| X)d\vthe.$ To better understand the algorithm, it will be interesting to see the concentration bound of the approximation.

\section{Summary} In this chapter, we propose a nonparametric adaptive importance sampling algorithm which leverages the nonparametric transforms of SVGD to iteratively improve the importance proposal $q_{\ell}(\vx).$  The $\KL$ divergence between the distribution $q_{\ell}(\vx)$ of the updated particles $\{\vx_i\}_{i=1}^\ell$ and the target distribution $p(\vx)$ is maximally decreased.
Our algorithm turns SVGD into a typical adaptive IS for more general inference tasks. Compared with traditional adaptive IS, our importance proposal is not restricted to the predefined specific distributional family, which might give poor approximation of the importance proposal to the target distribution. This is in contrast with our SteinIS. Our SteinIS can adaptively increase the approximation quality by increasing the number of particles. The $\KL$ divergence between our importance proposal and the target distribution can be decreased to be arbitrarily small. Empirical experiments on a target distribution which is not from any predefined specific distribution family demonstrates the better approximation of our importance proposal compared with the optimal importance proposal of traditional adaptive IS. Conditional on the particles in the leader particle sett, the particles in the follower particle set are independent. When the iteration is stopped at finite steps, our SteinIS can provide an unbiased estimation of the integreation $\E_p[f(\vx)]$; however, the original SVGD doesn't have such a unbiased estimation guarantee. Our SteinIS offers to estimate the partition function of the probability model where the original SVGD cannot be applied. Numerical experiments demonstrate that our SteinIS works efficiently
on the applications such as estimating the partition functions of graphical models such as Bernoulli restricted Boltzmann machine and evaluating the log-likelihood
of deep generative models. At the end of the chapter, we discuss one new importance sampling algorithm motivated from annealed importance sampling to ensure its applicability to the Bayesian setting when the target is the Bayesian posterior of a large dataset. Future research includes improving the computational and statistical efficiency in high dimensional cases and incorporating Hamiltonian
Monte Carlo into our SteinIS to derive more efficient algorithms.

Our SteinIS is inherited from SVGD and leverages the gradient information of the target distribution to construct the variable transform to steepest descent the $\KL$ divergence between the importance proposal and the target distribution. However, the gradient information of the target distribution $p(\vx)$ is not always available. In the next chapter, we will introduce one efficient approximate inference algorithm under the setting where the gradient information of the target distribution $p(\vx)$ is unavailable. We will develop a novel gradient-free sampling algorithm which only requires the availability of the evaluation of the target distribution $p(\vx).$
\chapter{Gradient-Free Sampling on Continuous Distributions\label{chap:gf}}
Sampling from high-dimensional complex probability distributions is a long-standing fundamental computational task in machine learning and statistics. We have introduced a sample-efficient adaptive importance sampling algorithm (SteinIS) in previous chapter, which is based on Stein variational gradient descent (SVGD) and provides an unbiased estimation of $\E_{\vx\sim p}[f(\vx)]$. Like most approximate inference algorithms in Markov chain Monte Carlo (MCMC) \citep{neal2011mcmc, hoffman2014no}, or variational inference \citep{blei2017variational, zhang2017advances}, SVGD and SteinIS require the gradient information of the target distributions. Starting from particles $\{\vx_i\}_{i=1}^n$ drawn from any distribution, SVGD iteratively updates the particles 
\begin{align}\label{chap:svgd:update11}
\!\!\vx_i \gets \vx_i  + \frac{\epsilon}{n} \Delta \vx_i, \text{~where~} \Delta \vx_i =\sum_{j=1}^n [\nabla \log p(\vx_j)k(\vx_j, \vx_i) +  \nabla_{\vx_j}k(\vx_j,\vx_i)],
\end{align}
where the $\KL$ divergence between the distribution of the updated particles and the target distribution is maximally
Unfortunately, gradient information of the target distribution is not always available in practice. In some cases, the distribution of interest is only available as a black-box density function and the gradient cannot be calculated analytically; in other cases, it may be computationally too expensive to calculate the gradient \citep{beaumont2003estimation, andrieu2009pseudo, filippone2014pseudo}.

In this chapter, we are going to extend SVGD to the gradient-free setting, where the gradient of the target distribution is unavailable or intractable.  Basically, we leverage the gradient of a surrogate distribution $\rho(\vx)$ and corrects the bias in the SVGD update with a form of importance weighting. The gradient-free update, motivated from the gradient-based update~\eqref{chap:svgd:update11}, is given as follows,
\begin{align}
\label{chap:gf:update}
\!\!\vx_i \gets \vx_i  + \frac{\epsilon}{n} \Delta \vx_i, \!~\text{where}~ \Delta \vx_i \propto\!\!\!
\sum_{j=1}^n 
\!\! w(\vx_j) \big[\nabla \log \rho(\vx_j) k( \vx_j, \vx_i) + \nabla_{\vx_j} k(\vx_j, \vx_i) \big], 
\end{align}
which replaces the true gradient $\nabla \log p(\vx)$ with a surrogate gradient $\nabla\log \rho(\vx)$ of an arbitrary auxiliary distribution $\rho(\vx)$, and then uses an importance weight $w(\vx_j):=\rho(\vx_j)/p(\vx_j)$ to correct the bias introduced by the surrogate distribution. In this chapter, we will provide theoretical analysis to justify the effectiveness of such gradient-free update. It is interesting to observe that replacing the kernel $k(\vx, \vx')$ in original SVGD with a new kernel,
$\wt{k}(\vx,\vx') = \frac{\rho(\vx)}{p(\vx)} k(\vx, \vx')\frac{\rho(\vx')}{p(\vx')}$, we exactly get the gradient-free update \eqref{chap:gf:update}. Therefore, our gradient-free update \eqref{chap:gf:update} inherits all nice properties of the gradient-based SVGD~
\citet{liu2017stein}.  

However, the performance of the gradient-free update \eqref{chap:gf:update} critically depends on the choice of the surrogate distributions. We provide some empirical guidance about how to choose a reasonable surrogate by conducting enough empirical experiments. We further propose a robust method to overcome the difficulty of choosing the surrogate distribution, which is motivated from annealed importance sampling and will be discussed in the main section of this chapter. The idea is that we apply gradient-free update to the intermediate distribution $p_\ell(\bd{x})$ that interpolate between the initial distribution $p_0(\vx)$ and the target distribution $p(\vx)$: 
$p_\ell(\bd{x}) \propto p_0(\bd{x})^{1-\beta_\ell} p(\bd{x})^{\beta_\ell}$,
 where $0= \beta_0< \beta_1< \cdots <\beta_T= 1$ is a set of temperatures. The initial particles can be drawn from $p_0(\vx)$. Instead of applying gradient-free update to $p(\vx),$ we set the intermediate distribution $p_\ell(\bd{x})$ as the target target and the surrogate distribution is constructed on the fly based on the current particles $\{\vx_i^{\ell}\}_{i=1}^n,$ which approximates $p_{\ell-1}(\bd{x})$ by our update. Therefore, the importance ratio $\rho(\vx)/p_{\ell}(\vx)$ is evaluated between two close distributions, which approximates $p_{\ell-1}(\vx)/p_{\ell}(\vx).$ Empirical experiments demonstrate the improved gradient-free update is robust and can be widely applied to perform gradient-free sampling and significantly outperform gradient-free MCMC algorithms.

{\bf Outline} We will first develop a gradient-free form of Stein's identity and gradient-free kernelized Stein discrepancy in Section 1. Based on this key observation, we develop our main gradient-free sampling algorithm which leverages the gradient information of the surrogate distribution and corrects the bias with a form of importance weighting in Section 2. We empirically investigate the optimal choice of the surrogate distributions and propose an annealed form of gradient-free SVGD in Section 3. We conduct experiments in Section 4 to verify the effectiveness of our proposed algorithms. We propose a gradient-free black-box importance sampling algorithm in Section 5.

\section{Gradient-Free Stein's Identity and Stein Discrepancy} 
\newcommand{\ellt}{\ell}
The standard SVGD integrates the advantages of both MCMC and variational inference to perform fast and sample-efficient approximate inference. But SVGD requires the gradient of the target $p(\vx)$ and cannot be applied when the gradient of the target distributions is unavailable. In this section, we propose a gradient-free variant of SVGD which replaces the true gradient with a surrogate gradient and corrects the bias introduced using an importance weight. 
We start with introducing a  gradient-free variant of Stein's identity and Stein discrepancy. 
 
We can generalize Stein's identity to make it depend on a surrogate gradient $\nabla_\vx \log \prop(\vx)$ of an arbitrary auxiliary distribution $\rho(\vx)$, instead of the true gradient $\nabla_\vx \log p(\vx)$. The idea is to use importance weights to transform Stein's identity of $\prop(\vx)$ into an identity regarding $p(\vx)$. Recall that the Stein's identity of $\prop(\vx)$:  
$$\E_{\vx\sim \prop} [ \steinbxtransp \ff(\vx)  ] =  0.$$ 
It can be easily seen that it is equivalent to the following \emph{importance weighted Stein's identity}: 
\begin{align} \label{equ:iwstein}
\E_{\vx\sim p} \bigg[\frac{\prop(\vx)}{p(\vx)} \steinbxtransp \ff(\vx) \bigg] = 0, 
\end{align}
which is already gradient free since it depends on $p(\vx)$ only through the value of $p(\vx)$, not the gradient. 
\eqref{equ:iwstein} holds for an arbitrary auxiliary distribution $\rho(\vx)$ which satisfies $\rho(\vx)/p(\vx) <\infty$ for any $\vx$.

Based on identity \eqref{equ:iwstein}, it is straightforward to define an 
importance weighted Stein discrepancy 
\begin{align}\label{equ:grds}
\D_{\F,\prop}(q~||~p) = \max_{\ff \in \F}\bigg\{ \E_{x\sim q}\bigg[\frac{\prop(\vx)}{p(\vx)} \steinbxtransp \ff(\vx) \bigg] \bigg\}, 
\end{align}
 which is gradient-free if $\rho$ does not depend on the gradient of $p$. 
Obviously, this includes the standard Stein discrepancy in Section~\ref{sec:svgd} as special cases: 
if $\prop = p$, then $\D_{\F,\rho}(q~||~p) = \D_\F(q~||~p)$, reducing to the original definition in \eqref{solvksd}, while if $\prop = q$, then $\D_{\F, \rho}(q~||~p) = \D_{\F}(p~||~q)$, which switches the order of $p$ and $q$.

It may appear that $\D_{\F, \rho} (q~||~p)$ strictly generalizes the definition~\eqref{solvksd} of Stein discrepancy. 
One of our key observations, however, shows that this is not the case. Instead, 
$\D_{\F, \rho} (q~||~p)$ can also be viewed as a special case of 
$\D_{\F} (q~||~p)$, 
by replacing $\F$ in \eqref{solvksd} with 
$$w \F := \{w(\vx) \ff(\vx) \colon \ff\in \F\},$$
where $w(\vx) = {\prop(\vx)}/{p(\vx)}$. 
\begin{thm}\label{pro:wphi}
Let $p(\vx)$, $\prop(\vx)$ be  positive differentiable densities and $w(\vx) = {\prop(\vx)}/{p(\vx)}$. We have 
\begin{align}\label{equ:ws}
 w(\vx)\steinbxtransp\ff(\vx)  = \steinpxtransp \big(w(\vx)\ff (\vx) \big).
\end{align}
Therefore,
$\S_{\F, \prop}(q~||~p)$ in \eqref{equ:grds}
is equivalent to 
\begin{align}
 \S_{\F, \prop}(q~||~p)  
 &  = \max_{\ff \in \F}\big\{ \E_{\vx\sim q} [\steinpxtransp \big( w(\vx)\ff(\vx)\big)]\big\} \label{sbf} \\
 & = \max_{\ff \in w\F} \big \{\E_{\vx\sim q}[\steinpxtransp \ff(\vx)]     \big \}  \label{sbf2}\\
 & = \S_{w\F}(q~||~p). \notag
\end{align}
\end{thm}
{\bf Proof:} The proof can be found in the appendix A. \hfill $\square$

Identity~\eqref{equ:ws} is interesting because it is \emph{gradient-free} (in terms of $\nabla_\vx\log p(\vx)$) from the left hand side,
but \emph{gradient-dependent} from the right hand side; 
this is because the $\nabla_\vx \log p$ term in $\steinpx$ is cancelled out when applying $\steinpx$ on the density ratio $w(\vx) = \rho(\vx)/p(\vx)$.

It is possible to further extend our method to take $\rho(\vx)$ and 
$w(\vx)$ to be general \emph{matrix-valued} functions,  
in which case the operator $\steinpxtransp (w(\vx)\phi(\vx))$ 
is called diffusion Stein operator in 
\citet{gorham2016measuring}, corresponding to various forms of Langevin diffusion when taking special values of $w(\vx)$.    
We leave it as future work to explore 
 $\rho(\vx).$ 
 
\section{Gradient-Free Sampling on Continuous Distributions}
Theorem~\ref{pro:wphi} suggests that by simply multiplying $\ff$ with an importance weight $w(\vx)$ (or replacing $\F$ with $w\F$), one can transform Stein operator $\steinpx$ to operator $\steinbx$, which depends on $\nabla \log \rho(\vx)$ instead of $\nabla \log p(\vx)$  (\emph{gradient-free}). 
\begin{algorithm}[t] %
\caption{Gradient-Free SVGD (GF-SVGD)}  
\label{alg:alg1}
\begin{algorithmic}
\STATE {\bf Input:} 
Target distribution $p(\bd{x})$; 
surrogate $\prop(\bd{x})$ and its score function $\bd{s}_{\prop}(\vx) := \nabla_\vx\log \prop(\vx).$ 
\STATE {\bf Goal:} Find particles  $\{\bd{x}_i\}_{i=1}^n$ to approximate $p.$ 
\STATE {\bf Initialize} particles $\{\vx^0_i\}_{i=1}^n$ from any distribution $q$.
\FOR{iteration $\ellt$}
\vspace{-1.\baselineskip}
\STATE
\hspace{-0.4\baselineskip}
\begin{align}
& \bd{x}_i^{\ellt+1} ~ \leftarrow ~ \bd{x}_i^\ellt  ~  + ~ \Delta\vx_i^\ellt, 
~~~\forall i = 1, \ldots, n,  ~~\text{where}~ \notag \\ 
&
\!\!\!\!\!\!\!\!\!\!\!\Delta\vx_i^\ellt = 
\frac{\epsilon_{\ellt,i}}{
Z_\ellt}\sum_{j=1}^n
w(\vx_j^\ellt)\big[ \bd{s}_{\prop}(\vx_j^\ellt) k( \vx_j^\ellt,  \vx_i^\ellt) + \nabla_{\vx_j} k(\vx_j^\ellt, \vx_i^\ellt) \big], \notag
\end{align}
where $w(\vx) := \prop(\vx)/p(\vx)$,  $Z_\ellt = \sum_{j=1}^n w(\vx_j^\ellt)$, 
and $\epsilon_{t,i}$ is a step size. 
\ENDFOR
\vspace{-.2\baselineskip}
\end{algorithmic}
\end{algorithm}

This idea can be directly applied to derive a gradient-free extension of SVGD, 
by updating the particles using velocity fields of form $w(\vx)\ff(\vx)$ from  space $w\F$:   
\begin{equation}
\label{gf:closed}
 \vx\gets \vx + \epsilon w(\vx) \ff^*(\vx),
\end{equation}
where $\ff^*$ maximzies the decrease rate of KL divergence, 
\begin{align}
\label{gradfreeKLmin}
\!\!\!\! \ff^*& 
\!=\! \argmax_{\ff \in \H} \!\!\left\{ \E_{q} [\steinpxtransp (w(\vx)\ff(\vx))], \mathrm{s.t.}~||\ff ||_{\H} \leq 1 \!\right\}\!. 
\end{align}
Similar to \eqref{solvksd}, 
we can derive a closed-form solution for \eqref{gradfreeKLmin} when $\H$ is RKHS. 
To do this, it is sufficient to recall that if $\H$ is an RKHS with kernel $k(\vx,\vx')$, then $w\H$ is also an 
RKHS, with an ``importance weighted kernel'' \citep{berlinet2011reproducing} 
\begin{align}\label{tildk}
    \tilde k(\vx,\vx') = w(\vx)w(\vx')k(\vx,\vx').
\end{align}
\begin{thm}
When $\Hd$ is an RKHS with kernel $k(\vx,\vx')$, the optimal solution of \eqref{sbf} is ${\ff}^*/||{\ff}^*||_\Hd,$ where 
\begin{align}
{\ff}^*(\cdot) 
& =\E_{\vx\sim q}[\steinpx(w(\vx)k(\vx, \cdot))] \label{newvel}
\\
& =\E_{\vx\sim q}[w(\vx) \steinbx k(\vx, \cdot)], \label{newvel2}
\end{align}
where the Stein operator $\steinbx$ is applied to variable $\vx$, $\steinbx k(\vx, \cdot)=\nabla_\vx \log \rho(\vx)k(\vx, \cdot)+\nabla_{\vx}k(\vx, \cdot).$
 Correspondingly, the optimal decrease rate of KL divergence in \eqref{gradfreeKLmin} equals the square of $\S_{\F, \rho}(q~||~p)$, which equals 
\begin{equation}
\label{newksd}
 \S_{\F, \prop}(q ~||~ p) = (\E_{\vx, \vx'\sim q}[w(\vx)w(\vx')\kappa_\prop(\vx,\vx')])^{\frac12},
\end{equation}
where the positive definite kernel $\kappa_\prop(\vx,\vx') = (\newsteinbx)^\top (\steinbx k(\vx,\vx'))$ and $\newsteinbx$ is the Stein operator applied to the surrogate distribution $\rho(\vx')$.
\label{thm:gf:disc}
\end{thm}
{\bf Proof:} The proof can be found in the appendix B. \hfill $\square$

The form in \eqref{newksd} allows us to estimate  $\S_{\F, \prop}(q~||~p)$ empirically either using U-statistics or V-statistics when $q(\vx)$ is observed through an i.i.d. sample, with the advantage of being gradient-free.  
Therefore, it can be directly applied to construct gradient-free methods for goodness-of-fit tests~\citep{liu2016kernelized, chwialkowski2016kernel} and black-box importance sampling~\citep{liu2016black} when the gradient of $p(\vx)$ is unavailable. 
We will discuss it in the following chapter. 

Theorem~\ref{thm:gf:disc} provides a novel way to iteratively update a set of particles $\{\vx_i\}_{i=1}^n$, which are drawn from any simple initial distribution,
\begin{align*}
\vx_i \gets \vx_i  + \frac{\epsilon}{n} \Delta \vx_i, ~\text{where}~ \Delta \vx_i \propto
\sum_{j=1}^n 
\! w(\vx_j) \big[\nabla \log \rho(\vx_j) k( \vx_j, \vx_i) + \nabla_{\vx_j} k(\vx_j, \vx_i) \big], 
\end{align*}
which replaces the true gradient $\nabla \log p(\vx)$ with a surrogate gradient $\nabla\log \rho(\vx)$ of an arbitrary auxiliary distribution $\rho(\vx)$, and then uses an importance weight $w(\vx_j):=\rho(\vx_j)/p(\vx_j)$ to correct the bias introduced by the surrogate distribution.

\paragraph{Comparison with SVGD Update} SVGD starts with a set of particles and iteratively updates the particles $\{\vx_i\}_{i=1}^n$ by 
\begin{align}
\label{gf:update11}
\!\!\vx_i \gets \vx_i  + \frac{\epsilon}{n} \Delta \vx_i, \text{~where~} \Delta \vx_i =\sum_{j=1}^n [\nabla \log p(\vx_j)k(\vx_j, \vx_i) +  \nabla_{\vx_j}k(\vx_j,\vx_i)], 
\end{align}
where $k(\vx,\vx')$ is any positive definite kernel; the term with the gradient $\nabla\log p(\vx)$ drives the particles to the high probability regions of $p(\vx)$,
and the term with $\nabla k(\vx, \vx')$ acts as a repulsive force to keep the particles away from each other to quantify the uncertainty. 

\paragraph{Monotone Decreasing of KL divergence} 
One nice property of the gradient-free SVGD is that the KL divergence between the updated distribution $q_\ell(\vx)$ and the target distribution $p(\vx)$ is monotonically decreasing. This property can be more easily understood by considering our iterative system in continuous evolution time as shown in \citet{liu2017stein}. 
Take the step size $\epsilon$ of the transformation defined in \eqref{update} to be infinitesimal, 
and define the continuous time $t = \epsilon \ell$. Then the evolution equation of random variable $\vx^t$ is governed by the following nonlinear partial differential equation~(PDE), 
\begin{equation}
\label{part}
\frac{d\bd{x}^t}{dt}=\mathbb{E}_{\bd{x}\sim{q_t}}[w(\vx)(\bd{s}_{\rho}(\bd{x})k(\bd{x},\bd{x}^t)+\nabla_{\bd{x}} k(\bd{x},\bd{x}^t))],
\end{equation}
where $t$ is the current evolution time and $q_t$ is the density function of $\vx^t.$ The current evolution time $t= \epsilon \ell$ when $\epsilon$ is small and $\ell$ is the current iteration. We have the following proposition: 
\begin{pro}
\label{gfpro3}
Suppose random variable $\bd{x}^t$ is governed by PDE \eqref{part}, then its density $q_t(\bd{x})$ is characterized by
\begin{equation}
\label{diffodegf}
\frac{\partial q_t}{\partial t}=-\mathrm{div}(q_t\mathbb{E}_{\bd{x}\sim{q_t}}[w(\vx)(\bd{s}_{\rho}(\bd{x}) k(\bd{x},\bd{x}^t)+\nabla_{\bd{x}} k(\bd{x},\bd{x}^t))]),
\end{equation}
where $\mathrm{div}(\bd{f})=\trace(\nabla \vv f) = \sum_{i=0}^d \partial f_i(\bd{x})/\partial x_i$, and  $\bd{f}=[f_1,\ldots, f_d]^\top.$ And the derivative of the $\KL$ divergence between the iterative distribution $q_t(\vx)$ and the target $p(\vx)$ satisfies that
\begin{equation}
\frac{d\KL(q_t, p)}{dt} = -  \E_{\vx, \vx'\sim q}[w(\vx)w(\vx')\kappa_\prop(\vx,\vx')] \le 0,
\end{equation}
where $\kappa_\prop(\vx,\vx')$ can be derived as
\begin{equation}
\label{chap:gf:ksd}
\begin{aligned}
\kappa_{\rho} (\bd{x},  \bd{y}) = & (\newsteinbx)^\top (\steinbx k(\vx,\vx'))= \bd{s}_{\rho}(\bd{x})^\top k(\bd{x},\bd{y})\bd{s}_{\rho}(\bd{y}) \\ & +\bd{s}_{\rho}(\bd{x})^\top \nabla_{\bd{y}}k(\bd{x},\bd{y}) 
+\bd{s}_{\rho}(\bd{y})^\top \nabla_{\bd{x}} k(\bd{x},\bd{y})+\nabla_{\bd{x}}\cdot(\nabla_{\bd{y}}k(\bd{x}, \bd{y})).
\end{aligned}
\end{equation}
\end{pro}
It is interesting to observe that replacing the kernel $k(\vx, \vx')$ in original SVGD with a new kernel,
\begin{equation}
\wt{k}(\vx,\vx') = \frac{\rho(\vx)}{p(\vx)} k(\vx, \vx')\frac{\rho(\vx')}{p(\vx')},     
\end{equation}
Proposition~\ref{gfpro3} is straightforward to derive from the derviation in SVGD~\citep{liu2017stein}.

Using the gradient-free form of $\ff^*$ in \eqref{newvel2}, 
we can readily derive a gradient-free SVGD update $\vx_i \gets \vx_i  + \Delta \vx_i$, with  
\begin{align}\label{dx}
\Delta \vx_i = 
\frac{\epsilon_i}{Z} \sum_{j=1}^n w(\vx_j) \steinbx k(\vx_j, \vx_i),
\end{align}
where the operator $\steinbx k(\vx_j,\vx_i)$ is applied on variable $\vx_j$, $$\steinbx k(\vx_j,\vx_i)=\nabla \log \rho(\vx_j) k( \vx_j, \vx_i) + \nabla_{\vx_j} k(\vx_j, \vx_i),$$ and we set $Z = n$, viewed as a normalization constant, and $\epsilon_i = \epsilon w(\vx_i)$, viewed as the step size of particle $\vx_i$ .  

Since $\epsilon_i/Z$ is a scalar, we can change it in practice without altering the set of fixed points of the update. 
In practice, because the variability of the importance weight $ w(\vx_i)$ can be very large, making the updating speed of different particles significantly different, we find it is empirically better to determine $\epsilon_i$ directly using off-the-shelf step size schemes such as Adam~\citep{kingma2014adam}. 

In practice, we also replace $Z={n}$ with a self-normalization factor $Z=\sum_{j=1}^n w(\vx_j)$ (see Algorithm~\ref{alg:alg1}). 
We find this makes tuning step sizes become easier in practice, and more importantly, 
avoids to calculate the normalization constant of either $p(\vx)$ or $\rho(\vx)$. 
This sidesteps the critically challenging problem of calculating the normalization constant and allows us to essentially choose $\rho(\vx)$ to be an arbitrary positive differentiable function once we can calculate its value and gradient. 

\paragraph{Choice of the Auxiliary Distribution $\prop(\vx)$}
Obviously, the performance of gradient-free SVGD (GF-SVGD) 
critically depends on the choice of the auxiliary distribution $\prop(\vx)$. 
Theoretically, gradient-free SVGD is just a standard SVGD with the importance weighted kernel $\tilde k(\vx, \vx')$. 
Therefore, the optimal choice of $\prop(\vx)$ is essentially the problem of choosing an optimal kernel for SVGD, which, unfortunately, is a difficult, unsolved problem.  

In this work, we take a simple heuristic that sets $\prop(\vx)$ to approximate $p(\vx)$. This is based on the justification that if the original kernel $k(\vx,\vx')$ has been chosen to be optimal or ``reasonably well'', 
we should take $\prop(\vx)\approx p(\vx)$ so that $\tilde k(\vx, \vx')$ is close to $k(\vx,\vx')$ and GF-SVGD will have similar performance as the original SVGD. 

In this way, the problem of choosing the optimal auxiliary distribution $\prop(\vx)$ and the optimal kernel $k(\vx, \vx')$ is separated, and different kernel selection methods can be directly plugged into the algorithm. In practice, we find that $\prop(\vx)\approx p(\vx)$ serves a  reasonable heuristic when using Gaussian RBF kernel $k(\vx,\vx')$. 
Interestingly, our empirical observation shows that a widely spread $\prop(\vx)$ tends to give better and more stable results than peaky $\prop(\vx)$. In particular, Figure~\ref{fig:gfgauss} in the experimental section shows that in the case when both $p(\vx)$ and $\rho(\vx)$ are Gaussian and RBF kernel is used,
the best performance is achieved when the variance of $\prop(\vx)$ is larger than the variance of $p$. 
In fact, the gradient-free SVGD update \eqref{dx}
still makes sense even 
when $\prop(\vx) = 1$, corresponding to an improper distribution with infinite variance: 
\begin{align}\label{invp}
\Delta \vx_i  = \frac{\epsilon_i}{Z} \sum_{j=1}^n \frac{1}{p(\vx_j)} \nabla_{\vx_j} k(\vx_j,\vx_i). 
\end{align}
This update is interestingly simple; it has only a repulsive force and relies on an inverse probability $1/p(\vx)$ to adjust the particles towards the target $p(\vx)$. 
We should observe that it is \emph{as general as} 
the GF-SVGD update \eqref{dx} (and hence the standard SVGD update \eqref{update11}), because if we replace $k(\vx,\vx')$ with $\prop(\vx)\prop(\vx')k(\vx,\vx')$, \eqref{invp} reduces back to \eqref{dx}.    

All it matters is the choice of the kernel function. 
With a ``typical'' kernel such as RBF kernel, we empirically find that the particles by the update \eqref{invp} can  estimate the mean parameter reasonably  well (although not optimally), but tend to  overestimate the variance because the repulsive force dominates; see  Figure~\ref{fig:gfgauss}.   

\begin{algorithm}[t]
\caption{Annealed SVGD (A-SVGD)}
\begin{algorithmic}
\STATE {\bf Inputs:} $p(\vx)$, distribution path $\{p_\ellt\}_{\ellt=1}^T$ with $p_T = p.$ 
\STATE {\bf Initialize} particles $\{\vx^0_i\}_{i=1}^n$ from any distribution. 
\FOR{iteration $\ellt =0, \cdots, T-1$}
\STATE Update the particles to get $\{\bd{x}_i^{t+1}\}_{i=1}^n$ by running the typical SVGD with $p_{\ellt+1}$ as the target for $m$ steps. 
\ENDFOR
\STATE {\bf Output:} $\{\bd{x}_i^{T}\}_{i=1}^n$ as an approximation of $p.$ 
\STATE {\bf Remark:} $m=1$ is sufficient when $T$ is large. 
\vspace{-.2\baselineskip}
\end{algorithmic}
\label{alg2}
\end{algorithm}

\section{Annealed Gradient-Free SVGD}
In practice, it may be difficult to directly find $\rho(\vx)$ that closely approximates the target $p$,  
causing the importance weights to have undesirably large variance and deteriorate the performance.  
In this section, we introduce an annealed GF-SVGD algorithm that overcomes the difficulty of choosing $\rho$ and improves the performance 
by iteratively approximating a sequence of distributions which interpolate the target distribution with a simple initial distribution. 
In the sequel, we first introduce the annealed version of the basic SVGD and then its combination with GF-SVGD.

\newcommand{\pz}{p_0}
\textbf{Annealed SVGD (A-SVGD)} is a simple combination of SVGD and simulated annealing, and has been discussed by \citet{liu2017steinpolicy} in the setting of reinforcement learning. 
Let $\pz(\vx)$ be a simple initial distribution. 
We define a path of distributions that interpolate between $\pz(\vx)$ and $p(\vx)$: 
 $$
 p_\ellt(\bd{x}) \propto \pz(\bd{x})^{1-\beta_\ellt} p(\bd{x})^{\beta_\ellt},
 $$
where $0= \beta_0< \beta_1< \cdots <\beta_T= 1$ is a set of temperatures. The following one-dimensional example \ref{fig:ais} illustrates the annealed distribution path from the initial distribution $\pz(\vx)$ and the target distribution $p(\vx).$
 \begin{figure}[H]
\centering
\includegraphics[width=.99\textwidth]{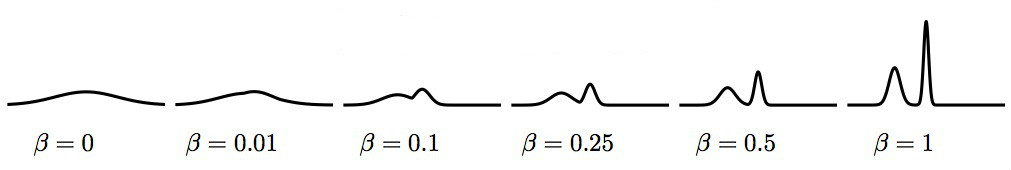}
\caption[Illustration of constructing the annealed distribution path from the initial distribution and the target distribution in 1D distribution.]{Illustration of constructing the annealed distribution path from the initial distribution $\pz(\vx)$ and the target distribution $p(\vx)$ in one-dimension case. $
 p_\ellt(\bd{x}) \propto \pz(\bd{x})^{1-\beta_\ellt} p(\bd{x})^{\beta_\ellt},$ with $\beta=0, 0.01, 0.1, 0.25, 0.5, 1.0$, respectively. \label{fig:ais}}
\end{figure}

Annealed SVGD starts from a set of particle $\{x_i^0\}_{i=1}^n$ drawn from $p_0$, and at the $\ellt$-th iteration, updates the particles so that $\{x_i^{\ellt+1}\}_{i=1}^n$ approximates the intermediate distribution $p_{\ellt+1}$ by running $m$ steps of SVGD with $p_{\ellt+1}$ as the target. See Algorithm~\ref{alg2}. In practice, $m=1$ is sufficient when $T$ is large.
\begin{algorithm}[t] %
\caption{Annealed GF-SVGD (AGF-SVGD)}  
\begin{algorithmic}
\STATE {\bf Input:}  Target distribution $p(\bd{x})$; initial distribution $\pz(\bd{x})$;  intermediate distributions  $\{p_t\}_{t=1}^T$. 
\STATE {\bf Goal:} Particles $\{\bd{x}_i\}_{i=1}^n$ to approximate $p(\bd{x}).$ 
\STATE {\bf Initialize} particles $\{\vx^0_i\}_{i=1}^n$ drawn from $p_0$.  
\FOR{iteration $\ellt =0,\cdots, T-1$}
\vspace{-1.\baselineskip}
\STATE
\hspace{-10\baselineskip}
\begin{align}
& \bd{x}_i^{\ellt+1}  \leftarrow  \bd{x}_i^{\ellt} + \Delta\vx_i^\ellt,  
~~\forall i = 1, \ldots, n,  ~~\text{where}~ \notag \\
& \!\!\!\!\!\!\!\!\!\!\!\!\!\!\!\!\!\!\! \Delta \vx_i^\ellt = 
\frac{\epsilon_{\ellt,i}}{
Z_\ellt}\sum_{j=1}^n
w_j^\ellt \big[ \vv s^\rho_{j, \ellt+1} k( \vx_j^\ellt, \vx_i^\ellt) + \nabla_{\vx_j} k(\vx_j^\ellt, \vx_i^\ellt)  \big], \notag\vspace{-1em} 
\end{align}
where $\vv s^\rho_{j, \ellt+1} = \nabla_\vx \log \rho_{\ellt+1}(\vx_j^\ellt)$ and $\rho_{\ellt+1}$ is defined in \eqref{bt}; $w_j^\ellt = \rho_{\ellt+1}(\vx_j^\ellt)/p_{\ellt+1}(\vx_j^\ellt)$, 
$Z_\ellt = \sum_{j=1}^n w_j^\ellt$. 
\ENDFOR
\STATE {\bf Output:} $\{\bd{x}_i^{T}\}_{i=1}^n$ to approximate $p.$
\vspace{.2\baselineskip}
\end{algorithmic}
\label{alg3}
\end{algorithm}
It is useful to consider the special case when $p_0=const$, 
and hence $p_\ellt(\bd{x}) \propto p(\bd{x})^{\alpha_\ellt}$, yielding an annealed SVGD update of form
$$
\Delta \vx_i
= \frac{\epsilon}{n}\sum_{j=1}^n [\nabla_\vx \log p(\vx_j) k(\vx_{j}, \vx_i) + \frac{1}{\beta_\ellt}\nabla_{\vx_j} k(\vx_{j}, \vx_i)], 
$$
where the repulsive force is weighted by the inverse temperature $1/\beta_\ellt$. 
As $\beta_\ellt$ increases from 0 to 1, the algorithm starts with a large repulsive force and gradually decreases it to match the temperature of the distribution of interest. 
This procedure is similar to the typical simulated annealing, but enforces the diversity of the particles using the deterministic repulsive force, instead of random noise. 

\begin{figure*}[ht]
\centering
\begin{tabular}{ccccc}
\includegraphics[height=0.2\textwidth]{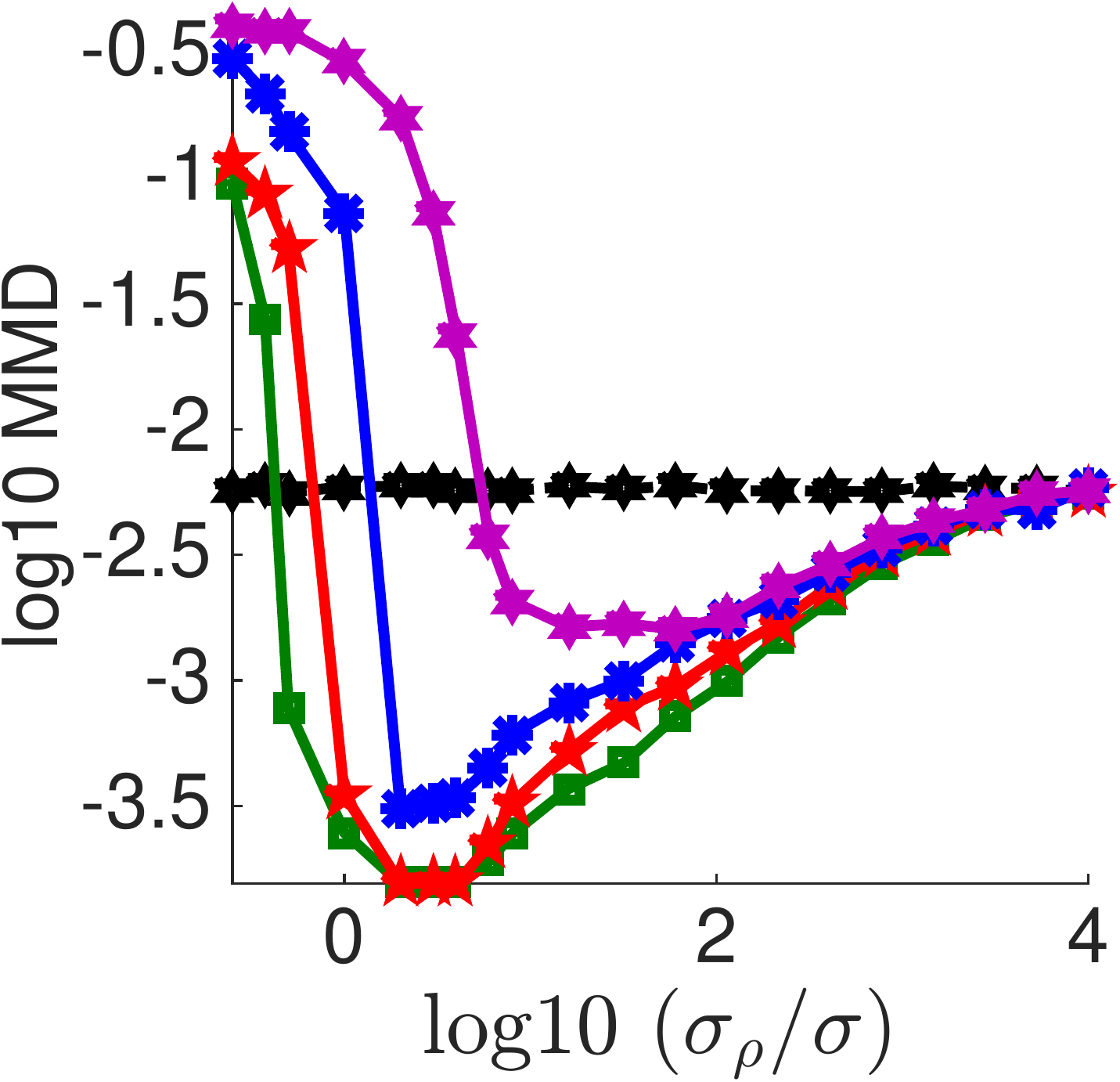}&\hspace{-0.6cm}
\includegraphics[height=0.2\textwidth]{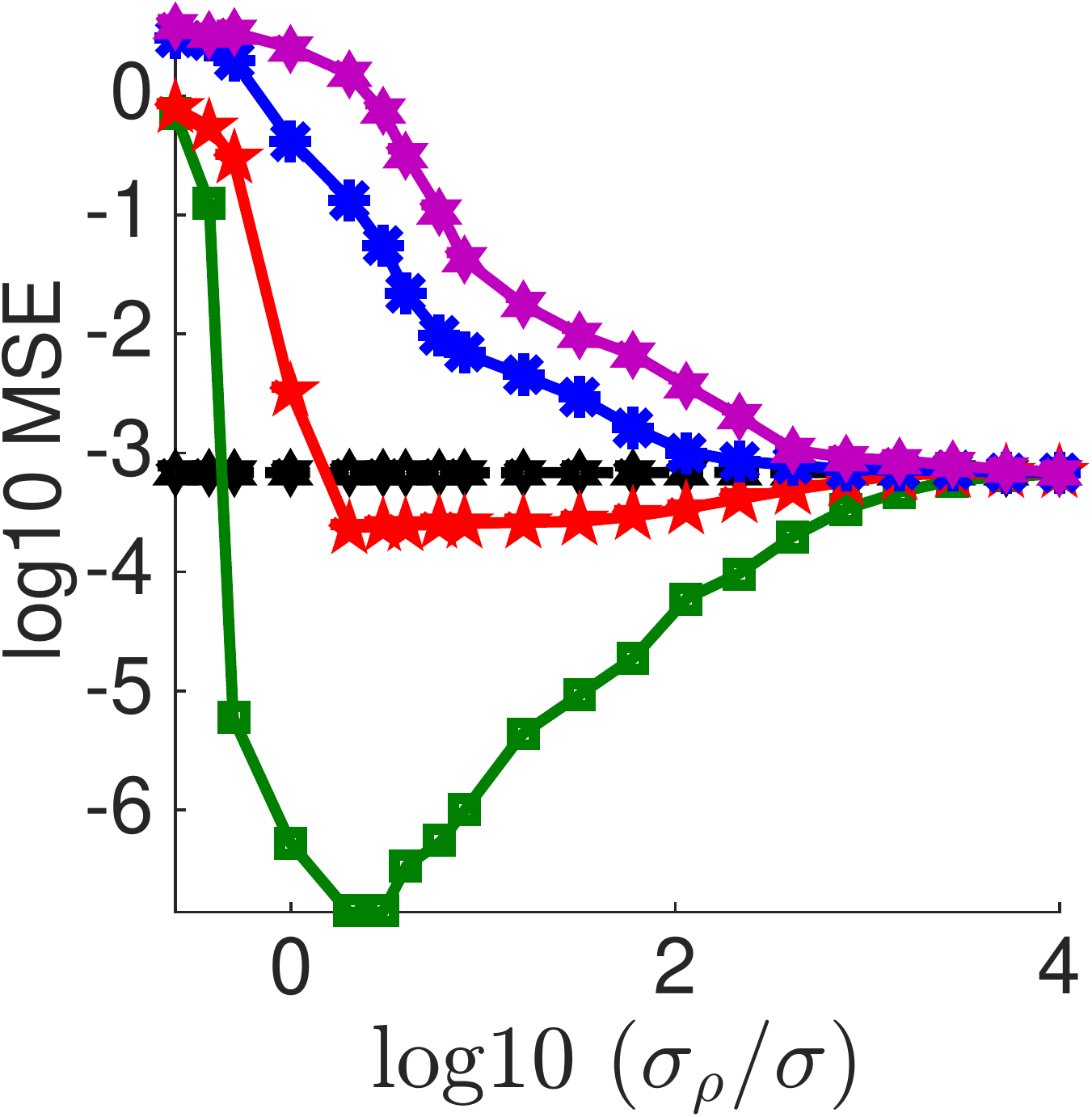} &\hspace{-0.6cm}
\includegraphics[height=0.2\textwidth]{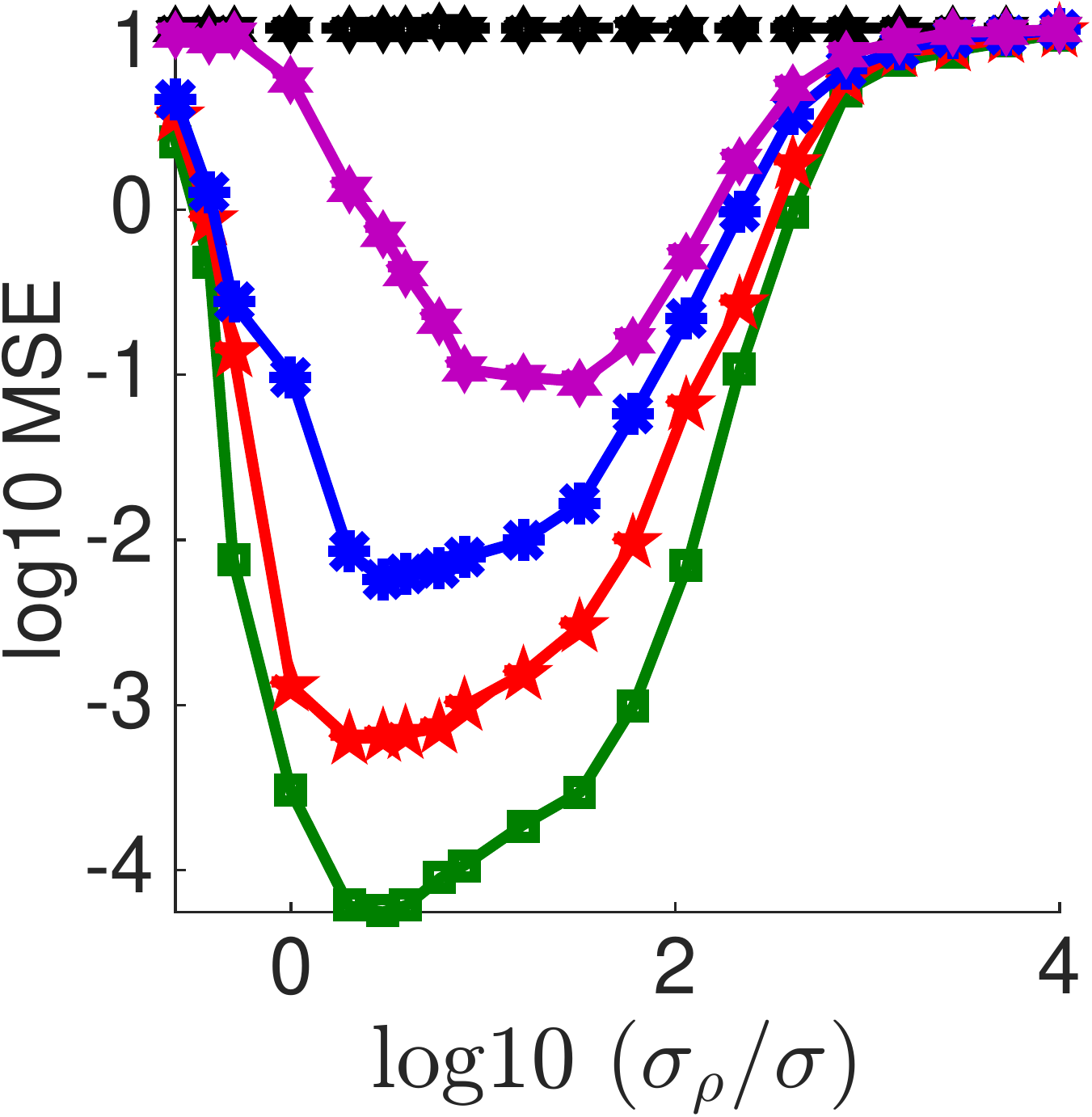} &\hspace{-0.6cm}
\raisebox{2em}{\includegraphics[height=0.1\textwidth]{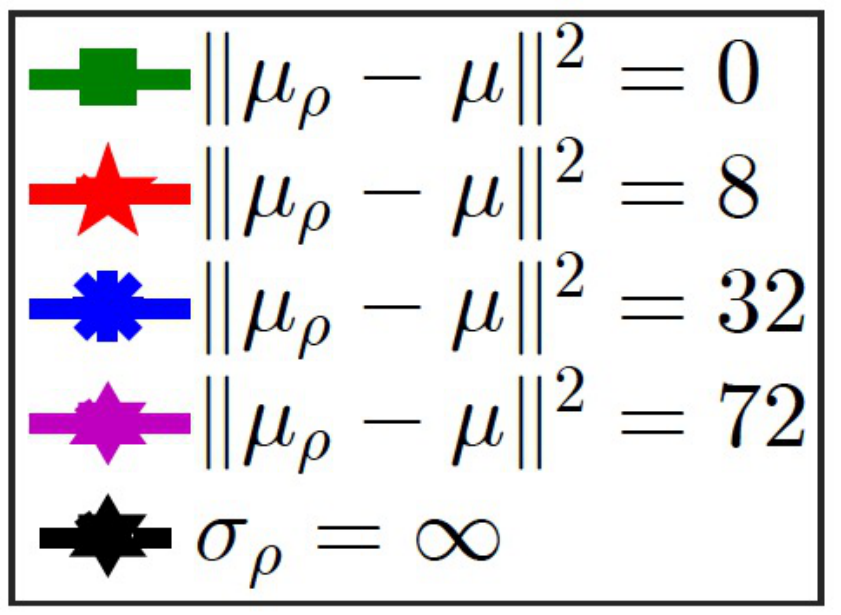}}&\hspace{-0.5cm}
\includegraphics[height=0.2\textwidth]{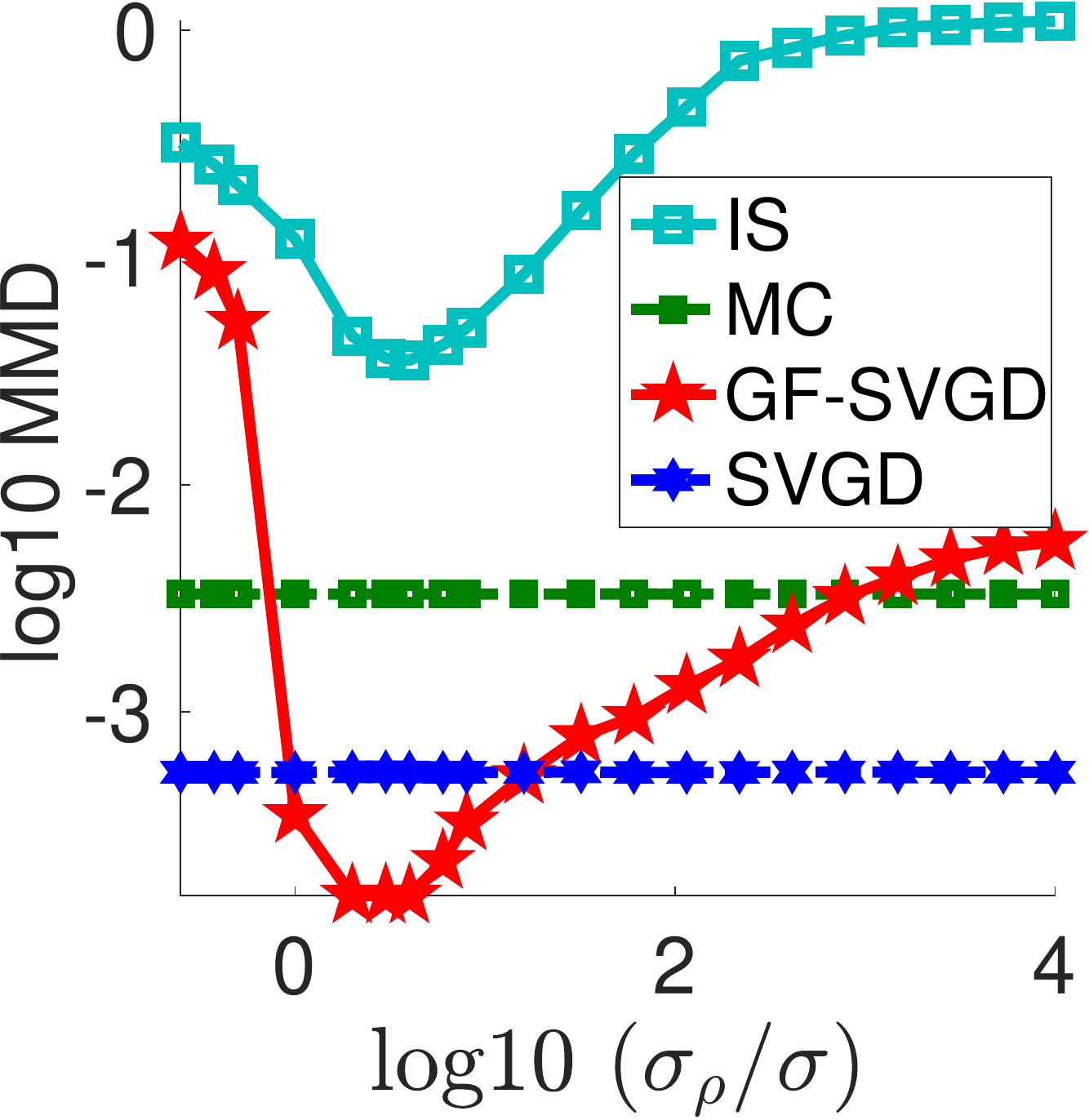} \\
 {\small (a) MMD} & {\small (b) Mean} &  {\small (c) Variance}&~~~ & {\small (d) MMD} \\
\end{tabular}
\caption[Results of GF-SVGD on 2D multivariate Gaussian distribution to investigate the choice of the surrogate distributions]{\small 
Results of GF-SVGD on 2D multivariate Gaussian distribution as we change the mean $\vv\mu_\rho$ and variance $\sigma_\rho$ of the surrogate $\rho(\vx)$. 
We can see that the best performance is achieved by matching the mean of  $\rho$ and the target $p$  ($\bd{\mu}_\rho = \bd{\mu}$),
and making $\sigma_\rho$ slightly larger than the variance $\sigma$ of $p$ (e.g., 
$\log10(\sigma_\rho/\sigma)\approx 0.5$ or 
$\sigma_\rho \approx 3 \sigma$). 
(d) uses the same setting with $||\vv\mu_\rho-\bd{\mu}||^2=8$, but also adds the result of exact Monte Carlo sampling, gradient-based SVGD, and importance sampling (IS) whose proposal is $\rho$, the same as the auxiliary distribution used by GF-SVGD shown in the red curve. We use $n=100$ particles in  this plot.
\label{fig:gfgauss}}
\end{figure*}

\newcommand{\kb}{k_\prop}
\textbf{Annealed Gradient-Free SVGD} (AGF-SVGD) is the gradient-free version of annealed SVGD 
which replaces the SVGD update with an GF-SVGD update. 
Specifically, at the $\ellt$-th iteration when we want to update the particles to match $p_{\ellt+1}$, 
we use a GF-SVGD update with auxiliary distribution $\prop_{\ellt+1} \approx p_{\ellt+1}$, which we construct by using a simple kernel curve estimation 
\begin{align}\label{bt}
\prop_{\ellt+1}(\vx) \propto \sum_{j=1}^n p_{\ellt+1}(\vx_j^{\ellt}) \kb(\vx_j^{\ellt}, \vx), 
\end{align}
where $\kb$ is a smoothing kernel (which does not have to be positive definite).  
Although there are other ways to approximate $p_{\ellt+1}$, 
this simple heuristic is computationally fast, and the usage of smoothing kernel makes $\prop_{\ellt+1}$ an \emph{over-dispersed} estimation which we show perform well in practice (see Figure~\ref{fig:gfgauss}). 
Note that here $\rho_{\ellt+1}$ is constructed to  \emph{fit smooth curve $p_{\ellt+1}$}, which leverages the function values of the distribution $p_{\ellt+1}(\vx)$ and is insensitive to the actual distribution of the current particles $\{\vx_j^{\ellt}\}$. 
It would be less robust to construct $\rho_{t+1}$ \emph{as a density estimator of distribution $p_{\ellt+1}$} because the actual distribution of the particles may deviate from what we  expect in practice. 

The procedure is organized in Algorithm~\ref{alg3}. Combining the idea of simulated annealing with gradient-free SVGD makes it easier to construct an initial surrogate distribution and estimate a good auxiliary distribution at each iteration, decreasing the variance of the importance weights. We find that it significantly improves the performance over the basic GF-SVGD for complex target distributions. 

\paragraph{Related Works on Gradient-Free Sampling Methods} 
Almost all gradient-free sampling methods employ some auxiliary (or proposal) distributions that are different, but sufficiently close to the target distribution, followed with some mechanisms to correct the bias introduced by the surrogate distribution. 
There have been a few number of bias-correction mechanisms underlying most of the gradient-free methods, 
including importance sampling, rejection sampling and the Metropolis-Hastings rejection trick. 
The state-of-the-art gradient-free sampling methods are often based on adaptive improvement of the proposals  
when using these tricks, this includes adaptive importance sampling and rejection sampling \citep[]{gilks1992adaptive, cappe2008adaptive, cotter2015parallel, han2017stein}, and adaptive MCMC methods \citep[e.g.,][]{sejdinovic2014kernel, strathmann2015gradient}.  
 
Our method is significantly different from these gradient-free sampling algorithms aforementioned in principle, with a number of key advantages. 
Instead of correcting the bias by either re-weighting or rejecting the samples from the proposal, which unavoidably reduces the effective number of usable particles,  
our method re-weights the SVGD gradient and steers the update direction of the particles in a way that compensates the discrepancy between the target and surrogate distribution, without directly reducing the effective number of usable particles. 

In addition, 
while the traditional importance sampling and rejection sampling methods require the proposals to be simple enough to draw samples from, 
our update does not require to draw samples from the surrogate $\rho$. We can set $\rho$ to be arbitrarily complex as long as we can calculate  $\rho(\vx)$ and its gradient.  
In fact, $\rho(\vx)$ does not even have to be a normalized probability, sidestepping the difficult problem of calculating the normalization constant.  

\section{Empirical Results}  
We test our proposed algorithms on both synthetic and real-world examples. 
We start with testing our methods on simple multivariate Gaussian and Gaussian mixture models, developing insights on the optimal choice of the auxiliary distribution. 

We then test AGF-SVGD on Gaussian-Bernoulli restricted Boltzmann machine (RBM)
and compare it with advanced gradient-free MCMC such as KAMH \citep{sejdinovic2014kernel} and KHMC  \citep{strathmann2015gradient}. Finally, we apply our algorithm to Gaussian process classification on 
real-world datasets. 

We use RBF kernel $k(\bd{x}, \bd{x}')=\exp(-\|\bd{x}-\bd{x}'\|^2/h)$ for the updates of our proposed algorithms and the kernel approximation in \eqref{bt}; the bandwidth $h$ is taken to be $h {=} \mathrm{med^2}/(2\log(n+1))$ where $\mathrm{med}$ is the median of the current $n$ particles. When maximum mean discrepancy (MMD) \citep{gretton2012kernel} is applied to evaluate the sample quality, RBF kernel is used and the bandwidth is chosen based on the median distance of the exact samples so that all methods use the same bandwidth for a fair comparison. Adam optimizer \citep{kingma2014adam} is applied to our proposed algorithms for accelerating convergence.

\begin{figure*}[ht]
\centering
\begin{tabular}{ccccc}
\includegraphics[height=0.2\textwidth]{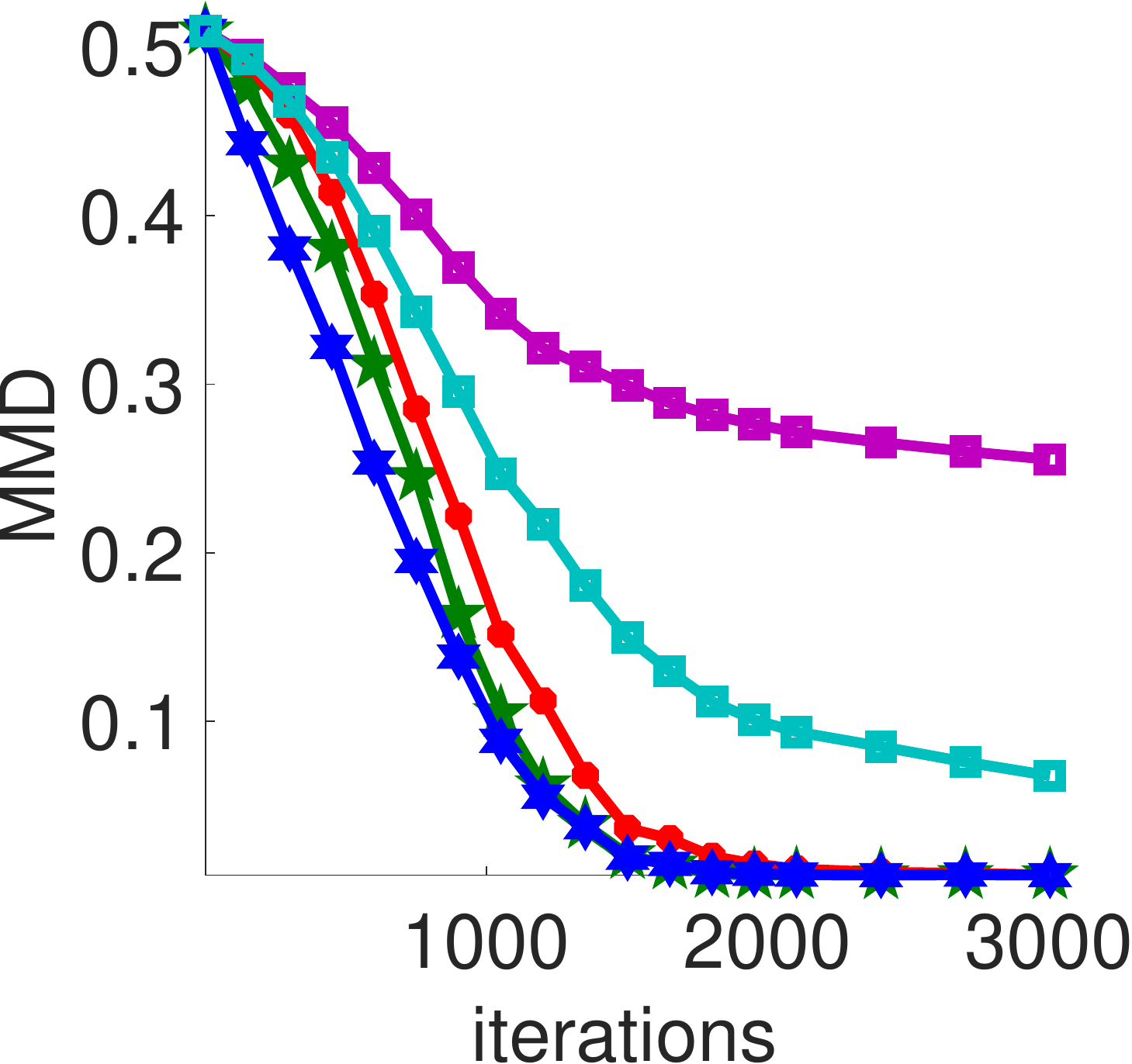}& \hspace{-0.6cm}
\includegraphics[height=0.2\textwidth]{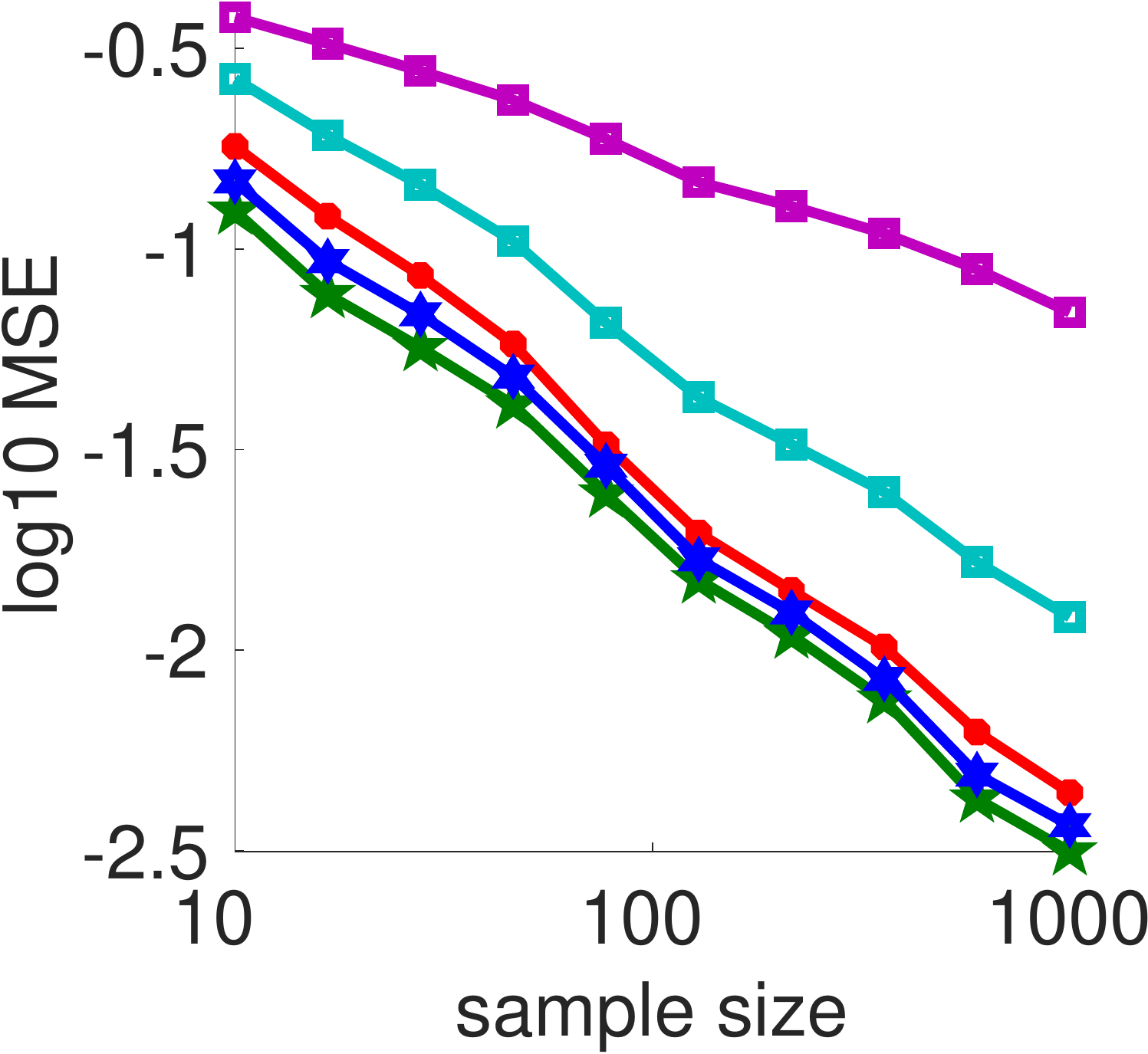}& \hspace{-0.6cm}
\includegraphics[height=0.2\textwidth]{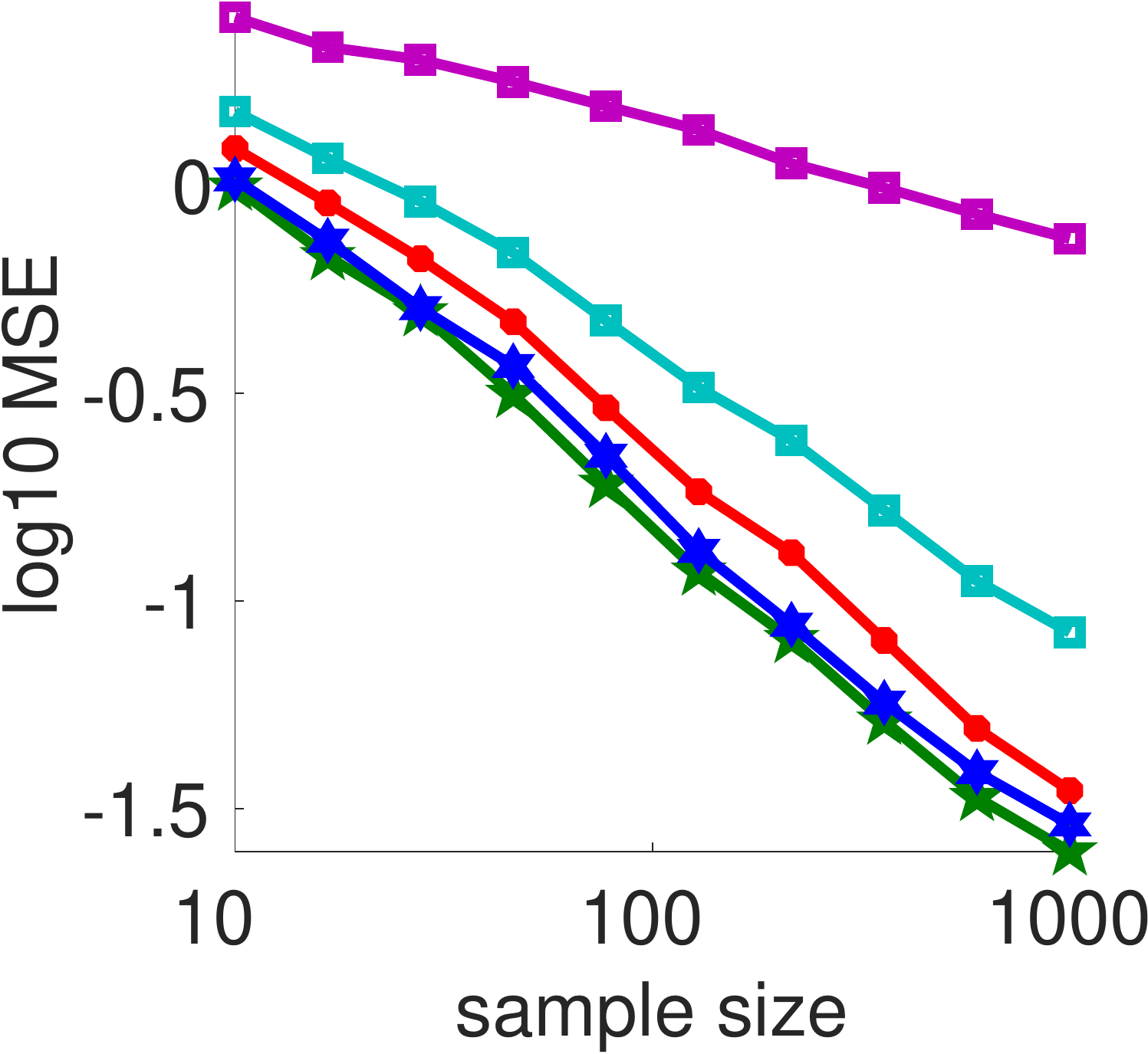} & \hspace{-0.6cm}
\includegraphics[height=0.2\textwidth]{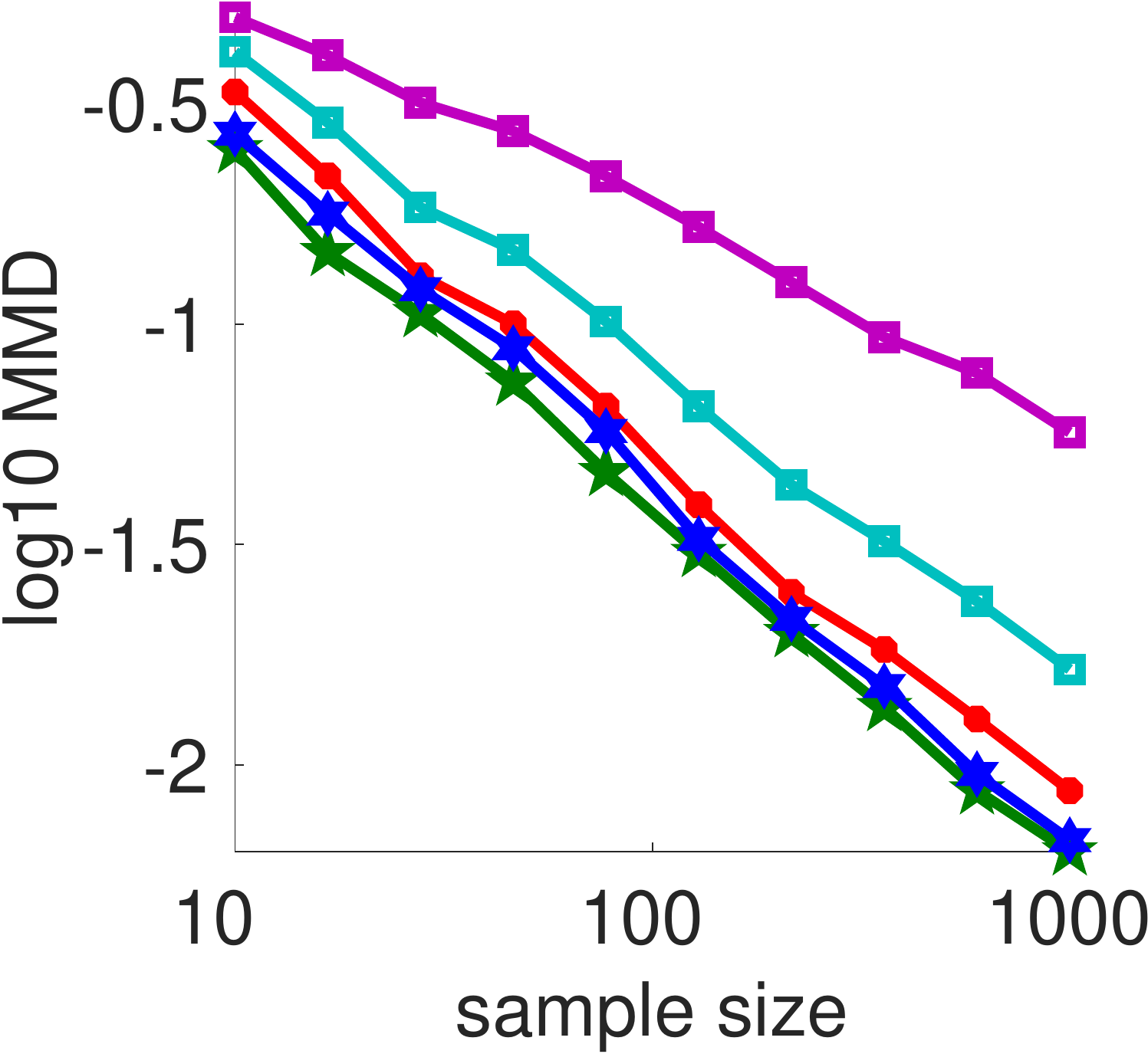}& \hspace{-0.6cm}
\raisebox{2.0em}{\includegraphics[height=0.1\textwidth]{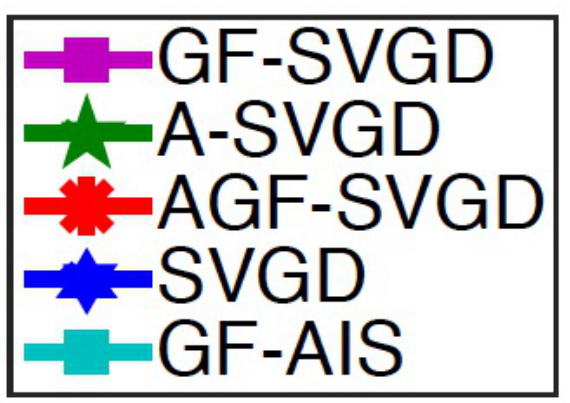}}\\
{\small (a) Convergence} & {\small (b) Mean} &  {\small (c) Variance} & {\small (d) MMD} \\
\end{tabular}
\caption[Performance of gradient-free SVGD on Gaussian mixture models with 10 random mixture components and 25 dimensions.]{\small Results on GMM with 10 random mixture components and 25 dimensions. 
(a): the convergence of MMD with fixed sample size of $n=200$. 
(b)-(c): the MSE vs. sample size when estimating the mean and variance using the particles returned by different algorithms at convergence. 
(d): the MMD between the particles of different methods and the true distribution $p$. In (b, c, d), 3000 iterations are used.
For GF-AIS, the sample size $n$ represents the number of parallel chains, 
and the performance is evaluated using the weighted average of the particles at the final iteration with their importance weights given by AIS.
\label{fig:gfgmm}}
\end{figure*}

\subsection{Simple Gaussian Distributions}   
We test our basic GF-SVGD in Algorithm \ref{alg:alg1} 
on a simple 2D multivariate Gaussian distribution to develop insights on the optimal choice of $\rho(\vx)$. 
We set a Gaussian target $p(\vx)=\mathcal{N}(\vx;\bd{\mu}, \sigma I)$ with fixed $\bd{\mu}=(0, 0)$ and $\sigma=2.0$, 
and an auxiliary distribution  $\rho(\vx)=\mathcal{N}(\vx;\bd{\mu}_\rho, \sigma_\rho I)$ where we vary the value of $\vv \mu_\rho$ and $\sigma_\rho$ in Figure~\ref{fig:gfgauss}.  We can see that the best performance is achieved by matching the mean of  $\rho$ and the target $p$  ($\bd{\mu}_\rho = \bd{\mu}$),
and making $\sigma_\rho$ slightly larger than the variance $\sigma$ of $p$ (e.g., 
$\log10(\sigma_\rho/\sigma)\approx 0.5$ or 
$\sigma_\rho \approx 3 \sigma$). 
(d) uses the same setting with $||\vv\mu_\rho-\bd{\mu}||^2=8$, but also adds the result of exact Monte Carlo sampling, gradient-based SVGD, and importance sampling (IS) whose proposal is $\rho$, the same as the auxiliary distribution used by GF-SVGD shown in the red curve. We use $n=100$ particles in  this plot.

The performance is evaluated based on MMD between GF-SVGD particles and the exact samples from $p(\vx)$ (Figure~\ref{fig:gfgauss}(a)), 
and the mean square error (MSE) of estimating $\vv\mu$ and $\sigma$ (Figure~\ref{fig:gfgauss}(b)-(c)).

Figure~\ref{fig:gfgauss} suggests a smaller difference in mean $\vv \mu_\rho$ and $\vv \mu$ generally gives better results,
but the equal variance $\sigma_\rho = \sigma$ does not achieve the best performance. 
Instead, it seems that $\sigma_\rho \approx 3 \sigma$ gives the best result in this particular case. 
This suggests by choosing $\rho$ to be a proper distribution that well covers the probability mass of $p$, it is possible to even outperform the gradient-based SVGD which uses $\rho=p$. 

Interestingly, even when we take $\sigma_\rho = \infty$, corresponding to the simple update in \eqref{invp} with $\rho=1$, 
the algorithm still performs reasonably well (although not optimally) in terms of MMD and mean estimation (Figure~\ref{fig:gfgauss}(a)-(b)). 
It does perform worse on the variance estimation (Figure~\ref{fig:gfgauss}(c)), 
and we observe that this seems to be because 
the repulsive force domains when $\sigma_\rho$ is large (e.g., when $\sigma_\rho=\infty$, only the repulsive term is left as shown in \eqref{invp}), and it causes the particles to be overly diverse, yielding an over-estimation of the variance. 
This is interesting because we have found that the standard SVGD with RBF kernel tends to underestimate the variance, 
and a hybrid of them may be developed to give a more calibrated variance estimation. 

In Figure \ref{fig:gfgauss}(d), we 
add additional comparisons with exact Monte Carlo (MC) which directly draws sample from $p$, and standard importance sampling (IS) with $\rho$ as proposal. 
We find that GF-SVGD provides much better results than the standard IS strategy with any $\rho$. 
In addition, GF-SVGD can even outperform the exact MC and the standard SVGD when auxiliary distribution $\rho$ is chosen properly (e.g., $\sigma_\rho\approx 3\sigma$). It is interesting to see with proper choice of auxiliary distribution $\rho$ ($\sigma_\rho\approx 3\sigma$), GF-SVGD can outperform SVGD in terms of the sample quality.

\subsection{Comparing GF-SVGD with IS and SVGD}
\begin{figure*}[h]
\centering
\scalebox{.99}{
\setlength{\tabcolsep}{0em}
\begin{tabular}{cccc}
\hspace{-.3cm} \includegraphics[height=0.19\textwidth]{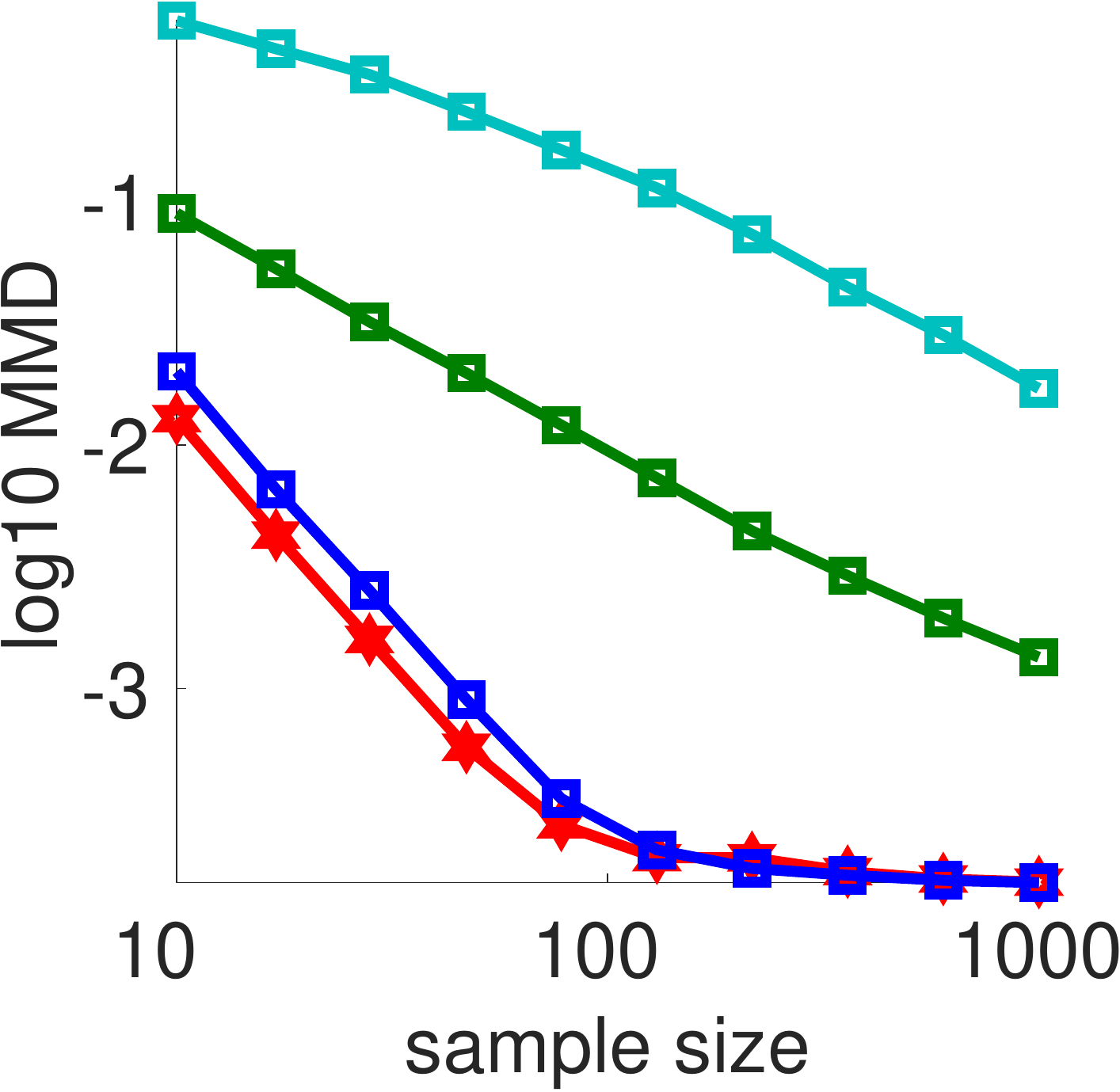}&
\includegraphics[height=0.19\textwidth]{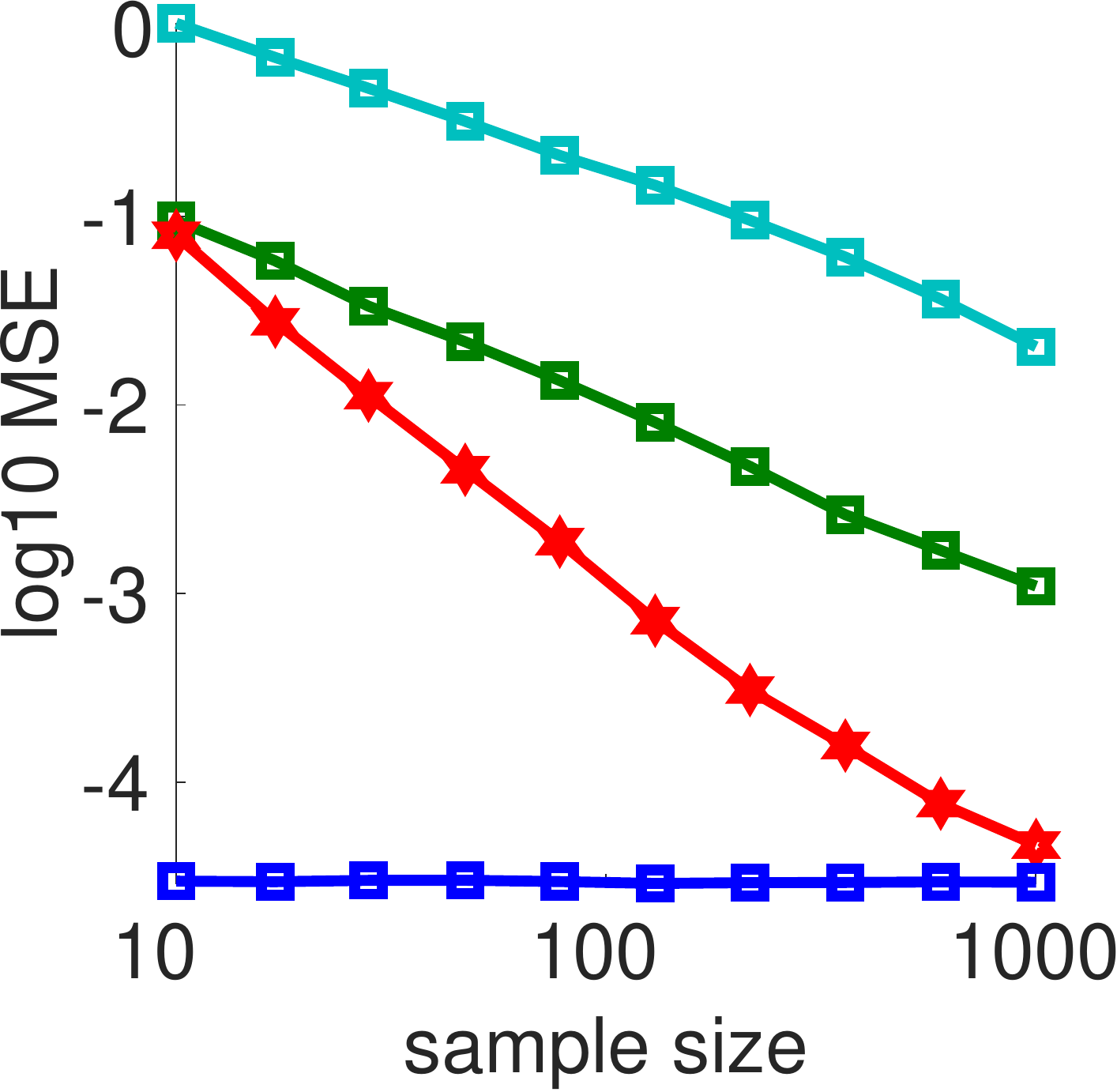} &
\includegraphics[height=0.19\textwidth]{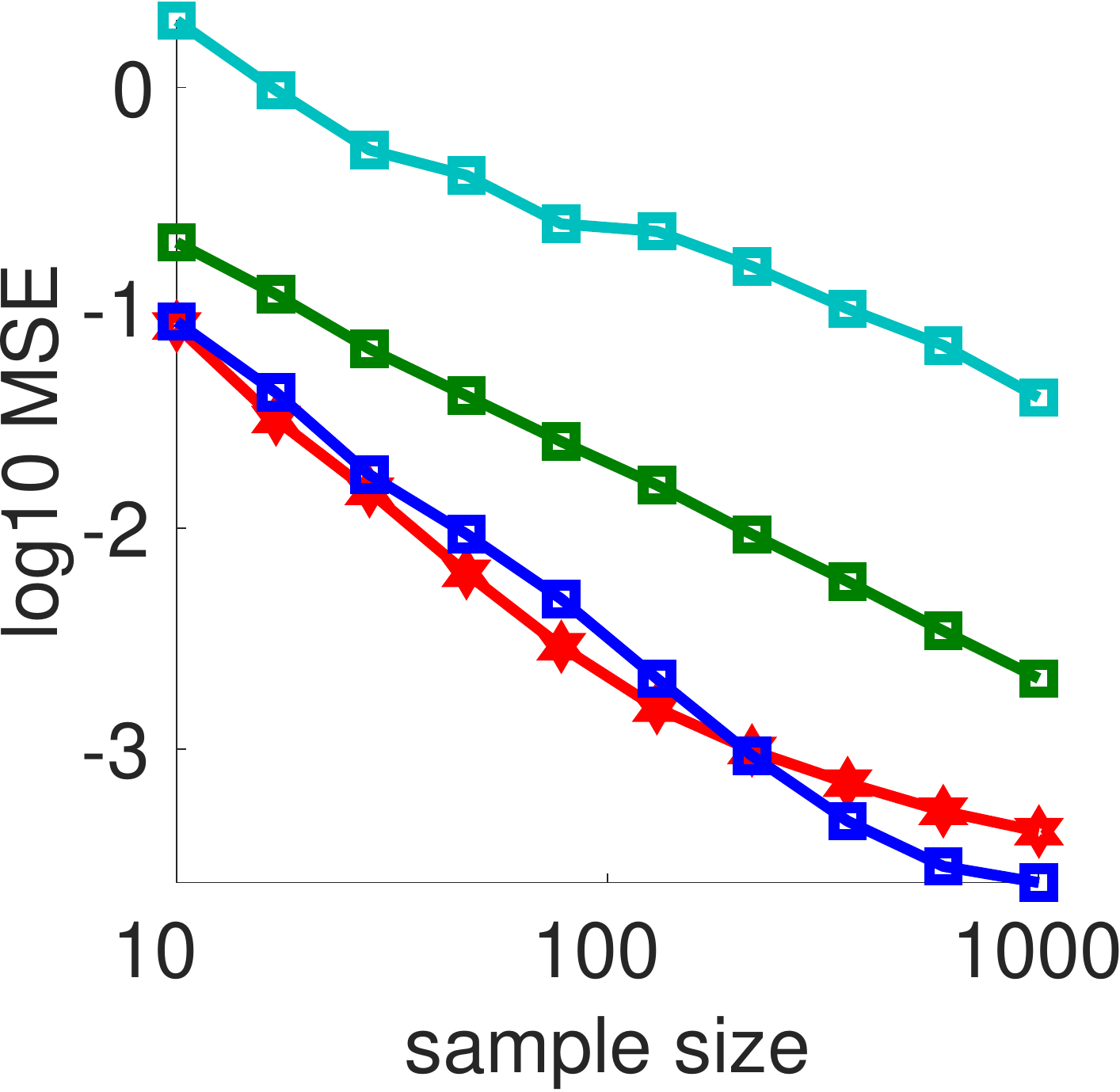} &
\includegraphics[height=0.19\textwidth]{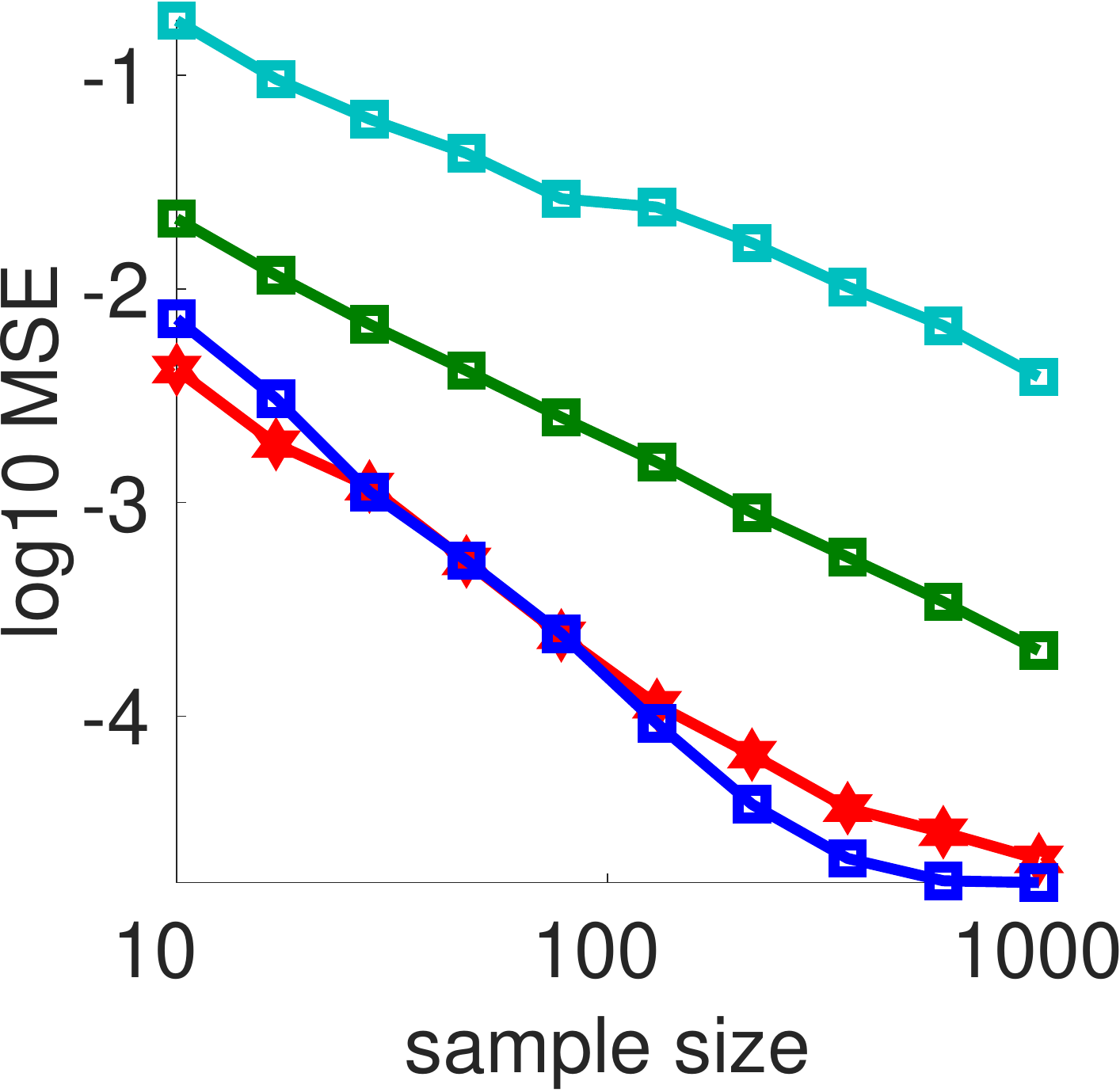} 
\raisebox{2em}{ \includegraphics[height=0.1\textwidth]{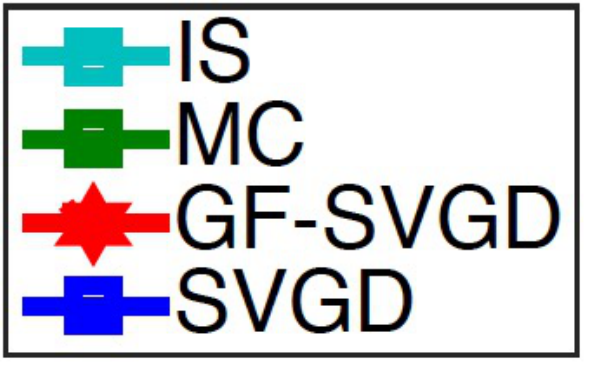}}\\
{\small (a) MMD}&  {\small (b) $E[\vx]$} & {\small (c) $E[\vx^2]$} &  {\small (d) $E[\cos(\nu\vx+c)]$}  \\
\end{tabular}
}
\caption[Performance comparison between GF-SVGD and SVGD on 2D Gauss distribution]{2D multivariate Gauss distribution. $p(\vx)=\mathcal{N}(\bd{\mu}, \sigma*I)$, $b(\vx)=\mathcal{N}(\bd{\mu}_0, \sigma_0*I)$ and $q(\vx)=\mathcal{N}(\bd{\mu}_q, \sigma_q*I)$. $\bd{\mu}=(0, 0)$and $\bd{\mu}_0=(-2, -2)$.  $\bd{\mu}_q=(-6, -6).$ Fix $\sigma=\sigma_0=\sigma_q=2.0$. Monte Carlo (MC) method means samples are directly drawn from $p$ and IS means samples are from $p_0$ and applies importance sampling to calculate the corresponding values. The initial particles for GF-SVGD and SVGD are drawn from $q$. We use $T=2000$ for GF-SVGD and SVGD. (a) shows MMD w.r.t. the iterations implemented. (b)-(d) shows MSE for estimating $\E_{p}[h(\vx)],$ where $h(\bd{x})=x_j,~x_j^2,~\cos(wx_j+c)$ with $\nu\sim \normal(0,1)$ and $c\in \mathrm{Uniform}(0,1)$ for $j=1, 2.$}
\label{Append:sam}
\end{figure*}

\begin{figure*}[h]
\centering
\scalebox{.99}{
\setlength{\tabcolsep}{0em}
\begin{tabular}{cccc}
\hspace{-.3cm} \includegraphics[height=0.19\textwidth]{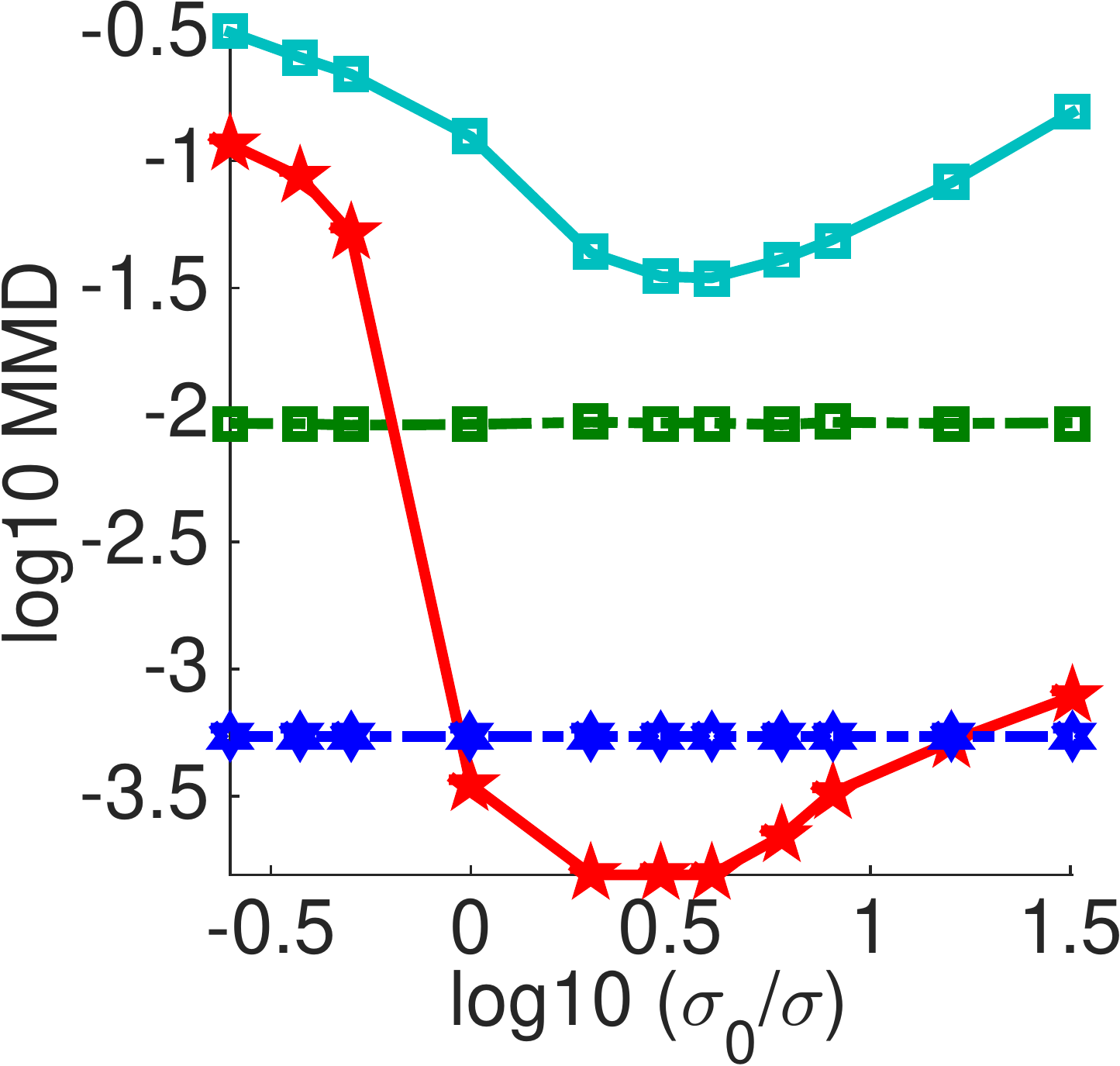}&
\includegraphics[height=0.19\textwidth]{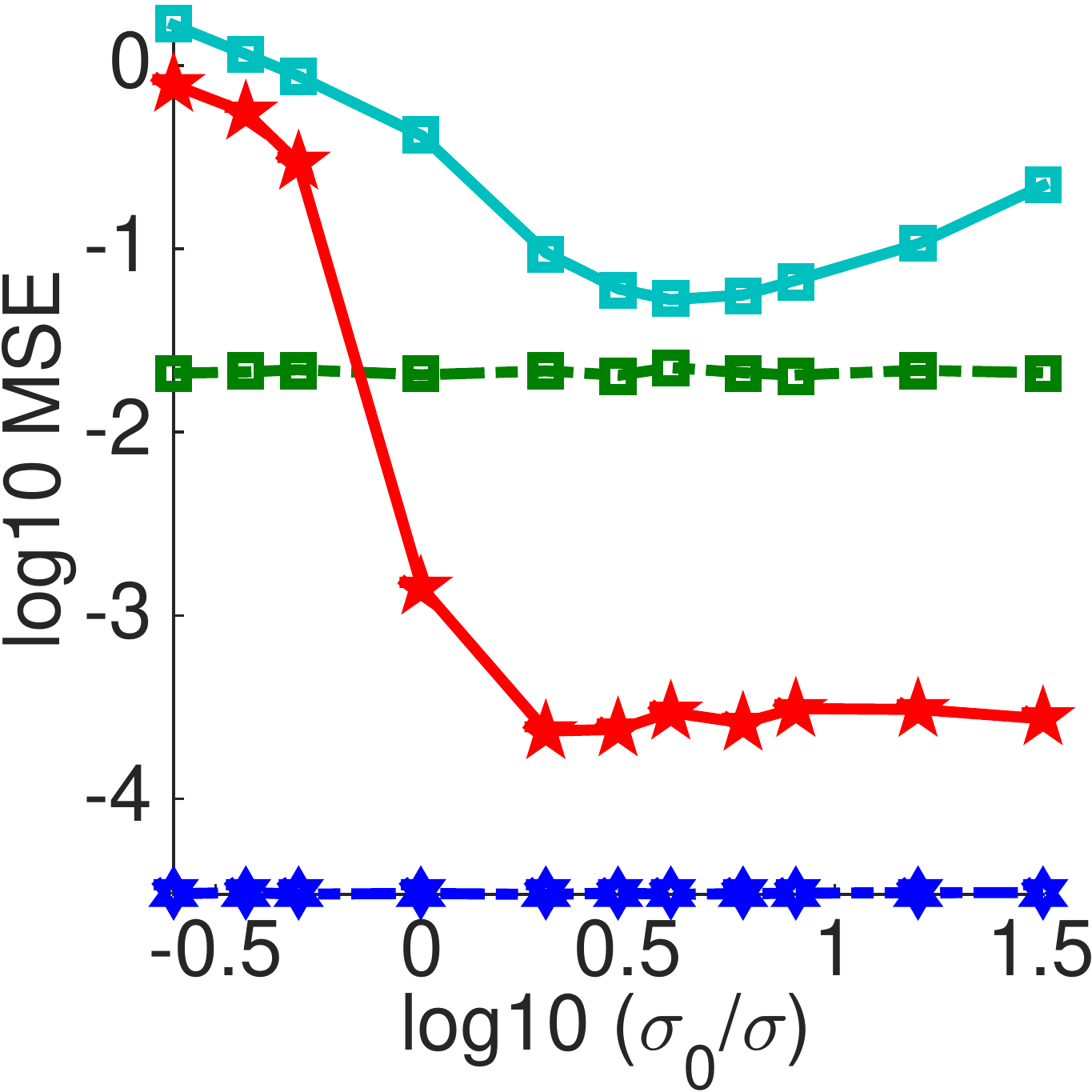} &
\includegraphics[height=0.19\textwidth]{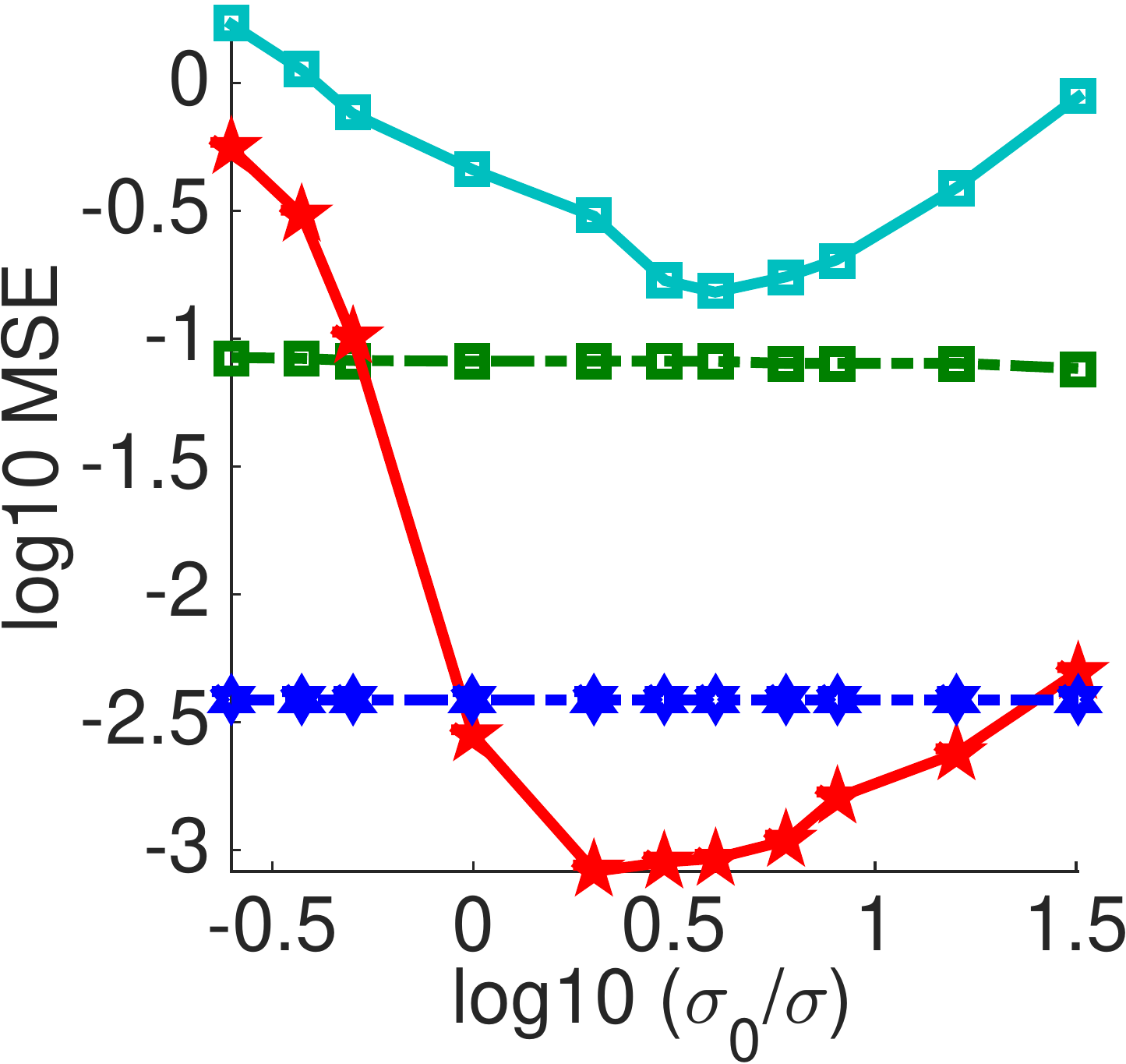} &
\includegraphics[height=0.19\textwidth]{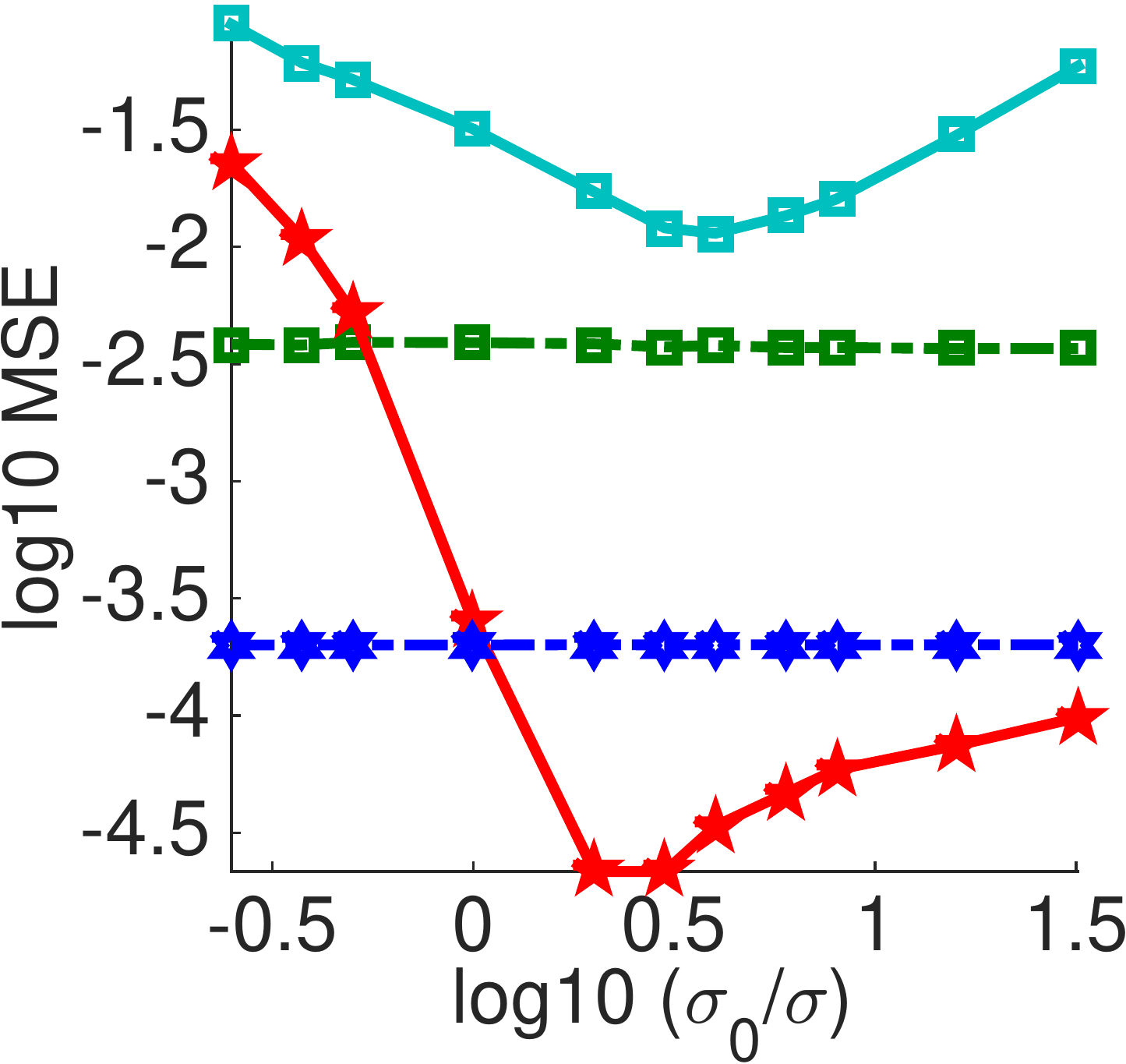} 
\raisebox{2em}{ \includegraphics[height=0.1\textwidth]{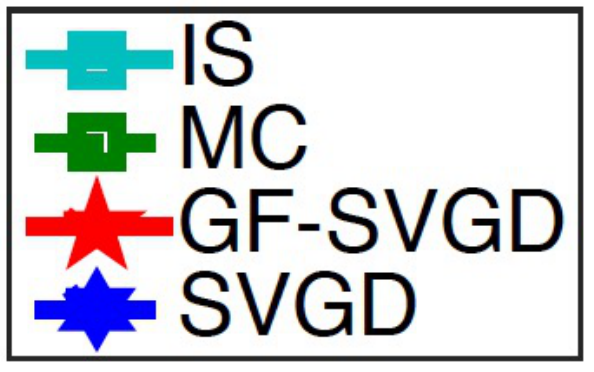}}\\
{\small (a) MMD}&  {\small (b) $E[\vx]$} & {\small (c) $E[\vx^2]$} &  {\small (d) $E[\cos(\nu\vx+c)]$}  \\
\end{tabular}
}
\caption[More empirical investigation of the choice of the surrogate distribution in GF-SVGD on 2D Gaussian distribution]{More empirical investigation of the choice of the surrogate distribution in GF-SVGD on 2D multivariate Gaussian distribution. $p(\vx)=\mathcal{N}(\bd{\mu}, \sigma*I)$, $b(\vx)=\mathcal{N}(\bd{\mu}_0, \sigma_0*I)$ and $q(\vx)=\mathcal{N}(\bd{\mu}_q, \sigma_q*I)$. $\bd{\mu}=(0, 0)$, $\bd{\mu}_0=(-2, -2)$ and $\bd{\mu}_q=(-6, -6).$ Fix $\sigma=\sigma_q=2.0$. Change $\sigma_0.$ The number of particles for all methods is 100. The initial particles for both SVGD and GF-SVGD are drawn from distribution $q$. We use $T=2000$ for both GF-SVGD and SVGD. (a) shows MMD w.r.t. the iterations implemented. (b)-(d) shows MSE for estimating $\E_{p}[h(\vx)],$ where $h(\bd{x})=x_j,~x_j^2,~\cos(wx_j+c)$ with $\nu\sim \normal(0,1)$ and $c\in \mathrm{Uniform}(0,1)$ for $j=1, 2.$}
\label{Append:sigma}
\end{figure*}

In the following, we compare our GF-SVGD with standard IS and vanilla SVGD on 2-dimensional multivariate Gauss in Fig.~\ref{Append:sam} and Fig.~\ref{Append:sigma}. The target distribution is $p(\vx)=\mathcal{N}(\vx; \bd{\mu}, \sigma*I)$, the proposal distribution is $\rho(\vx)=\mathcal{N}(\vx; \bd{\mu}_0, \sigma_0*I)$ and the initial particles of GF-SVGD and SVGD are drawn from $q(\vx)=\mathcal{N}(\vx; \bd{\mu}_q, \sigma_q*I)$ for experiments in Fig.~\ref{Append:sam} and Fig.~\ref{Append:sigma}.  We need to calculate the MMD between importance sampler and the target $p.$ We illustrate it in the following. As MMD between $(p_0, p)$ is
$$\mathrm{MMD}(p_0, p)=\E_{\vx\sim p_0, \vx'\sim p_0}[k(\vx, \vx')] -2\E_{\vx\sim p_0, \vy\sim p}[k(\vx, \vy)]+\E_{\vy\sim p, \vy'\sim p}[k(\vy, \vy')]$$ 
When we have samples from $q(\vx)$, we want to calculate MMD between $p_0(\vx)$ and $p(\vx)$. We can use samples from $q(\vx)$ and derive a importance weighted MMD,
\begin{align*}
\mathrm{MMD}(p_0, p) &=\E_{\vx\sim q, \vx'\sim q}[\frac{p_0(\vx)}{q(\vx)}k(\vx, \vx') \frac{p_0(\vx')}{q(\vx')}] \\
&-2\E_{\vx\sim q, \vy\sim p}[\frac{p_0(\vx)}{q(\vx)} k(\vx, \vy)]+\E_{\vy\sim p, \vy'\sim p}[k(\vy, \vy')].    
\end{align*}

Let ${w(\vx_i)=p_0(\vx_i)/p(\vx_i)}, \{\vx_i\}\sim q(\vx),$ $\{\vy_j\}_{j=1}^M\sim p(\vx)$ and $\hat{w}(\vx_i)=\frac{w(\vx_i)}{\sum_i w(\vx_i)},$ then the importance weighted MMD between $p_0(\vx)$ and $p(\vx)$ is calculated as follows,
\begin{align}
\label{ImpWeig:MMD}
\wt{\mathrm{MMD}}(p_0, p)\approx & \sum_{i, j}\hat{w}(
\vx_i) k(\vx_i, \vx_j)\hat{w}(\vx_j) \\ \notag
& - \frac{2}{M} \sum_{j=1}^M \sum_i  \hat{w}(\vx_i) k(\vx_i, \vy_j)+\frac{1}{M^2}\sum_{i, j} k(\vy_i, \vy_j).\\ \notag
\end{align}

As we can see from Fig.~\ref{Append:sam}, under the setting $\|\bd{\mu}-\bd{\mu}_0\|=2$ and $\sigma_0=\sigma=2.0$, GF-SVGD performs much better than IS when IS uses $b$ as importance sampler, the iterations of our GF-SVGD   progressively refine the importance proposal~\citep{han2017stein}. It is interesting to observe that our GF-SVGD has almost the same performace as vanilla SVGD except on estimating $\E[\vx]$. The interesting fact that SVGD has very low mean square error (MSE) of estimating $\E[\vx]$ when the target $p$ is Gaussian distribution deserves further investigation. 

In Fig. ~\ref{Append:sigma}, we fix $\|\bd{\mu}-\bd{\mu}_0\|=2$ and $\sigma=2.0$ and change $\sigma_0$. It is interesting to observe that the performance of GF-SVGD increases to certain threshold and then decreases as we increase $\sigma_0$. We also empirically check that even when $\sigma_0 = 10^6\cdot\sigma$, GF-SVGD still converges to the stationary fixed points but the MSEs of estimating $\E[\vx^2]$ and $\E[\cos(\nu\vx+c)]$ are somewhat large. We can see that with proper choice of $\sigma_0$, GF-SVGD can further improve the sample efficiency of SVGD in low dimensions in terms of MMD, $\E[\vx^2]$ and $\E[\cos(\nu\vx+c)]$ (SVGD already has much better accuracy than the exact Monte Carlo samples). It is also interesting to see that the performance of importance sampling (IS) with importance weight $w(\vx)=b(\vx)/p(\vx)$ also increases first when $\sigma_0$ increases to almost the same threshold. Then as when further increases $\sigma_0$, the performance of IS also decreases.

\subsection{Gaussian Mixture Models (GMM)}
We test GF-SVGD and AGF-SVGD on a 25-dimensional GMM with 10 randomly generated mixture components, 
$p(\vx)=\frac{1}{10}\sum_{i=1}^{10}\mathcal{N}(\vx;\bd{\mu}_i, I)$, 
with each element of $\vv \mu_i$ is drawn from $\mathrm{Uniform}([-1,1])$.  
The auxiliary distribution $\rho(\vx)$ is a multivariate Gaussian  $\rho(\vx) = \normal(\vx; \vv \mu_\rho, \sigma_\rho I)$, with fixed $\sigma_\rho = 4$ and each element of $\vv \mu_\rho$ drawn from $\mathrm{Uniform}([-1,1])$. 
For AGF-SVGD, we set
its initial distribution $p_0$ to equal the  
$\rho$ above in GF-SVGD. Let us now give the description of details for each experiments in Fig.~\ref{fig:gfgmm}. Fig.~\ref{fig:gfgmm}(a) shows the convergence of MMD with fixed sample size of $n=200$. 
Fig.~\ref{fig:gfgmm}(b) and Fig.~\ref{fig:gfgmm}(c) shows the mean square error with respect to the sample size when estimating the mean and variance using the particles returned by different algorithms at convergence. 
Fig.~\ref{fig:gfgmm}(d) shows the maximal mean discrepancy(MMD) between the particles of different methods and the true distribution $p$. In Fig.~\ref{fig:gfgmm}(b, c, d), 3000 iterations are used. For comparison, we also tested a gradient-free variant of annealed importance sampling (GF-AIS) \citep{neal2001annealed} with a transition probability constructed by Metropolis-adjusted Langevin dynamics, 
in which we use the same temperature scheme as our AGF-SVGD, and the same surrogate gradient $\nabla_{\vx} \log \rho_\ellt$ defined in \eqref{bt}.  For GF-AIS, the sample size $n$ represents the number of parallel chains, 
and the performance is evaluated using the weighted average of the particles at the final iteration with their importance weights given by AIS.

Figure~\ref{fig:gfgmm}(a) shows the convergence of MMD vs. the number of iterations 
of different algorithms with a particle size of $n = 200$, 
and Figure~\ref{fig:gfgmm}(b)-(d) shows the converged performance as the sample size $n$ varies. 
It is not surprising to see that that standard SVGD converges fastest since it uses the full gradient information of the target  $p$. 
A-SVGD converges slightly slower in the beginning but catches up later; this is because that it uses increasingly more gradient information from $p$. 
GF-SVGD performs significantly worse, which is expected because it does not leverage the gradient information. 
However, it is encouraging that annealed GF-SVGD, which also leverages no gradient information, performs much better than GF-SVGD, only slightly worse than the gradient-based SVGD and A-SVGD. 

%
GF-AIS returns a set of particles with importance weights, so we use weighted averages when evaluating the MMD and the mean/variance estimation. 
This version of GF-AIS is highly comparable to our AGF-SVGD since both of them use the same annealing scheme and surrogate gradient.  %
However, Fig.~\ref{fig:gfgmm} shows that AGF-SVGD still significantly outperforms GF-AIS. 
%

\begin{figure*}[ht]
\centering
\begin{tabular}{cccc}
\hspace{-.3cm} \includegraphics[height=0.22\textwidth]{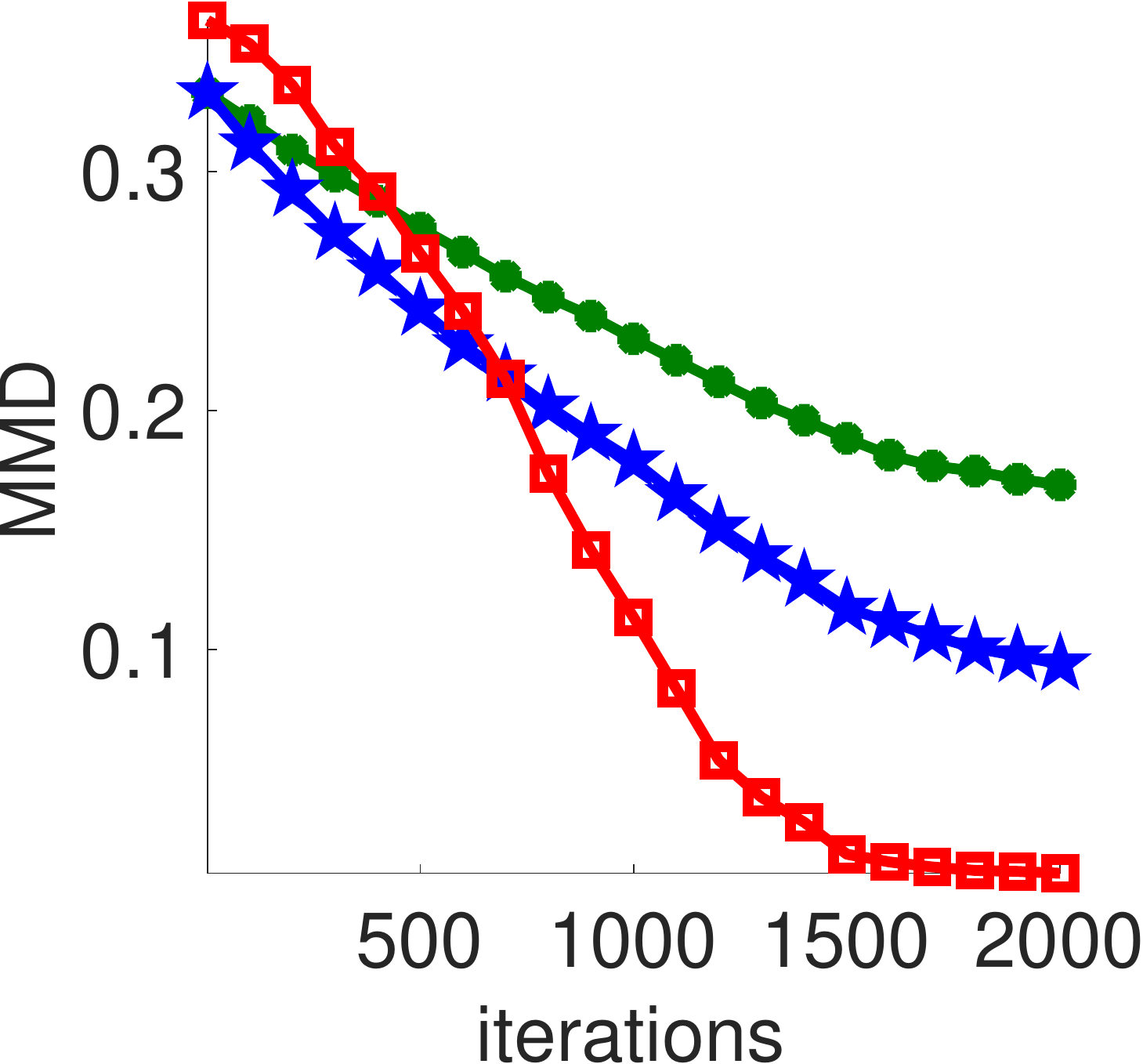}&
\includegraphics[height=0.22\textwidth]{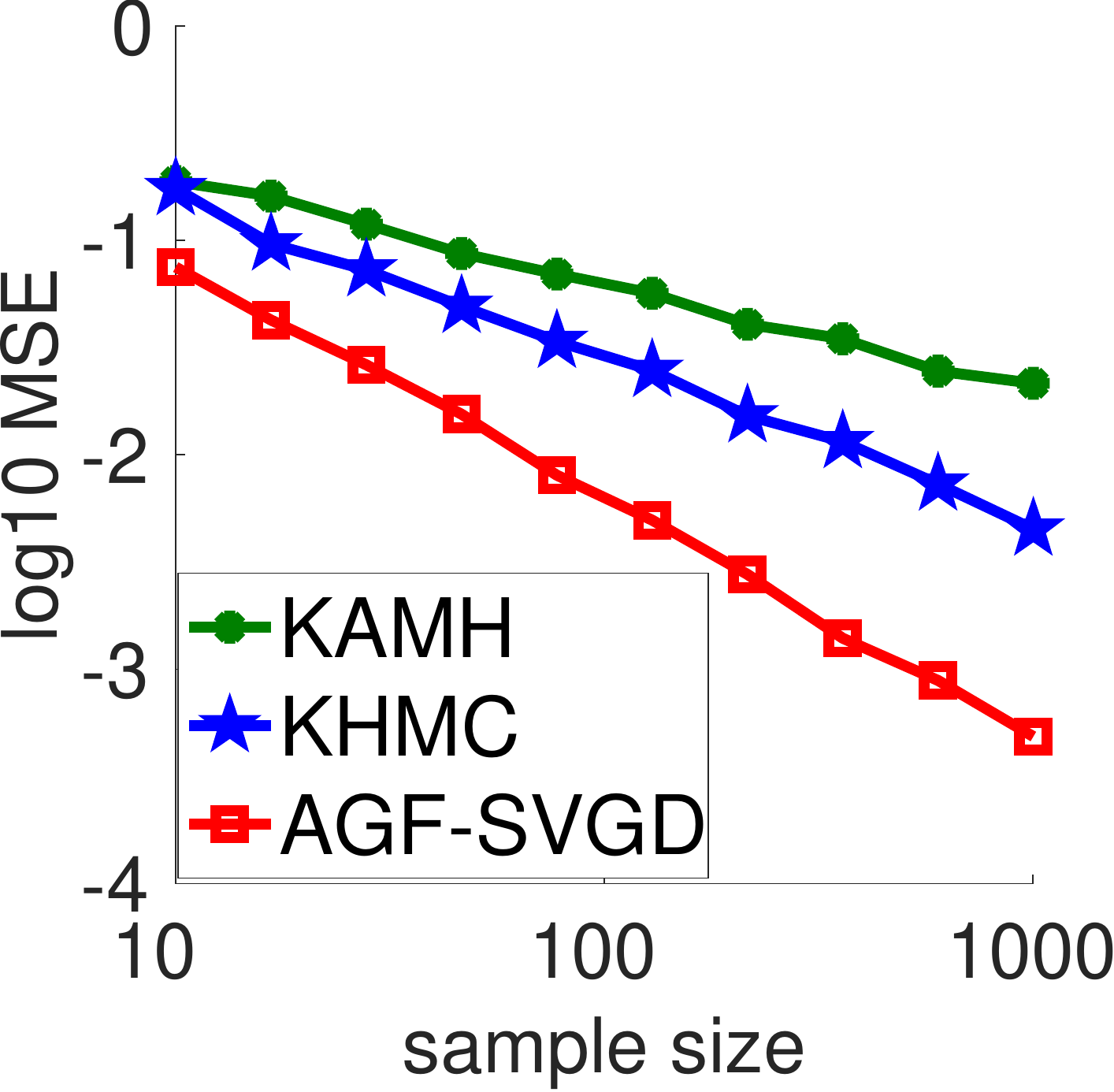} &
\includegraphics[height=0.22\textwidth]{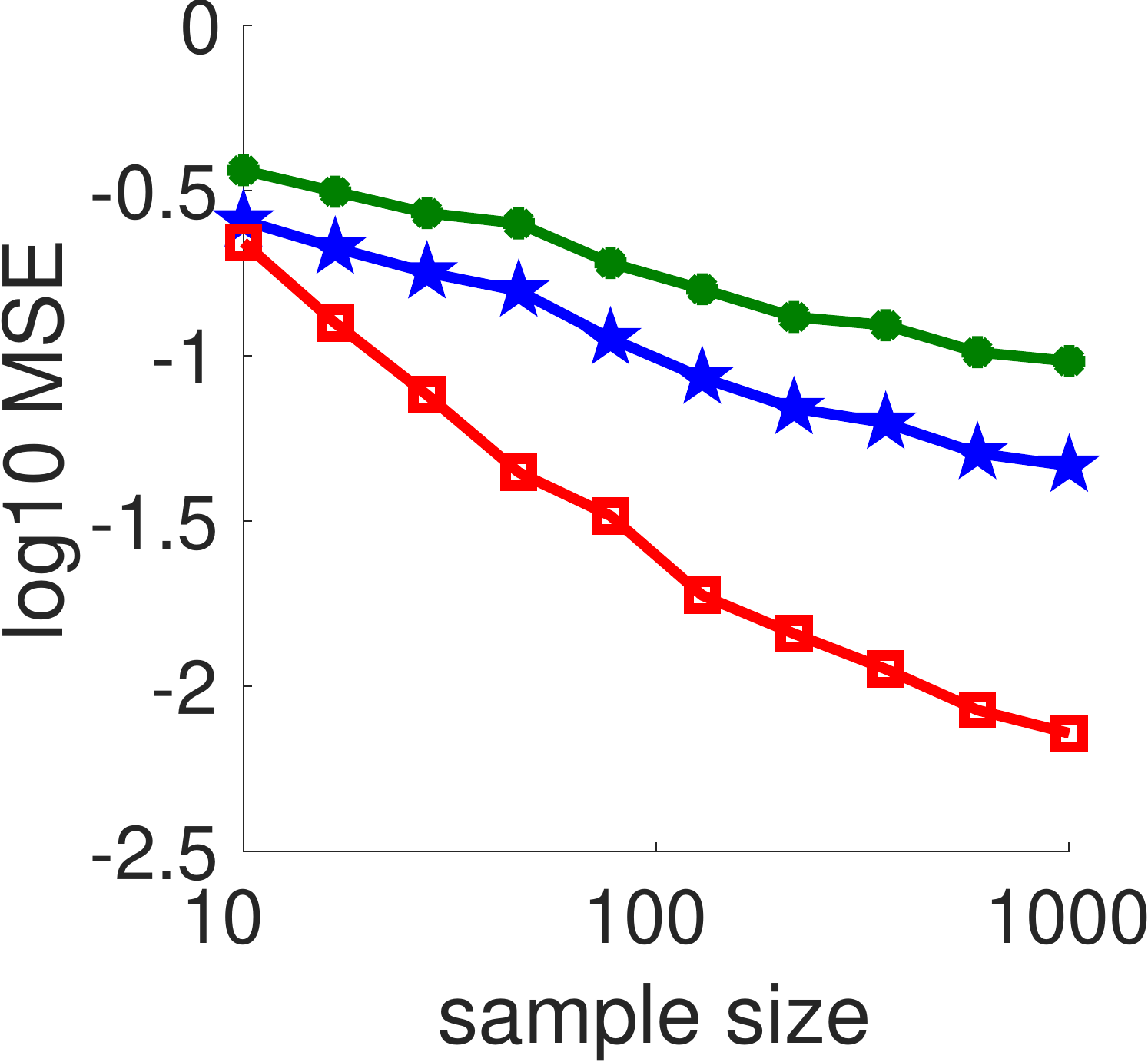} &
\includegraphics[height=0.22\textwidth]{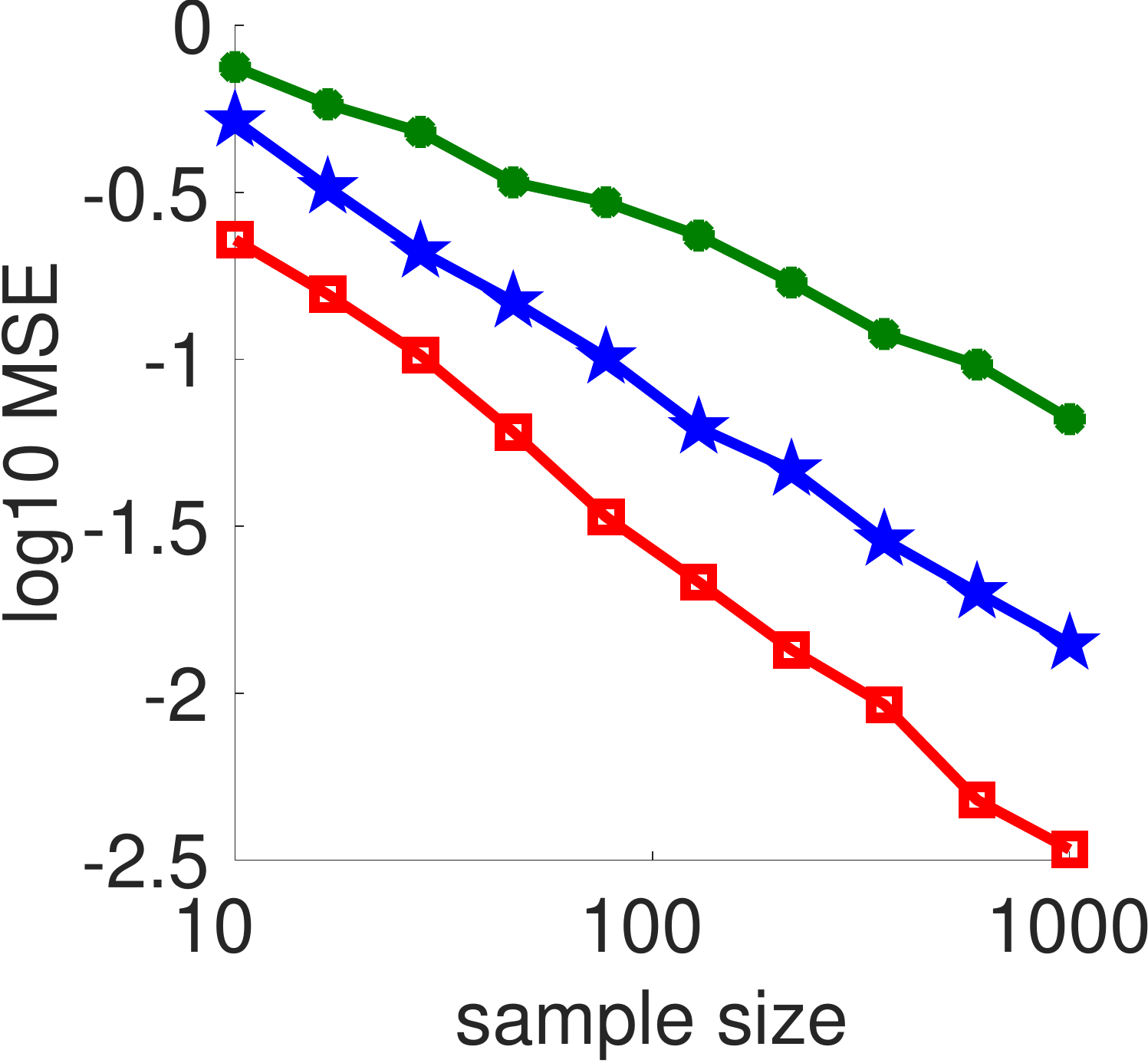}\\
 {\small (a) Convergence} & {\small (b) Mean} &  {\small (c) Variance} & {\small (d) MMD} \\
\end{tabular}
\caption[Performance of gradient-free sampling methods on Gauss-Bernoulli RBM]{\small Gauss-Bernoulli RBM with $d=20$ and $d'=10$. (a): the convergence of MMD with $n=100$ for all the algorithms. The evaluations of MMD of KAMH and KHMC in (a) starts from the burn-in steps of the typical MH algorithm.
(b)-(c): the MSE vs. sample size when estimating the mean and variance using the particles returned by different algorithms at 2000 iterations.  
(d): the MMD between the particles of different algorithms and the true distribution $p$ at 2000 iterations.
\label{fig:gfrbm}}

\end{figure*}

\subsection{Gauss-Bernoulli Restricted Boltzmann Machine}
We further compare AGF-SVGD with two recent baselines 
on Gauss-Bernoulli RBM, defined by 
\begin{equation}
p(\bd{x}) \propto \sum_{\vv h}\exp(\bd{x}^\top B\bd{h}+\vv c_1^\top\bd{x}+\vv c_2^\top\bd{h}-\frac{1}{2}\| \bd{x}\|_2^2),
\end{equation}
where $\vx\in \RR^d$ and $\vv h\in \{\pm1 \}^{d'}$ is a binary latent variable. 
By marginalizing the hidden variable $\vv h$, we can see that $p(\bd{x})$ is a special GMM with $2^{d'}$ components. In our experiments, we draw the parameters $\vv c_1$ and $\vv c_2$ from standard Gaussian and select each element of $B$  randomly from $\{\pm 0.5\}$ with equal probabilities. We set the dimension $d$ of $\vx$ to be 20 and the dimension $d'$ of $\vv h$ to be 10 so that $p(\vx)$ is a 20-dimensional GMM with $2^{10}$ components, for which it is still feasible to draw exact samples by brute-force for the purpose of evaluation. For AGF-SVGD, we set the initial distribution to be $p_0(\vx)=\mathcal{N}(\vx; \bd{\mu}, \sigma I)$, with  $\bd{\mu}$ drawn from $\mathrm{Uniform}([1, 2])$ and $\sigma =3.$

We compare our AGF-SVGD with two recent gradient-free methods: KAMH \citep{sejdinovic2014kernel} and KHMC \citep{strathmann2015gradient}. Both methods are advanced MCMC methods that adaptively improves the transition proposals based on kernel-based approximation from the history of Markov chains. The detailed description of each experiment is provided in the following. Fig.~\ref{fig:gfrbm}(a) shows the convergence of MMD with $n=100$ for all the algorithms. The evaluations of MMD of KAMH and KHMC in (a) starts from the burn-in steps of the typical MH algorithm.
Fig.~\ref{fig:gfrbm}(b) and Fig.~\ref{fig:gfrbm}(c) shows the mean square error (MSE) w.r.t the sample size when estimating the mean and variance using the particles returned by different algorithms at 2000 iterations.  
Fig.~\ref{fig:gfrbm}(d) shows the maximal mean discrepancy~(MMD) between the particles of different algorithms and the true distribution $p$ at 2000 iterations. 

For a fair comparison with SVGD, we run $n$ parallel chains of KAMH and KHMC and take the last samples of $n$ chains for estimation.  In addition, we find that both KAMH and KHMC require a relatively long burn-in phase before the adaptive proposal becomes useful. In our experiments, we use 10,000 burn-in steps for both KAMH and KHMC, and exclude the computation time of burn-in when comparing the convergence speed with GF-SVGD in Figure~\ref{fig:gfrbm}; this gives KAMH and KHMC much advantage for comparison, and the practical computation speed of KAMH and KHMC is much slower than our AGF-SVGD. From Figure~\ref{fig:gfrbm} (a), we can see that our AGF-SVGD converges fastest to the target $p$, even when we exclude the 10,000 burn-in steps in KAMH and KHMC. Fig.~\ref{fig:gfrbm} (b, c, d) shows that our AGF-SVGD performs the best in terms of the accuracy of estimating the mean, variance and MMD. 

\subsection{Gaussian Process Classification}
We apply our AGF-SVGD to sample hyper-parameters from marginal posteriors of Gaussian process (GP) binary classification. 
Consider a classification of predicting binary label $y \in \{\pm1\}$ from feature $\vv z$. 
We assume $y$ is generated by a latent Gaussian process $f(\vv z)$, $p(y|\vv z) = 1/(1+\exp(-y f(\vv z )))$ and 
$f$ is drawn from a GP prior $f\sim GP(0, k_{f, \bd{\theta}})$, where $k_{f,\vv \theta}$ is the GP kernel indexed by a hyperparameter $\vv \theta$. In particular, we assume 
$k_{f,\vv{\theta}}(\vv z, \vv z') = \exp(-\frac12||(\vv z-\vv z')./\exp(\vv \theta)||^2)$, where $./$ denotes the element-wise division and $\vv\theta$ is a vector of the same size as $\vv z$. 
Given a dataset $Y = \{y_i\}$ and $Z = \{\vv z_i\}$, we are interested in drawing samples from the posterior distribution $p(\vv\theta|Z,Y)$. Note that the joint posterior of $(\bd{\theta}, f)$ is 
$$
p(\vv{\theta}, f | Z,Y) =  p(Y | f, Z) p(f|\bd{\theta}) p(\bd{\theta}). 
$$
Since it is intractable to exactly calculate the marginal posterior of $\vv\theta$, we approximate it by 
\begin{equation}
\label{estimate}
\hat{p}(\vv{\theta} |Z,Y):= p(\vv\theta)  \frac{1}{m}\sum_{i=1}^{m}
\frac{p(Y| f^i, Z) p(f^i |\bd{\theta}) }{q(f^i|\bd{\theta})}, 
\end{equation} 
where $\{f^{i}\}_{i=1}^m$ is drawn from a proposal distribution $q(f\mid\bd{\theta})$, which is constructed by an expectation propagation-based approximation of $p(f|\vv\theta,Z,Y)$ following %
\citet{filippone2014pseudo}.

\begin{figure}[ht]
\centering
\begin{tabular}{cc}
\includegraphics[height=0.26\textwidth]{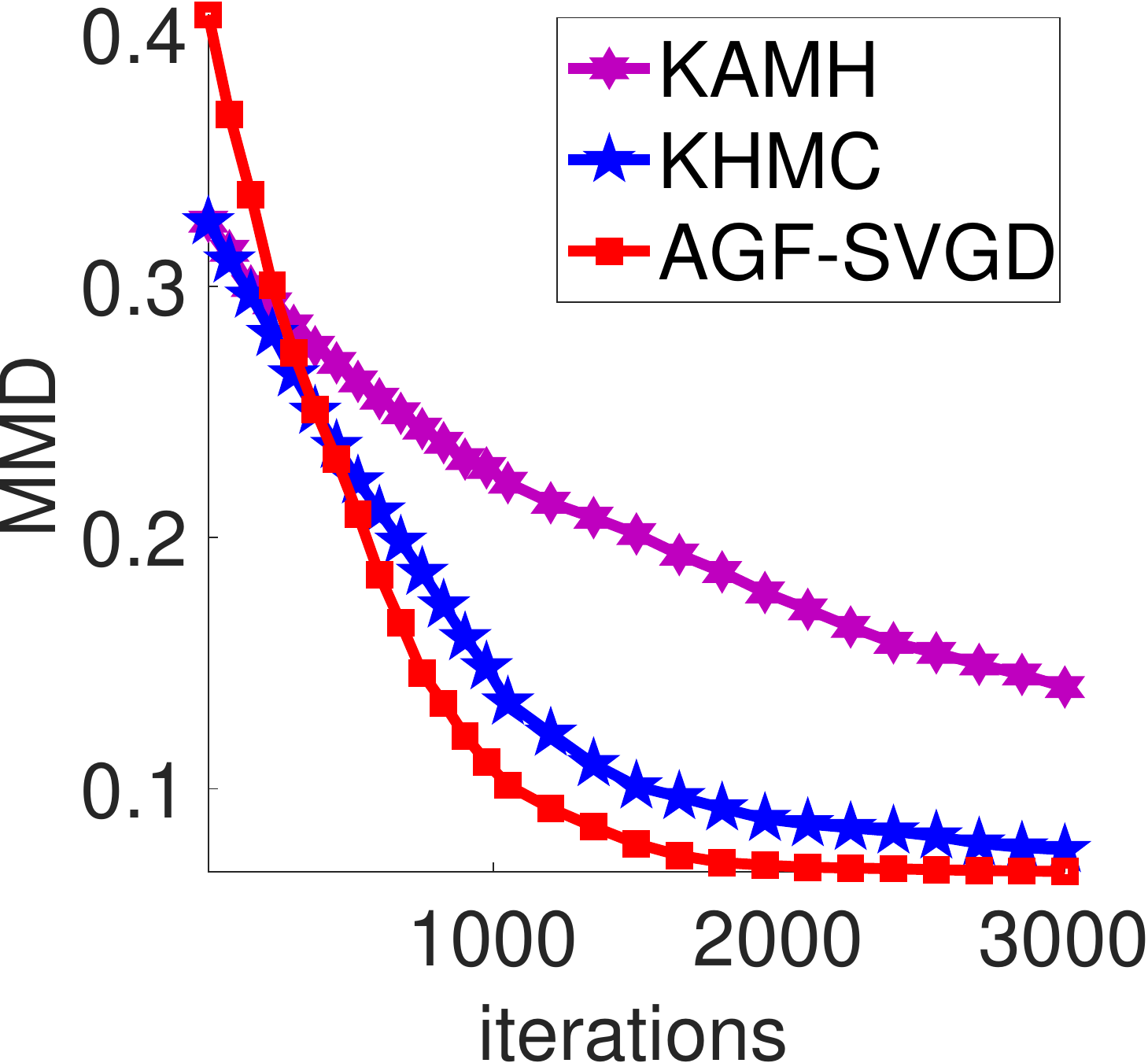} &
\includegraphics[height=0.26\textwidth]{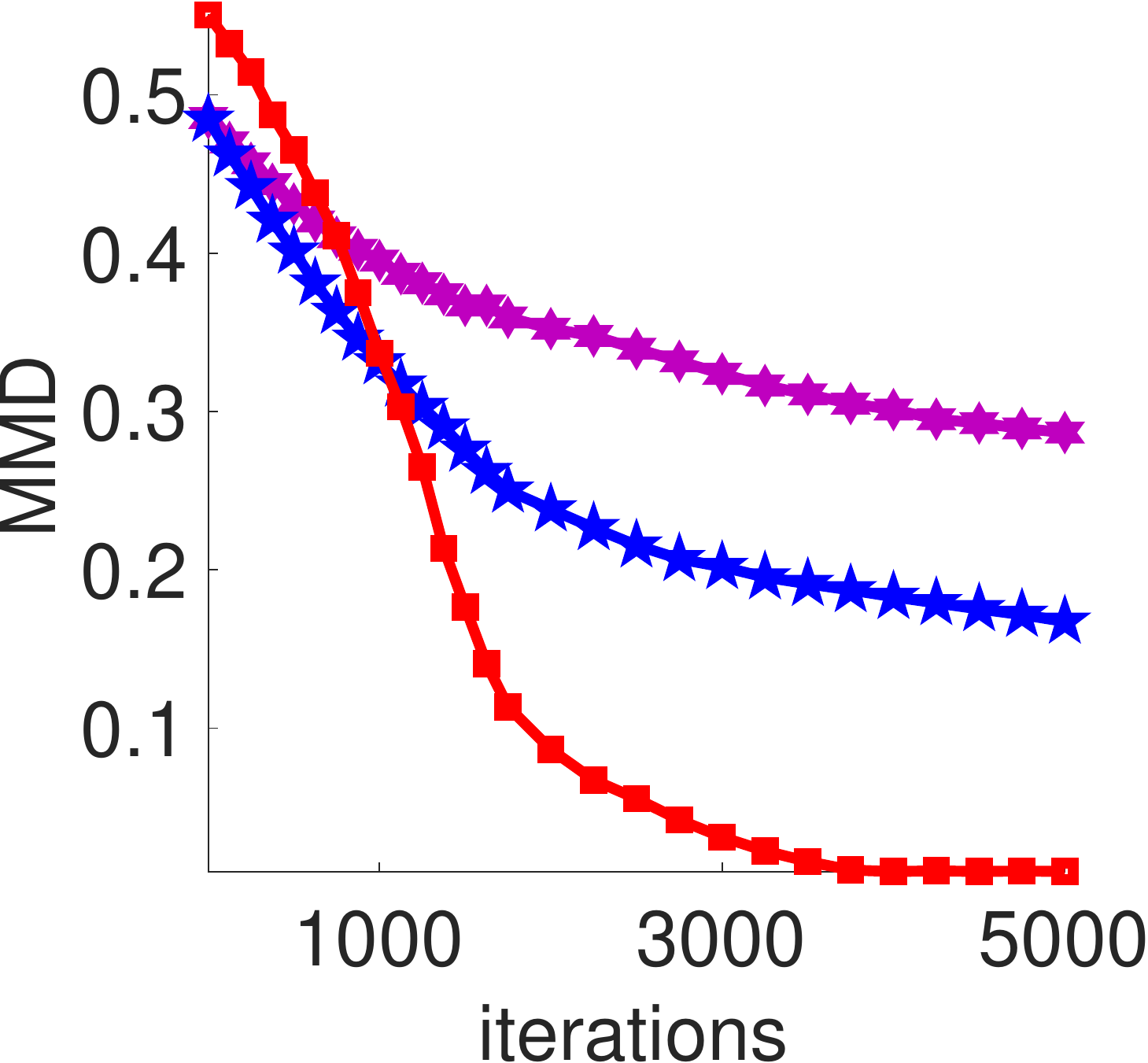} \\
{\small (a)  Glass dataset} &
{\small (b)  SUSY dataset} 
\end{tabular}
\caption[Performance of gradient-free sampling methods from the marginal posteriors on GP classification]{\small Sampling from the marginal posteriors on GP classification for Glass dataset (a) and SUSY dataset (b). We use a sample size of $n=200$ for all methods.}
\label{fig:real}
\end{figure}

We run multiple standard Metropolis-Hastings chains to obtain  ground truth samples from $p(\bd{\theta}\mid D)$, 
 following the procedures in section 5.1 of \citet{sejdinovic2014kernel} and Appendix D.3 of \citet{strathmann2015gradient}. 
We test the algorithms on Glass dataset and SUSY dataset in Figure~\ref{fig:real} from UCI repository \citep{asuncion2007uci} for which the dimension of $\bd{\theta}$ is $d = 9$ and $d=18$, respectively. 
We initialize our algorithm with draws from $p_0(\vx)=\mathcal{N}(\vx; \bd{\mu}, \sigma I)$ where $\sigma = 3$ and each element of $\bd{\mu}$ is drawn from $\mathrm{Uniform}([-1, 1])$. 
For KAMH and KHMC, 
we again run $n$ parallel chains and 
initialize them 
with an initial burn-in period of 6000 steps which {is \emph{not} taken into account in evaluation.}
Figure~\ref{fig:real} shows that AGF-SVGD again converges faster than KAMH and KHMC, even without the additional burn-in period.

\paragraph{Conclusions and Discussions}
We derive a gradient-free extension of Stein's identity and Stein discrepancy and propose a novel gradient-free sampling algorithm. The $\KL$ divergence between the iterated distribution and the target distribution is maximally decreased with a new kernel $\rho(\vx)/p(\vx)k(\vx, \vx')\rho(\vx')/p(\vx')$ in RKHS, which is in contrast with the decrease rate of $\KL$ divergence in SVGD with kernel $k(\vx, \vx')$. It is interesting to observe that the gradient-free SVGD performs even better than the original gradient-based SVGD in some experiments with the same number of iterations, where both methods have the same initialization of particles. We expect the reason is that the decrease rate of gradient-free SVGD is even larger than that of gradient-based SVGD in these settings. Future direction includes theoretical investigation of optimal choice of the auxiliary proposal with which we may leverage the gradient of the target to further improve the sample efficiency over the standard SVGD.  Our gradient-free sampling framework provides a powerful tool to perform statistical inference on the target distributions whose gradients are unavailable or intractable. The gradient-free kernelized Stein discrepancy we derive can be applied to perform the goodness-of-fit test~\citep{liu2016kernelized} and black-box importance sampling~\citep{liu2016black} when the gradients of the target distributions are unavailable or intractable.

\section{Gradient-Free Black-Box Importance Sampling}
In the last section of this chapter, we are going to introduce a new algorithm which basically equips any given set of particles $\{\vx_i\}_{i=1}^n$ with importance weights $\{u_i\}_{i=1}^n$ so that  
\begin{equation}
\label{def:bbis}
\sum_{i=1}^n u_i h(\bd{x}_i) \approx \E_{\vx\sim p} [h(\bd{x})], 
\end{equation}
for general test function $h(\bd{x})$. In the following, we first discuss the idea of black-box importance sampling~\citep{liu2016black}, which requires the gradient information of the target distribution $p(\vx).$ Then we introduce a gradient-free version of black-box importance sampling, motivated from gradient-free kernelized Stein discrepancy derived in Theorem~\ref{thm:gf:disc}. 

Let $k(\bd{x}, \bd{y})$ be the kernel of the reproducing kernel Hilbert space~(RKHS) $\mathcal{H}_d.$ Based on the Stein's identy, it is easy to know the following identity,
\begin{equation}
\label{iden:zero}
\E_{\bd{x}\sim p}[\kappa_p(\bd{x}, \bd{y})] = 0,~\text{for}~ \bd{y}\in\mathcal{X}   
\end{equation}
where the positive definite kernel $\kappa_p(\bd{x}, \bd{y})$ is defined as 
\begin{equation}
\begin{aligned}
\kappa_p (\bd{x},  \bd{y}) = & \bd{s}_p(\bd{x})^\top k(\bd{x},\bd{y})\bd{s}_p(\bd{y}) +\bd{s}_p(\bd{x})^\top \nabla_{\bd{y}}k(\bd{x},\bd{y}) \\
&+\bd{s}_p(\bd{y})^\top \nabla_{\bd{x}} k(\bd{x},\bd{y})+\nabla_{\bd{x}}\cdot(\nabla_{\bd{y}}k(\bd{x}, \bd{y})).
\end{aligned}
\end{equation}
Replace the expectation $\E_p[\cdot]$ in \eqref{iden:zero} with the expectation $\E_q[\cdot]$ of a different distribution $q,$ \eqref{iden:zero} will not be zero, which provides a discrepancy measure between $q(\vx)$ and $p(\vx),$
\begin{equation}
\label{ksd}
\mathbb{S}(q, p) =\E_{\bd{x},\bd{y}\sim q}[\kappa_p(\bd{x}, \bd{y})]\ge 0,
\end{equation}
where the square of KSD $\mathbb{S}(q, p)=0$ if and only if $q(\vx)$ equals $p(\vx).$

\paragraph{Black-Box Importance sampling (BBIS)}
Let $\{\bd{x}_i\}_{i=1}^n$ be a set of points in $\R^d$ and we want to find a set of weights $\{u_i\}_{i=1}^n$, $u_i \in \R$, such that the weighted sample $\{\bd{x}_i, u_i\}_{i=1}^n$ closely approximates the target distribution $p(\bd{x})$ in the sense that
\begin{equation}
\label{isproblem}
\sum_{i=1}^n u_i h(\bd{x}_i) \approx \E_{\vx\sim p} [h(\bd{x})], 
\end{equation}
for general test function $h(\bd{x}).$
For this purpose, we define an empirical version of the KSD in \eqref{ksd} to measure the discrepancy between $\{\bd{x}_i, u_i\}_{i=1}^n$ and $p(\bd{x})$,
\begin{equation}
\label{def:bbisksd}
\mathbb{S}(\{\bd{x}_i, u_i\}, ~ p) = \sum_{i,j=1}^n u_i u_j \kappa_p(\bd{x}_i, \bd{x}_j) = \vv u ^\top \bd{K}_p \vv u,    
\end{equation}
where $\bd{K}_p = \{ \kappa_p(\bd{x}_i, \bd{x}_j) \}_{i,j=1}^n$ and $\vv u = \{ u_i \}_{i=1}^n$. BBIS solves the problem \eqref{isproblem} by minimizing the discrepancy $\mathbb{S}(\{\bd{x}_i, u_i\}, ~ p)$,
\begin{align}
\label{equ:vw}
\begin{split}
\hat {\vv u} =  \argmin_{\vv u}\bigg\{ \vv u^\top \bd{K}_p \vv u,  ~~  s.t.~~ \sum_{i=1}^n u_i = 1, ~~~ u_i \geq 0\bigg\}.
\end{split}
\end{align}
Solving the quadratic programming \eqref{equ:vw} does not require how the particles $\{\vx_i\}_{i=1}^n$ are generated, and hence gives a black-box importance sampling. 
\paragraph{Gradient-Free BBIS}
Consider the target distribution $$p(\bd{x})\propto \rho(\bd{x})/w(\vx), \quad \bd{x}\in \mathcal{X},$$
where $\rho(\bd{x})$ is surrogate distribution with tractable gradient information and $w(\bd{x})$ is a weight function whose gradient is unavailable or intractable, $w(\bd{x})=\rho(\bd{x})/p(\bd{x}).$ In the following, we are going to derive the gradient-free KSD by directly replacing the kernel $k(\bd{x},\bd{y})$ with a distribution-informed kernel $\widetilde{k}(\bd{x}, \bd{y})=w(\vx)k(\bd{x}, \bd{y})w(\bd{y}).$  

\begin{thm}
\label{gf-ksd:bbis}
Replace the kernel $k(\bd{x},\bd{y})$ with the kernel $\widetilde{k}(\bd{x}, \bd{y})=w(\vx)k(\bd{x}, \bd{y})w(\bd{y})$ in RKHS $\mathcal{H}_d,$ the KSD can be rewritten as follows,  
\begin{equation}
\label{bbis:ksd}
\wt{\mathbb{S}}(q, p) =\E_{\bd{x},\bd{y}\sim q}[\wt{\kappa}_p(\bd{x}, \bd{y})]\ge 0,
\end{equation}
where $\wt{\kappa}_{p}(\bd{x}, \bd{y})$ satisfies $
\wt{\kappa}_{p}(\bd{x}, \bd{y}) =w(\bd{x})\kappa_{\rho} (\bd{x},  \bd{y})w(\bd{y}),$ \\
\begin{equation*}
\begin{aligned}
\kappa_{\rho} (\bd{x},  \bd{y}) = & \bd{s}_{\rho}(\bd{x})^\top k(\bd{x},\bd{y})\bd{s}_{\rho}(\bd{y}) +\bd{s}_{\rho}(\bd{x})^\top \nabla_{\bd{y}}k(\bd{x},\bd{y}) \\
&+\bd{s}_{\rho}(\bd{y})^\top \nabla_{\bd{x}} k(\bd{x},\bd{y})+\nabla_{\bd{x}}\cdot(\nabla_{\bd{y}}k(\bd{x}, \bd{y})),
\end{aligned}
\end{equation*}
which does not require the gradient of the target distribution $p(\vx).$
\end{thm}

Motivated from \cite{liu2016black}, we can equip any given set of particles $\{\vx_i\}_{i=1}^n$ with importance weights $\{u_i\}_{i=1}^n$ so that \eqref{def:bbis} gives a good approximation. $\{u_i\}_{i=1}^n$ can be evaluated by solving a quadratic programming, which does not require the gradient information of the target distribution $p(\vx),$
\begin{align}
\label{equ:gfbbis:pro}
\begin{split}
\hat {\vv u} =  \argmin_{\vv u}\bigg\{ \vv u^\top \wt{\bd{K}}_p \vv u,  ~~  s.t.~~ \sum_{i=1}^n u_i = 1, ~~~ u_i \geq 0\bigg\},
\end{split}
\end{align}
where $\wt{\bd{K}}_p = \{ w(\vx_i)\wt{\kappa}_{\rho}(\bd{x}_i, \bd{x}_j)w(\vx_j) \}_{i,j=1}^n$ and $\vv u = \{ u_i \}_{i=1}^n.$

\begin{lem}
\label{lem:bbis:appr}
The approximation error \eqref{def:bbis} can be bounded by 
\begin{equation}
|\sum_{i=1}^n u_i h(\vx_i) - \E_p[h(\vx)]| \le C_h\sqrt{\wt{\mathbb{S}}(\{\vx_i, u_i\}, p)}    
\end{equation}
where $\wt{\mathbb{S}}(\{\bd{x}_i, u_i\}, p) = \sum_{i,j=1}^n u_i w(\vx_i) \kappa_{\rho}(\bd{x}_i,\bd{x}_j)w(\vx_j) u_j$ and $C_h=\|h-\E_p[h]\|_{\mathcal{H}_d},$ which depends on $h(\vx)$ and $p(\vx)$ but not on $\{\bd{x}_i, u_i\}_{i=1}^n$.  
\end{lem}

Lemma~\ref{lem:bbis:appr} provides the approximation error of our proposed method~\eqref{equ:gfbbis:pro}. Under some mild conditions, the approximation error satisfies $|\sum_{i=1}^n u_i h(\vx_i) - \E_p[h(\vx)]|=O(n^{-\frac{1}{2}}),$ which indicates that the mean square error converges with rate $n^{-\frac{1}{2}}.$ In practice, we can first run GF-SVGD to get a set of particles $\{\vx_i\}_{i=1}^n$ to approximate the target distribution $p(\vx)$ and then run GF-BBIS to further refine the particles $\{\vx_i\}_{i=1}^n$ with the importance weight $\{u_i\}_{i=1}^n$ so that $\{\bd{x}_i, u_i\}_{i=1}^n$ provides a better approximation of the integration \eqref{def:bbis}. 

\paragraph{Alpha-Weighted KSD} In practice, it is also possible to incorporate the gradient information of the target distributions into the kernel of RKHS to improve the performance of black-box importance sampling. Instead of applying kernel $w(\vx)k(\vx, \vx')w(\vx'),$ where $w(\vx)=\rho(\vx)/p(\vx),$ it might be beneficial to use a new kernel
\begin{equation}
\wt{k}(\vx, \vx')=p(\vx)^{\alpha}k(\vx, \vx')  p(\vx')^{\alpha}. 
\end{equation}
Based on this new kerenl, we can a new form of kernelized Stein discrepancy between $q$ and $p$ (alpha-weighted KSD) , whose square is given as $\wt{\mathcal{S}}(q, p) = E_{\vx, \vx'\sim q}[\wt{\kappa}_p(\vx, \vx')],$
where $\wt{\kappa}_p(\vx, \vx')$ is defined as 
\begin{align}
\wt{\kappa}_p (\bd{x},  \bd{x}') & = p(\bd{x})^\alpha p(\bd{x}')^\alpha \big[(\alpha + 1)^2 \bd{s}_p(\bd{x})^\top k(\bd{x},\bd{x}')\bd{s}_p(\bd{y})+(\alpha + 1) \bd{s}_p(\bd{x})^\top \nabla_{\bd{y}} k(\bd{x},\bd{x}') \\ \notag
&+(\alpha + 1) \bd{s}_p(\bd{x}')^\top \nabla_{\bd{x}} k(\bd{x},\bd{x}')+\nabla_{\bd{x}'}\cdot(\nabla_{\bd{x}} k(\bd{x}, \bd{x}'))\big].      
\end{align}
It is interesting to investigate the cases when the BBIS induced from alpha-weighted KSD outperforms the original BBIS~\cite{liu2016black}.

\section{Summary} We provide a unified framework to sample from the target distribution whose gradient information is unavailable or intractable. Starting from any set of particles $\{\vx_i\}_{i=1}^n,$ GF-SVGD iteratively transports the particles $\{\vx_i\}_{i=1}^n$ to approximate the target distribution $p(\vx).$ GF-SVGD leverages the gradient of the surrogate distribution $\rho(\vx)$ and corrects the bias with a form of importance weight $\rho(\vx)/p(\vx).$ The $\KL$ divergence between the distribution of the updated particles $\{\vx_i\}_{i=1}^n$ and the target distribution $p(\vx)$ is proven to be maximally decreased in the functional space. The performance of GF-SVGD critically depends on the choice of the surrogate distributions. We empirically investigate the choice of the surrogate distributions and have found that the surrogate distributions which have wide variance tend to perform better. Both theoretical justifications and empirical experiments are provided to demonstrate the effectiveness of our provided gradient-free sampling algorithm. We further improve the gradient-free sampling algorithm, which is motivated from annealed importance sampling, by applying the gradient-free update to the intermediate distribution $p_\ell(\bd{x})$ that interpolate between the initial distribution $p_0(\vx)$ and the target distribution $p(\vx).$ The initial particles can be drawn from $p_0(\vx)$. Instead of applying gradient-free update to $p(\vx),$ we set the intermediate distribution $p_\ell(\bd{x})$ as the target target and the surrogate distribution is constructed on the fly based on the current particles $\{\vx_i^{\ell}\}_{i=1}^n,$ which approximates $p_{\ell-1}(\bd{x})$ by our update. Therefore, the importance ratio $\rho(\vx)/p_{\ell}(\vx)$ is evaluated between two close distributions, which approximates $p_{\ell-1}(\vx)/p_{\ell}(\vx).$ Empirical experiments demonstrate the improved gradient-free update is robust and can be widely applied to perform gradient-free sampling in various methods with different dimensions. In the end, we provide a gradient-free black-box importance sampling algorithm, which equips any given set of particles $\{\vx_i\}_{i=1}^n$ with importance weights $\{u_i\}_{i=1}^n$ so that \eqref{def:bbis} gives a good approximation. The theoretical approximation error is provided. 

In the next two chapters, we will leverage the gradient-free SVGD and the gradient-free KSD we develop in this chapter to propose the sampling algorithm and the goodness-of-fit test on discrete distributions. The basic idea is to transform the discrete-valued distributions to the corresponding continuous-valued distributions by a simple form of transformation, which are non-differentiable in finite states. In the next chapter, we apply the gradient-free SVGD to the continuous-valued distributions for sampling and use the inverse transform to get the discrete-valued samples.

\chapter{Sampling from Discrete Distributions\label{chap:disc}}
Discrete probabilistic models provide 
a powerful framework for capturing complex phenomenons and patterns, 
such as conducting logic, symbolic reasoning\citep{holland1981exponential}, natural language processing\citep{johnson2007bayesian} and computer vision\citep{sutton2012introduction}. 
However, probabilistic inference of high dimensional discrete distribution is in general NP-hard and requires highly efficient approximate inference tools. Traditionally, approximate inference in discrete models is performed by either Gibbs sampling and Metropolis-Hastings algorithms, or deterministic variational approximation, such as belief propagation, mean field approximation and variable elimination methods \citep{wainwright2008graphical, dechter1998bucket}.  
However, both of these two types of algorithms have their own critical weaknesses: Monte Carlo methods provides theoretically consistent sample-based (or particle) approximation, but are  typically slow in practice, 
while deterministic approximation 
are often much faster in speed, 
but does not provide progressively better approximation like Monte Carlo methods offers. 
New methods that integrate the advantages of the two methodologies is a key research challenge; see, for example, \citep{liu2015probabilistic, lou2017dynamic, ahn2016synthesis}. 

Stein variational gradient descent (SVGD) \citep{liu2016stein} has been shown a powerful approach for approximate inference on large scale distributions. However, existing forms of SVGD are designed for continuous-valued distributions and requires the availability and tractability of the gradient information of the target distributions. Gradient-free SVGD leverages the gradient information of a surrogate distribution and corrects the bias with a form of importance weights, which does not require the gradient information of the continuous-valued target distribution. In this chapter, we leverage the power of SVGD for the inference of discrete distributions. Our idea is to transform discrete distributions to piecewise continuous distributions, on which gradient-free SVGD, a variant of SVGD that leverages a differentiable surrogate distribution to sample non-differentialbe continuous distributions, is readily applied to perform inference. 
To do so, 
we design a simple yet general framework for transforming discrete distributions to equivalent continuous distributions, 
which is specially tailored for our purpose, 
so that we can conveniently and effectively construct the differentiable surrogates needed for GF-SVGD. 
\paragraph{Outline} This chapter is organized as follows. We first discuss the background and some discrete distributional sampling baselines. Then we introduce our main algorithm to sample from discrete distribution. We provide empirical experiments on discrete graphical probability models to demonstrate the effectiveness of our proposed algorithm. 

\section{Background and Other Discrete Sampling Algorithms}

We start with briefly introducing gradient-free SVGD 
\citep{han2018stein}, which works for nondifferentiable functions by introducing a differentiable surrogate function. 
Let $p(\vx)$ be a differentiable density function supported on $\R^d$. 
The goal of GF-SVGD is to find a set of samples 
$\{ \vy^i\}_{i=1}^n$ (called ``particles'') to approximate $p(\vx)$ in the sense that 
$$
\lim_{n\to\infty}\frac{1}{n}\sum_{i=1}^n f(
\vx_i) =\E_p [f(\vx)],  
$$
for general test functions $f(\vx)$ without using the gradient information of the target $p(\vx)$. Note that when this holds for all bounded continuous functions, the empirical distribution of the particles is called weakly converges to $p(\vx)$. 

GF-SVGD achieves this by starting 
with a set of initial particles $\{  \vy^i\}_{i=1}^n$, and iteratively updates the particles by  
\begin{align}\label{equ:xxii}
\vy^i  \gets \vy^i +  \epsilon \ff^*(\vy^i),  ~~~~ \forall i = 1, \ldots, n,  
\end{align}
where $\epsilon$ is a step size, and 
$\ff\colon \RR^d \to \RR^d$ is a velocity field  chosen to drive the  particle distribution closer to the target.
Assume the distribution of the particles at the current iteration is $q$, 
and $q_{[\epsilon\ff]}$ is the distribution of the updated particles $\vy^\prime = \vy + \epsilon \ff(\vy)$. 
The optimal choice of $\ff$ can be framed as the following optimization problem:  
\begin{align}\label{equ:ff00}
\ff^* =   \argmax_{\ff \in \F}  \bigg\{  -   \frac{d}{d\epsilon} \KL(q_{[\epsilon\ff]} ~|| ~ p) \big |_{\epsilon = 0}  \bigg\}, 
\end{align}
where $\F$ is a set of candidate  velocity fields, and  $\ff$ is chosen in $\F$ to maximize the decreasing rate on the KL divergence between the particle distribution and the target. 

\paragraph{Gradient-free SVGD} (GF-SVGD) \citep{han2018stein} extends SVGD to the setting when the gradient of the target distribution does not exist or is unavailable. The key idea is to replace the gradient of differentiable surrogate distribution $\rho(\vy)$  whose gradient exists and can be calculated easily, 
and leverage it for sampling from $p(\vy)$  
using a mechanism similar to importance sampling. 

The derivation of GF-SVGD is based on the following key observation, 
\begin{equation}
\label{obj:iwksd}
 w(\vy)\steinbxtransp\ff(\vy)  = \steinpxtransp \big(w(\vy)\ff (\vy) \big).
\end{equation}
where $w(\vy)=\rho(\vy)/p(\vy).$ Eq. \eqref{obj:iwksd} indicates that the Stein operation w.r.t. $p$, which requires the gradient of the target $p$,  can be transferred to the Stein operator of a surrogate distribution $\rho$, which does not depends on the gradient of $p$.  
Based on this observation, 
GF-SVGD modifies \label{equ:klstein00} to optimize the following object,
\begin{align}
\label{gradfreeKLmin}
\!\!\!\! \ff^*& 
\!= \argmax_{\ff \in \F} \{ \E_{q} [\steinpxtransp (w(x)\ff(\vy))]\}. 
\end{align}
Similiar to the derivation in SVGD, the optimization problem \eqref{gradfreeKLmin} can be analytically solved; in practice, GF-SVGD derives a gradient-free update as $\vy^i\leftarrow \vy^i+\frac{\epsilon}{n}\Delta \vy^i,$ where
\begin{align}
\label{update222}
\!\!\!\!\!\! \Delta \vy^i \propto
\!\sum_{j=1}^n 
\!w(\vy_j) \big[\nabla \log \rho(\vy^j) k( \vy^j, \vy^i) + \nabla_{\vy^j} k(\vy^j, \vy^i) \big], 
\end{align}
which replaces the true gradient $\nabla \log p$ with a surrogate gradient $\nabla\log \rho$, and then uses an importance weight $w(\vy_j):=\rho(\vy^j)/p(\vy^j)$ to correct the bias introduced by the surrogate. In practice, the weights $\{\vy_j\}_{j=1}^n$ might have very large variance when the surrogate $\rho(\vy)$ is different from the target $p(\vy).$ The key ingredient reduces to design an effective surrogate to approximate the target distribution reasonably well so that the variance of $\{\vy_j\}_{j=1}^n$ is small enough for the convergence of the gradient-free SVGD update. Although it might be challenging to construct a good approximation of the continuous-valued target distributions, it is interesting to observe that it is easy to construct an effective surrogate in dicrete-valued distributions by leveraging their discrete structures.

\citet{han2018stein} observed that  GF-SVGD can be viewed as a special case of SVGD with an ``{importance weighted}'' kernel, 
\begin{equation*}
\wt{k}(\vy, \vy')=\frac{\rho(\vy)}{p(\vy)}k(\vy, \vy')\frac{\rho(\vy')}{p(\vy')}.    
\end{equation*}
Therefore, GF-SVGD inherits
the theorectical justifications of SVGD  \citep{liu2017stein}. GF-SVGD is proposed to applied to continuous-valued distributions. The goal of this work is to further develop GF-SVGD into a key inference tool for discrete distributions, by proposing a simple yet powerful method to transform discrete distributions to piecewise continuous distributions, which can be efficiently handled by GF-SVGD with easily constructed surrogate distribution. 


However, because SVGD only works for continuous distributions, a key open question is if it is possible to exploit it for more efficient inference of discrete distributions.  





We apply our proposed algorithm to a wide range of discrete distributions, 
Ising models, restricted Boltzmann machines, as well as challenging real-world problems drawn from  
UAI approximate inference competitions and learning binarized neural networks. 
We find that our proposed algorithm significantly outperforms 
traditional inference algorithms for discrete distributions. 

In particular, our algorithm is shown to be provide a promising tool for ensemble learn of binarized neural network (BNN) in whichs both weights and activations functions are binarized.  
Learning BNNs have been shown to be a highly challenging problem, because standard backpropagation can not be applied.  
We cast the learning BNN as a Bayesian inference problem of drawing a set of samples (which in turns forms an ensemle predictor) of the posterior distribution of weights,
and apply our SVGD-based algorithm for efficient inference. 
We show that our method outperforms other widely-used ensemble method such as bagging and AdaBoost in achieving  highest accuracy with the same ensemble size.   

{\bf Related work} 
Markov chain Monte Carlo (MCMC) is routinely used to generate samples from posterior distributions. Among a variety of MCMC algorithms, Hamiltonian Monte Carlo~(HMC) promises a better scalability and has enjoyed wide-ranging successes as one of the most reliable approaches in general settings~\citep{neal2011mcmc, gelman1998simulating}, which are originally targeted for sampling from discrete distributions. However, a fundamental limitation of HMC is the lack of support for
discrete parameters. The difficulty of extending HMC from the continuous distribution to the discrete distribution comes from the fact that the construction of HMC proposals relies on a numerical solvers which should satisfy the volume preserving property. The idea of transforming 
the inference of discrete 
distributions to continuous distributions have been widely studied, which, however, mostly concentrate on 
leveraging the power of Hamiltionian Monte Carlo (HMC); see, for example, \citep{afshar2015reflection, nishimura2017discontinuous, pakman2013auxiliary, zhang2012continuous, dinh2017probabilistic}. 

Our work is instead motivated by leveraging the power of SVGD, which allows us to derive 
a novel framework for fast and deterministic sampling of inference discrete distributions.
Our framework of transforming discrete distributions to piecewise continuous distribution is similiar to \citet{nishimura2017discontinuous}, 
but is more general and tailored for the application of GF-SVGD. 
Our empirical results show that our method outperforms both traditional algorithms such as Gibbs sampling and discontinuous HMC  \citep{nishimura2017discontinuous}. 





\section{Sampling from Discrete Distribution}

This section introduces the main idea of this work,  
provides a simple yet powerful way for leveraging GF-SVGD as a key inference tool for discrete distributions. 
This is done by  
converting discrete distributions to  piecewise continuous distributions, which can be efficiently handled by GF-SVGD. 

Assume we are interested in sampling from a given discrete distribution $p_*(\vz)$, defined on a finite discrete set $\mathcal Z=\{a_1,\ldots, a_K\}$.  We may assume each $a_i$ is a $d$-dimensional vector of discrete values, so that $\mathcal Z$ is a product space.  
Our idea is to construct a piecewise continuous  distribution $p_c(\vx)$ for $\vx\in  \RR^d$, and a map $\Gamma\colon \RR^d \to \mathcal Z$, 
such that the distribution of $\vz = \Gamma(\vx)$ is $p_*$  when $\vx\sim p_c$. In this way, we can apply GF-SVGD on $p_c$ to get a set of samples $\{\vx^i\}_{i=1}^n$ and apply transform $\vz^i = \Gamma(\vx^i)$ to get samples $\{\vz^i\}$ from $p_*(\vz).$ 

\begin{mydef}
A piecewise continuous distribution $p_c(\vx)$ on $\RR^d$ and map $\Gamma\colon \RR^d 
\to \mathcal Z$ is called to form a 
\textbf{continuous parameterization} of  $p_*$, if $\vz = \Gamma(\vx)$ follows $p_*(\vz)$ when $x\sim p_c$. 
\end{mydef}

Following this definition, we have the following immediate result. 
\begin{pro}
 $p_c$ and $\Gamma$ form a continuous parameterization of discrete distribution $p_*$ on $\mathcal Z =\{a_1, \ldots, a_K\}$, iff 
\begin{align}\label{pstar}
p_*(a_i) = \int_{\RR^d} 
p_c(\vx) \ind [ a_i = \Gamma(\vx)] d\vx, 
\end{align}
for all $i = 1,\ldots, K$. 
Here $\ind(\cdot)$ is the 0/1 indicator function, $\ind(t) =0$ iff $t = 0$  and $\ind(t)=1$ if otherwise. 
\end{pro}

\begin {algorithm*}[h]
\caption {Gradient-free SVGD for Discrete Distributions} 
\label{alg:alg1}  
\begin {algorithmic}
\STATE {\bf Goal}: 
Approximate a given distribution $p_*(\vz)$ on a finite discrete set $\mathcal Z$. \vspace{.5em}  
\STATE 1) Decide a base distribution $p_0(\vx)$ on $\RR^d$ (such as Gaussian distribution), and a map $\Gamma\colon \RR^d \to \mathcal Z$ which partitions $p_0(\vx)$ evenly. Construct a 
\textbf{piecewise continuous distribution 
$p_c(\vx)$} by \eqref{equ:pc}:
$$
p_c(\vx) = p_0(\vx) p_*(\Gamma(\vx)). 
$$
\STATE 2) Construct a \textbf{differentiable surrogate} of $p_c(\vx)$, for example, by $\rho(\vx) \propto  p_0(\vx) \tilde p_*(\tilde \Gamma(\vx)),$ where $\tilde{p}_*$ and $\tilde \Gamma$ are smooth approximations of $p_*(\vz)$ and $\Gamma$, respectively.  \vspace{.5em}  
\STATE 3) Run gradient-free SVGD on $p_c$ with differentiable surrogate $\rho$: Starting from an initial $\{\vx^i\}_{i=1}^n$ and repeat 
$$
\vx^i \gets \vx^i + \epsilon \sum_{j=1}^n
w_j(\nabla\rho(\vx) k(\vx^j, \vx^i) + \nabla_{\vx^j} k(\vx^j, \vx^i)). 
$$
where $w_j = {\rho(\vx^j)}/{p_c(\vx^j)}$, and $k(\vx,\vx')$ is a positive definite kernel.  \vspace{.5em}  
\STATE 
4) Calculate $\vz^i = \Gamma(\vx^i)$ and output sample 
$\{\vz^i \}_{i=1}^n$ for approximating the distribution $p_*(\vz)$. 
\end {algorithmic}
\end {algorithm*}

Given a discrete distribution $p_*$, there are many different continuous parameterizations. And exact samples of $p_c$ yields exact samples of $p_*$ following the definition. 
For the purpose of approximation, 
the best continuous parameterization should  be constructed so that the continuous distribution $p_c$ can be efficiently sampled using continuous inference algorithms, which is gradient-free SVGD in our case. 

Here we introduce a simple yet powerful framework for constructing continuous parameterizations, 
which naturally comes with good differentiable surrogates with which   gradient-free SVGD can perform   efficiently.  
Our procedure is also highly general and allows us to construct continuous parameterization for different discrete distributions in an almost automatic fashion. 

Our method starts with choosing a simple base distribution $p_0(\vx)$ on $\RR^d$, which we take to be the  Gaussian distribution in most cases, and a map $\Gamma$ that \emph{evenly partition} $p_0$. 

\begin{mydef} A map $\Gamma \colon \Z \to \RR^d$ is said to 
\textbf{evenly partition} 
 $p_0$ if we have  
\begin{align} \label{even}
\int_{\RR^d} 
p_0(\vx) \ind[a_i = \Gamma(x)] d\vx = \frac{1}{K}, 
\end{align}
for $i=1,\ldots K$. 
Following \eqref{pstar},  
this is equivalent to saying that 
$p_0$ and $\Gamma$ forms a continuous  relaxation of the uniform distribution $p_*(a_i) = 1/K$. 
\end{mydef}

For simple $p_0$ such as standard Gaussian distributions, it is straightforward to construct even partitions using the quantiles of $p_0$.  
For example,
in one dimensional case $(d=1)$, we can evenly partition any continuous $p_0(\vx)$, $\vx\in \RR$ by  
\begin{align}\label{equ:gamma1D}
\Gamma(\vx) = a_i ~~~~~
\text{if ~~ $x \in [\eta_{i-1}, ~~ \eta_{i})$}, 
\end{align}
where $\eta_i$ denotes the $i/K$-th quantile of distribution $p_0$. 
In multi-dimensional cases ($d>1$) and when $p_0$ is a product distribution:
$$
p_0(\vx) = \prod_{j=1}^d p_{0,i}(x_i). 
$$
One can easily show that an even partition can be constructed by concatenating one-dimensional even partitions: 
$$
\Gamma\left (
\begin{bmatrix}
x_1\\
\vdots\\
x_d
\end{bmatrix}
\right) = 
\begin{bmatrix}
\Gamma_{1}(x_1)\\
\vdots\\
\Gamma_{d}(x_d)
\end{bmatrix}, 
$$
where $\Gamma_{i}(\cdot)$ an even partition of $p_{0,i}$.  
A particularly simple case is when $\vz$ are binary vectors, that is, $\mathcal Z = \{\pm1 \}^d$, in which case $\Gamma(\vx) = \sign(\vx)$ evenly partitions any distribution $p_0$ that is symmetric around the origin. 

Given an even partition of $p_0$, we can conveniently construct 
a continuous parameterization of an arbitrary discrete distribution $p_*$ by 
\emph{weighting each bin of the partition with corresponding probability in $p_*$}, that is, we may  construct $p_c$ by 
\begin{align}\label{equ:pc}
p_c(\vx) \propto p_0(\vx) p_*(\Gamma(\vx)),  
\end{align} 
where $p_0(\vx)$ is weighted by $p_*(\Gamma(\vx))$, the probability
of the discrete value $\vz=\Gamma(\vx)$ that $\vx$ maps. 
\begin{pro}
Assume $\Gamma$ is an even partition of $p_0(\vx)$, and $p_c(\vx) = K p_0(\vx) p_*(\Gamma(\vx))$, where $K$ severs as a normalization constant, then $(p_c, ~ \Gamma)$ is a continuous parameterization of $p_*$. 
\end{pro}
\begin{proof}
We just need to verify that \eqref{pstar} holds. 
\begin{align*}
    & \int p_c(\vx) \ind[a_i = \Gamma(\vx)] d\vx  \\
    & = K \int p_0(\vx) p_*(\Gamma(\vx)) \ind[a_i = \Gamma(\vx)] d\vx   \\
    & = K \int p_0(\vx) p_*(a_i) \ind[a_i = \Gamma(\vx)] d\vx \\
    & = K p_*(a_i) \int p_0(\vx)  \ind[a_i = \Gamma(\vx)] d\vx \\
    & = p_*(a_i), 
\end{align*} 
where the last step follows \eqref{even}. 
\end{proof}
This provides a simple and general approach for constructing continuous parameterizations of arbitrary discrete distributions. 
Given such a construction ,  it is also convenient to construct differentiable surrogate of $p_c(\vx)$ in \eqref{equ:pc} for gradient-free SVGD. 
To do so, note that the non-differetiable part of $p_c$ comes from $p_*(\Gamma(x))$, and hence 
we can construct differentiable  surrogate of $p_c(\vx)$ by simply removing $p_*(\Gamma(x))$ (so that $\rho=p_0$), 
or approximate it with some smooth approximate of it, based on properties of $p_*(\vz)$ and $\Gamma$. 
See Algorithm~\ref{alg:alg1} for the summarization of our main procedure. As the piecewise continuous distribution $p_c(\vx)$ has at most $K-1$ points which is non-differentiable, it is expected that the updated particles when applying GF-SVGD will theoretically converge to the target with mild conditions. Let us now illustrate our constructions using examples. 
\begin{figure}[tbh]
\centering
\begin{tabular}{ccc}
\includegraphics[width=0.34\textwidth]{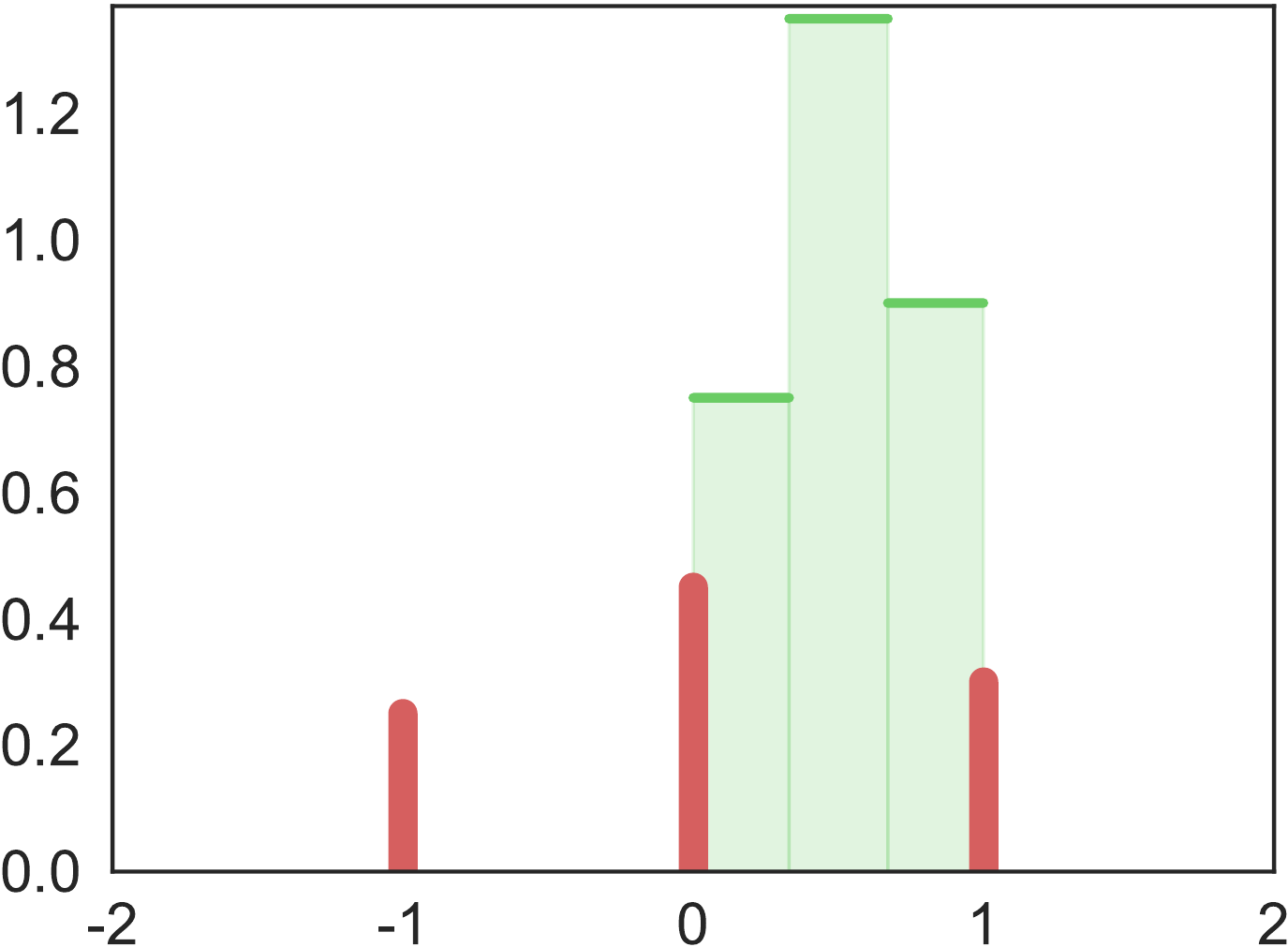} &
\hspace{-.5cm}
\includegraphics[width=0.34\textwidth]{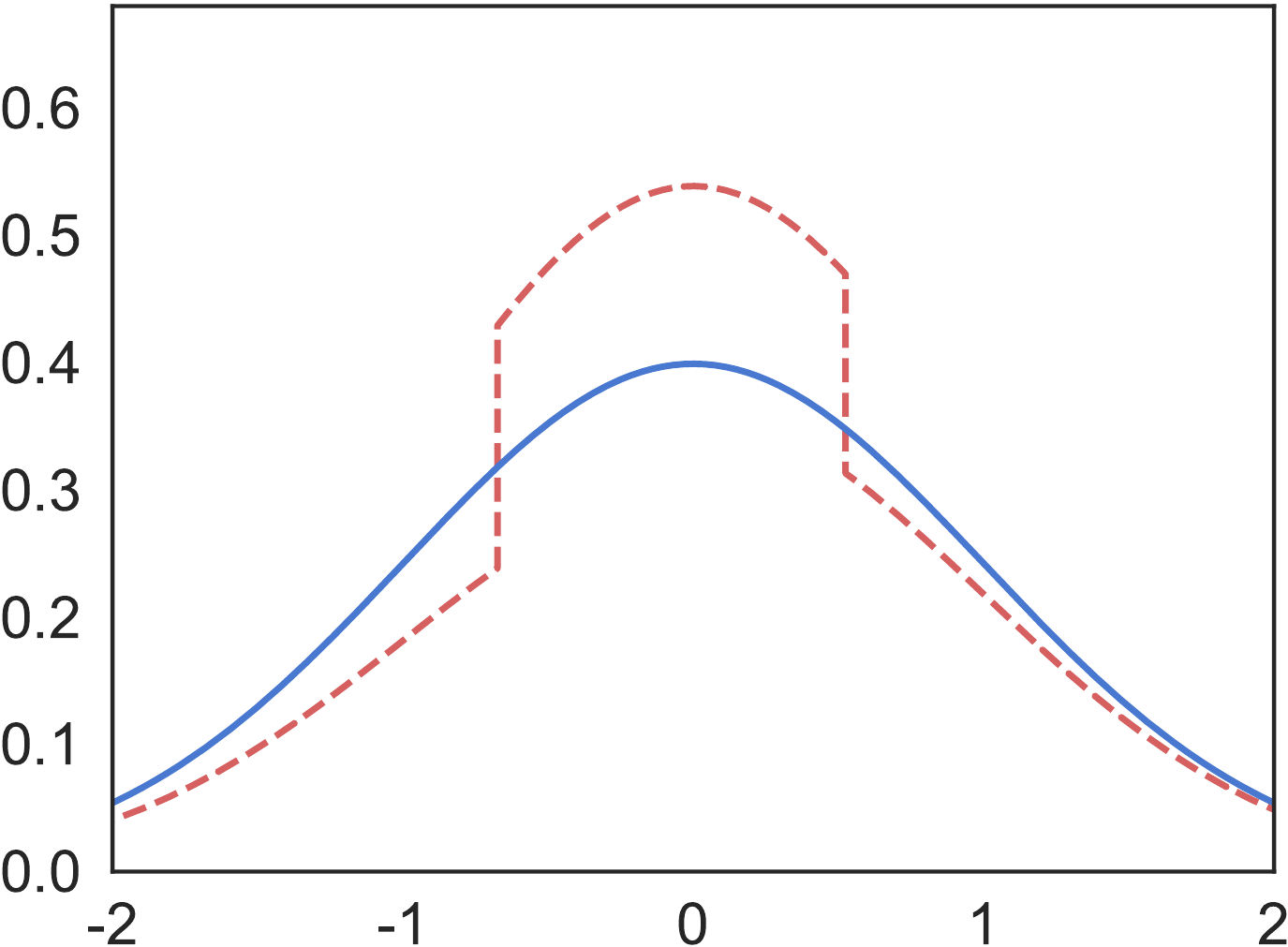} &
\hspace{-.5cm}
\raisebox{3.9em}{\includegraphics[height=0.15\textwidth]{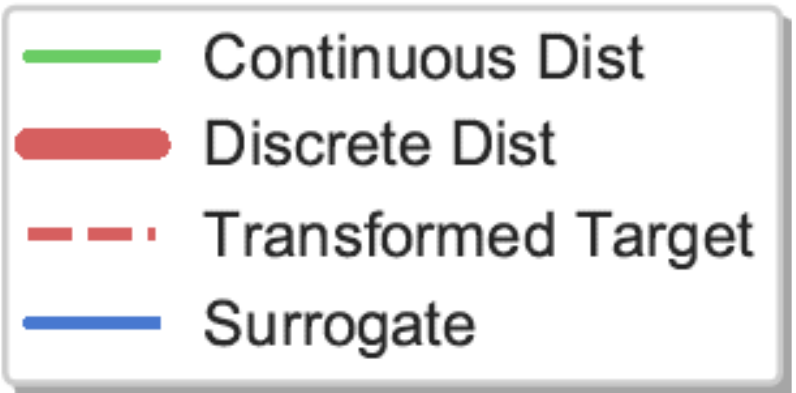}} \\
{\small discrete-valued distribution} & {\small continuous-valued distribution}\\
\end{tabular}
\caption[Illustration of our discrete distributional sampling algorithm]{Illustrating 
the constructing of $p_c$ (red line) 
of a three-state distribution $p_*$ (green bars). The blue dash line represents the base distribution we use, which is a standard Gaussian distribution.}
\label{fig:definition} 
\end{figure} 

\paragraph{1D Categorical Distribution}
Consider the simple discrete distribution $p_*$ show in Figure~\ref{fig:definition}, which takes values in 
$\{-1, 0, 1\}$ with probabilities $\{0.25, 0.45, 0.3\}$, respectively.  
We take the standard Gaussian distribution (blue dash) as the base distribution $p_0$, and obtain 
a continuous parameterization $p_c$ using \eqref{equ:pc}, 
in which $p_0(x)$ is weighted by the  probabilities of $p_*$ in each bin.  
Note that $p_c$ is a piecewise continuous distribution. 
In this case, we may naturally choose the base distribution $p_0$ as the  differentiable surrogate function to draw samples from $p_c$ when using gradient-free SVGD. 

In the following, we will empirically investigate the choices of the transform and provide a simple yet practically powerful transform, which will be demonstrated by a number of probability models in the experimental section.

\subsection{Investigation of the Choice of Transform}
\begin{figure}[tbh]
\centering
\begin{tabular}{cc}
\includegraphics[width=0.46\textwidth]{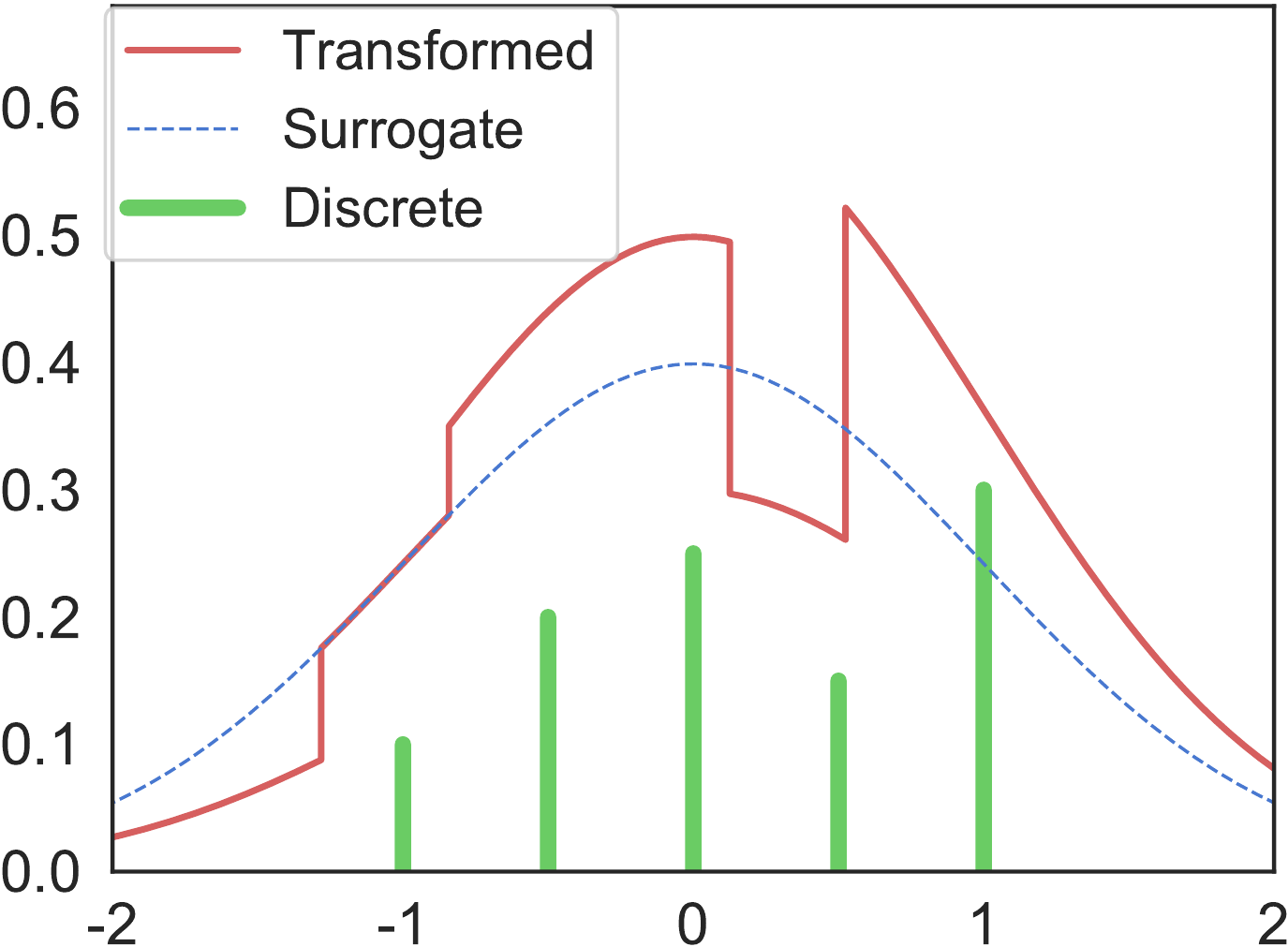} & \hspace{-.5cm}
\includegraphics[width=0.46\textwidth]{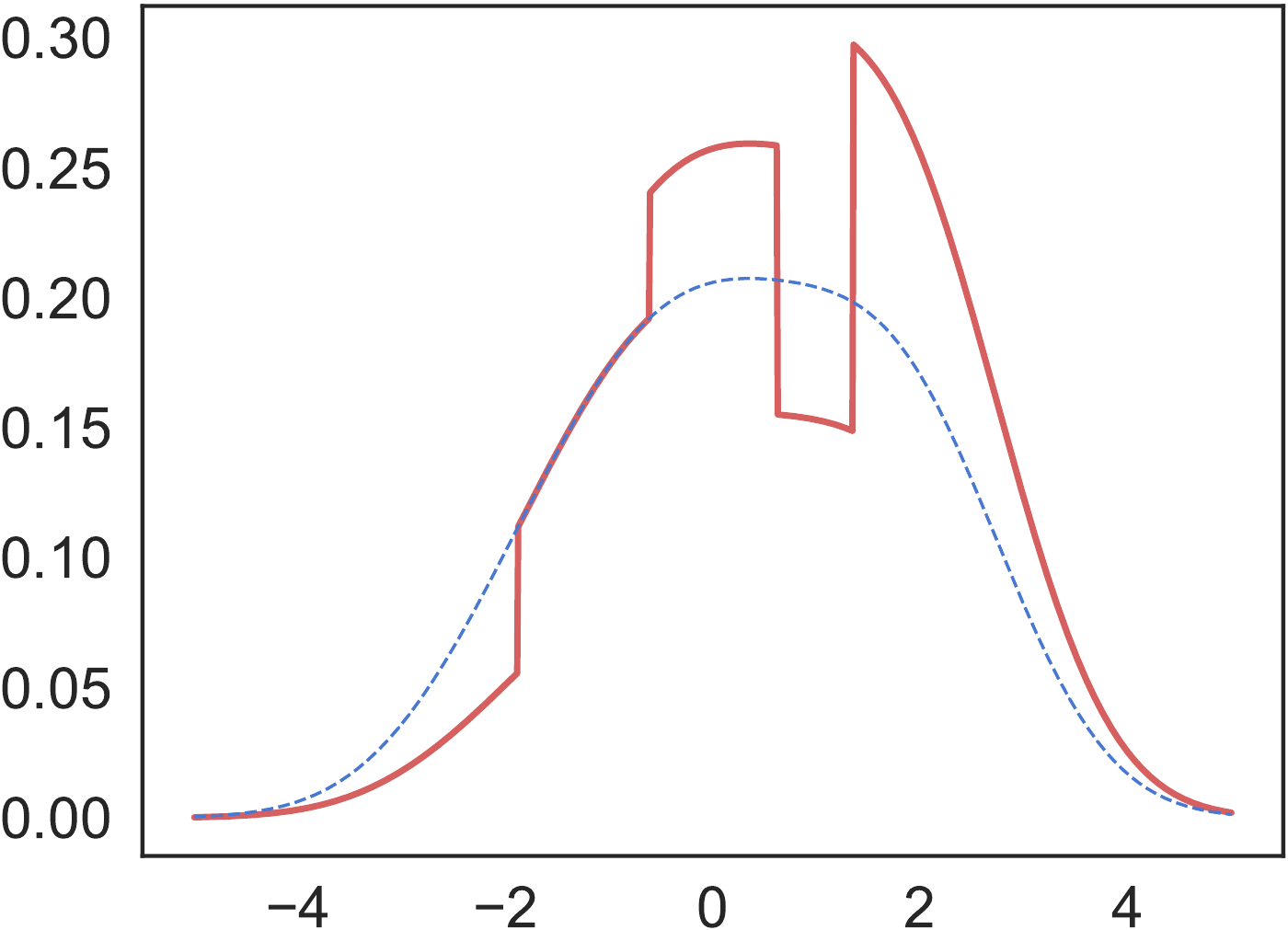} \\
{\small (a), Base $p_0(x)=\mathcal{N}(x; 0, 1)$} & {\small (b), $\bd{\mu}=(-2,-1,0,1,2) $}  \\
\includegraphics[width=0.46\textwidth]{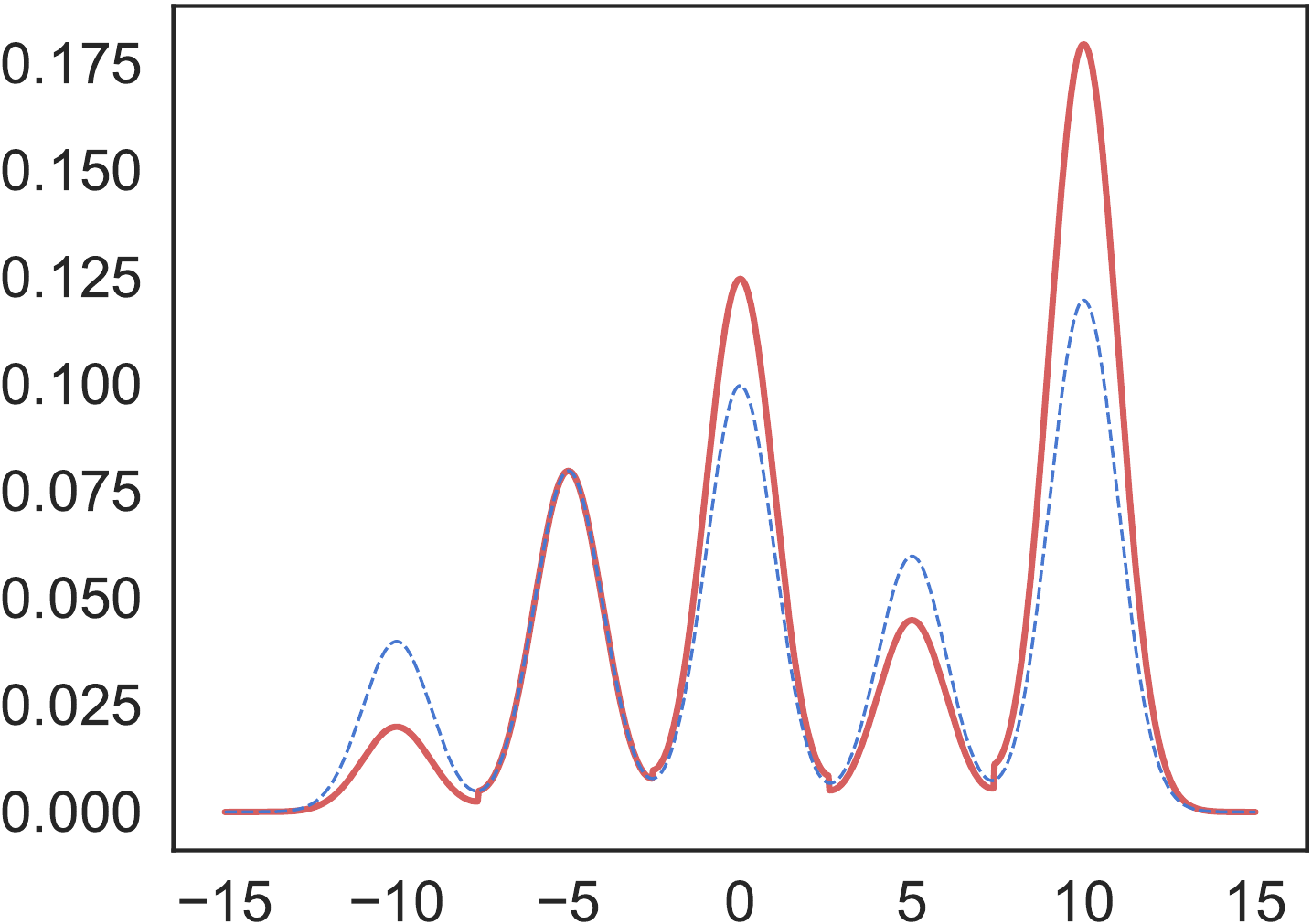} &\hspace{-.5cm}
\includegraphics[width=0.46\textwidth]{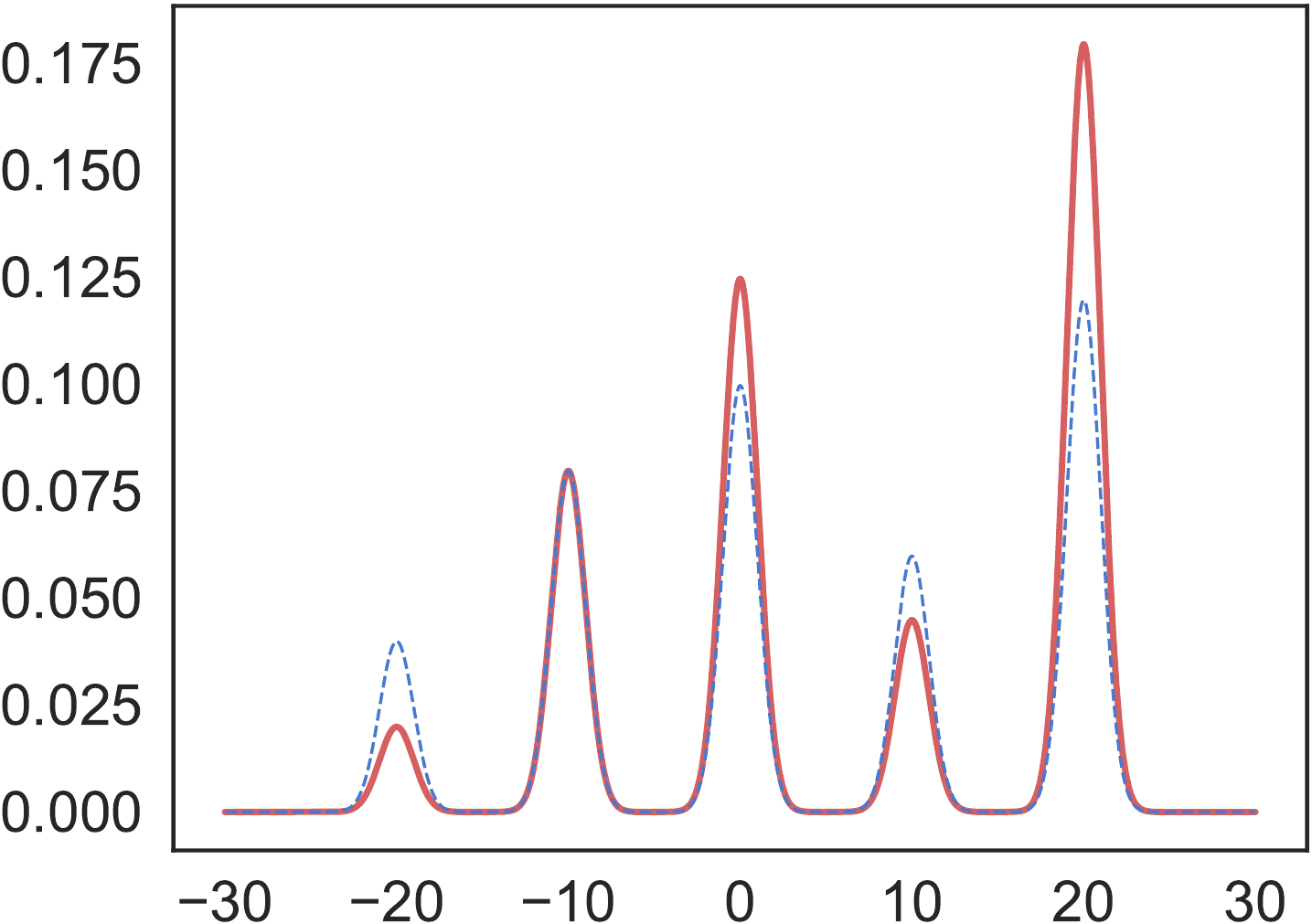} \\
{\small (c), $\bd{\mu}=(-10,-5,0,5,10)$} & {\small (d), $\bd{\mu}=(-20,-10,0,10,20)$} 
\end{tabular}
\caption[Empirical investigation of the choice of transform when applying GF-SVGD to transformed discrete distributions]{Illustrating the construction of $p_c(x)$ (red line) of a five-state discrete distribution $p_*$ (green bars) and the choice of transform. $p_*(z)$ takes values $[-2,-1,0,1,2]$ with probabilities $[p_1, p_2, p_3, p_4, p_5]=[0.1, 0.2, 0.25, 0.15, 0.3]$ respectively. $K=5.$ The dash blue is the surrogate using base $p_0$. Let $p(y)$, $y\in [0, 1)$ be the stepwise density, $p(y\in [\frac{i-1}{K}, \frac{i}{K}))=p_i$, for $i=1,\cdots,K$. In (b, c, d), the base is chosen as $p_0(x)=\sum_{i=1}^5 p_i\mathcal{N}(x; \mu_i, 
1.)$ and $\bd{\mu}=(\mu_1,\mu_2,\mu_3, \mu_4, \mu_5).$ The base $p_0(x)$ in (a) can be seen as $\bd{\mu}=(0., 0., 0., 0., 0.).$
Let $F(x)$ be the c.d.f. of $p_0(x).$ With variable transform $x=F^{-1}(y)$, the transformed target is $p_c(x)=p(F(x)) p_0(x).$ \label{fig:transf}}
\end{figure}

There are many choices of the base function $p_0$ and the transform. We empirically investigate the optimal choice of the transform on categorical distribution in Fig.~\ref{fig:transf}. In Fig.~\ref{fig:transf}(b, c, d), the base is chosen as $p_0(x)=\sum_{i=1}^5 p_i\mathcal{N}(x;\mu_i, 
1.0)$ for different $\bd{\mu}.$ The base $p_0(x)$ in Fig.~\ref{fig:transf}(a) can be seen as $\bd{\mu}=(0., 0., 0., 0., 0.).$ We observe that with simple Gaussian base in Fig.~\ref{fig:transf}(a), the transformed target is easier to draw samples, compared with the multi-modal target in Fig.~\ref{fig:transf}(c, d), which is empirically believed that Gaussian-like distribution is easier to sample than sampling from multi-modal distributions. This suggests that Gaussian base $p_0$ is a simple but powerful choice as its induced transformed target is easy to sample by GF-SVGD. As shown in Fig.~\ref{fig:gfgauss}, even if $\rho(\vx)=p(\vx),$ which reduces to Vanilla SVGD, the update of SVGD is inferior with other choice of $\rho(\vx).$ Therefore, it is challenging to find the optimal surrogate as we cannot find a uniform metric to measure the optimality of the surrogate and the transform. Nevertheless, we find a simple yet practical surrogate and transform to perform GF-SVGD on discrete distributions.

\paragraph{Binary Ising Models} 
Consider a binary Ising model of form 
$$
p_*(z) = \exp(b^\top z -\frac{1}{2} z^\top A z ), 
$$
where $z \in \{\pm 1\}^d$ and $b \in \RR^d$, $A\in \RR^{d\times d}$ are the model parameters. 
We can take the base distribution $p_0$ to be any zero-mean Gaussian distribution, e.g., $p_0(x)\propto \exp(-\lambda x^\top x/2)$, where $x\in \RR^d$ and 
$\lambda > 0$ is a inverse variance parameter,
and take $\Gamma(x) = \sign(x)$, 
which obviously evenly partition $p_0$. Following \eqref{equ:pc}, we have 
\begin{align*} 
p_c(x) 
& \propto p_0(x) p_*(\Gamma(x)),  \\
&\!\!\!\!\!\!\!\! \propto \exp(-\frac{1}{2}\lambda x^\top  x 
+ b^\top \sign(x)-\frac{1}{2} \sign(x)^\top A ~\sign(x)). 
\end{align*}
In this case, it is convenient to construct a differentiable surrogate of $p_c$ by simply dropping the $\sign(\cdot)$ function: 
$$
\rho(x):= 
\exp(b^\top x-\frac{1}{2} x^\top (A+\lambda I)x ). 
$$
The  $\lambda$ can be properly chosen to match the scale of $A$.

\begin{figure*}[ht]
\centering
\subfigure[0th iteration]{\label{fig:binaryiter0}
\includegraphics[width=0.176\linewidth]{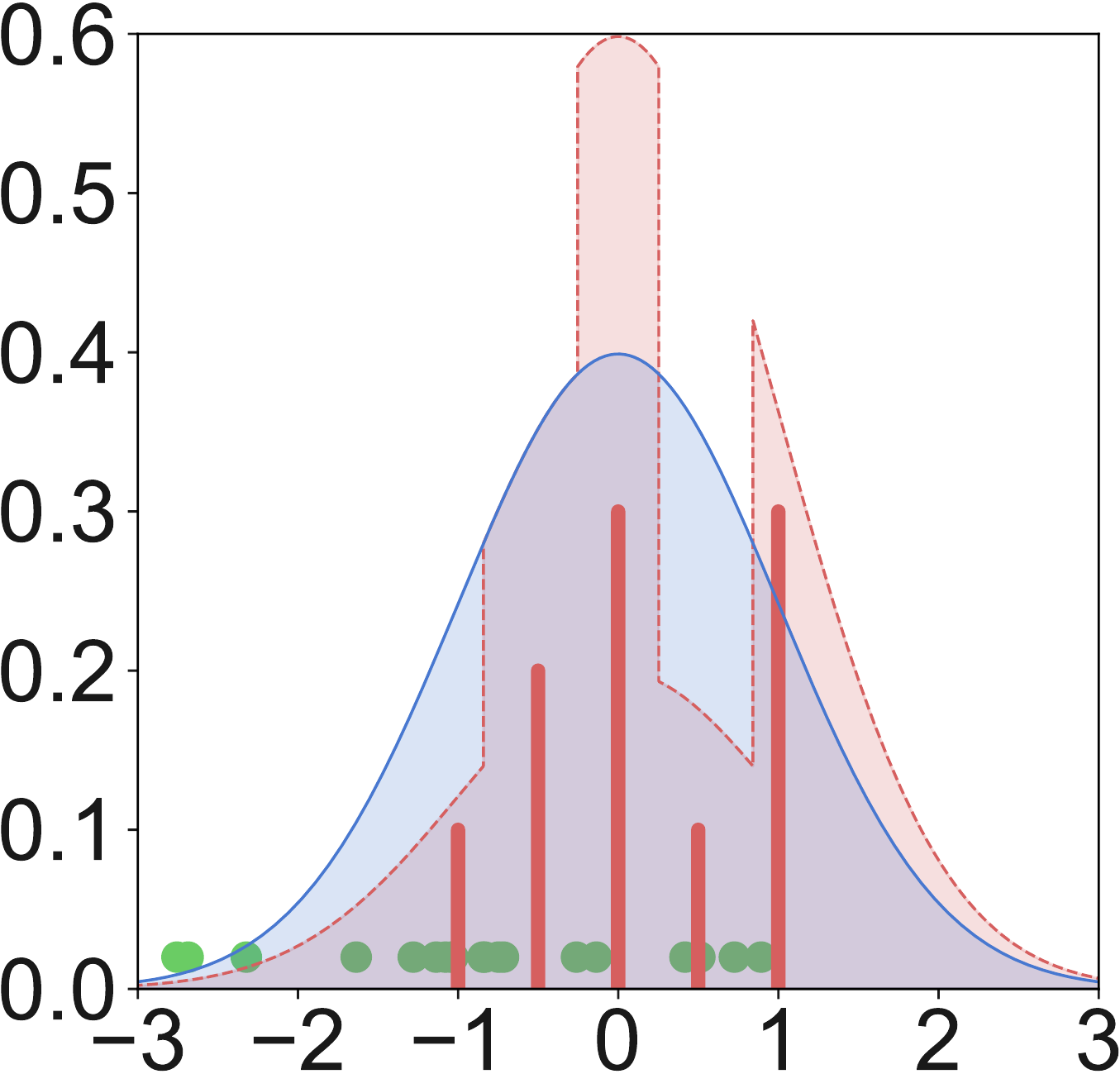}}
\subfigure[25th iteration]{\label{fig:binaryiter250}
\includegraphics[width=0.18\linewidth]{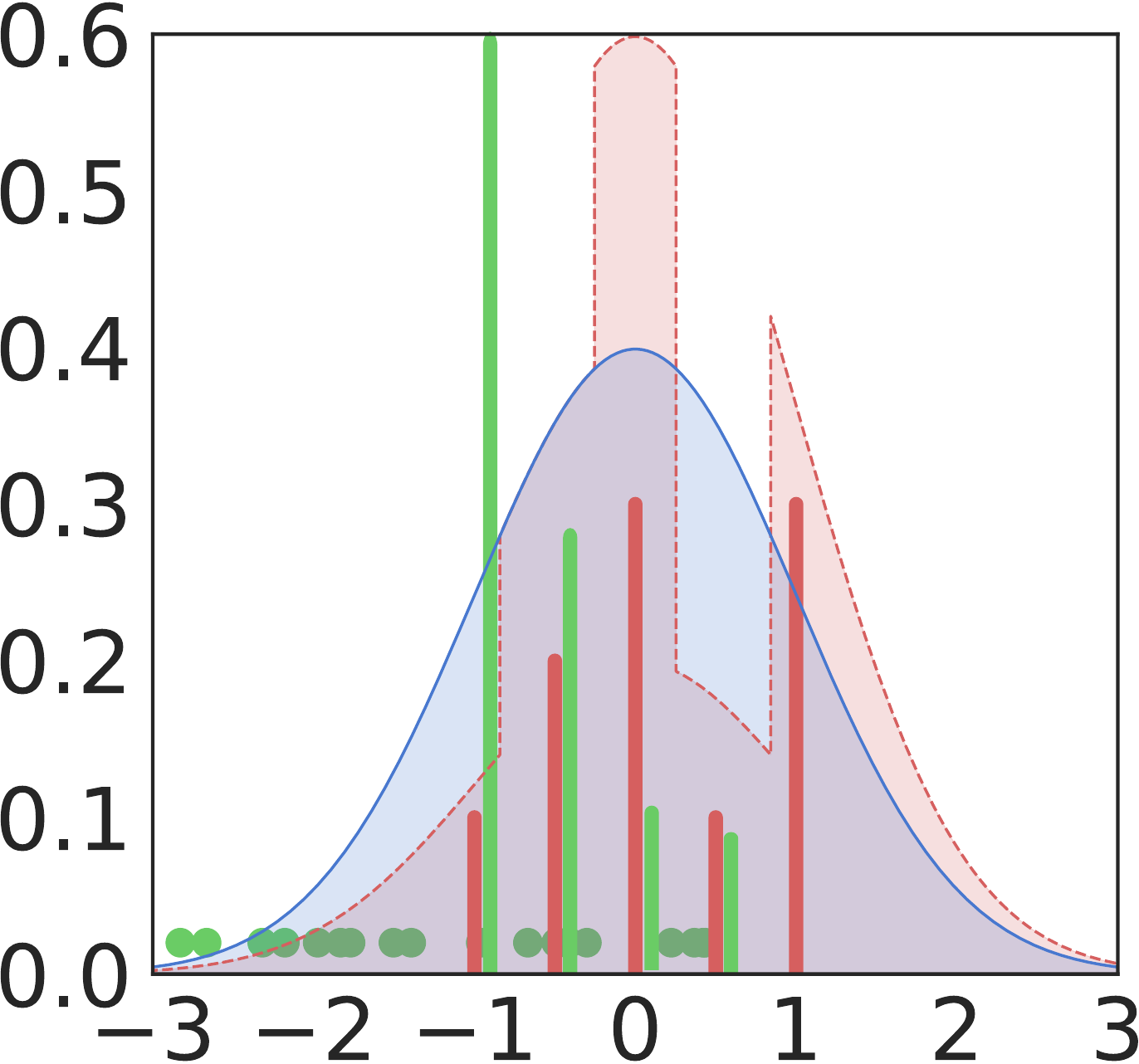}}
\subfigure[50th iteration]{\label{fig:binaryiter500}
\includegraphics[width=0.18\linewidth]{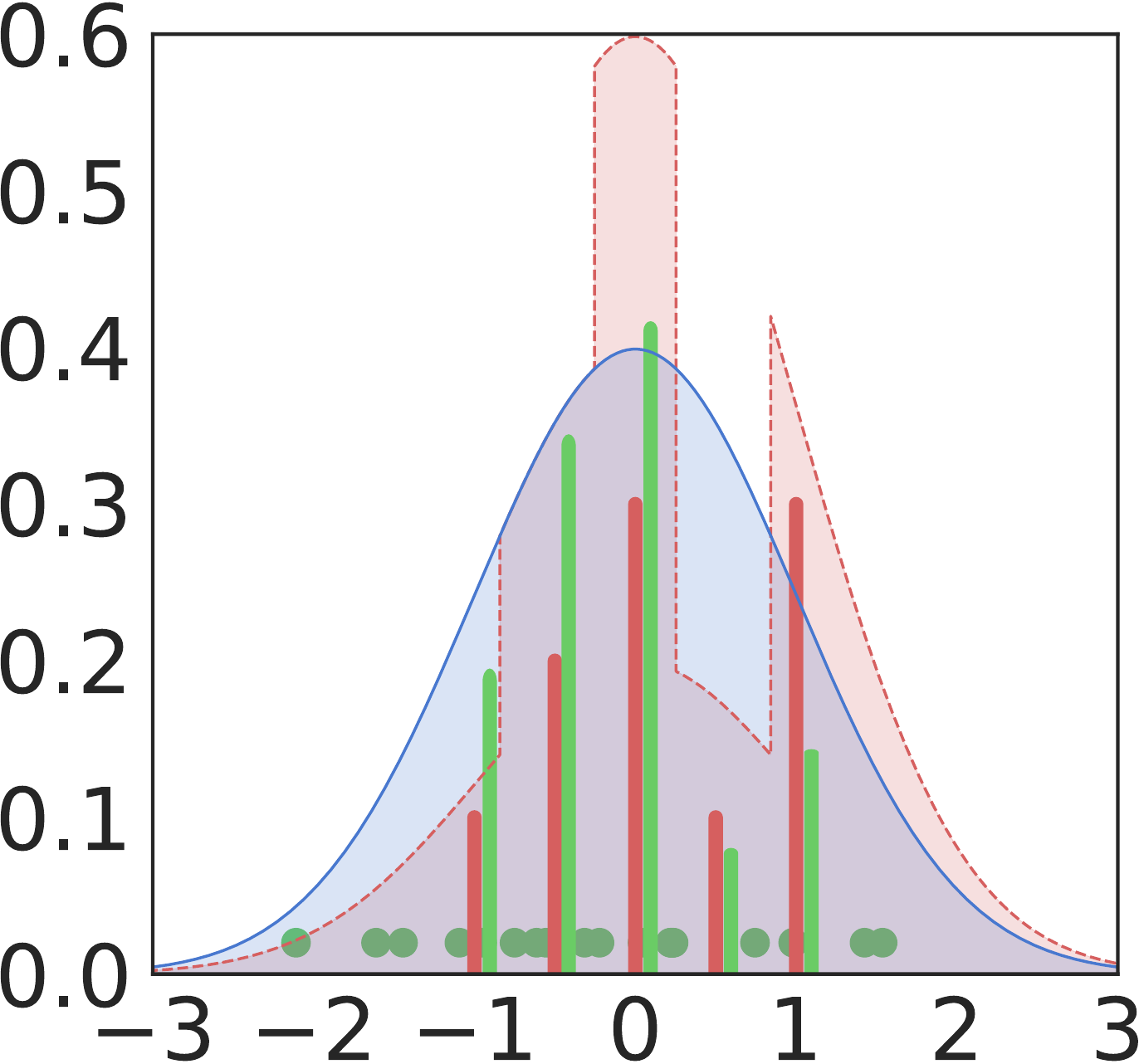}}
\subfigure[100th iteration]{\label{fig:binaryiter1000}
\includegraphics[width=0.18\linewidth]{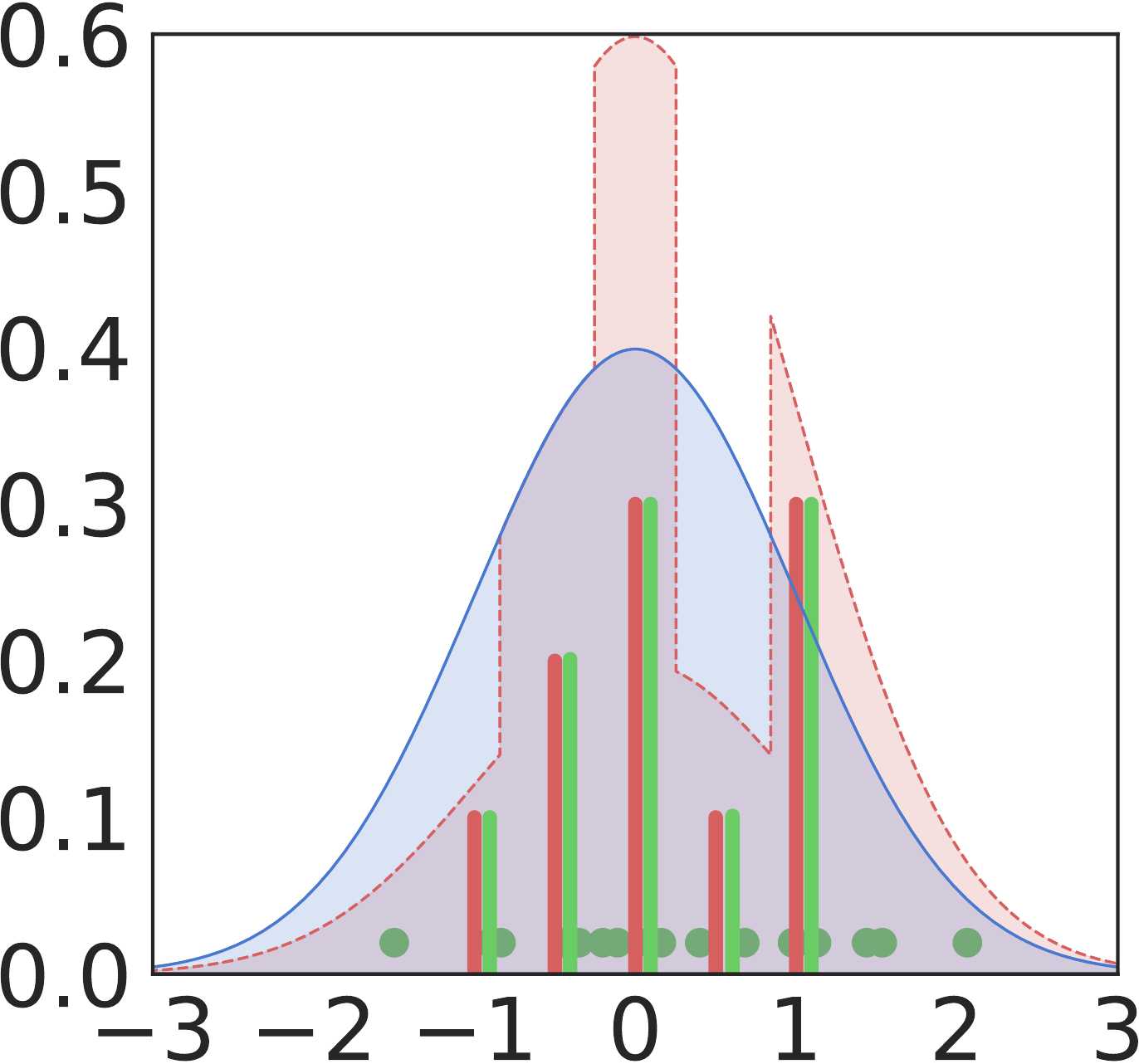}}
\subfigure{\label{fig:movement_legend}
\includegraphics[width=0.19\linewidth]{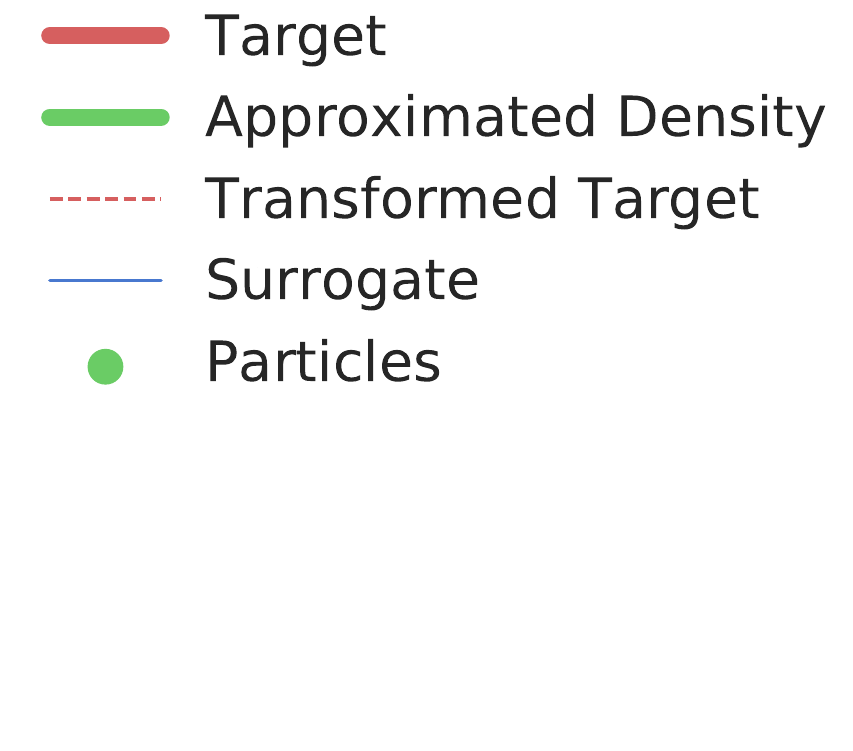}}
\caption[Illustration of the proposed method on 1D categorical distribution]{Evolution of real-valued particles $\{x^i\}_{i=1}^n$ (in green dots) by our discrete sampler in Alg.\ref{alg:alg1} on a one-dimensional categorical distribution. (a-d) shows particles $\{x^i\}$ at iteration 0, 10, 50 and 100 respectively. The categorical distribution is defined on states $z\in\{-1,-0.5, 0, 0.5, 1\}$ denoted by $a_1$, $a_2$, $a_3$, $a_4$, $a_5$, with probabilities $\{0.1, 0.2, 0.3, 0.1, 0.3\}$ denoted by $c_1$, $c_2$, $c_3$, $c_4$, $c_5$, respectively. $p_*(z=a_i)=c_i$. The base function is $p_0(x)$, shown in blue line. The transformed target to be sampled $p_c(x)\propto p_0(x)p_*(\Gamma(x))$, where $\Gamma(x)=a_i$ if $x\in[\eta_{i-1}, \eta_i)$ and $\eta_i$ is $i/5$-th quantile of standard Gaussian distribution. The surrogate distribution $\rho(x)$ is chosen as $p_0(x)$. We obtain discrete samples $\{z^i\}_{i=1}^n$ by $z^i=\Gamma(x^i)$.}
\label{fig:cat}
\vspace{-0.1in}
\end{figure*}

\section{Empirical Experiments}
We apply our algorithm to a number of large scale discrete distributions to  demonstrate its empirical effectiveness. We start with  illustrating our algorithm on sampling  from a simple one-dimensional categorical distribution. We then apply our algorithm to sample from discrete Markov random field,  Bernoulli restricted Boltzman machine and models from UAI approximation inference competition. 
Finally, we apply our method to learn  ensemble models of binarized neural networks (BNN). 

We use RBF kernel $k(x, x')=\exp(-\|x-x'\|^2/h)$ for the updates of our proposed algorithms; the bandwidth $h$ is taken to be $h {=} \mathrm{med^2}/(2\log(n+1))$ where $\mathrm{med}$ is the median of the current $n$ particles. Adam optimizer \citep{kingma2014adam} is applied to our proposed algorithms for accelerating convergence.

\begin{figure}[h]
\begin{tabular}{ccc}
\includegraphics[width=0.32\textwidth]{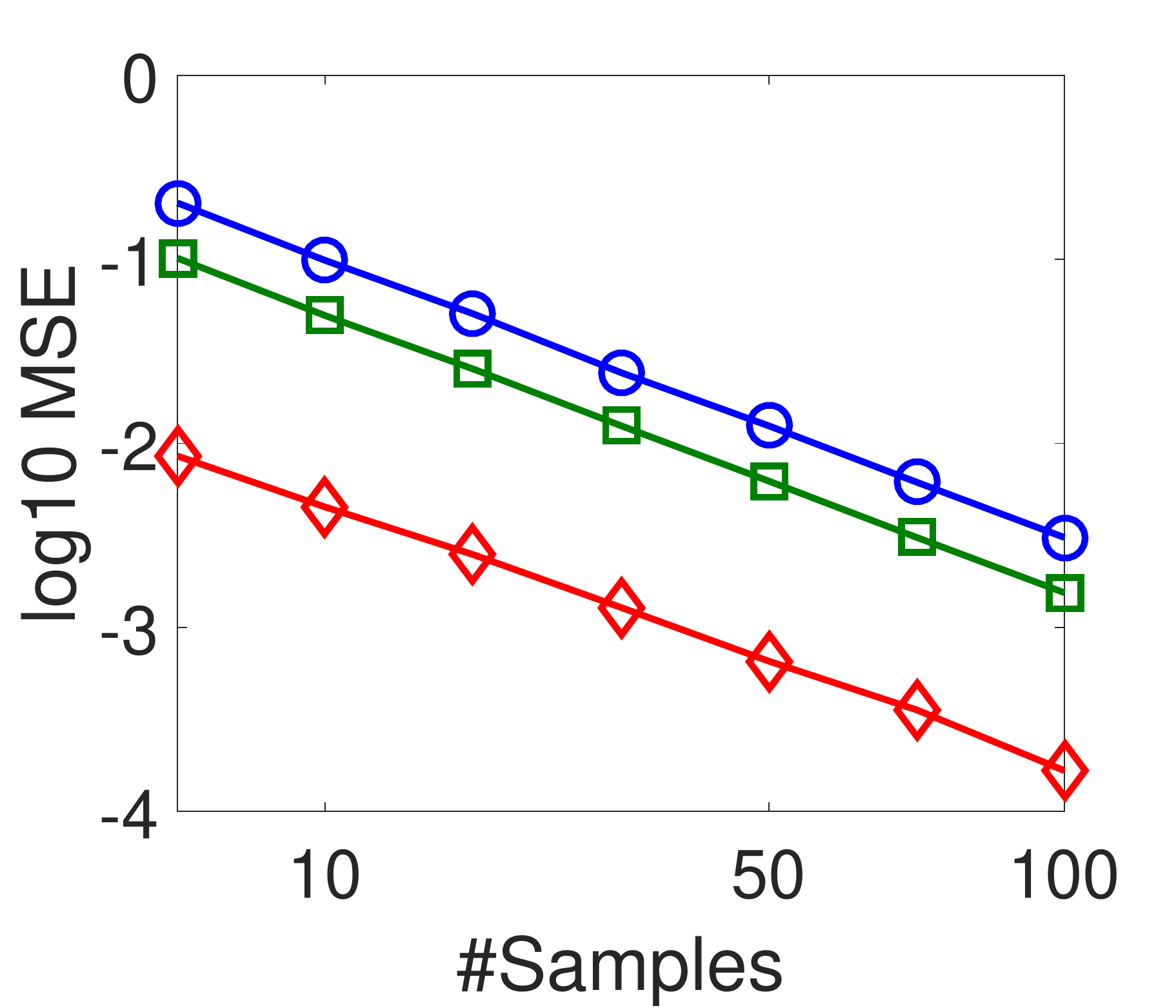} &
\includegraphics[width=0.32\textwidth]{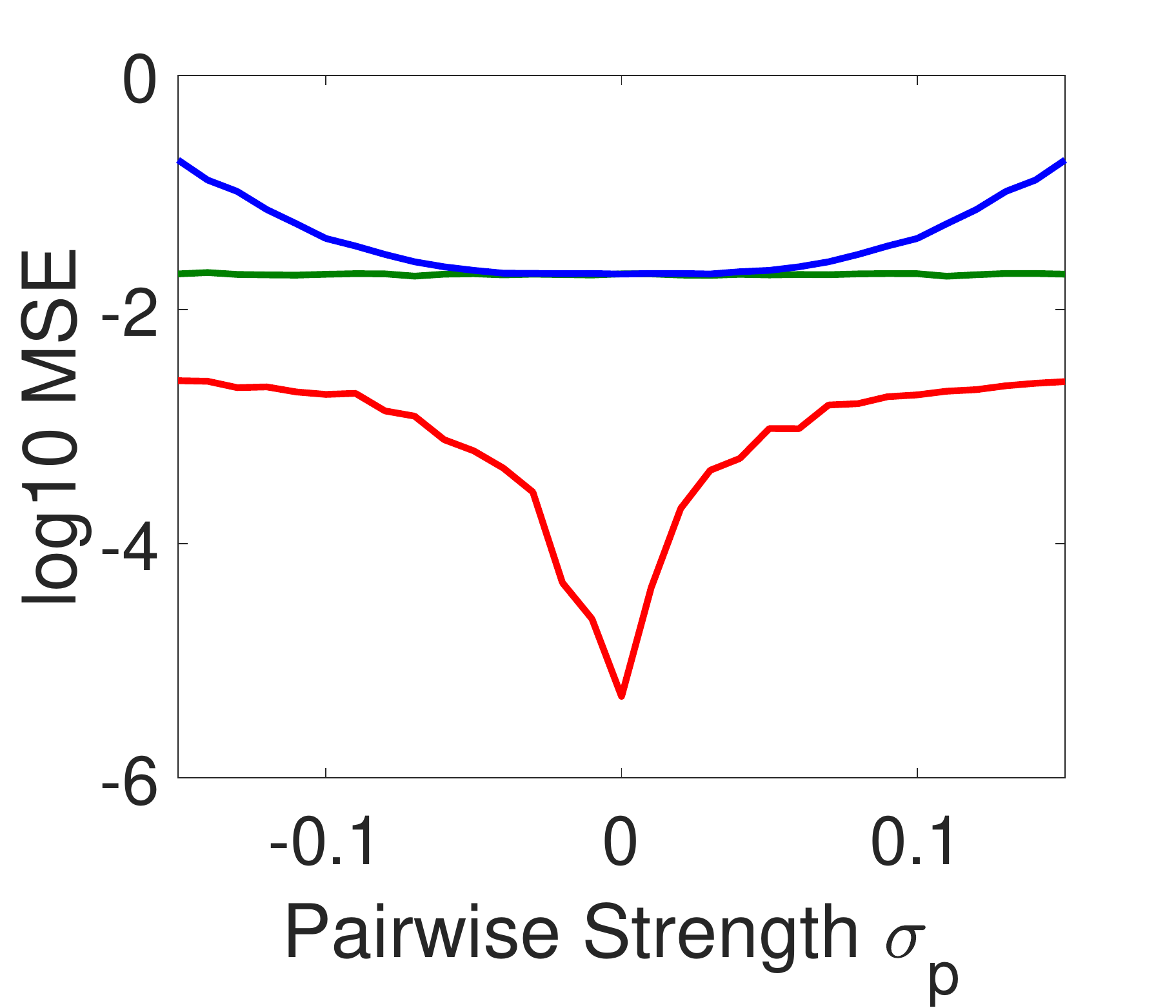} &
\raisebox{-.5em}{\includegraphics[width=0.2\textwidth]{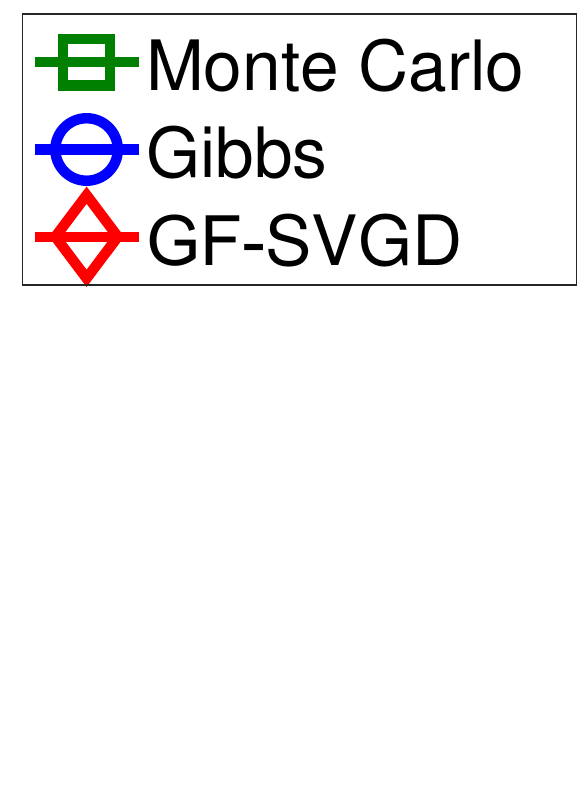}} \\
{\small (a) Fixed $\sigma_p$} &
{\small (b) Fixed $n$} &
{}
\end{tabular}
\caption[Sampling on binary Ising model evaluated with MSE metric]{Performance of different methods on the Ising model with 10x10 grid. Set $\sigma_s=0.01$ in both (a) and (b). We compute the MSE for estimating mean $\E[\vz]$ in each dimension.  In (a), we fix $\sigma_p=0.1$ and vary the sample size $n$. In (b), we fix the sample size $n=20$ and vary $\sigma_p$ from -0.15 to 0.15. In both (a) and (b) we evaluate the $\log\mathrm{MSE}$ based on 200 trails.}
\label{fig:ising}
\end{figure}


\subsection{Categorical Distribution}
We apply our algorithm to sample from one-dimensional categorical distribution $p_*(z)$ 
shown in the red bars in Fig.~\ref{fig:cat}, 
defined on $\mathcal Z := \{-1,-0.5, 0, 0.5, 1\}$ with corresponding  probabilities $\{0.1, 0.2, 0.3,0.1, 0.3\}$.  The blue dash line is the surrogate distribution $\rho(y)=p_0(x)$, where the base function $p_0(x)$ is the p.d.f. of standard Gaussian distribution. The red dash line is the transformed piecewise continuous density $p_c(x)\propto p_0(x)p_*(\Gamma(x))$, where $\Gamma(x)=a_i$ if $x\in[\eta_{i-1}, \eta_i)$ and $\eta_i$ is $i/5$-th quantile of standard Gaussian distribution. We apply Algorithm~\ref{alg:alg1} to draw a set of samples $\{x^i\}_{i=1}^n$ (shown in green dots) to approximate the transformed target distribution. Then we can obtain a set of samples $\{z^i\}_{i=1}^n$ by $z^i=\Gamma(x^i))$, to get an approximation of the original categorical distribution. 
As shown in Fig~\ref{fig:cat},
the empirical distribution
 of the discretized sample 
 $\{z^i\}_{i=1}^n$ (shown in green bars) 
aligns closely with the true distribution (the red bars) when the algorithm converges (e.g., at the 100th iteration). 

 

\subsection{Ising Model}
The Ising model~\citep{ising1924beitrag} is a widely used model in Markov random field. Consider an (undirected) graph $G=(V, E)$, where each vertex $i\in V$ is associated with a binary spin, which consists of $\vx=(x_1,\cdots,x_d)$. The probability mass function is $p(\vx)=\frac{1}{Z}\sum_{(i,j)\in E} \theta_{ij}x_ix_j$, $x_i\in\{-1, 1\}$, $\theta_{ij}$ is edge potential and $Z$ is the normalization constant, which is infeasible to calculate when $d$ is high.  

As shown in Section 3, it is easy to map $z$ to the piecewise continuous distribution of $x$ in each dimension. We take $\Gamma(x)=\sign(x)$, with the  transformed target $p_c(x)\propto p_0(x)p_*(\sign(x))$. The base function $p_0(x)$ is taken to be the standard Gaussian distribution on $\RR^d$.  
We apply GF-SVGD to sample from $p_c(x)$ with the surrogate $\rho(x)=p_0(x)$. The initial particles $\{x^i\}$ is sampled from $\mathcal{N}(-2, 1)$ and update $\{x^i\}$ by 500 iterations. We obtain $\{z^i\}_{i=1}^n$ by $z^i=\Gamma(x^i)$, which approximates the target model $p_*(z)$. We compared our algorithm with both Exact Monte Carlo (MC) and Gibbs sampling which is iteratively sampled over each coordinate and use same initialization (in terms of $z=\Gamma(x)$) and number of iterations as ours. 

Fig.~\ref{fig:ising}(a) shows the log MSE over the log sample size. With fixed $\sigma_s$ and $\sigma_p$, our method has the smallest MSE and the MSE has the convergence rate $\mathcal{O}(1/n)$. The correlation $\sigma_{p}$ indicates the difficulty of inference. As $|\sigma_{p}|$ increases, the difficulty increases. As shown in Fig.~\ref{fig:ising}(b), our method can lead to relatively less MSE in the chosen range of correlation. It is interesting to observe that as $\sigma_p\rightarrow0$, our method significantly outperforms MC and Gibbs samplimg.

\begin {algorithm*}[tbh]
\caption {GF-SVGD on training BNN}
\label{alg:GF-SVGDonBNN}
\begin {algorithmic}
\STATE {\bf Inputs}: training set $D$ and testing set $D_{\mathrm{test}}$
\STATE {\bf Outputs}: classification accuracy on testing set.
\STATE {\bf Initialize} full-precision models $\{\vw^i\}_{i=1}^n$ and its binary form $\{\vw_b^i\}_{i=1}^n$ where $\vw^i_b=\sign(\vw^i)$.
\WHILE{not converge}
\STATE -Sample $n$ batch data $\{D_i\}_{i=1}^n.$
\STATE -Calculate the true likelihood $p_c(\vw^i; D_i)\propto p_*(\sign(\vw^i);D_i)p_0(x)$ 
\STATE -Relax $\vw^i_b$ with $\sigma(\vw^i)$
            \STATE -Relax each sign activation function to the smooth function defined in \eqref{binary:approx} to get $\wt{p}$
            \STATE -Calculate the surrogate likelihood $\rho(\vw^i;D_i)\propto\wt{p}(\sigma(\vw^i);D_{i})p_0(x)$
            \STATE -$\vw^i \leftarrow \vw^i+\Delta \vw^i$, $\forall i=1,\cdots, n,$ where $\Delta \vw^i$ is defined in \eqref{bnn:update}.
            \STATE -Clip $\{\vw^i\}$ to interval $(-1, 1)$ for stability.
\ENDWHILE            
\STATE  -Calculate the probability output by softmax layer $p(\vw^i_b;D_{\mathrm{test}})$
\STATE -Calculate the average probability $f(\vw_b;D_{\mathrm{test}})\leftarrow \sum_{i=1}^{n} p(\vw^i_b;D_{\mathrm{test}})$
\STATE {\bf Output} test accuracy from $f(\vw_b;D_{\mathrm{test}}).$
\end {algorithmic}
\end {algorithm*}

\subsection{Bernoulli Restricted Boltzman Machine}
Bernoulli restricted Boltzman Machine (RBM) \citep{hinton2002training} is an undirected graphical model consisting of a bipartite graph between visible variables $\vz$ and hidden variables $h.$ In a Bernoulli RBM, the joint distribution of visible units $\vz \in \{-1, 1\}^d$ and hidden units $h \in \{-1, 1\}^M$ is given by 
\begin{equation}
p(\vz, h)\propto \exp(-E(\vz, h))   
\end{equation}
where $E(\vz, h)=-(\vz^\top W h+\vz^\top b+h^\top c)$, $W\in \mathbb{R}^{d\times M}$ is the weight, $b\in\mathbb{R}^d$ and $c\in\mathbb{R}^M$ are the bias. Marginalizing out the hidden variables $h,$ the probability mass function of $\vz$ is given by $p(\vz)=\frac{1}{\Omega}\exp(-E(\vz)),$ with free energy $E(\vz)=-\vz^\top b-\sum_k\log(1+\varphi_k),$ where $\varphi_k = \exp(W_{k*}^\top \vz + c_k)$ and $W_{k*}$ is the k-th row of $W.$ 

\begin{figure}
\begin{tabular}{cc}
\includegraphics[width=0.4\textwidth]{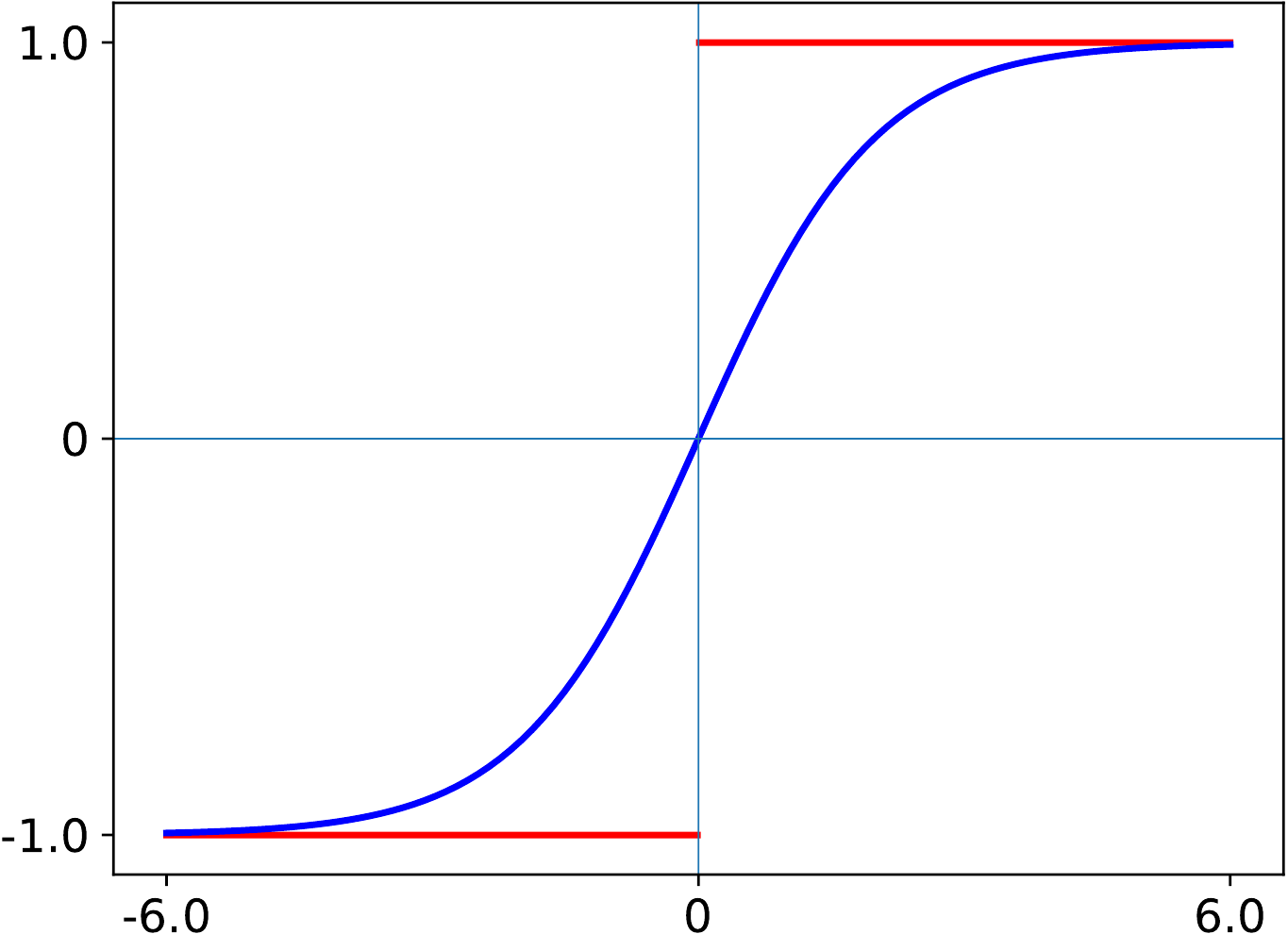} &
\hspace{-.3cm}
\includegraphics[width=0.4\textwidth]{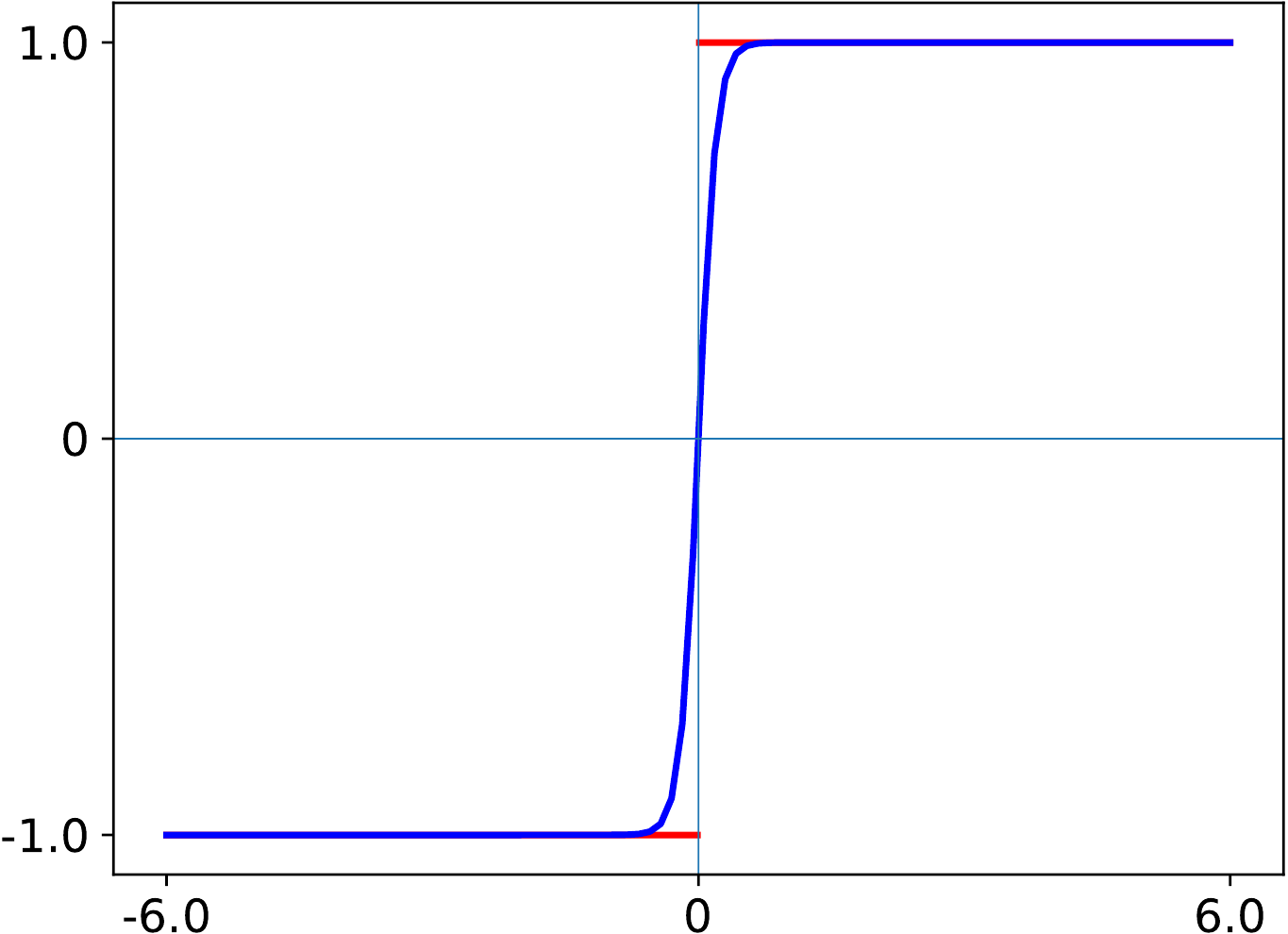} \\
{\small (a), $y=\sigma(x)$} & {\small (b), $y=\sigma(10x)$} \\
\end{tabular}
\caption[Illustration the approximation of the relaxation function of $\sign$ function]{Illustration the approximation of the relaxation function of $\sign$ function. (b) provides a better approximation than (a) by introducing a temperature parameter.\label{sign:approx}}
\end{figure}

The base function $p_0(x)$ is the product of the p.d.f. of the standard Gaussian distribution over the dimension $d.$ Applying the map $z=\Gamma(x)=\sign(x)$, the transformed piecewise continuous target is $p_c(x)\propto p_0(x)p_*(\sign(x)).$ Different from previous example, we construct a simple and more powerful surrogate distrubtion $\rho(x)\propto \wt{p}(\sigma(\vy))p_0(x)$ where $\wt{p}(\sigma(\vy))$ is differentiable approximation of $p_*(x)$ and $\sigma(x)$ is defined as 
\begin{equation}
\label{binary:approx}
\sigma(x)=\frac{2}{1+\exp(-x)}-1,    
\end{equation} 
and $\sigma(x)$ approximates $\sign(x).$ Intuitively, it relaxes the target to a differentiable surrogate with tight approximation, which is plotted in Fig.~\ref{sign:approx}.

We compare our algorithm with Gibbis sampling and discontinuous HMC(DMHC, \citet{nishimura2017discontinuous}). In Fig.~\ref{fig:discreterbm}, $W$ is drawn from $N(0, 0.05)$, both $b$ and $c$ are drawn from $N(0, 1).$
With $10^5$ iterations of Gibbs sampling, we draw 500 parallel chains to take the last sample of each chain to get 500 ground-truth samples. We run Gibbs, DHMC and GF-SVGD at 500 iterations for fair comparison. In Gibbs sampling, $p(\vz\mid h)$ and $p(h\mid\vz)$ are iteratively sampled. In DHMC, a coordinate-wise integrator with Laplace momentum is applied to update the discontinuous states. We calculate the maximum mean discrepancy (MMD, \citet{gretton2012kernel}) between the ground truth sample and the sample drawn by different methods. The kernel used in MMD is the exponentiated Hamming kerenl from \citet{yang2018goodness}, defined as,
$k(\vz, \vz')=\exp(-H(\vz, \vz')),$ where $H(\vz, \vz'):=\frac{1}{d}\sum_{i=1}^d \mathbb{I}_{\{z_i \neq z_i'\}}$ is normalized Hamming distance. We perform experiments by fixing $d=100$ and varying sample size in Fig.~\ref{fig:discreterbm}(a) and fixing $n=100$ and varying $d$, the dimension of visible units. Fig.~\ref{fig:discreterbm}(a) indicates that the samples from our method match the ground truth samples better in terms of MMD. Fig.~\ref{fig:discreterbm}(b) shows that the performance of our method is least sensitive to the dimension of the model than that of Gibss and DHMC. Both  Fig.~\ref{fig:discreterbm}(a) and  Fig.~\ref{fig:discreterbm}(b) show that our algorithm converges fastest.     

\begin{figure}[h]
\centering
\begin{tabular}{cc}
\includegraphics[height=0.25\textwidth]{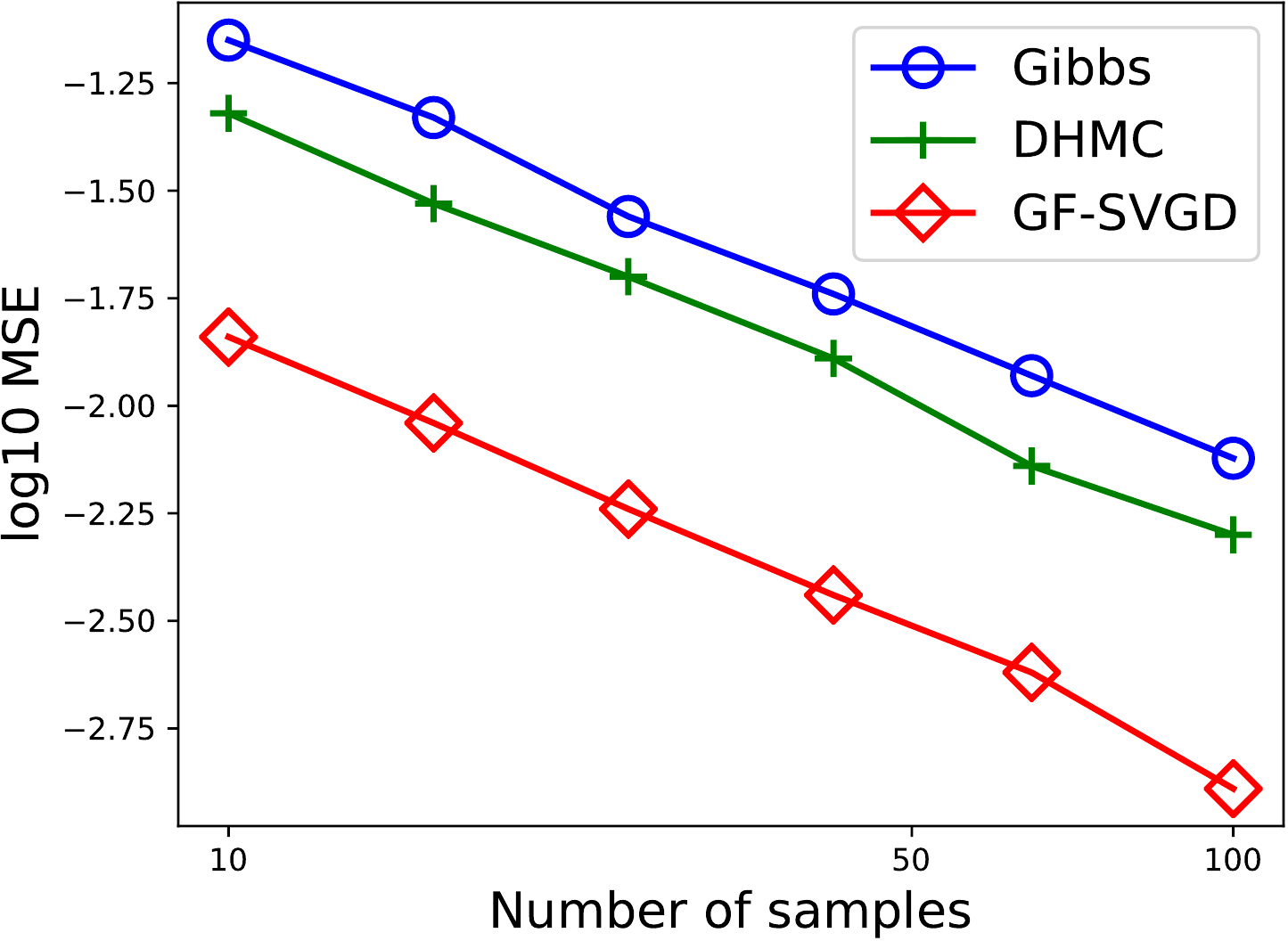}&
\includegraphics[height=0.25\textwidth]{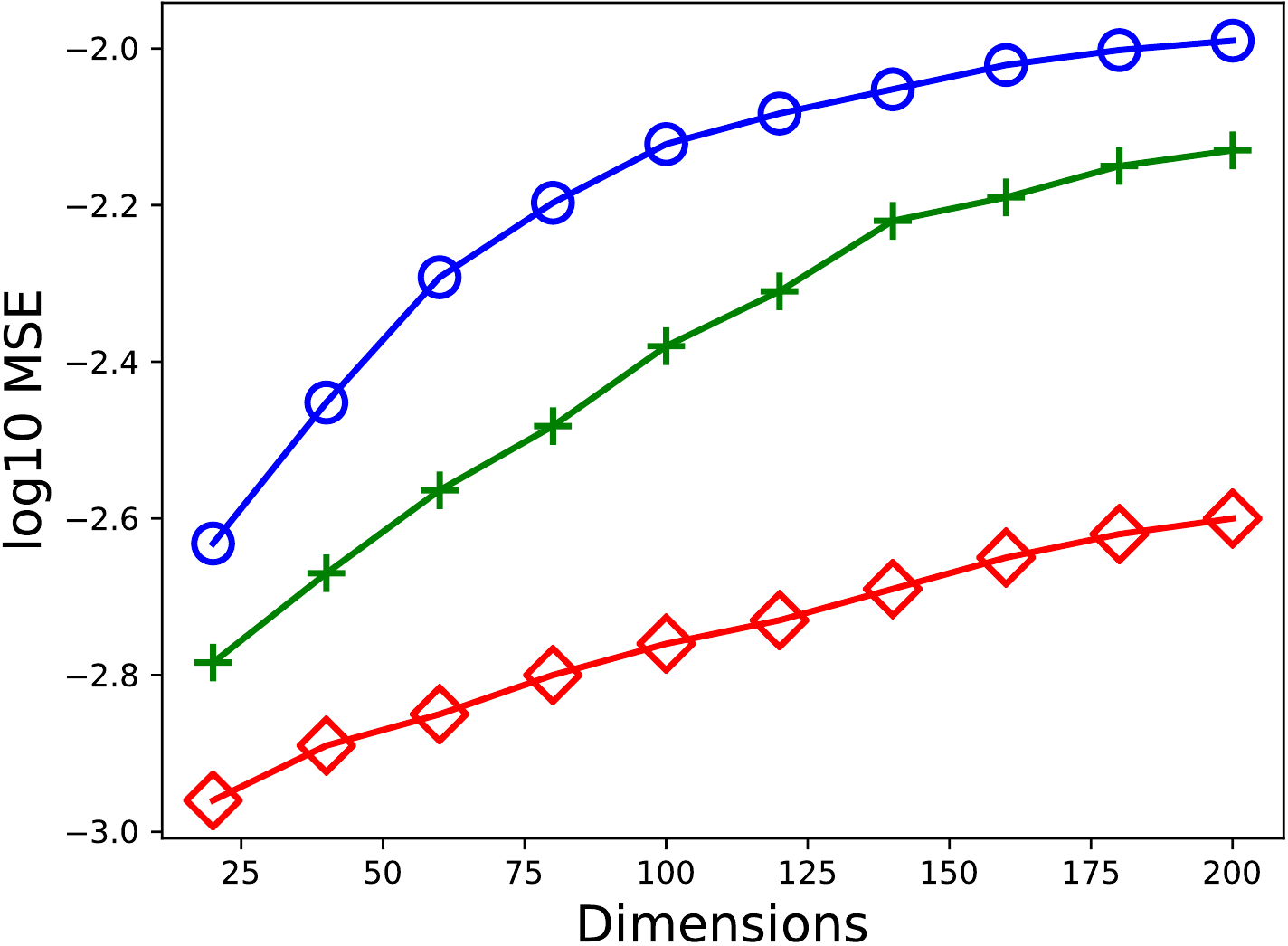}\\ 
{\small (a)  Fix dimension} &
{\small (b)  Fix sample size}
\end{tabular}
\caption[Sampling on Bernoulli RBM evaluated with MSE metric.]{\small Bernoulli RBM with number of visible units $M=25$. In (a), we fix the dimension of visible variables $d=100$ and vary the number of samples $\{\vz^j\}_{j=1}^n$. In (b), we fix the number of samples $n=100$ and vary the dimension of visible variables $d$. We evaluate the MSE between the estimator and the ground truth quantity. \label{fig:discreterbm}}
\end{figure}

\begin{figure}[h]
\centering
\begin{tabular}{cc}
\includegraphics[height=0.26\textwidth]{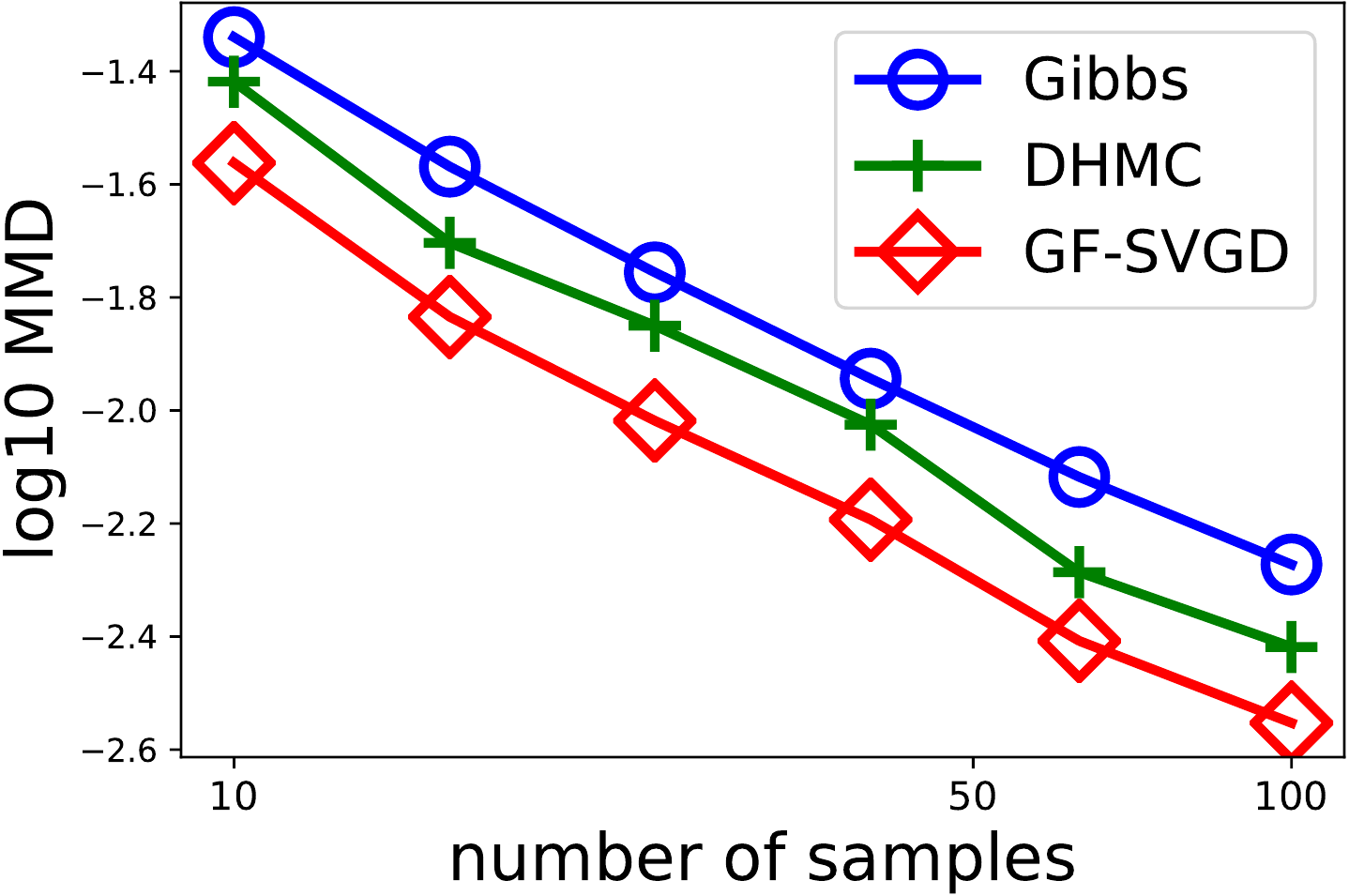} &
\includegraphics[height=0.26\textwidth]{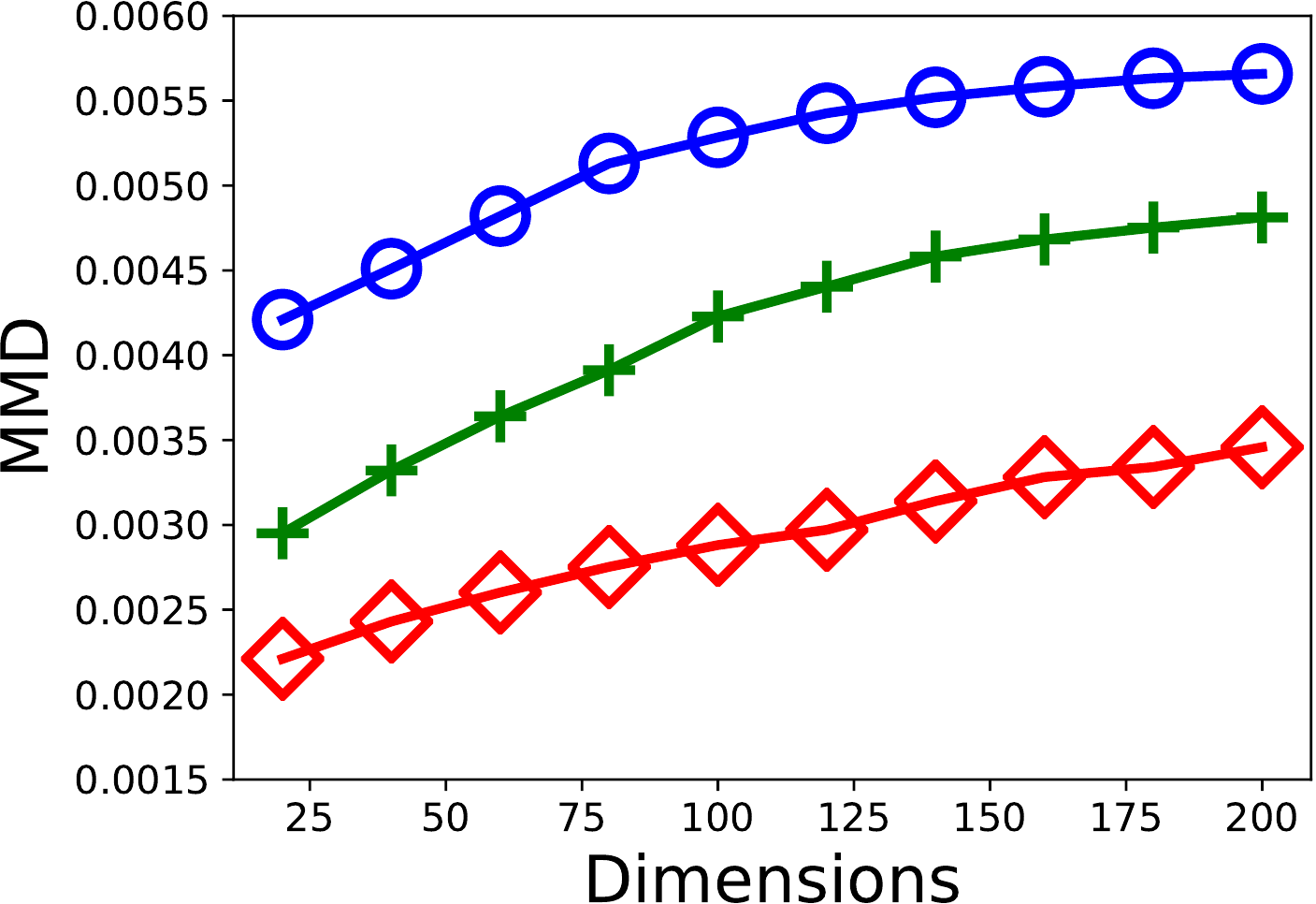} \\ 
{\small (a)  Fix dimension} &
{\small (b)  Fix sample size}
\end{tabular}
\caption[Sampling on Bernoulli RBM evaluated with MMD metric]{\small Bernoulli RBM with $M=25$. In (a), we fix the dimension of visible variables $d=100$ and vary the number of samples $n$. In (b), we fix $n=100$ and vary $d$. We calculate the MMD between the sample of different methods and the ground-truth sample. \label{fig:discreterbm}}
\end{figure}

\section{Ensemble Learning on Binarized Neural Networks}
We slightly modify our algorithm to the application of training binarized neural network (BNN), where both the weights and activation functions are binary $\pm 1$. BNN has been studied extensively because of its fast computation, energy efficiency and low memory cost \citep{rastegari2016xnor, hubara2016binarized, darabi2018bnn+, zhu2018binary}. The challenging problem in training BNN is that the gradients of the weights cannot be backpropagated through the binary activation functions because the gradients are zero almost everywhere. \citet{bengio2013estimating} proposes to use the gradients of identity function, Relu or leaky Relu to approximate the gradients of binary activation functions, which are known as straight-through estimators. While these estimators work well in some cases, their theorectical understanding is largely unexplored although some initial theorectical result in the simple setting has been proposed recently \citep{yin2018understanding}.  

\begin{figure}[tbh]
\centering
\begin{tabular}{c}
\includegraphics[width=0.9\textwidth]{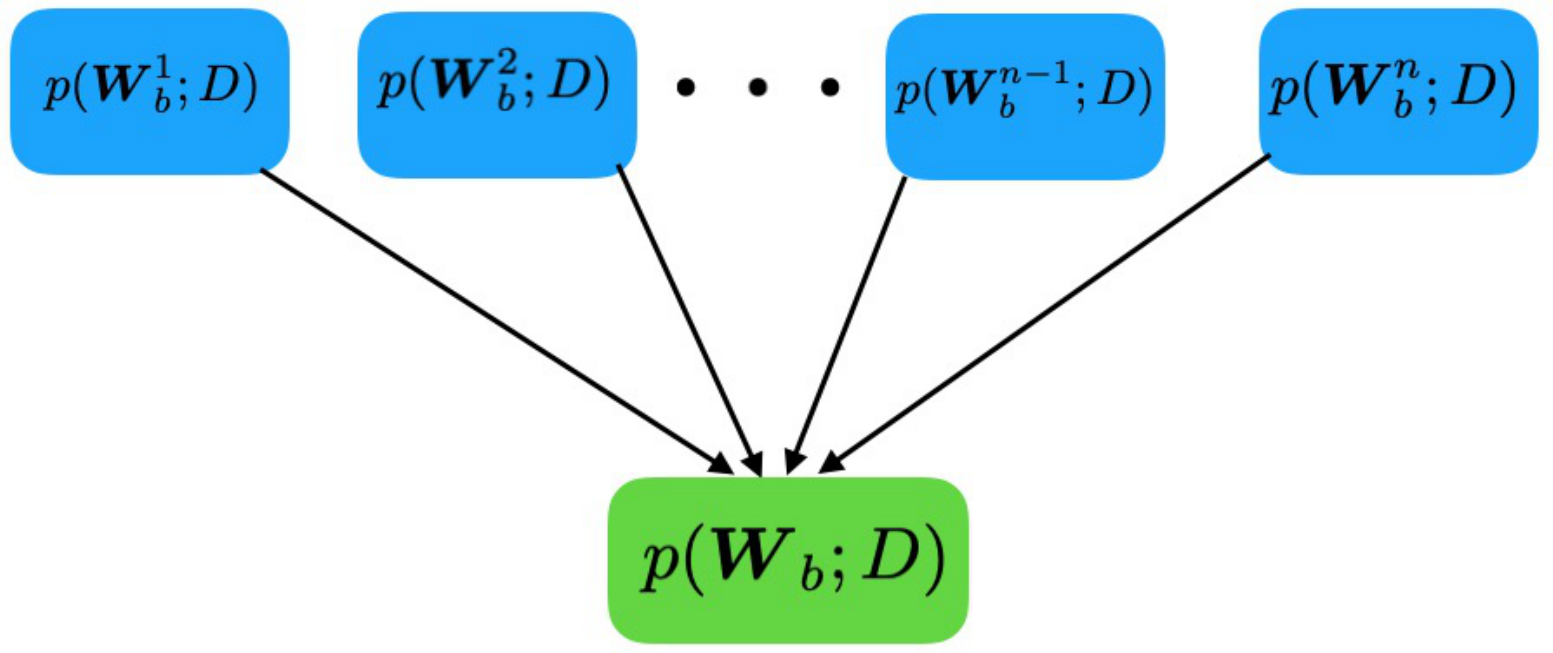} 
\end{tabular}
\caption[Illustration of our principled BNN ensemble learning]{Ensemble learning by drawing n samples $\{p(\bd{W}^i_b; D)\}_{i=1}^n$ from the posterior $p(\bd{W}_b; D);$ prediction model $p(\bd{W}_b; D)=\frac{1}{n}\sum_{i=1}^n p(\bd{W}^i_b; D).$
\label{fig:ensem}} 
\end{figure} 

We train an ensemble of $n$ neural networks (NN) with the same architecture ($n\ge 2$). Let $\vw^i_b$ be the binary weight of model $i$, for $i=1,\cdots, n$, and $p_*(\vw^i_b;D)$ be the target probability model with softmax layer as last layer given the data $D$. Learning the target probability model is framed as drawing $n$ samples $\{\vw^i_b\}_{i=1}^n$ to approximate the posterior distribution $p_*(\vw_b; D)$. 
\begin{figure}
    \begin{tabular}{cc}
\includegraphics[height=0.4\textwidth]{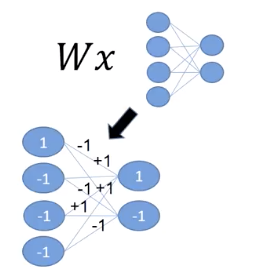} & 
\includegraphics[height=0.34\textwidth]{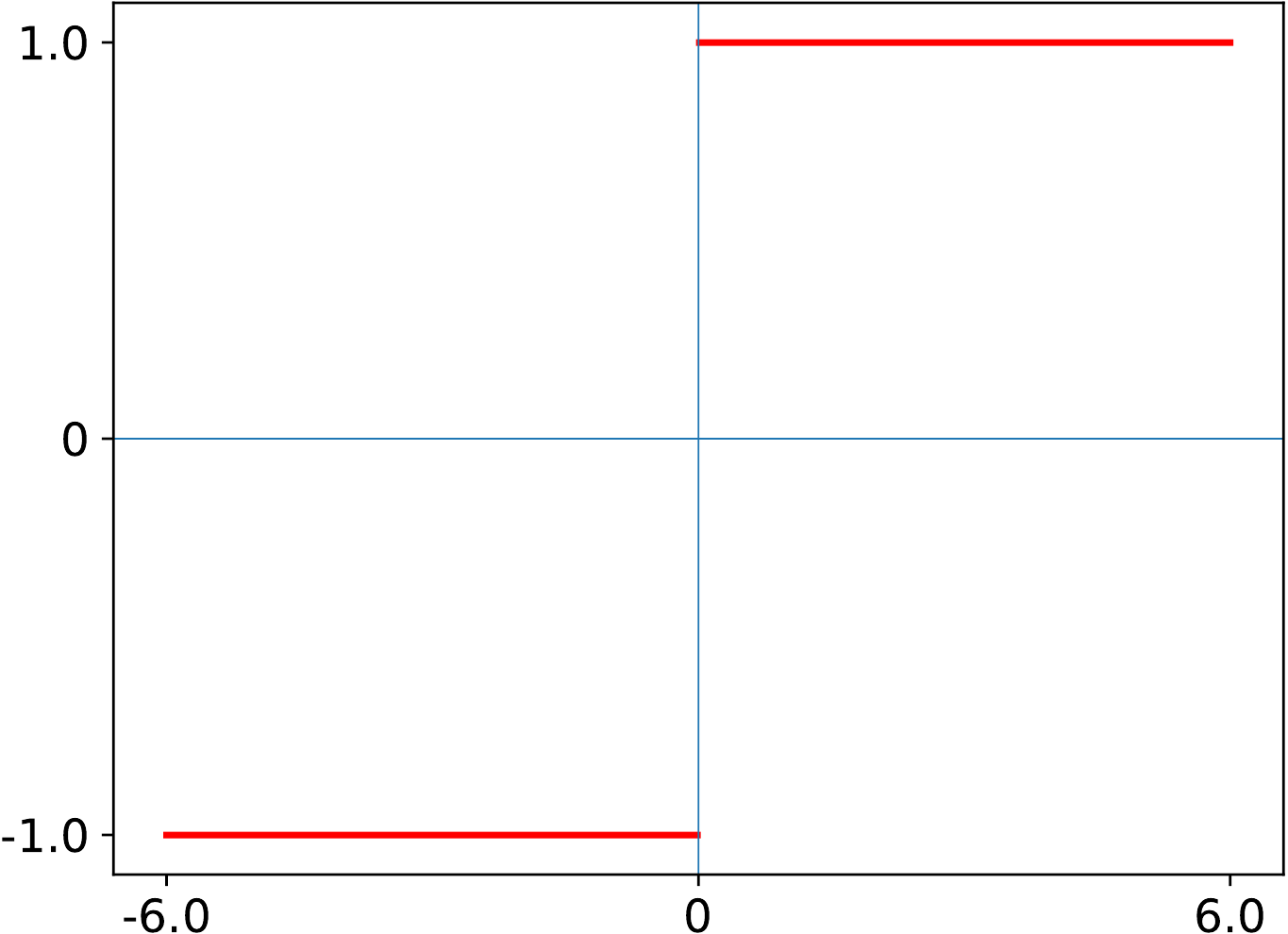} \\ 
{(a), BNN} & {(b), Activation}
\end{tabular}
\caption[Illustration of binarized neural networks with binary weights and activation functions]{Illustration of  binarized neural networks(BNN) with binary weights and activation functions.}
\end{figure}

We apply multi-dimensional quantile transformation $\vF$ to transform the original discrete-valued target to the target distribution of real-valued $\vw\in\R^d$. Let $p_0(w)$ be the base function, which is the product of the p.d.f. of the standard Gaussian distribution over the dimension $d.$ Based on the derivation in Section 3, the distribution of $\vw$ has the form $p_c(\vw;D)\propto p_*(\sign(\vw);D)p_0(\vw)$ with weight $\vw$ and the $\sign$ function is applied to each dimension of $\vw$. To backpropagate the gradient to the non-differentiable target, we construct a surrogate probability model $\rho(\vw;D)$ which approximates $\sign(\vw)$ in the transformed target by $\sigma(\vy)$ and relax the binary activation function $\{-1, 1\}$ by $\sigma$, where $\sigma$ is defined by \eqref{binary:approx}, denoted by $\wt{p}(\sigma(\vw);D)p_0(\vw)$. Here $\wt{p}(\sigma(\vw);D)$ is a differentiable approximation of $p_*(\sign(\vw);D).$  Then we apply GF-SVGD to update $\{\vw^i\}$ to approximate the transformed target distribution of $p_c(\vw;D)$ of $\vw$ as follows, 
$\vw^i \leftarrow \vw^i+\frac{\epsilon_{i}}{\Omega}\Delta \vw^i$, $\forall i=1,\cdots, n,$
\begin{equation}\label{bnn:update}
 \Delta \vw^i \!\! \leftarrow \!\! \! \sum_{j=1}^n \! \gamma_j [\nabla_{\vw}\log \rho(\vw^j;\!D_i)k(\vw^j\!,\!\vw^i)
             +\!\nabla_{\vw^j} k(\vw^j\!,\!\vw^i)]   
\end{equation}
where $D_i$ is batch data and $\mu_j =\rho(\vw^j; D_i)/p_c(\vw^j;D_i)$, $H(t) \overset{\mathrm{def}}{=}\sum_{j=1}^n \mathbb{I}(\mu_j\ge t)/n$, $\gamma_j= (H(w_j))^{-1}$ and $\Omega=\sum_{j=1}^n \gamma_j$. Note that we don't need to calculate the cumbersome term $p_0(w)$ as it can be canceled from the ratio between the surrogate distribution and the transformed distribution.  In practice, we find a more effective way to estimator this density ratio denoted by $\gamma_j$. Intuitively, this corresponds to assigning each particle a weight according to the rank of its density ratio in the population. After training the model, we make a prediction on test data $D$ by linearly averaging, 
\begin{equation}
p(\vw_b;D)=\frac1n \sum_{i=1}^n p(\vw_b^i;D).  
\end{equation}
Algorithm \ref{alg:GF-SVGDonBNN} can be viewed as a new form of the ensemble method for training neural networks models with discrete parameters by drawing a set of samples $\{\vw_b^i \}$ from the posterior $p(\vw_b;D).$

\begin{figure}
\centering
\begin{tabular}{cc}
\includegraphics[height=.3\textwidth]{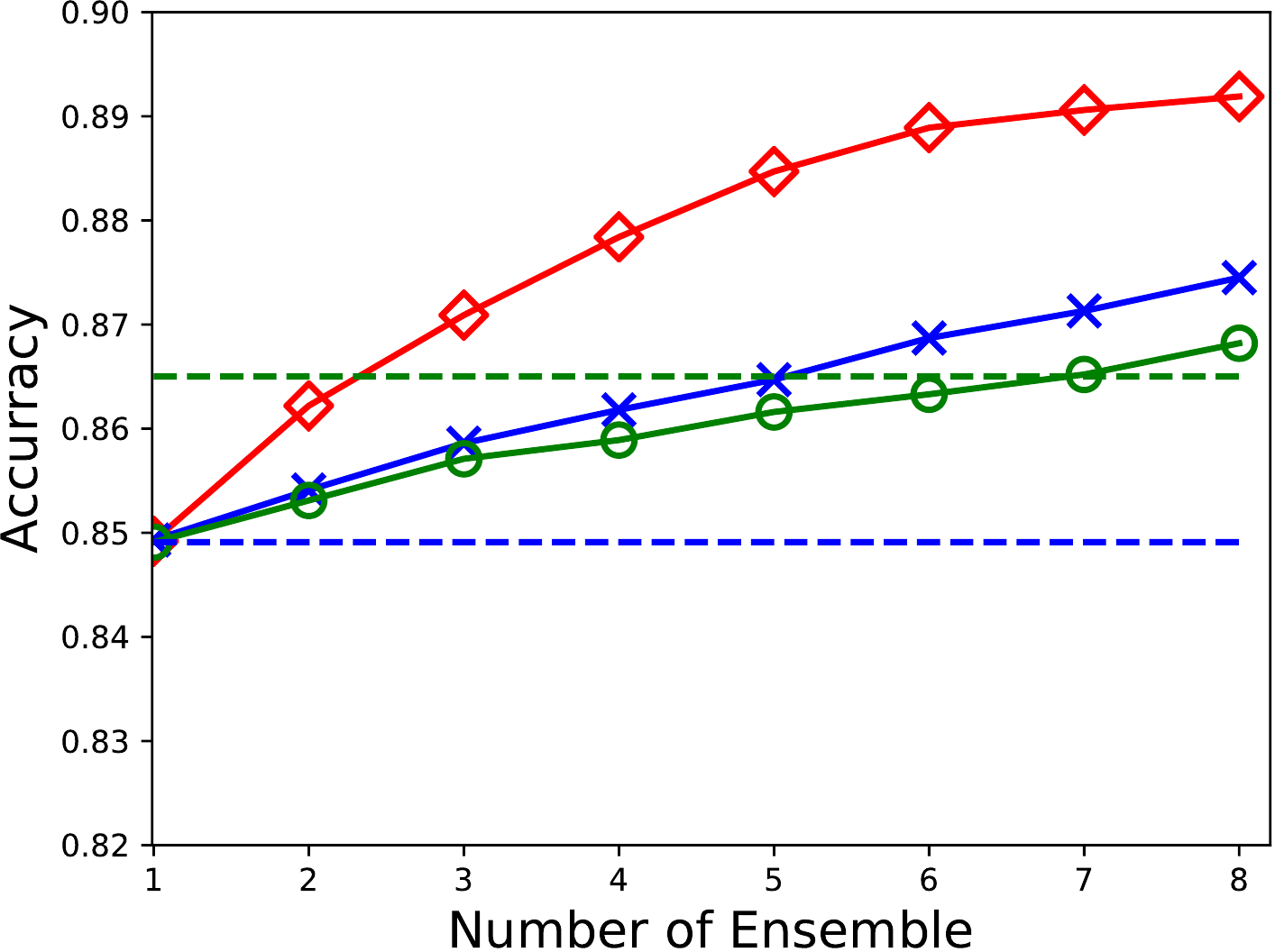}&
\raisebox{2em}{\includegraphics[height=0.15\textwidth]{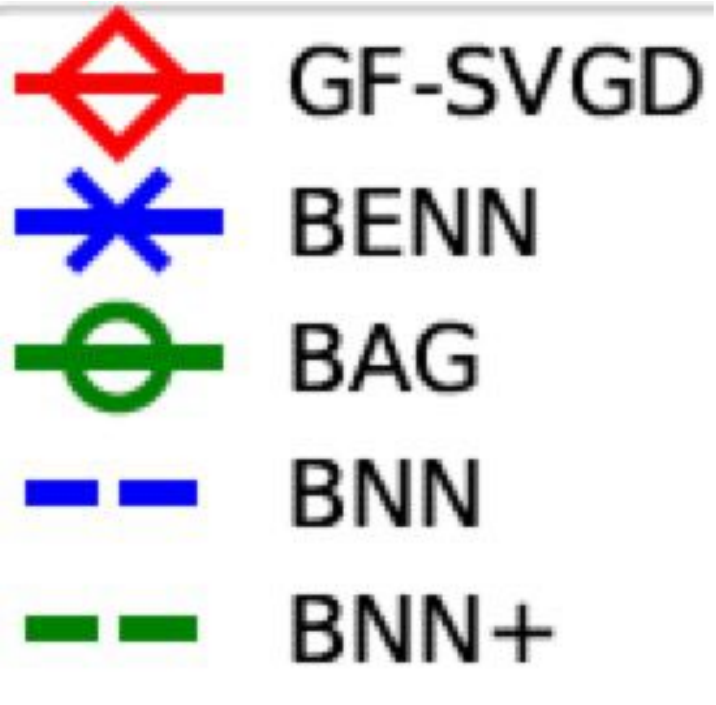}}
\end{tabular}
\caption[Performance comparison of different methods on binarized AlexNet on CIFAR10 dataset]{Comparison of different methods using AlexNet with binarized weights and activations on CIFAR10 dataset. We compare our GF-SVGD with BNN \citep{hubara2016binarized}, BNN+\citep{darabi2018bnn+} and BENN \citep{zhu2018binary}. "BAG" denote models are independently trained and linearly averaged the softmax output for prediction. Performance is based on the accuracy of different models w.r.t. ensemble size $n$ on test data.}
\label{fig:bnn}
\end{figure}

We test our ensemble algorithm by using AlexNet \citep{krizhevsky2012imagenet} on CIFAR-10 dataset. We use the same setting for AlexNet as that in \citet{zhu2018binary}, where the detail can be found in Appendix A. We compare our ensemble algorithm with typical ensemble method using bagging and AdaBoost (BENN, \citet{zhu2018binary}), BNN \citep{hubara2016binarized} and BNN+\citep{darabi2018bnn+}. Both BNN and BNN+ are trained on a single model with same network structure. From Fig.~\ref{fig:bnn}, we can see that all three ensemble methods (GF-SVGD, BAG and BENN) improve test accuracy over one single model (BNN and BNN+). To use the same setting for all methods, we don't use data augmentation or pre-training. Our ensemble method has the highest accuracy among all three ensemble methods. This is because our ensemble model are sufficiently interactive at each iteration during training and our models $\{\vw^i\}$ in principle are approximating the posterior distribution $p(\vw;D).$


\section{Summary} In this chapter, we propose a new sampling algorithm to sample from the discrete-valued target distributions $p_*(\vz)$. SVGD and GF-SVGD were originally proposed to continuous-valued target distributions. GF-SVGD leverages the gradient information of the surrogate distribution $\rho(\vx)$ and corrects the bias with a form of importance weights. To exploit GF-SVGD to perform sampling on discrete-valued distributions, we first transform the discrete-valued distributions to the corresponding continuous-valued distributions. We propose a simple yet powerful framework for transforming discrete distributions to equivalent piecewise continuous distribution, on which the differentiable surrogate distribution in GF-SVGD is easy to construct. Our empirical results show that our method consistently outperforms traditional algorithms such as Gibbs sampling and discontinuous Hamiltonian Monte Carlo on various challenging benchmarks of discrete graphical models.

In addition, we demonstrate that our method provides a promising tool for learning an
ensemble model of binarized neural network (BNN), outperforming other widely used ensemble methods on learning binarized AlexNet on CIFAR-10 data. We frame learning an ensemble BNN on dataset $D$ as drawing a set of models $\{ p(\vw_b^i;D)\}_{i=1}^n$ from the posterior $p(\vw_b;D).$ Then to make a prediction on test data, we just linearly average 
$$
p(\vw_b;D)=\sum_{i=1}^n p(\vw_b^i;D).
$$ 
Since $p(\vw_b^i;D)$ is the softmax layer of the last layer in each neural network and normalized, we just need to sum their probability without introducing 1/n to average them. Our ensemble algorithm is welled justified from the Bayesian perspective. Our ensemble method of learning BNN provides a new way to train BNN. Future research includes applying our ensemble method to training BNN with larger networks such as VGG net and  larger dataset such as ImageNet dataset.

In the next chapter, we will leverage our derived gradient-free kernelized Stein discrepancy to perform the goodness-of-fit test on discrete distributions by first transforming the discrete distribution to its continuous counterpart using the transformation constructed in this chapter. 
\chapter{Goodness-of-fit testing on Discrete Distributions\label{chap:gof}}
We have introduced two algorithms in Chapter \ref{chap:is} and Chapter \ref{chap:gf} to perform approximate inference on continuous-valued distributions and one algorithm to sample from discrete-valued distributions Chapter \ref{chap:disc}. The fundamental problem for these three algorithms to tackle is to draw a set of samples $\{\vx_i\}_{i=1}^n$ to estimate $\E_p[f(\vx)]$ for any interested function $f(\vx).$ In this chapter, we will leverage the results in previous chapters to perform a different task, the goodness-of-fit test on discrete distributions. Goodness-of-fit testing measures how well the model $p_*(\vz)$ fits the observed data $\{\vz_i\}_{i=1}^n$, which is drawn from some unknown distribution $q(\vz)$. Goodness-of-fit test usually performs null hypothesis $H_0: q = p$ versus alternative hypothesis $H_1: q \neq p.$ Then gradient-free kernelized Stein discrepancy in Chapter~\ref{chap:gf} can be applied to construct some justified statistics to perform goodness-of-fit testing. The details of our proposed method will be introduced in the following section.

Classical goodness-of-fit tests on discrete distributions includes $\chi^2$ test~\citep{pearson1900x}, the Kolmogorov-Smirnov test~\citep{kolmogorov1933sulla, smirnov1948table} and the Anderson-Darling test~\citep{anderson1954test}. These tests usually assume the model is fully specified and easy to calculate, which cannot be applied to modern complex models with intractable normalization constants.

\section{Goodness-of-fit testing Algorithm}
In this section, we first review gradient-free KSD and then illustrate how it can be leveraged to propose our goodness-of-fit testing algorithm on discrete distributions~\citep{han2020stein}. As the gradient-free KSD is applied to continuous-valued distributions, we first transform the target distribution $p_*(\vz)$ and the data to a continuous-valued distribution by one-by-one transform. Then we perform the goodness-of-fit test to the transformed data and distribution by naturally choosing gradient-free KSD.

\paragraph{Goodness-of-fit Test on Discrete Distributions} 
We are given i.i.d. samples $\{\vz_i\}_{i=1}^n$ from some unknown distribution $q_*(\vz),$ $\vz \in \{\va_1,\va_2,\cdots, \va_{K_q}\},$ where $K_q$ is the number of discrete states in $q_*$,  and a candidate discrete distribution $p_*(\vz)$, $\vz \in \{\va_1,\va_2,\cdots, \va_{K_p}\},$ where $K_p$ is the number of discrete states in $p$. $K_q$ might not be equal to $K_p.$
We would like to measure the goodness-of-fit of the model $p_*(\vz)$ to the observed data $\{\vz_i\}_{i=1}^n.$ We conduct hypothesis test as follows:
\begin{equation*}
\text{null hypothesis}~ H_0:q_*=p_*~ \mathrm{vs.}~ \text{alternative hypothesis} ~H_1:q_*\neq p_*.    
\end{equation*}

\subsection{Gradient-Free KSD} 
The gradient-free KSD leverages the gradient information of the surrogate distribution and corrects the bias in KSD \citep{liu2016kernelized, gong2019quantile, wang2019stein} with a form of importance weights.
As shown in Theorem~\ref{bbis:ksd} in Chapter \ref{chap:gf}, with the choice of kernel $w(\vx)k(\vx,\vx')w(\vx')$ in RKHS $\H_d$, the square of the gradient-free KSD between $q(\vx)$ and $p(\vx)$ is 
\begin{equation}
\label{gof:ksd}
\wt{\mathcal{S}}(q, p) = \E_{\vx, \vx'\sim q}[w(\vx)k_{\rho}(\vx, \vx') w(\vx')],  
\end{equation}
where $w(\vx)=\rho(\vx)/p(\vx)$ and  $\kappa_{\rho}(\bd{x},  \bd{x}')$ is defined as,
\begin{align}
\label{imp:kernel}
\!\! \kappa_{\rho}(\bd{x},  \bd{x}')\!\! & = \!\! \bd{s}_{\rho}(\bd{x})^\top k(\bd{x},\bd{x}')\bd{s}_{\rho}(\bd{x}')
+\bd{s}_{\rho}(\bd{x})^\top \nabla_{\bd{x}'}k(\bd{x},\bd{x}') \\ \notag
 & \!\! +\bd{s}_{\rho}(\bd{x}')^\top \nabla_{\bd{x}}  k(\bd{x},\bd{x}')\!\! +\!\!\nabla_{\bd{x}}\!\cdot\!(\nabla_{\bd{x}'}k(\bd{x}, \bd{x}')),
\end{align}
$\bd{s}_{\rho}(\bd{x})$ is the score function of the surrogate distribution $\rho(\vx).$
Note that in order to calculate the gradient-free KSD between $q(\vx)$ and $p(\vx)$, we only need the evaluation of $p(\vx)$ and the gradient information of the surrogate distribution $\rho(\vx)$. The gradient-free KSD can be mapped back to the original KSD by choosing the kernel in \eqref{gof:ksd} as $k(\vx,\vx')/(w(\vx)w(\vx'))$. Therefore, the gradient-free KSD inherits all theoretical propoerties of the original KSD~\citep{liu2016kernelized} and is a natural choice for goodness-of-fit test.

In order to apply GF-KSD to goodness-of-fit test on discrete distributions, we need to transform the discrete-valued distribution and discrete data to continuous-valued distribution respectively. In the following, let us first review the key steps to transform a discrete-valued distribution to the corresponding continuous-valued distribution. Then we will discuss procedures of transforming discrete data to the corresponding continuous-valued data. 

Let $p_*(\vz)$ discrete distribution, defined on a finite discrete set $\mathcal Z=\{\va_1,\ldots, \va_K\}$. Each $\va_i$ is a $d$-dimensional vector of discrete values. 
Now we review our idea to construct a piecewise continuous-valued distribution $p_c(\vx)$ for $\vx\in  \RR^d$, and a map $\Gamma\colon \RR^d \to \mathcal Z$, 
such that the distribution of $\vz = \Gamma(\vx)$ is $p_*$ when $\vx\sim p_c$, which has been discussed in Chapter~\ref{chap:disc}.

\begin{mydef}
A piecewise continuous distribution $p_c$ on $\RR^d$ and map $\Gamma\colon \RR^d 
\to \mathcal Z$ is called to form a 
\textbf{continuous parameterization} of  $p_*$, if $\vz = \Gamma(\vx)$ follows $p_* $ when $\vx\sim p_c$. 
\end{mydef}

\paragraph{Even Partition} 
Our method starts with choosing a simple base distribution $p_0$, 
which can be the standard Gaussian distribution.  
We then construct a map $\Gamma$ that \emph{evenly partition} $p_0$ into
several regions with equal probabilities. 
\begin{mydef} A map $\Gamma \colon \Z \to \RR^d$ is said to 
\textbf{evenly partition} 
 $p_0$ if we have  
\begin{align} \label{even}
\int_{\RR^d} 
p_0(\vx) \ind[\va_i = \Gamma(\vx)] d\vx = \frac{1}{K}, 
\end{align}
for $i=1,\ldots K$. This is equivalent to saying that 
$p_0$ and $\Gamma$ forms a continuous  relaxation of the uniform distribution $q_*(\va_i) = 1/K$. 
\end{mydef}

For simple $p_0$ such as standard Gaussian distributions, it is straightforward to construct even partitions using the quantiles of $p_0(\vx)$.  
For example,
in the one dimensional case $(d=1)$, we can evenly partition any continuous $p_0(\vx)$, $\vx\in \RR$ by  
\begin{align}\label{equ:gamma1D}
\Gamma(\vx) = \va_i ~~~~~
\text{if ~~ $\vx \in [\eta_{i-1}, ~~ \eta_{i})$}, 
\end{align}
where $\eta_i$ denotes the $i/K$-th quantile of distribution $p_0$. 
In multi-dimensional cases ($d>1$) and when $p_0$ is a product distribution:
\begin{equation}
\label{multiconti:surr}
p_0(\vx) = \prod_{i=1}^d p_{0,i}(x_i).     
\end{equation}

One can easily show that an even partition can be constructed by concatenating one-dimensional even partition: $\Gamma(\vx)=(\Gamma_{1}(x_1),\cdots, \Gamma_{d}(x_d)),$ where $\vx=(x_1,\cdots,x_d)$ and $\Gamma_{i}(\cdot)$ an even partition of $p_{0,i}$.  
A particularly simple case is when $\vz$ is a binary vector, i.e., $\mathcal Z = \{\pm1 \}^d$, in which case $\Gamma(\vx) = \sign(\vx)$ evenly partitions any distribution $p_0$ that is symmetric around the origin. 

\paragraph{Weighting the Partitions} 
Given an even partition of $p_0$, we can conveniently construct 
a continuous parameterization of an arbitrary discrete distribution $p_*$ by 
\emph{weighting each bin of the partition with corresponding probability in $p_*$}, that is, we may  construct $p_c(\vx)$ by 
\begin{align}\label{equ:pc}
p_c(\vx) \propto p_0(\vx) p_*(\Gamma(\vx)),  
\end{align} 
where $p_0(\vx)$ is weighted by $p_*(\Gamma(\vx))$, the probability
of the discrete value $\vz=\Gamma(\vx)$ that $\vx$ maps.

\begin{mydef} 
The even partition naturally defines a corresponding {\bf \emph{stepwise distribution}} $p(\bd{y})$, $\bd{y}\in [0, 1)^d,$ as follow: let $F(\vx)$ is c.d.f. of $p_0(\vx)$ (Gaussian or GMM p.d.f.) in each dimension. If for any $\bd{y}$, there exists unique $\vx=F^{-1}(\bd{y})$, $\vx\in[\eta_{i-1}, \eta_i),$  $p(\bd{y}):=p_*(\va_i).$ \end{mydef}

\paragraph{Derivation of $p_c$ from variable transform formula} $p_c$ can be derived from the formula of invertible variable transform. The distribution of $\vx$, $\vx=F^{-1}(\bd{y}),$ is $p(F(\vx))\mathrm{det}(F'(\vx))$, i.e.,%
\begin{equation}
\label{def:vari:transf}
p_c(\vx) \! =p(F(\vx))\prod_{i=1}^d p_{0,i}(\vx)=p_{0}(\vx)p(F(\vx)). 
\end{equation}

\begin{figure}[h]
\centering
\begin{tabular}{c}
\includegraphics[width=0.7\textwidth]{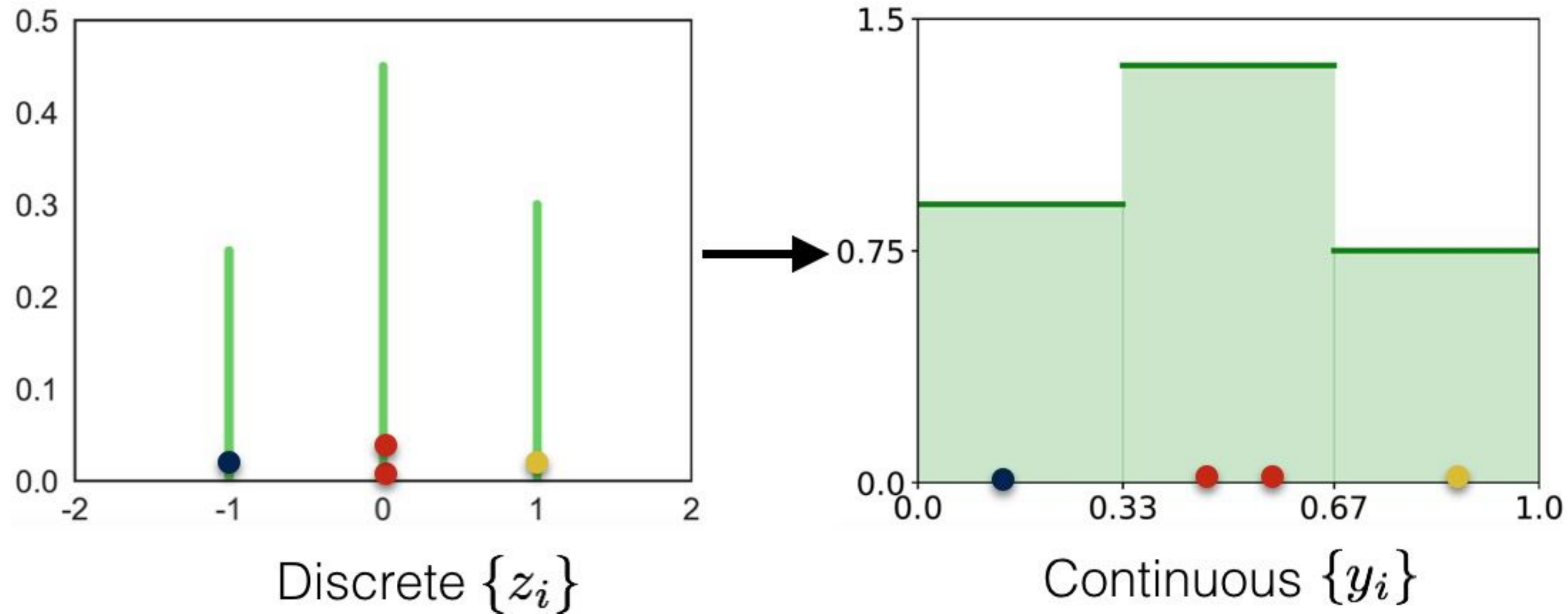}\\
\end{tabular}
\caption[Illustration of transforming the discrete data to continuous data in 1D categorical distribution]{Illustration of transforming the discrete data to continuous data in 1D categorical distribution. The unknown categorical distribution has three states. We correspond each discrete state in the left figure to each unique interval in the right figure. Different colors in left figure means data at different discrete states. Given one sample in left figure, we randomly sample one continuous-valued sample in its corresponding interval in right figure. \label{fig:tran:dis:cont}}
\end{figure}

\paragraph{Transform Discrete Samples to Continuous Samples} In order to apply continuous-valued GF-KSD, we need to transform samples $\{\vz_i\}_{i=1}^n$ and the target distribution $p_*(\vz)$ to the corresponding continuous-valued samples and the continuous-valued distribution. In order to do so, 
Our idea is to transform the testing of discrete distributions $q_* = p_*$ to their continuous parameterizations. 
Let $\Gamma$ be a even partition of a base distribution $p_0$, 
and $p_c$ and $q_c$ are the continuous parameterizations of $p_*$ and $q_*$ following our construction, respectively, that is, 
\begin{align*} 
p_c(\vx) \propto p_0(
\vx) p_*(\Gamma(\vx)), &&
q_c(\vx) \propto p_0(\vx) q_*(\Gamma(\vx)). 
\end{align*}
Obviously, $p_c = q_c$  implies that $p_* = q_*$ (following the definition of continuous parameterization). 
This allows us to transform the problem to a goodness-of-fit test of continuous distributions, which is achieved by testing if the gradient-free KSD 
\label{imp:ksd} equals zero, $H_0:q_c=p_c$ vs. $H_1:q_c\neq p_c.$   

In order to implement our idea, we need to convert the discrete sample $\{\vz_i\}_{i=1}^n$ from $q_*$ to a continuous sample $\{\vx_i\}_{i=1}^n$ from the corresponding (unknown) continuous distribution $q_c$. 
To achieve this goal, note that when $\vx \sim q_c$ and $\vz = \Gamma(\vx)$, 
the posterior distribution $\vx$ of giving $\vz = \va_i$ equals 
$$
q(\vx ~|~ \vz = \va_i) \propto  p_0(\vx) \ind(\Gamma(\vx) = \va_i), 
$$
which corresponds to sampling a truncated version of $p_0$ inside the region defined $\{\vx\colon~\Gamma(\vx) = \va_i\}$. 
This can be implemented easily for the simple choices of $p_0$ and $\Gamma$. 
For example,
in the case when $p_0$ is the product distribution in  \eqref{multiconti:surr} 
and $\Gamma$ is the concatenation of the quantile-based partition in  \eqref{equ:gamma1D}, 
we can sample $\vx ~|~ \vz = \va_i$ by sample $\vy$ from $\mathrm{Uniform}([\eta_{i-1}, \eta_i)^d)$ and obtain $\vx$ by $\vx = F^{-1}(\vy)$ where $F^{-1}$ is the inverse CDF of $p_0$. 

In the following, we will illustrate the way of transforming the discrete data to continuous data in a number of detailed procedures to make it easier to understand. Let $F$ be the c.d.f. of Gaussian base density $p_0.$ Let us first illustrate how to transform one-dimensional samples $\{z_i\}_{i=1}^n$ to continuous samples.
\begin{enumerate}
  \vspace{-.3cm}
    \item Given discrete data $\{z_i\}_{i=1}^n.$ 
    Let $\{a_j\}_{j=1}^K$ are possible discrete states. Assume $K$ is large so that for any $z_i,$ we have $z_i=a_j$ for one $j.$
    \vspace{-.2cm}
    \item For any $ z_i$ such as $z_i=a_j$, randomly sample $y_i\in [\frac{j-1}{K}, \frac{j}{K}).$ We obtain data $\{y_i\}_{i=1}^n.$ 
    \vspace{-.2cm}
    \item Apply $x=F^{-1}(y),$ we obtain data $\{x_i\}_{i=1}^n.$
    \vspace{-.3cm}
\end{enumerate}
Fig.~\ref{fig:tran:dis:cont} illustrates this procedure in 1D categorical distribution with three states.

For $\vx=(x^1,\cdots, x^d),$ let $F(\vx)=(F_1(x^1),\cdots, F_d(x^d)$, where each $F_i$ is the c.d.f. of Gaussian density $p_{0,i}(x^i).$ We apply the above one-dimensional transform to each dimension of $\{\vz_i\}_{i=1}^n,$ $\vz_i=(z_i^1, \cdots, z_i^d).$ We can easily obtain the continuous data $\{\vx_i\}_{i=1}^n.$


\begin{equation*}
q_*(\vz)\equiv p_*(\vz)~\mathrm{iff}~q_c(\vx)\equiv p_c(\vx).  
\end{equation*}
Therefore, the problem reduces to perform the hypothesis test$H_0:q_c=p_c$ vs. $H_1:q_c\neq p_c.$

With the one-to-one transform $F^{-1}$ and variable transform formula, the transformed distribution has the form
\begin{equation}
p_c(\vx) = p(F(\vx))\prod_{i=1}^d F'(x^i),    
\end{equation}
where $\vx=(x^1,\cdots,x^d).$ Now the original goodness-of-fit test between $q_*(\vz)$ and $p_*(\vz)$ reduces to perform the hypothesis test 
 \begin{equation*}
\text{null hypothesis}~ H_0:q_c=p_c~ \mathrm{vs.} ~\text{alternative hypothesis}~ H_1:q_c\neq p_c.    
\end{equation*}   

Let $p_0$ be the base function (the product of p.d.f. of Gaussian distribution w.r.t. dimension) and $w(\vx)=\wt{p}_c(\vx)/p_c(\vx)$, $\wt{p}_c(\vx)$ is a relaxation of $p_c(\vx)$.  
With the surrogate $\rho(\vx)=p_0(\vx)\wt{p}(F(\vx)),$ it is easy to apply the GF-KSD to the transformed discrete distributions between $q_c(\vx)$ and $p_c(\vx)$. The square of GF-KSD between $q_c(\vx)$ and $p_c(\vx)$ has the following form,
\begin{equation}
\label{ksd}
\wt{\mathbb{S}}(q_c, p_c) =\E_{\bd{x},\vx'\sim q_c}[w(\bd{x})\kappa_{\rho}(\bd{x}, \vx')w(\vx')],
\end{equation}
where $\kappa_{\rho}$ is defined in \eqref{imp:kernel} with $p_c(\vx)=p_0(\vx)p(F(\vx))$ and  $\rho(\vx)=p_0(\vx)\wt{p}(F(\vx)).$ 

\begin{figure}
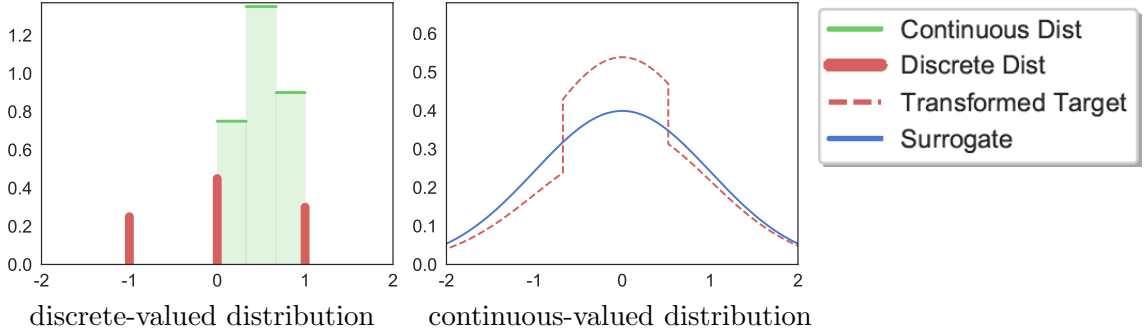

\centering
\begin{tabular}{ccc}
\includegraphics[width=0.34\textwidth]{figures/cat_def13.pdf} &
\hspace{-.5cm}
\includegraphics[width=0.34\textwidth]{figures/cat_def23.pdf} &
\hspace{-.5cm}
\raisebox{3.9em}{\includegraphics[height=0.14\textwidth]{figures/cat_legend2.pdf}} \\
{\small discrete-valued distribution} & {\small continuous-valued distribution}\\
\end{tabular}
\caption[Illustration of our goodness-of-fit test method on discrete distributions]{Illustrating 
the construction of $p_c(\vx)$ (red dash line) 
of a three-state discrete distribution $p_*(\vz)$ (red solid lines). The blue line represents the base distribution $p_0(\vx)$, which is a standard Gaussian distribution.\label{fig:def:gof}}
\end{figure}

Let us use the categorical distribution in Fig.~\ref{fig:def:gof} to illustrate. The categorical distribution defines on states $z=-1, ~0, ~1$ with probability $0.25, ~0.45, ~0.3$ respectively, which is shown in red in left figure of Fig.~\ref{fig:def:gof}. The discrete distribution is bijectively corresponded to a piecewise continuous-valued distribution $p(y)$ (defined within $[0, 1]$), which is shown in green in left figure of Fig.~\ref{fig:def:gof}. The data  $\{z_i\}_{i=1}^n$ can also be transformed to $\{y_i\}_{i=1}^n.$ By applying the one-to-one transform $F^{-1}$ ($F$ is the c.d.f. of the standard Gaussian distribution), we get the transformed target $p_c,$ which is shown in dashed red in right figure of Fig.~~\ref{fig:def:gof}. We can choose the surrogate distribution $\rho(x)$ as the p.d.f. of of the standard Gaussian distribution.

\begin {algorithm}[h]
\caption {Goodness-of-fit testing by GF-KSD (GF-KSD)} 
\label{alg:alg2}  
\begin {algorithmic}
\STATE {\bf Input}: Sample $\{\vz_i\}_{i=1}^n\sim q_*$ and its corresponding continuous-valued $\{\vx_i\}_{i=1}^n\sim q_c$, and null distribution $p_c$. Base function $p_0(\vx)$ and bootstrap sample size $m$.
\STATE {\bf Goal}: Test $H_0:q=p_c$ vs. $H_1:q\neq p_c$.
\STATE -Compute test statistics $\hat{\mathbb{S}}$ by \eqref{emp:ksd}.
\STATE -Compute m bootstrap sample $\hat{\mathbb{S}}_i^*$ by \eqref{boot:ksd}, $i=1,\cdots,m.$
\STATE -Reject $H_0$ with significance level $\alpha$ if the percentage of $\{\hat{\mathbb{S}}^*_i\}_{i=1}^m$ that satisfies $\hat{\mathbb{S}}^*>\hat{\mathbb{S}}$ is less than $\alpha.$
\end {algorithmic}
\end {algorithm}

With $\{\vx_i\}_{i=1}^n$ from $q_c(\vx)$, the GF-KSD between $q_c(\vx)$ and $p_c(\vx)$ can be estimated by the U-statistics,
\begin{equation}
\label{emp:ksd}
\hat{\mathbb{S}}(q_c, p_c) =\frac{1}{(n-1)n}\sum_{1\le i\neq j\le n} w(\vx_i)\kappa_{\rho}(\bd{x}_i, \vx_j)w(\vx_j) 
\end{equation}
\begin{lem}
\label{gof:lem}
Let $k(\vx,\vx')$ be a positive definite kernel. Suppose $\|p(\vx)(s_{q}(\vx)-s_{\rho}(\vx)) \|_2^2<\infty,$ and 
$\wt{\mathcal{S}}(q, p) = \E_{\vx, \vx'\sim q}[w(\vx)k_{\rho}(\vx, \vx') w(\vx')]< \infty,$ we have:
\begin{enumerate}
\item If $q\neq p,$ then $\sqrt{n}(\hat{\mathbb{S}}(q, p)- \mathbb{S}(q, p) )\rightarrow \mathcal{N}(0, \sigma_u^2)$ in distribution with the variance $\sigma_u^2=\mathrm{var}_{\vx\sim q}(\E_{\vx'\sim} [w(\vx)k_{\rho}(\vx, \vx') w(\vx')]),$ and $\sigma_u^2\neq 0.$
\item If $q=p,$ then $\sigma_u^2=0.$ And we have
$$n\hat{\mathbb{S}}(q, p)\rightarrow \sum_{j=1}^\infty c_j(Z_j^2-1), ~ \text{in distribution,} $$
where $\{Z_j\}$ are i.i.d. standard Gaussian random variable, and $\{c_j\}$ are the eigenvalues of kernel $w(\vx)k_{\rho}(\vx, \vx') w(\vx')$ under distribution $p.$
\end{enumerate}
\end{lem}
Proof: The detailed explanation and proof can be found on the appendix~\ref{append:go}.

{\bf Bootstrap Sample} The asymptotic distribution of $\hat{\mathbb{S}}(q_c, p_c)$ under the null hypothesis cannot be evaluated. In order to perform goodness-of-fit test, we draw random multinomial weights $u_1, \cdots, u_n\sim \mathrm{Multi}(n;1/n,\cdots,1/n),$ and calculate
\begin{equation}
\label{boot:ksd}
\hat{\mathbb{S}}^*(q_c, p_c) =\sum_{i\neq j } (u_i\!-\!\frac1n)w(\vx_i) \kappa_{\rho}(\bd{x}_i, \vx_j)w(\vx_j) (u_j-\frac1n).     
\end{equation}
We repeat this process by $m$ times and calculate the critical values of the test by taking the $(1-\alpha)$-th quantile, denoted by $\gamma_{1-\alpha}$, of the bootstrapped statistics $\{\hat{\mathbb{S}}_i^*(q_c, p_c)\}.$ 
\begin{pro}
\label{gof:pro}
Suppose the conditions in \ref{gof:lem} hold. For any fixed $q_c\neq p_c,$ the limiting power of the test that rejects the null hypothesis $q_c \neq p_c$ when $\hat{\mathbb{S}}^*(q_c, p_c)\ge\gamma_{1-\alpha}$
is one, which means the test is consistent in power against any fixed $q_c\neq p_c.$
\end{pro}
The proof is similar to the procedure in Proposition 4.2~\citet{liu2016kernelized}.  The Proposition~\ref{gof:pro} theoretically justifies the correctness of our proposed goodness-of-fit testing algorithm. The whole procedure is summarized in Alg.~\ref{alg:alg2}.

\section{Empirical Results}
In this section, we conduct some empirical experiments to demonstrate the effectiveness of our proposed goodness-of-fit testing algorithm and compare with two baselines, discrete kernelized Stein discrepancy~(\cite{yang2018goodness}, DKSD) and maximum mean discrepancy(\cite{gretton2012kernel}, MMD). We provide numerical comparison of the goodness-of-fit test with baselines on binary Ising model and then Bernoulli restricted Boltzmann machine~(RBM) to demonstrate the effectiveness of our proposed algorithm. The performance of all algorithms are based on type-II error rate (false negative rate).

\subsection{Binary Ising Model} The Ising model~\citep{ising1924beitrag} is a widely used model in Markov random field. Consider an (undirected) graph $G=(V, E)$, where each vertex $i\in V$ is associated with a binary spin, which consists of $\vx=(x_1,\cdots,x_d)$. The probability mass function is $p(\vx)=\frac{1}{Z}\sum_{(i,j)\in E} \theta_{ij}x_ix_j$, $x_i\in\{-1, 1\}$, $\theta_{ij}$ is edge potential and $Z$ is the normalization constant, which is infeasible to calculate when $d$ is high.  

\begin{figure*}[h]
\begin{center}
\begin{tabular}{cc}
\includegraphics[width=0.45\textwidth]{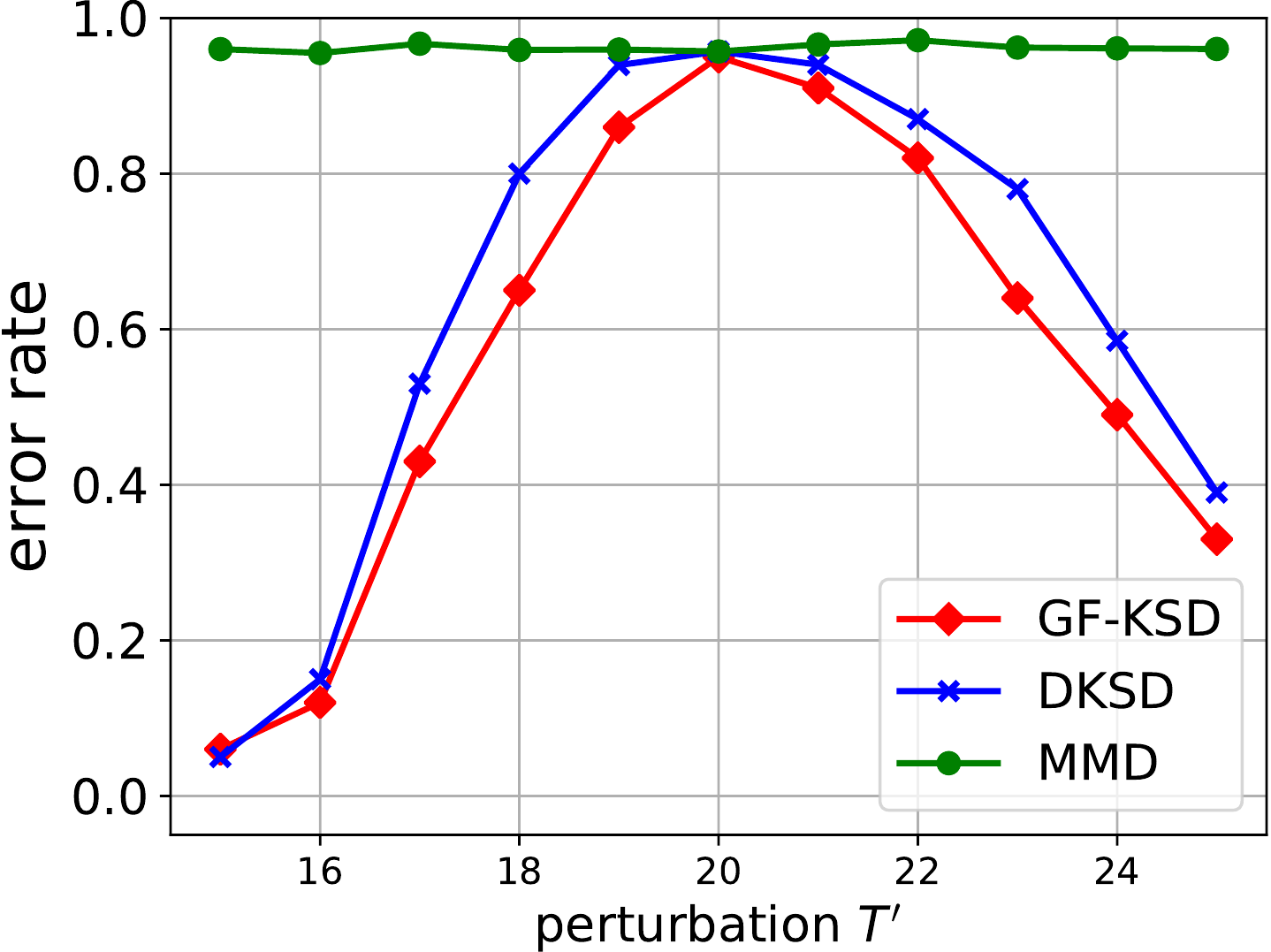} &
\includegraphics[width=0.45\textwidth]{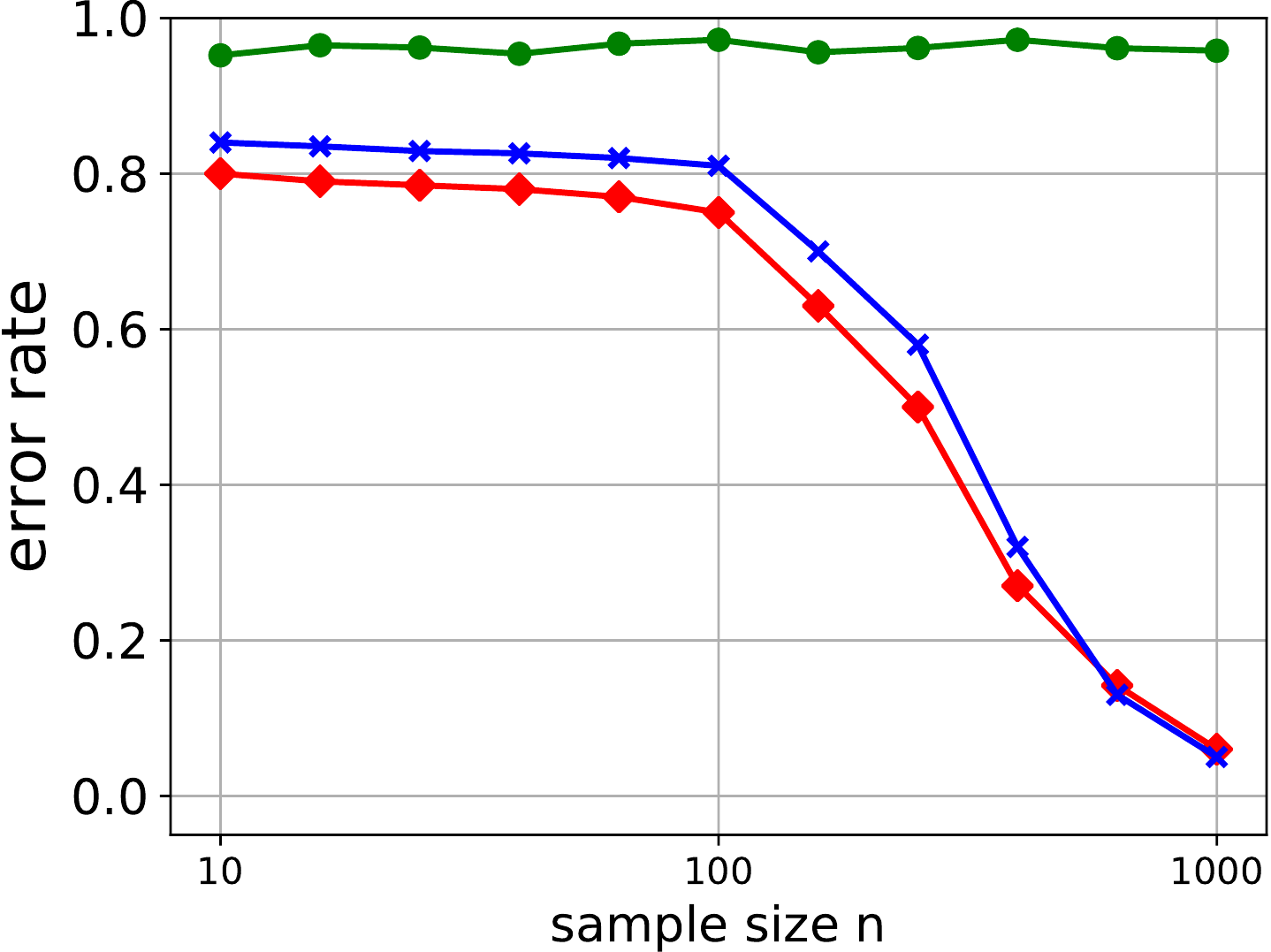} \\
{\small (a) $T=20$~(n=1000)} &
{\small (b) $T=20$ and $T'=15$}\\
\end{tabular}
\caption[Performance of goodness-of-fit test methods on binary Ising model]{Goodness-of-fit testing on Ising model with significant level $\alpha=0.05$. In (a, b), $p_*$ and $q_*$ has temperature $T$ and $T'$ respectively. In (a) we vary the parameters of $q_*$. We fix the models and vary the sample size $n$ in (b). We test $H_0:q_*=p_*$ vs. $H_1:q_*\neq p_*.$ \label{fig:gof}}
\end{center}
\end{figure*}

In Fig.~\ref{fig:gof}(a, b), $p_*$ and $q_*$ has temperature $T$ and $T'$ respectively. In (a) we vary the parameters $T'$ of $q_*$ and fix the parameter $T$ of $p_*$. Fig.~\ref{fig:gof}(a) shows that when the difference between $T$ and $T'$ are in some range $[\alpha, \beta]$, $\alpha>0, \beta>0$, our method has lower False negative rate than DKSD~\citep{yang2018goodness}. When $\alpha$ go to zeros, all methods have the same high error rate. When $\beta$ increases, the performance difference between ours and DKSD is small. In all settings, the MMD always keep perform worst, which indicates that MMD is not a goodness-of-fit test algorithm for such discrete probability models. We fix the models $p_*$ and $q_*$ and vary the sample size $n$ in (b). We test $H_0:q_*=p_*$ vs. $H_1:q_*\neq p_*.$ MMD keeps performing worst. When the sample size is small, our GF-KSD performs better than DKSD, which indicates that our GF-KSD is more sample-efficient. As the
sample size increases, the gap between our GF-KSD and DKSD becomes smaller and smaller.

\subsection{Bernoulli restricted Boltzmann Machine} Bernoulli RBM\citep{hinton2002training} is an undirected graphical model consisting of a bipartite graph between visible variables $\vz$ and hidden variables $\vh.$ In a Bernoulli RBM, the joint distribution of visible units $\vz\in \{-1, 1\}^d$ and hidden units $h \in \{-1, 1\}^M$ is given by 
\begin{equation}
p(\vz, \vh) \propto \exp(-E(\vz, \vh))   
\end{equation}
where $E(\vz, \vh)=-(\vz^\top W \vh+\vz^\top b+\vh^\top c)$, $W\in \mathbb{R}^{d\times M}$ is the weight, $b\in\mathbb{R}^d$ and $c\in\mathbb{R}^M$ are the bias. Marginalizing out the hidden variables $\vh,$ the probability mass function of $\vz$ is given by $p(\vz) \propto \exp(-E(\vz)),$ with free energy $E(\vz)=-\vz^\top b-\sum_k\log(1+\varphi_k),$ where $\varphi_k = \exp(W_{k*}^\top \vz + c_k)$ and $W_{k*}$ is the k-th row of $W.$

\begin{figure*}[h]
\begin{center}
\begin{tabular}{cc}
\includegraphics[width=0.45\textwidth]{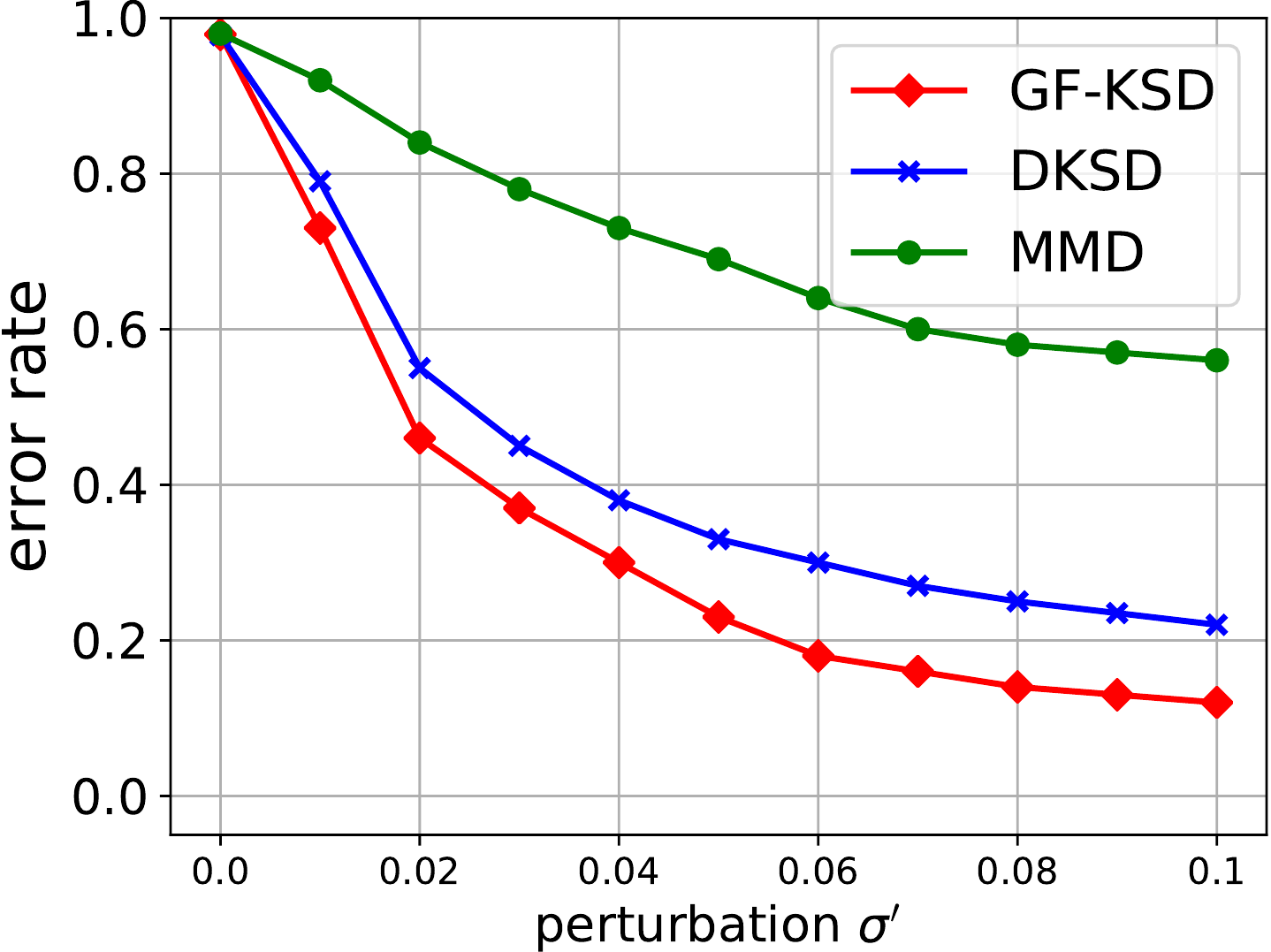} &
\includegraphics[width=0.45\textwidth]{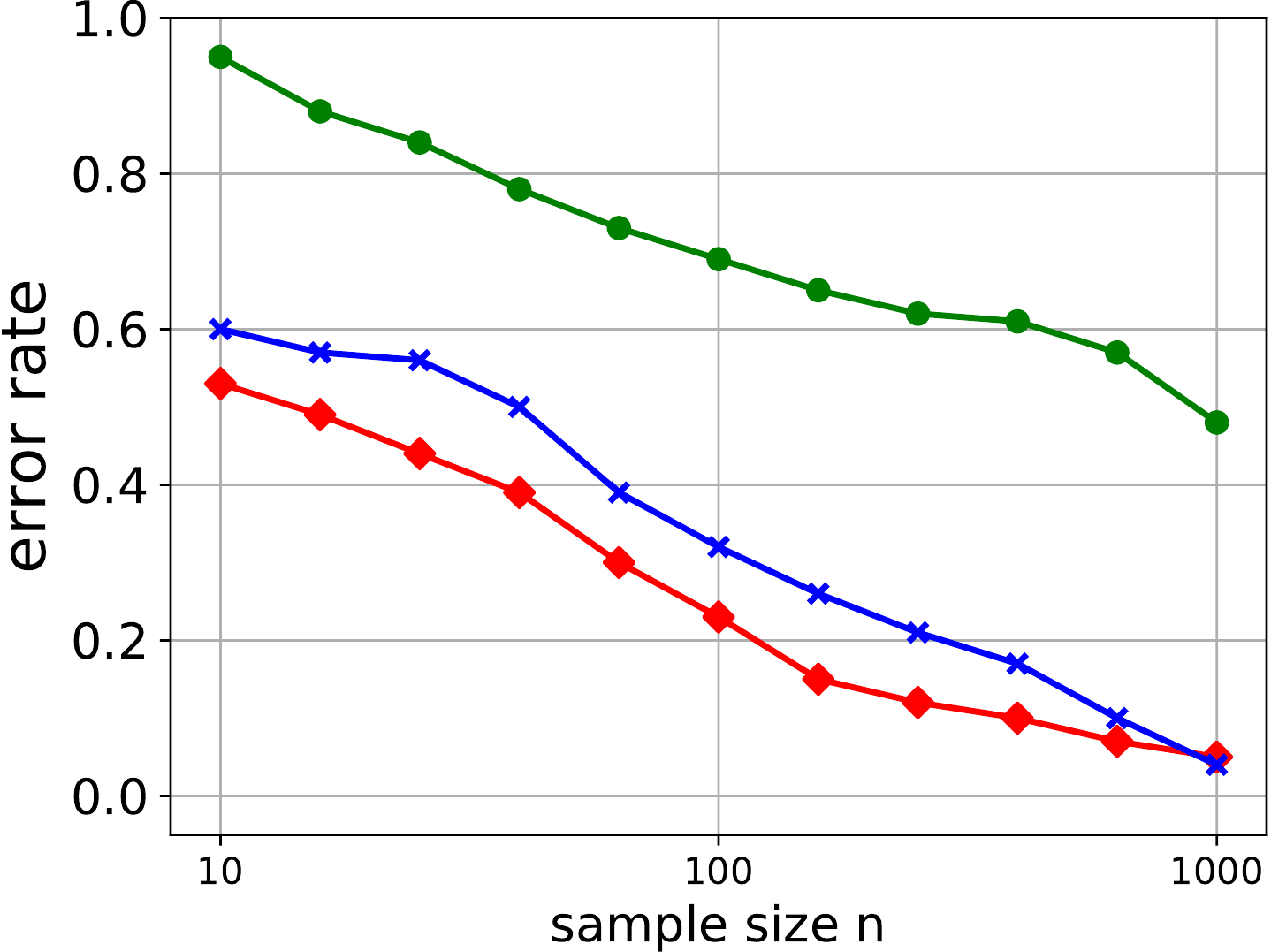}\\
{\small (a) $\sigma=0$~(n=100)}&
{\small (b) $\sigma=0$ and $\sigma'=15$} \\
\end{tabular}
\caption[Performance of goodness-of-fit test methods on Bernoulli RBM]{Goodness-of-fit testing on Bernoulli RBM (a, b) with significant level $\alpha=0.05$. In (a, b), $p_*$ has $W\sim \mathcal{N}(0, 1/M)$ and $q_*$ has $W+\epsilon$, where $\epsilon\sim \mathcal{N}(0, \sigma').$ $b$ and $c$ in $p_*$ and $q_*$ are the same. In (a) we vary the parameters of $q_*$. We fix the models and vary the sample size $n$ in (b). \label{fig:gof:rbm}}
\end{center}
\end{figure*}

We perform the goodness-of-fit tests on Bernoulli RBM in Fig. \ref{fig:gof}, which shows the false negative error. In Fig.~\ref{fig:gof:rbm}(a, b), $p_*$ has $W\sim \mathcal{N}(0, 1/M)$ and $q_*$ has $W+\epsilon$, where $\epsilon\sim \mathcal{N}(0, \sigma').$ $b$ and $c$ in $p_*$ and $q_*$ are the same. In Fig.~\ref{fig:gof:rbm}(a) we vary the parameters of $q_*$. We fix the models and vary the sample size $n$ in Fig.~\ref{fig:gof:rbm}(b). Fig. \ref{fig:gof:rbm}(a) shows that as $\sigma'$ increases, our GF-KSD performs bettern than DKSD. When $\sigma'$ is close to zero ($p_*$ and $q_*$ are almost the same), the type-II error rate is high, which is expected. In Fig.~\ref{fig:gof:rbm}(b), our GF-KSD performs better than DKSD when the sample size $n$ is small. As the sample size $n$ increases, the gap between ours and DKSD becomes smaller and smaller. In all cases, MMD performs the worst. 

\section{Summary} 
In this chapter, we propose an efficient algorithm to perform goodness-of-fit tests on discrete distributions. Our method leverages the gradient information of a surrogate distribution and corrects the bias with a form of importance weight to compute the gradient-free KSD for goodness-of-fit tests. The surrogate distribution can be constructed in an arbitrary form. On discrete distributions, the surrogate distributions can be easily constructed by exploiting the structure of the discrete distributions with simple smoothness trick. It is interesting to theoretically investigate the effective choice of the surrogate distributions. We verify the effectiveness of our proposed algorithm on two widely-used discrete graphical models, binary Ising model and Bernoulli restricted Boltzmann machine. Our proposed algorithm provides a new perspective to perform goodness-of-fit tests on discrete distributions. We expect our algorithm can be widely applied to goodness-of-fit tests on discrete distributions. 

Up to the current chapter, we have introduced four efficient approximate inference algorithms and one goodness-of-fit testing method. The non-parametric importance sampling algorithm adaptively improves the importance proposal, which leverages the gradient information of the target distribution. In Chapter~\ref{chap:gf}, we propose gradient-free SVGD, which doesn't require the gradient information of the target distribution. The $\KL$ divergence between the distribution of the updated particles and the target distribution is maximally decreased in a functional space. We propose gradient-free black-box importance sampling, which equips a set of samples with importance weights so that $\sum_{i=1}^n w(\vx_i) h(\vx_i)$ can be applied to estimate $\E_{\vx\sim p}[h(\vx)].$ In Chapter~\ref{chap:disc}, we propose a new algorithm to sample from discrete distributions, which leverages the gradient-free SVGD to sample from the corresponding piecewise continuous-valued target distributions of the discrete distributions. 

We have finished all approximate inference algorithms. In the next chapter, we will leverage some widely used approximate inference tools to perform some applications. We will provide one example in distributed learning, where we will 
We propose an importance-weighted estimator to reduce the variance in bootstrapped samples to estimate one integral when we try to find a global model, which has a minimal sum of $\KL$ divergence with each local model.

\chapter{Distributed Model Aggregation by Pseudo Importance Sampling\label{chap:boot}}
We have introduced two sample-efficient approximate inference algorithms on continuous distributions to estimate $\E_{\vx\sim p}[f(\vx)]$ with the gradient and without the gradient of the target distribution $p(\vx)$. We have also introduced one sampling algorithm on discrete distributions. The fundamental problem we have been solving is to find an efficient set of samples $\{\vx_i\}_{i=1}^n$ to estimate the integration $\E_{\vx\sim p}[f(\vx)].$ In this chapter, we provide an application in distributed model aggregation, where the setting is slightly different from that of the approximate inference. In the following, we will briefly illustrate the problem we are going to solve in this chapter and provide an algorithm whose key idea is motivated from the tools widely used in approximate inference. 

In distributed, or privacy-preserving learning, we are often given a set of probabilistic models estimated from different local repositories $\{p(\boldsymbol{x}|\boldsymbol{\hat{\theta}}_k)\}_{k=1}^d$, where $d$ is the number of local machines and $\boldsymbol{\hat{\theta}}_k$ is the parameter of the probabilistic local model, and asked to combine them into a single model $p(\boldsymbol{x}|\boldsymbol{\hat{\theta}})$ that gives efficient statistical estimation. We focuses on a \emph{one-shot} approach for distributed learning, in which we first learn a set of local models $\{p(\boldsymbol{x}|\boldsymbol{\hat{\theta}}_k)\}_{k=1}^d$ from local machines, and then combine them in a fusion center to form a single model that integrates all the information in the local models. This approach is highly efficient in both computation and communication costs,
but casts a challenge problem in designing statistically efficient combination strategies. A simple method is to linearly average the parameters of the local models, $\boldsymbol{\hat{\theta}}=\frac{1}{d}\sum_{j=1}^d \boldsymbol{\hat{\theta}}_j$. Many studies have been focused on a simple \emph{linear averaging} method that linearly averages the parameters of the local models \citep[e.g.,][]{zhang2012communication, zhang2013information, liu2015robust, rosenblatt2014optimality, connamacher2020rankboost, liu2017analysis, liu2019bandit}. Although nearly optimal asymptotic error rates can be achieved, this simple method tends to degenerate in practical scenarios for models with non-convex log-likelihood or non-identifiable parameters (such as latent variable models and neural models), and is not applicable at all for models with non-additive parameters (e.g., when the parameters have discrete or categorical values, the number of parameters in local models are different, or the parameter dimensions of the local models are different). 

The more meaningful and interpretable way is to find a joint model $p(\boldsymbol{x}|\boldsymbol{\hat{\theta}})$, which satisfies that the sum of the divergence between the single model $p(\boldsymbol{x}|\boldsymbol{\hat{\theta}})$ and the learned local model $p(\boldsymbol{x}\mid \boldsymbol{\hat{\theta}}_k)$ in distributional space. For example, we can use the $\KL$ divergence to measure the difference between two distributions and our goal reduces to find a global model $p(\boldsymbol{x}|\boldsymbol{\hat{\theta}})$ such as 
the sum of the $\KL$ divergence between $p(\boldsymbol{x}|\boldsymbol{\hat{\theta}})$ and $p(\boldsymbol{x}\mid \boldsymbol{\hat{\theta}}_k)$, $\boldsymbol{\hat{\theta}}=\argmax_{\vthe} \sum_{j=1}^d \KL(p(\boldsymbol{x}\mid \boldsymbol{\hat{\theta}}_k)|| p(\vx|\vthe)),$ is minimized, which finds a joint model $p(\boldsymbol{x}|\boldsymbol{\hat{\theta}})$ that minimizes the sum of Kullback-Leibler (KL) divergence to all the local models $p(\boldsymbol{x}|\boldsymbol{\hat{\theta}}_k)$ and is called the \emph{KL-averaging} method. Some elementary results of the \emph{KL-averaging} method can be found ~\citep{liu2014distributed, merugu2003privacy, hanvideo, han2018deep, lombardo2019deep, }. The \emph{KL-averaging} method directly combines the local models into a global model in the distributional perspective, instead of the parameters, which overcome all these practical limitations of the \emph{linear averaging} aforementioned. The exact \emph{KL-averaging} is not computationally tractable because of the intractability of calculating $\KL$ divergence; a practical approach is to draw (bootstrap) samples from the given local models, and then learn a combined model based on all the bootstrap data. The problem we are interested in solving is equivalent to the following optimization problem, $\boldsymbol{\hat{\theta}}=\argmax_{\vthe}\sum_{k=1}^d\int p(\boldsymbol{x}\mid \boldsymbol{\hat{\theta}}_k)\log
p(\boldsymbol{x}\mid \boldsymbol{\theta})d\boldsymbol{x}.$ In most applications, the integration $\int p(\boldsymbol{x}\mid \boldsymbol{\hat{\theta}}_k)\log
p(\boldsymbol{x}\mid \boldsymbol{\theta})d\boldsymbol{x}$ is not available in an analytical form, which casts a challenge optimization. To solve such an optimization problem, one more practical strategy is to generate bootstrap samples $\{\vv{\widetilde{x}}_j^k\}_{j=1}^n$ from each local model $p(\boldsymbol{x}\mid \boldsymbol{\hat{\theta}}_k)$, where $n$ is the number of the bootstrapped samples drawn from each local model $p(\boldsymbol{x}\mid \boldsymbol{\hat{\theta}}_k)$, and use the typical Monte Carlo method to estimate the integration~\citet{liu2014distributed, han2016bootstrap, jun2015numerical}, $\argmax_{\vthe}\sum_{k=1}^d\sum_{j=1}^n \log
p(\vv{\widetilde{x}}_j^k \mid \boldsymbol{\theta}).$ Typical gradient descent method can be applied to solve this optimization to obtain a joint model. Unfortunately, the bootstrap procedure introduces additional noise and can significantly deteriorate the performance of the learned joint model. To reduce the variance induced from the bootstrapped samples and improve the performance of the learned joint model, we introduce two variance-reduced techniques to more efficiently combine the local models, including a weighted M-estimator
that is both statistically efficient and practically powerful. The weighted M-estimator method to reduce the variance of the bootstrapped samples is motivated from the importance sampling technique widely used in approximate inference. Both theoretical and
empirical analysis is provided to demonstrate our proposed methods. Empirical results justify the correctness of our theoretical analysis. 

The outline of this chapter is organized as follows: we first discuss the background and the problem we are going to solve in one-shot distributed learning; we then propose two main algorithms to solve the problem and provide theoretical analysis of our proposed methods; we conduct empirical experiments to verify the correctness of our theoretical analysis and to demonstrate the effectiveness of our proposed methods on real datasets. 

\section{Background}
Modern data science applications increasingly involve learning complex probabilistic models over massive datasets. In many cases, the datasets are distributed into multiple machines at different locations, between which communication is expensive or restricted; 
this can be either because the data volume is too large to store or process in a single machine, or due to privacy constraints
as these in healthcare or financial systems.
There has been a recent growing interest in developing
\emph{communication-efficient} algorithms for probabilistic learning with distributed datasets; see e.g., \citet{boyd2011distributed, zhang2012communication, dekel2012optimal, liu2014distributed, rosenblatt2014optimality, dai2019opaque, huang2019taxable, huang2018costly, dailife, han2017high} and reference therein. In the following, we first frame the problem in a mathematical way with the introduction of commonly-used notations. Then we discuss some baseline methods and analyze their advantages and disadvantages under different settings.


\begin{figure}
\centering
 \includegraphics[width=.8\textwidth]{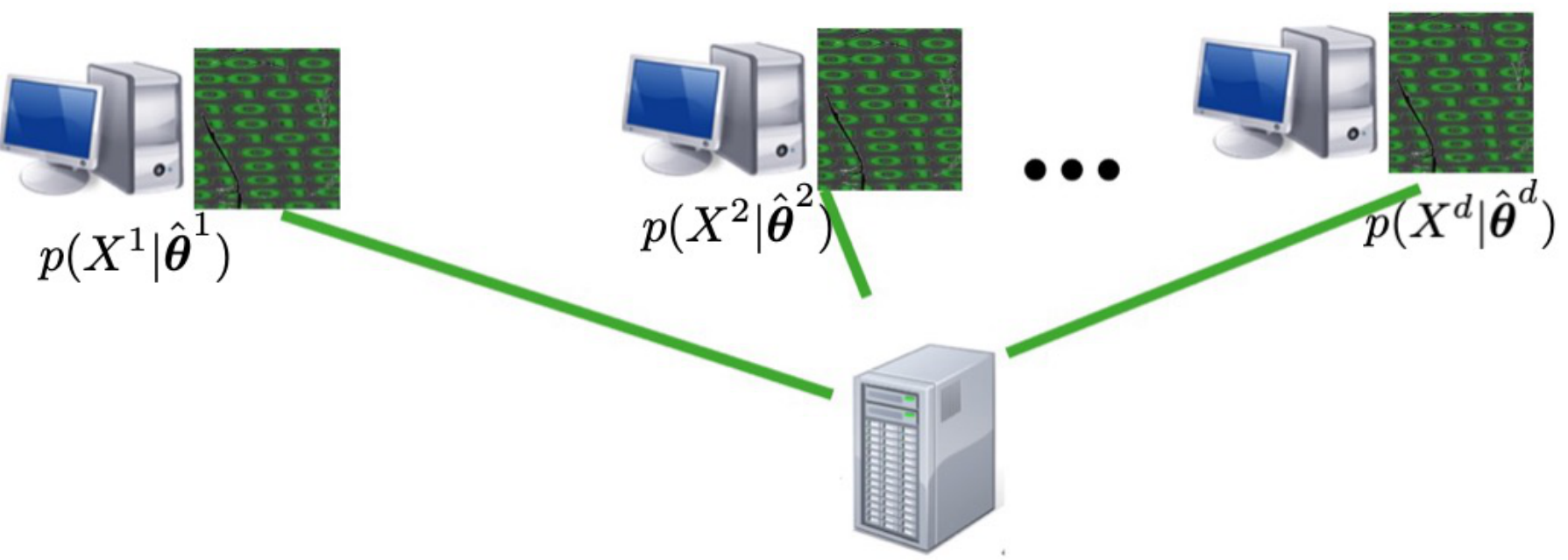}
\vspace{-.2cm}
\caption[Illustration of the one-shot communication in distributed learning]{Illustration of the one-shot communication in distributed learning. Each local machine learns a probability model. The dataset is evenly partitioned $X=X^1\cup X^2\cup\cdots\cup X^d.$}
\end{figure}

\paragraph{Problem Setting}
\label{sec:background}
Suppose we have a dataset $X=\{\boldsymbol{x}_j, ~ j=1,2,...,N\}$ of size $N$, \emph{i.i.d.} drawn from a probabilistic model $p(\vx | \vv{\theta}^*)$ within a parametric family $\mathcal{P}=\{p(\boldsymbol{x}|\boldsymbol{\theta}):\boldsymbol{\theta}\in\Theta\}$; here $\boldsymbol{\theta}^*$ is the unknown true parameter that we want to estimate based on $X$.
In the distributed setting, the dataset $X$ is partitioned into $d$ disjoint subsets, $X=\bigcup_{k=1}^d X^k$, where $X^
k$ denotes the $k$-th subset which we assume is stored in a local machine.
For simplicity, we assume all the subsets have the same data size ($N/d$).
The traditional maximum likelihood estimator (MLE) provides a natural way for estimating the true parameter $\boldsymbol{\theta}^*$
based on the whole dataset $X$,
\begin{align} \label{globalMLE}
\text{Global MLE:}\quad \boldsymbol{\hat{\theta}}_{\mathrm{mle}}=\argmax_{\boldsymbol{\theta}\in\Theta}\sum_{k=1}^d\sum_{j=1}^{N/d}\log p(\vx^k_j\mid\boldsymbol{\theta}),
\quad\text{where } X^k=\{\vx^k_j\}.
\end{align}
However, directly calculating the global MLE is challenging due to the distributed partition of the dataset.
Although distributed optimization algorithms exist \citep[e.g.,][]{boyd2011distributed, shamir2013communication},
they require iterative communication between the local machines and a fusion center,
which can be very time consuming in distributed settings, for which
the number of communication rounds
forms the main bottleneck (regardless of the amount of information communicated at each round).

We instead consider a simpler \emph{one-shot} approach that first learns a set of local models based on each subset, and then send them to a fusion center in which
they are combined into a global model that captures all the information. We assume each of the local models is estimated using a MLE based on subset $X^k$ from the $k$-th machine:  %
\begin{align}\label{equ:localmle}
\text{Local MLE:}\quad \boldsymbol{\hat{\theta}}_k=\argmax_{\boldsymbol{\theta}\in\Theta}\sum_{j=1}^{N/d}\log p(\boldsymbol{x}^k_j\mid\boldsymbol{\theta}),~~\text{where}~~ k\in [d]=\{1,2,\cdots,d\}.
\end{align}

The major problem is how to combine these local models into a global model.
The simplest way is to linearly average all local MLE parameters:
$$\text{Linear Average:}\quad \boldsymbol{\hat{\theta}}_{\mathrm{linear}}=\frac{1}{d}\sum_{k=1}^d\boldsymbol{\hat{\theta}}_k.$$
Comprehensive theoretical analysis has been done for $ \boldsymbol{\hat{\theta}}_{\mathrm{linear}}$ \citep[e.g.,][]{zhang2012communication, rosenblatt2014optimality}, which show that it has an asymptotic MSE of $\E||\boldsymbol{\hat{\theta}}_{\mathrm{linear}} - \vv{\theta}^* ||^2 = O(N^{-1})$.
In fact, it is equivalent to the global MLE $ \boldsymbol{\hat{\theta}}_{\mathrm{mle}}$ up to the first order $O(N^{-1})$, and
several improvements have been developed to improve the second order term \citep[e.g.,][]{zhang2012communication, huang2015distributed}. 

Unfortunately,  the linear averaging method can easily break down in practice,
or is even not applicable when the underlying model is complex.
For example, it may work poorly when the likelihood has multiple modes,
or when there exist non-identifiable parameters for which different parameter values correspond to a same model (also known as the label-switching problem);
models of this kind include latent variable models and neural networks, and appear widely in machine learning. 
In addition, the linear averaging method is obviously not applicable when the local models have different numbers of parameters (e.g., Gaussian mixtures with unknown numbers of components),
or when the parameters are simply not additive (such as parameters with discrete or categorical values).
Further discussions on the practical limitations of the linear averaging method can be found in \citet{liu2014distributed}.

All these problems of linear averaging can be well addressed by a
 \emph{KL-averaging} method which averages the model (instead of the parameters)
 by finding a geometric center of the local models in terms of KL divergence \citep{merugu2003privacy, liu2014distributed}.
Specifically, it finds a model $p( \vv x| \vv{\theta}_{\mathrm{KL}}^*)$ where  $\vv{\theta}_{\mathrm{KL}}^*$  is obtained by 
\begin{equation}
\label{boot:KL:opt}
\vv{\theta}_{\mathrm{KL}}^* = \argmin_{\vv\theta} \sum_{k=1}^d\KL(p(\boldsymbol{x}|\boldsymbol{\hat{\theta}}_k)\mid\mid p(\boldsymbol{x}|\boldsymbol{\theta}))
\end{equation}
The optimization of \eqref{boot:KL:opt} is equivalent to,
\begin{equation}
\label{KLdivmax}
\text{Exact KL Estimator: }\quad \boldsymbol{\theta}_{\KL}^*=\argmax_{\boldsymbol{\theta}\in\Theta} \bigg\{ \eta(\boldsymbol{\theta})
\equiv \sum_{k=1}^d\int p(\boldsymbol{x}\mid \boldsymbol{\hat{\theta}}_k)\log
p(\boldsymbol{x}\mid \boldsymbol{\theta})d\boldsymbol{x} \bigg\}.
\end{equation}
\citet{liu2014distributed} studied the theoretical properties of the KL-averaging method, and showed that
 it exactly recovers the global MLE, that is, $\boldsymbol{\theta}_{\KL}^*={\boldsymbol{\hat\theta}}_{\mathrm{mle}}$, when the distribution family is a full exponential family,
 and achieves an optimal asymptotic error rate (up to the second order) among all the possible combination methods of $\{\vv{\hat\theta}_k\}$.
Despite the attractive properties,
the exact KL-averaging is not computationally tractable except for very simple models.
\citet{liu2014distributed} suggested a naive \emph{bootstrap} method for approximation:
it draws \emph{parametric bootstrap} sample $\{\boldsymbol{\widetilde{x}}^k_j\}_{j=1}^{n}$ from each local model $p(\boldsymbol{x}|\boldsymbol{\hat{\theta}}_k)$, $k\in [d]$ and use it to approximate each integral in \eqref{KLdivmax}. 
The optimization in \eqref{KLdivmax} then reduces to a tractable one,
\begin{equation}
\label{KLdivmaxapprox}
\text{KL-Naive Estimator:} \quad\boldsymbol{\hat{\theta}}_{\mathrm{KL}}=  \argmax_{\boldsymbol{\theta}\in\Theta}  \bigg \{ \hat{\eta}(\boldsymbol{\theta}) \equiv    \frac{1}{n} \sum_{k=1}^d \sum_{j=1}^n \log p(\boldsymbol{\widetilde{x}}^k_j\mid \boldsymbol{\theta}) \bigg\}.
\end{equation}
Intuitively, we can treat each $\widetilde X_k = \{\boldsymbol{\widetilde{x}}^k_j\}_{j=1}^{n}$ as an approximation of the original subset $X^k  = \{\boldsymbol{{x}}^k_j\}_{j=1}^{N/d}$, and hence can be used to approximate the global MLE in \eqref{globalMLE}.

 Unfortunately, as the theoretical results shown in the next section, the accuracy of  $\boldsymbol{\hat{\theta}}_{\KL}$ critically depends on the bootstrap sample size $n$, and one would need $n$ to be nearly as large as the original data size $N/d$ to make $\boldsymbol{\hat{\theta}}_{\mathrm{KL}}$ achieve the baseline asymptotic rate $O(N^{-1})$ that the simple linear averaging achieves; this is highly undesirably since $N$ is often assumed to be large in distributed learning settings.

\section{Importance-Weighted Estimator to Bootstrapped Model Aggregation}
\label{sec:two}
In order to reduce the variance induced from the bootstrapped samples in ~\eqref{KLdivmaxapprox}, we propose two variance reduction techniques for improving the KL-averaging estimates and discuss their theoretical and practical properties.
We start with a concrete analysis on the $\KL$-naive estimator $\boldsymbol{\hat{\theta}}_{\KL},$ which was missing in \citet{liu2014distributed}. Then we introduce a baseline control variates estimator method by leveraging the correlation of bootstrap samples $\{\boldsymbol{\widetilde{x}}^k_j\}_{j=1}^n$ among different local models $p(\boldsymbol{x}|\boldsymbol{\hat{\theta}}_k),$ $k=1,\cdots, d.$ While such method can only be used under the setting as that of the linear-averaging method, which limits its application. To overcome such a limitation, we finally propose a $\KL$-weighted estimator method, which overcomes all limitations of linear-averaging method.

The theoretical results provided in this chapter is based on the following assumptions of the probability model.
\begin{ass}
\label{assump}
 1. $\log p(\boldsymbol{x}\mid\boldsymbol{\theta}),$
 $\frac{\partial\log p(\boldsymbol{x}\mid\boldsymbol{\theta})}{\partial\boldsymbol{\theta}},$
 and $\frac{\partial^2\log p(\boldsymbol{x}\mid\boldsymbol{\theta})}{\partial\boldsymbol{\theta}\partial\boldsymbol{\theta}^\top}$
 are continuous for $\forall~ \boldsymbol{x}\in\mathcal{X}$ and $\forall ~\boldsymbol{\theta}\in\Theta;$ 2. $\frac{\partial^2\log p(\boldsymbol{x}\mid\boldsymbol{\theta})}{\partial\boldsymbol{\theta}\partial\boldsymbol{\theta}^\top}$ is positive definite and $C_1\leq \|\frac{\partial^2\log p(\boldsymbol{x}\mid\boldsymbol{\theta})}{\partial\boldsymbol{\theta}\partial\boldsymbol{\theta}^\top}\|\leq C_2$ in a neighbor of $\boldsymbol{\theta}^*$ for $\forall ~x\in\mathcal{X}$, and $C_1$, $C_2$ are some positive constants.
\end{ass}
\subsection{$\KL$-naive estimator to distributed model aggregation}
In the following, we first prove ${\boldsymbol{\hat\theta}}_{\KL}$ provides a consistent estimation of the ground truth minimized model estimator $\boldsymbol{\theta}_{\KL}^*$ and the mean square error (MSE) between ${\boldsymbol{\hat\theta}}_{\KL}$ and $\boldsymbol{\theta}_{\KL}^*$ has an error rate $O(\frac{1}{dn}),$ where $d$ is the number of the local machines and $n$ is the number of the bootstrapped samples.
\begin{thm}
\label{boot:thm1}
\label{naivemethod}
 Under Assumption \ref{assump},  ${\boldsymbol{\hat\theta}}_{\KL}$ is a consistent estimator of $\boldsymbol{\theta}_{\KL}^*$ as $n\to\infty$, and
 $$\mathbb{E}({\boldsymbol{\hat\theta}}_{\KL}-\boldsymbol{\theta}_{\KL}^*)=o(\frac{1}{dn}),\quad \mathbb{E}\|{\boldsymbol{\hat\theta}}_{\KL}-\boldsymbol{\theta}_{\KL}^*\|^2=O(\frac{1}{dn}),$$
 where $d$ is the number of machines and $n$ is the bootstrap sample size for each local model $p(\boldsymbol{x}\mid\boldsymbol{\hat{\theta}}_k)$.
\end{thm}
The proof of Theorem \ref{boot:thm1} is provided in the Appendix \ref{append:boot}.
Based on the MSE between the exact $\KL$ estimator $\boldsymbol{\theta}_{\KL}^*$ and the true parameter $\boldsymbol{\theta}^*$ provided in \citet{liu2014distributed}, it is easy to derive that the MSE between $\boldsymbol{\hat{\theta}}_{\KL}$ and the true parameter $\boldsymbol{\theta}^*$ satisfies
\begin{equation}
\label{globalmse}
\mathbb{E}\|\boldsymbol{\hat{\theta}}_{\KL}-\boldsymbol{\theta}^*\|^2\approx\mathbb{E}\|\boldsymbol{\hat{\theta}}_{\KL}-\boldsymbol{\theta}^*_{\KL}\|^2+\mathbb{E}\|\boldsymbol{\theta}_{\KL}^*-\boldsymbol{\theta}^*\|^2=O(N^{-1}+(dn)^{-1}).
\end{equation}
To guarantee the MSE rate between $\boldsymbol{\hat{\theta}}_{\KL}$ and $\boldsymbol{\theta}^*$ has the rate order $O(N^{-1})$, as what is achieved by the simple linear averaging, we need to draw $d n \gtrsim N$ bootstrap data points in total, which is undesirable since $N$ is often assumed to be very large by the assumption of distributed learning setting (one exception is when the data is distributed due to privacy constraint, in which case $N$ may be relatively small).

Therefore, it is a critical task to develop more accurate methods that can reduce the noise introduced by the bootstrap process.
In the sequel, we introduce two variance reduction techniques to achieve this goal.
One is based a (linear) control variates method that improves $\boldsymbol{\hat{\theta}}_{\KL}$ using a linear correction term,
and another is a \emph{multiplicative} control variates method that modifies the M-estimator in \eqref{KLdivmaxapprox} by assigning each bootstrap data point with a positive weight to cancel the noise.
We show that both method achieves a higher $O(N^{-1} + (dn^2)^{-1})$ rate under mild assumptions,
while the second method has more attractive practical advantages.

\subsection{Proposed Baseline Control Variates Estimator}
\label{sec:control}
The control variates method is a technique for variance reduction on Monte Carlo estimation \citep[e.g.,][]{wilson1984variance}.
It introduces a set of correlated auxiliary random variables with known expectations or asymptotics (referred as the control variates), to balance the variation of the original estimator.
In our case, since each bootstrapped subsample $\widetilde X^k =\{\boldsymbol{\widetilde{x}}^k_j\}_{j=1}^n$ is know to be drawn from the local model $p(\boldsymbol{x} \mid\boldsymbol{\hat\theta}_k)$, we can construct a control variate by re-estimating the local model based on $\widetilde X^k$:

\begin{align}
\text{Bootstrapped Local MLE:}\quad\boldsymbol{\widetilde{\theta}}_k=\argmax_{\boldsymbol{\theta}\in\Theta}\sum_{j=1}^n\log p(\boldsymbol{\widetilde{x}}^k_j\mid\boldsymbol{\theta}),\quad \mathrm{for}~~ k\in [d],\label{tildethea}
\end{align}
where $\boldsymbol{\widetilde{\theta}}_k$ is known to converge to $\boldsymbol{\hat{\theta}}_k$ asymptotically.
This allows us to define the following control variates estimator:

\begin{equation}
\label{KLControl}
\text{KL-Control Estimator:}\quad \boldsymbol{\hat{\theta}}_{\KL-C}=\boldsymbol{\hat{\theta}}_{\KL}+\sum_{k=1}^d\boldsymbol{\mathfrak{B}}_k(\boldsymbol{\widetilde{\theta}}_k-\boldsymbol{\hat{\theta}}_k),
\end{equation}
where $\boldsymbol{\mathfrak{B}_k}$ is a matrix chosen to minimize the asymptotic variance of $ \boldsymbol{\hat{\theta}}_{\KL-C}$;
our derivation shows that the asymptotically optimal $\boldsymbol{\mathfrak{B}_k}$ has a form of

\begin{equation}
\label{scorecoeff}
\boldsymbol{\mathfrak{B}}_k=-(\sum_{k=1}^dI(\boldsymbol{\hat{\theta}}_k))^{-1}I(\boldsymbol{\hat{\theta}}_k), \quad k\in [d],
\end{equation}
where $I(\boldsymbol{\hat{\theta}}_k)$ is the empirical Fisher information matrix of the local model $p(\vv x \mid \boldsymbol{\hat{\theta}}_k)$.
Note that this differentiates our method  from the typical control variates methods where $\boldsymbol{\mathfrak{B}}_k$ is instead estimated using empirical covariance between the control variates and the original estimator (in our case, we can not directly estimate the covariance because $\boldsymbol{\hat{\theta}}_{\KL}$ and  $\boldsymbol{\widetilde{\theta}}_k$ are not averages of i.i.d. samples).
The procedure of our method is summarized in Algorithm \ref{alg:kl-control}.
Note that the form of \eqref{KLControl} shares some similarity with the one-step estimator in \citet{huang2015distributed}, but
\citet{huang2015distributed} focuses on improving the linear averaging estimator, and is different from our setting.

We analyze the asymptotic property of the estimator $\boldsymbol{\hat{\theta}}_{\KL-C}$,  and summarize it as follows.
\begin{thm}
\label{boot:control}
 Under Assumption (\ref{assump}), $\boldsymbol{\hat{\theta}}_{\KL-C}$ is a consistent estimator of $\boldsymbol{\theta}_{\KL}^*$ as $n\to\infty,$
 and its asymptotic MSE is guaranteed to be smaller than the KL-naive estimator ${\boldsymbol{\hat\theta}}_{\KL}$, that is,
 $$
  n \mathbb{E}\|\boldsymbol{\hat{\theta}}_{\KL-C}-\boldsymbol{\theta}_{\KL}^*\|^2 < n  \mathbb{E}\|{\boldsymbol{\hat\theta}}_{\KL}-\boldsymbol{\theta}_{\KL}^*\|^2, ~~~~~~ \text{as}~~ n\to \infty.
  $$
  In addition, when $N> n\times d$,  the $\boldsymbol{\hat{\theta}}_{\KL-C}$ has \emph{``zero-variance''} in that
   $\mathbb{E}\|{\boldsymbol{\hat\theta}}_{\KL}-\boldsymbol{\theta}_{\KL}^*\|^2=O((dn^2)^{-1})$.
   Further, in terms of estimating the true parameter, we have
\begin{equation}
\label{globalklc}
\mathbb{E}\|\boldsymbol{\hat{\theta}}_{\KL-C}-\boldsymbol{\theta}^*\|^2=O(N^{-1}+(dn^{2})^{-1}).
\end{equation}
 \end{thm}
The proof is in the Appendix \ref{append:boot}.

From \eqref{globalklc}, we can see that the MSE between $\boldsymbol{\hat{\theta}}_{\KL-C}$ and $\boldsymbol{\theta}^*$ reduces to $O(N^{-1})$
as long as $n \gtrsim (N/d)^{1/2}$, which is a significant improvement over the KL-naive method which requires $n  \gtrsim N/d$.
When the goal is to achieve an $O(\epsilon)$ MSE, we would just need to take $n \gtrsim 1/(d\epsilon)^{1/2}$ when $N > 1/\epsilon$, that is,
$n$ does not need to increase with $N$ when $N$ is very large.

 Meanwhile, because $\boldsymbol{\hat{\theta}}_{\mathrm{KL}-C}$ requires a linear combination of $\vv{\hat \theta}_k$,
 $\vv{\widetilde \theta}_k$ and $\vv{\hat \theta}_{\mathrm{KL}}$, it carries the practical drawbacks of the linear averaging estimator as we discuss in Section~\ref{sec:background}.
 %
This motivates us to develop another \emph{KL-weighted} method shown in the next section, which achieves the same asymptotical efficiency as $\boldsymbol{\hat{\theta}}_{\mathrm{KL}-C}$, while still
inherits all the practical advantages of \emph{KL-averaging}.
%
\begin{algorithm}[tb]
\caption{KL-Control Variates Method for Combining Local Models}
\label{alg:kl-control}
\begin{algorithmic}[1]
\STATE {\bfseries Input:} Local model parameters $\{\boldsymbol{\hat{\theta}}_k\}_{k=1}^d$.
\STATE {Generate bootstrap data $\{\boldsymbol{\widetilde{x}}^k_j\}_{j=1}^n$ from each $p(\boldsymbol{x}|\boldsymbol{\hat{\theta}}_k)$, for $k\in [d]$.}
\STATE{Calculate the KL-Naive estimator, $\boldsymbol{\hat{\theta}}_{\KL}=\argmax_{\boldsymbol{\theta}\in\Theta}\sum_{k=1}^d\frac{1}{n}\sum_{j=1}^n \log p(\boldsymbol{\widetilde{x}}^k_j| \boldsymbol{\theta}).$}
\STATE {Re-estimate the local parameters $\widetilde{\boldsymbol{\theta}}_k$ via \eqref{tildethea} based on the bootstrapped data subset $\{\boldsymbol{\widetilde{x}}^k_j\}_{j=1}^n$}, ~ for $k \in [d]$.
\STATE{Estimate the empirical Fisher information matrix $I(\boldsymbol{\hat{\theta}}_k)=\frac{1}{n}\sum_{j=1}^n \frac{\partial{\log p(\boldsymbol{\widetilde{x}}_j^k|\boldsymbol{\hat{\theta}}_k)}}{\partial{\boldsymbol{\theta}}}{\frac{\partial{\log p(\boldsymbol{\widetilde{x}}_j^k|\boldsymbol{\hat{\theta}}_k)}}{\partial{\boldsymbol{\theta}}}}^\top$, for $k\in [d]$.}
\STATE{\bfseries Ouput:}
The parameter $\boldsymbol{\hat{\theta}}_{\KL-C}$ of the combined model is given by \eqref{KLControl} and \eqref{scorecoeff}.
\end{algorithmic}
\end{algorithm}

\subsection{KL-Weighted Estimator}
\label{sec:kl-weighted}
Our {KL-weighted} estimator is based on directly modifying the M-estimator for $\boldsymbol{\hat{\theta}}_{\mathrm{KL}}$ in \eqref{KLdivmaxapprox},
by assigning each bootstrap data point $\vv{\widetilde{x}}_j^k$ a positive weight
according to the probability ratio $p(\vv{\widetilde{x}}_j^k  \mid  \vv{\hat{\theta}}_k ) / p(\vv{\widetilde{x}}_j^k \mid \vv{\widetilde{\theta}}_k)$ of the actual local model $p(x | \vv{\hat{\theta}}_k)$ and the re-estimated model $p(x |\vv{\widetilde{\theta}}_k)$ in \eqref{tildethea}.
Here the probability ratio acts like a \emph{multiplicative} control variate \citep{nelson1987control}, which has the advantage of being positive and applicable to non-identifiable, non-additive parameters. Our KL-weighted estimator is defined as
\begin{equation}
\label{KLweigthed}
{\boldsymbol{\hat\theta}}_{\KL-W} =
\argmax_{\boldsymbol{\theta}\in\Theta} \bigg\{ \widetilde{\eta}(\boldsymbol{\theta})  \equiv \sum_{k=1}^d\frac1n\sum_{j=1}^n
\frac{p(\boldsymbol{\widetilde{x}}_j^k|\boldsymbol{\hat{\theta}}_k)}{p(\boldsymbol{\widetilde{x}}_j^k| \boldsymbol{\widetilde{\theta}}_k)}\log p(\boldsymbol{\widetilde{x}}_j^k|\boldsymbol{\theta}) \bigg\}.
\end{equation}
We first show that this weighted estimator $\widetilde{\eta}(\boldsymbol{\theta})$ gives a more accurate estimation of $\eta(\boldsymbol{\theta})$ in \eqref{KLdivmax} than the straightforward estimator $\hat{\eta}(\boldsymbol{\theta})$ defined in \eqref{KLdivmaxapprox} for any $\boldsymbol{\theta}\in\Theta$. 
\begin{lem}
As $n\to\infty$, $\widetilde{\eta}(\boldsymbol{\theta})$ is a more accurate estimator of $\eta(\boldsymbol{\theta})$ than $\hat{\eta}(\boldsymbol{\theta})$, in that
\begin{equation}
n \mathrm{Var}(\widetilde{\eta}(\boldsymbol{\theta}))\leq  n \mathrm{Var}(\hat{\eta}(\boldsymbol{\theta})), ~~~~~\text{as }n\to\infty,~~
\quad \text{for any } \boldsymbol{\theta}\in\Theta.
\end{equation}
\end{lem} 
This estimator is motivated by \citet{henmi2007importance} in which the same idea is applied to reduce the asymptotic variance in importance sampling. 
Similar result is also found in \citet{hirano2003efficient}, in which  it is shown that a similar weighted estimator with estimated propensity score is more efficient than the estimator using true propensity score in estimating the average treatment effects. 
Although being a very powerful tool, results of this type seem to be not widely known in machine learning, except several applications in semi-supervised learning \citep{sokolovska2008asymptotics, kawakita2013semi}, and off-policy learning \citep{li2015toward}.

We go a step further to analyze the asymptotic property of our weighted M-estimator $\boldsymbol{\hat{\theta}}_{\KL-W}$ that maximizes $\widetilde{\eta}(\boldsymbol{\theta})$. It is natural to expect that the asymptotic variance of $\boldsymbol{\hat{\theta}}_{\KL-W}$ is smaller than that of $\boldsymbol{\hat{\theta}}_{\KL}$ based on maximizing $\hat{\eta}(\boldsymbol{\theta})$; this is shown in the following theorem. 
\begin{thm}
\label{boot:thm3}
Under Assumption~\ref{assump}, $\boldsymbol{\hat{\theta}}_{\KL-W}$ is a consistent estimator of $\boldsymbol{\theta}_{\KL}^*$ as $n\to\infty,$ and has a better asymptotic variance than $\boldsymbol{\hat{\theta}}_{\KL}$, that is, 
$$
 n \mathbb{E}\|\boldsymbol{\hat{\theta}}_{\KL-W}-\boldsymbol{\theta}_{\KL}^*\|^2  \le  n \mathbb{E}\|\boldsymbol{\hat{\theta}}_{\KL}-\boldsymbol{\theta}_{\KL}^*\|^2,
~~~~~ \text{when $n \to \infty$}.
$$
When $N>n\times d$, we have $\mathbb{E}\|\boldsymbol{\hat{\theta}}_{\KL-W}-\boldsymbol{\theta}_{\KL}^*\|^2=O(({dn^2})^{-1})$ as $n\to\infty.$
Further, its MSE for estimating the true parameter $\vv{\theta}^*$ is
\begin{align}
\label{globalklw}
\mathbb{E}\|\boldsymbol{\hat{\theta}}_{\KL-W}-\boldsymbol{\theta}^*\|^2
=O(N^{-1}+(dn^2)^{-1}).
\end{align}
\end{thm}
The proof is in Appendix C.
This result is parallel to Theorem~\ref{boot:control} for the linear control variates estimator $\boldsymbol{\hat{\theta}}_{\KL-C}$.
 Similarly, it reduces to an $O(N^{-1})$ rate once we take $n \gtrsim  (N/d)^{1/2}$.

Meanwhile,  unlike the linear control variates estimator, 
$\boldsymbol{\hat{\theta}}_{\KL-W}$ inherits all the practical advantages of KL-averaging:
it can be applied whenever the KL-naive estimator can be applied, including for models with non-identifiable parameters, or with different numbers of parameters. The implementation of $\boldsymbol{\hat{\theta}}_{\KL-W}$ is also much more convenient (see Algorithm~\ref{alg:kl-weighted}), since it does not need to calculate the Fisher information matrix as required by Algorithm~\ref{alg:kl-control}.   
\begin{algorithm}[tb]
\caption{KL-Weighted Method for Combining Local Models}
\label{alg:kl-weighted}
\begin{algorithmic}[1]
\STATE {\bfseries Input:} Local MLEs $\{\boldsymbol{\hat{\theta}}_k\}_{k=1}^d$.
\STATE {Generate bootstrap sample $\{\boldsymbol{\widetilde{x}}^k_j\}_{j=1}^n$ from each $p(\boldsymbol{x}|\boldsymbol{\hat{\theta}}_k)$, for $k\in[d].$}
\STATE {Re-estimate the local model parameter $\boldsymbol{\widetilde{\theta}}_k$ in \eqref{tildethea} based on bootstrap subsample $\{\boldsymbol{\widetilde{x}}^k_j\}_{j=1}^n$, for each $k\in[d].$}
\STATE {\bfseries Output:}  The parameter $\boldsymbol{\hat{\theta}}_{\KL-W}$ of the combined model is given by \eqref{KLweigthed}.
\end{algorithmic}
\end{algorithm}

\section{Empirical Experiments}\label{sec:empirical}
We study the empirical performance of our proposed two methods on both simulated and real world datasets. 
We first numerically verify the convergence rates predicted by our theoretical results using simulated data,
and then demonstrate the effectiveness of our methods in a challenging setting when the number of parameters of the local models are different as decided by Bayesian information criterion (BIC).
Finally, we conclude our experiments by testing our methods on a set of real world datasets.

The models we tested include
 probabilistic principal components analysis (PPCA), mixture of PPCA and Gaussian Mixtures Models (GMM). PPCA model is  defined with the help of a hidden variable $\vv t$,  $$ p(\vv x  ~| ~ \vv \theta) = \int p(\boldsymbol{x} ~|~ \boldsymbol{t}; ~  \vv\theta)  p(\vv t ~|~  \vv \theta) d \vv t,$$
where
$p(\boldsymbol{x}\mid\boldsymbol{t};~ \vv\theta)=\mathcal{N}( \vv x;  ~ \boldsymbol{\mu}+W\boldsymbol{t},\sigma^2),$ and the distribution of hidden variable $ \boldsymbol{t}$ is $p (\boldsymbol{t} \mid \vv\theta) =  \mathcal{N}(\vv t; ~ \boldsymbol{0}, \boldsymbol{I})$ and $\vv \theta = \{\boldsymbol{\mu}, ~W, ~\sigma^2\}$.
The mixture of PPCA is $p(\vv x \mid \vv\theta) = \sum_{s=1}^m\alpha_s p_s(\vv x \mid \vv\theta_s)$, where $\vv\theta = \{\alpha_s, \vv\theta_s\}_{s=1}^m$ and each $p_s(\vv x \mid \vv\theta_s)$ is a PPCA model.

GMM is given by
 $p(\boldsymbol{x}\mid \boldsymbol{\theta})=\sum_{s=1}^m\alpha_s\mathcal{N}(\boldsymbol{\mu}_s, \Sigma_s)$ where $\mathcal{N}(\boldsymbol{\mu}_s, \Sigma_s)$ is a multivariate Gaussian distribution and the parameters are $\boldsymbol{\theta}=(\alpha_s, \boldsymbol{\mu}_s, \Sigma_s).$
 
Because all these models are latent variable models with unidentifiable parameters, the direct linear averaging method is not applicable.
For GMM, it is still possible to use a \emph{matched linear averaging} which matches the mixture components of the different local models by minimizing a symmetric $\KL$ divergence; the same idea can be used on our linear control variates method to make it applicable to GMM. On the other hand, because the parameters of PPCA-based models are unidentifiable up to arbitrary orthonormal transforms, linear averaging and linear control variates can no longer be applied easily. We use expectation maximization (EM) to learn the parameters in all these three models. 
\subsection{Numerical Verification of the Convergence Rates}
 We start with verifying the convergence rates
 in \eqref{globalmse}, \eqref{globalklc} and \eqref{globalklw}
 of MSE $\E||\vv{\hat\theta} -\vv \theta^*||^2$ of
 the different estimators for estimating the true parameters.
 Because there is also an non-identifiability problem in calculating the MSE,
we again use the symmetric KL divergence to match the mixture components,
and evaluate the MSE on $WW^\top$ to avoid the non-identifiability w.r.t. orthonormal transforms.

\begin{figure}[htb]
\begin{centering}
\begin{tabular}{ccc}
\includegraphics[height=0.28\textwidth]{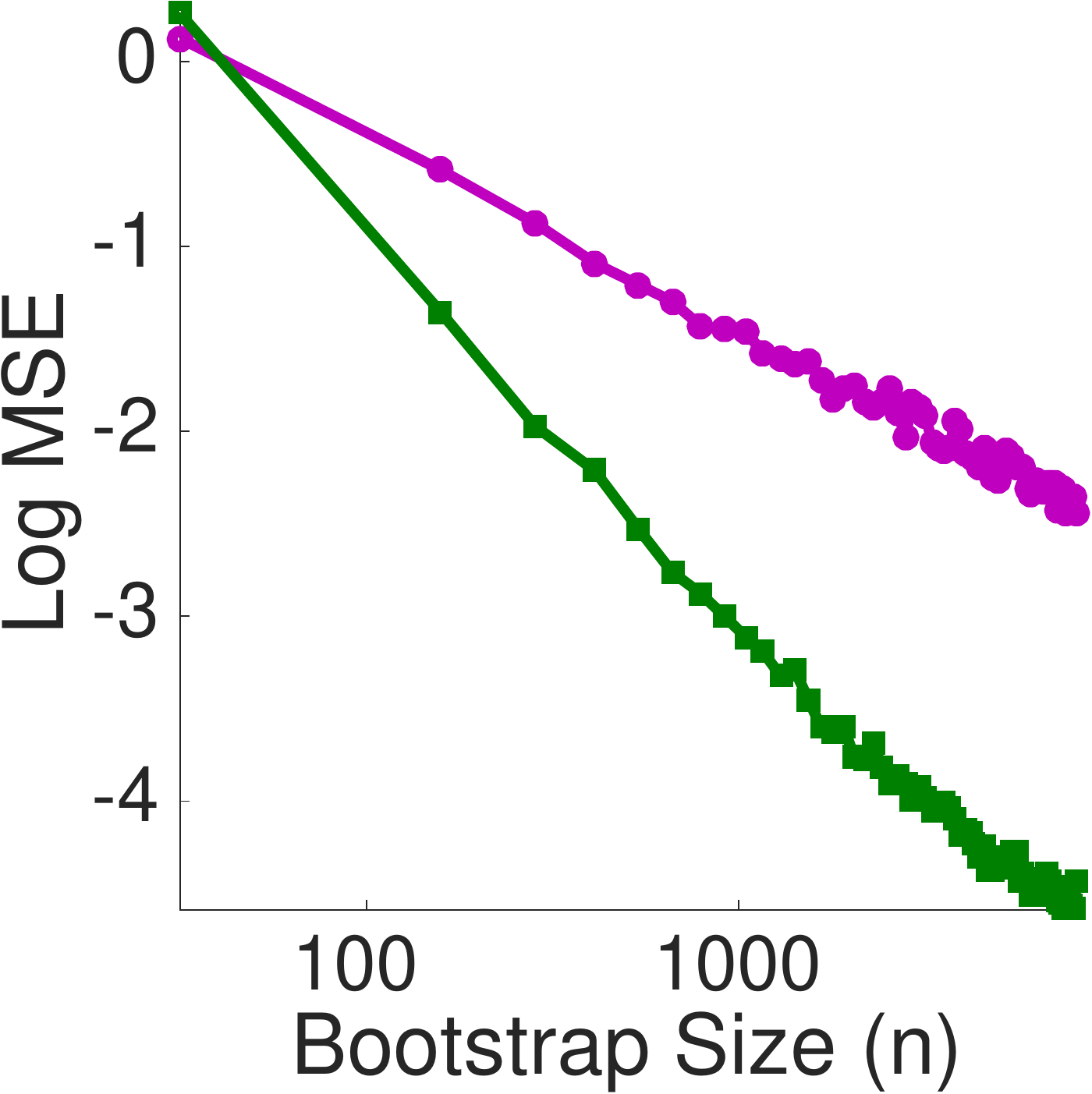}&
\includegraphics[height=0.28\textwidth]{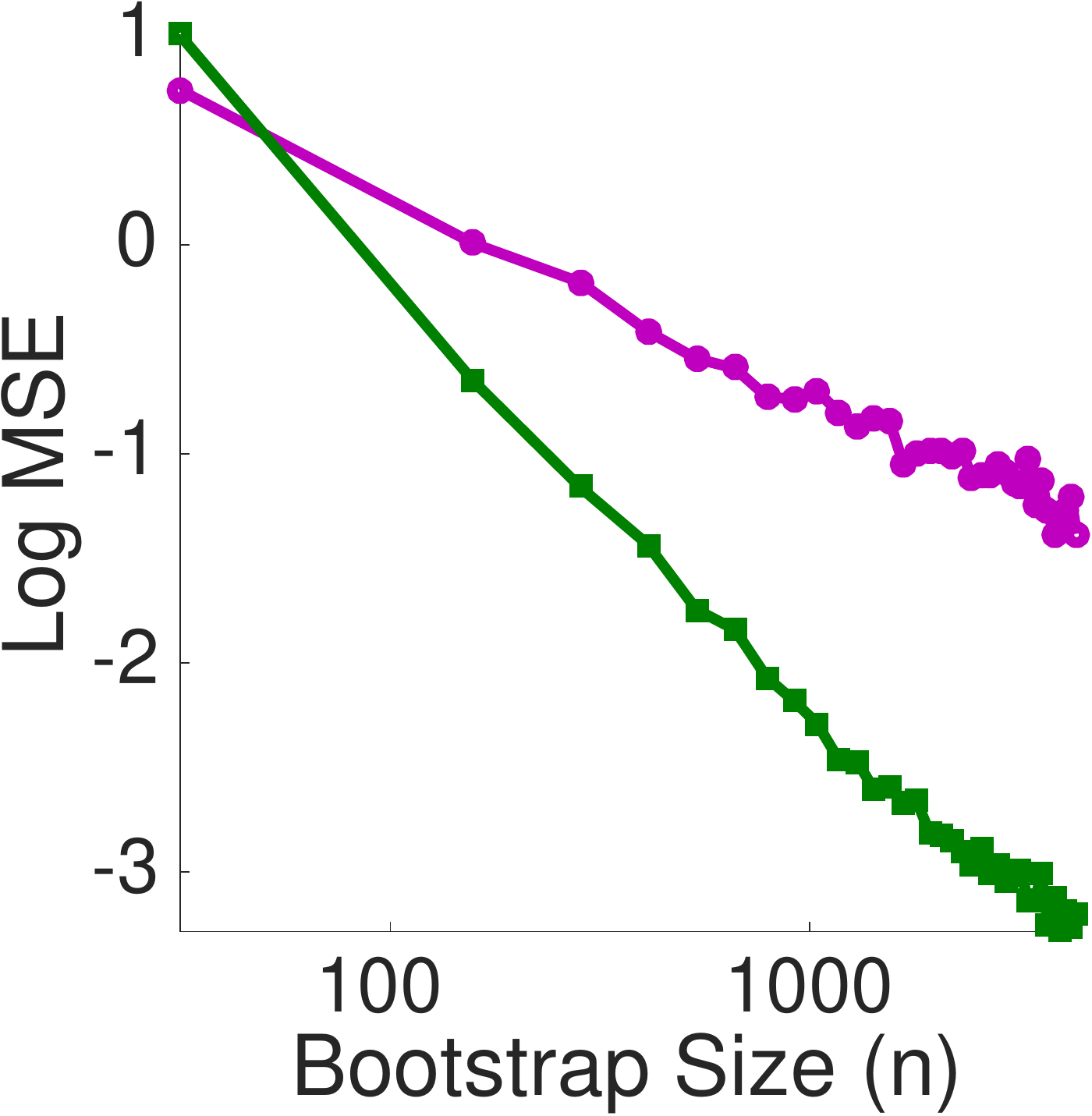}&
\includegraphics[height=0.28\textwidth]{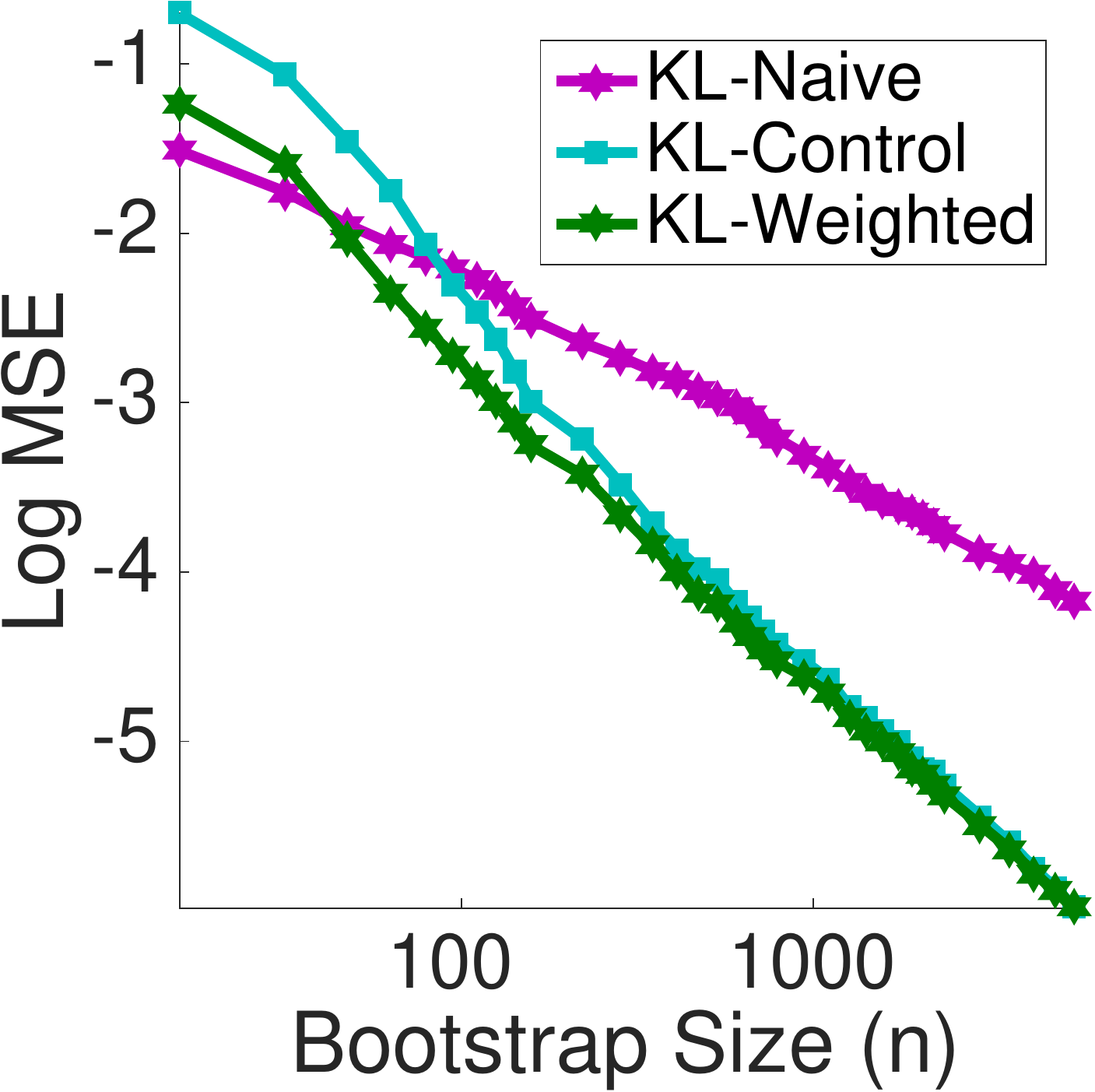} \\
{\small (a) PPCA} &
{\small (b) Mixture of PPCA}&
{\small (c) GMM }
\end{tabular}
\caption[Verification experiments of proposed methods on PPCA, mixture of PPCA and GMM with simulated data]{
Results on different models with simulated data when we change the bootstrap sample size $n$, with fixed $d=10$ and $N=6\times10^7$. \label{fig:simple}}
\end{centering}
\end{figure}

 To verify the convergence rates w.r.t. $n$, we fix $d$ and let the total dataset $N$ be very large so that $N^{-1}$ is negligible. The dimensions of the PPCA models in (a)-(b) are 5, and that of GMM in (c) is 3.
The numbers of mixture components in (b)-(c) are 3.
Linear averaging and KL-Control are not applicable for the PPCA-based models, and are not shown in (a) and (b).
 %
Figure~\ref{fig:simple} shows the results when we vary $n$,
where we can see that the MSE of
KL-naive ${\boldsymbol{\hat\theta}}_{\KL}$ is $O(n^{-1})$
while that of KL-control $\hat{\boldsymbol{\theta}}_{\KL-C}$ and KL-weighted $\hat{\boldsymbol{\theta}}_{\KL-W}$ are $O(n^{-2})$;
both are consistent with our results in \eqref{globalmse}, \eqref{globalklc} and \eqref{globalklw}.

\begin{figure}[h]
\begin{centering}
\begin{tabular}{cccc}
\hspace{-6mm}
\includegraphics[width=0.23\textwidth]{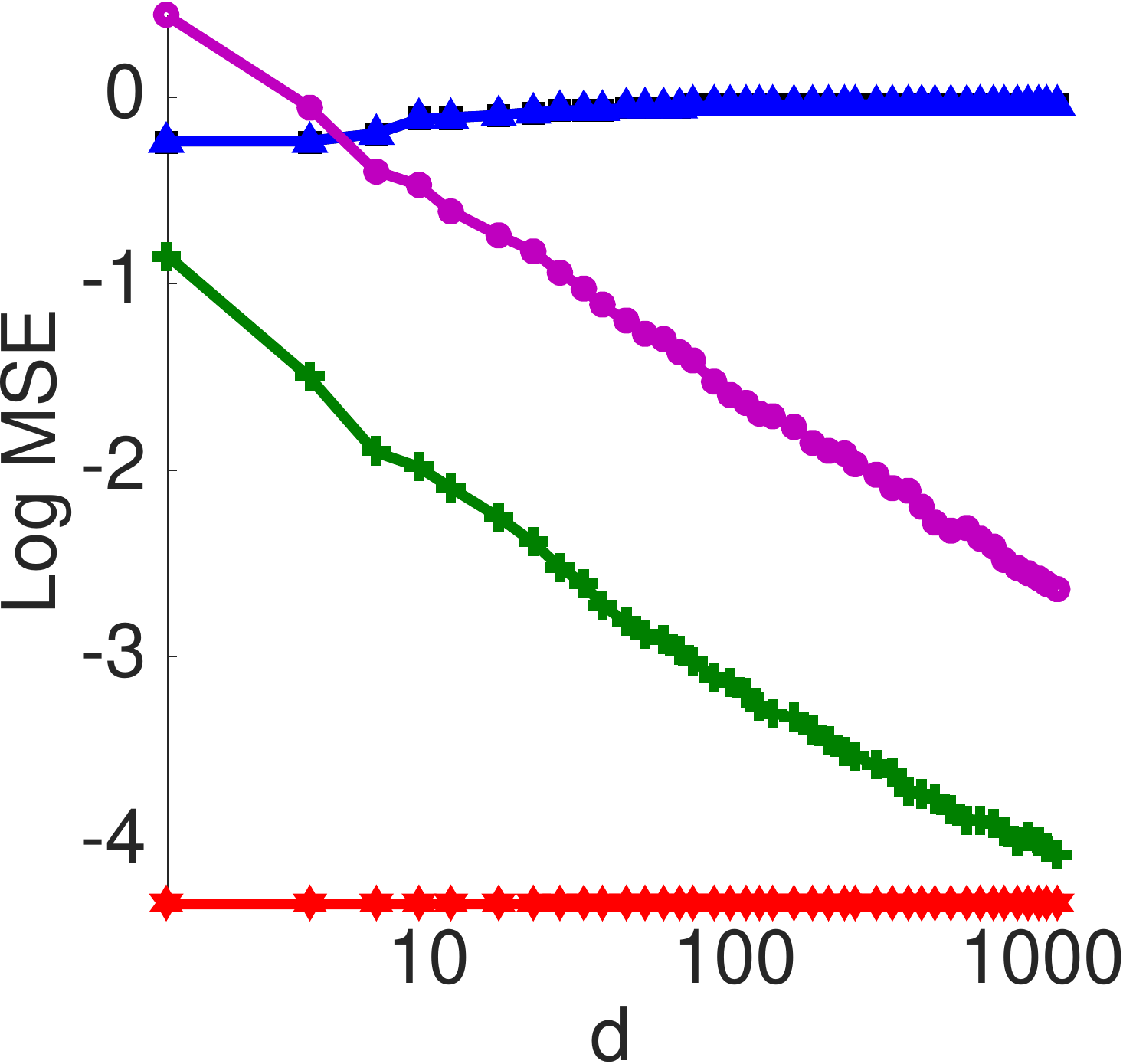} &\hspace{-.4cm}
\includegraphics[width=0.23\textwidth]{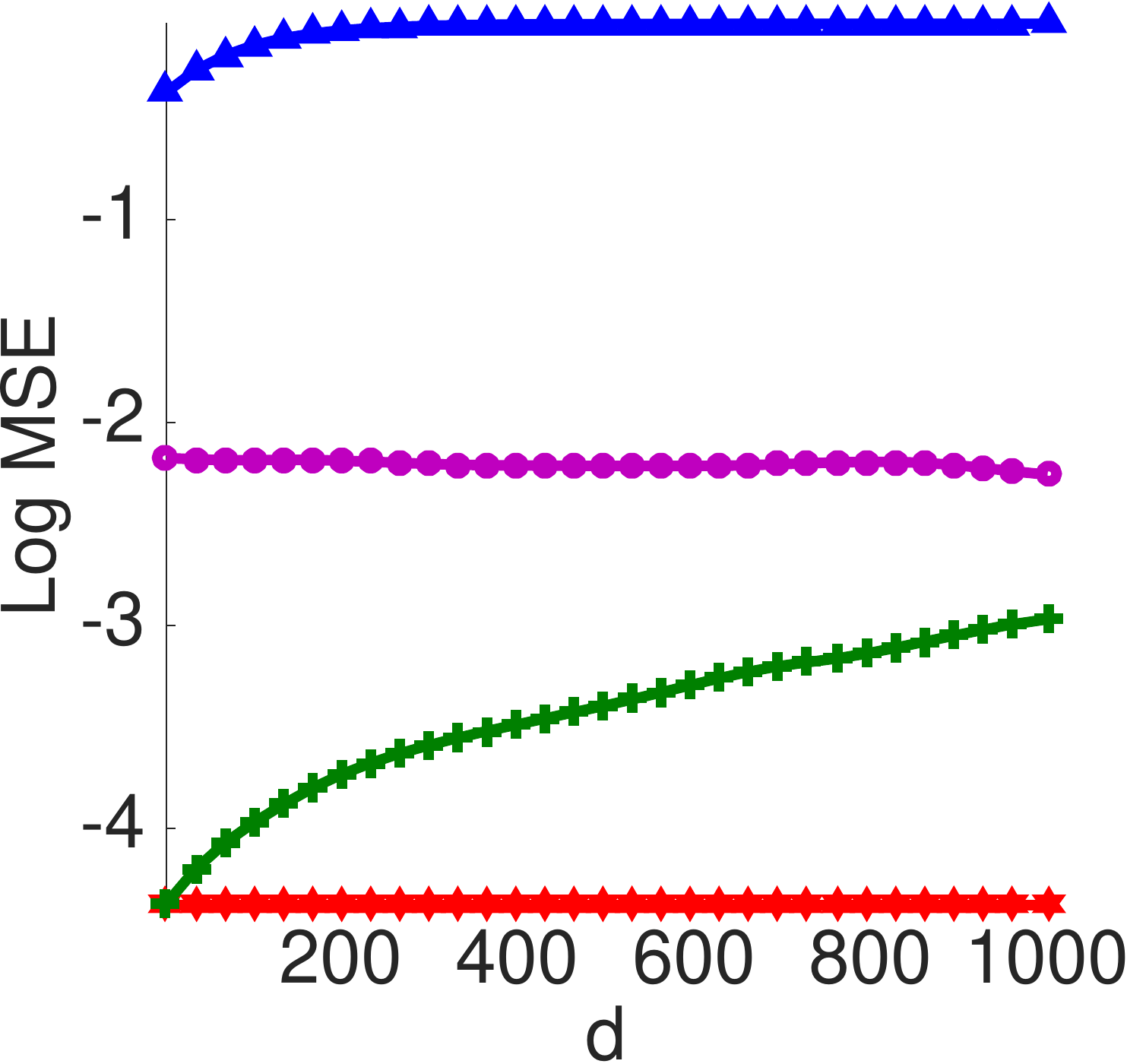} &\hspace{-.8cm}
\includegraphics[width=0.23\textwidth]{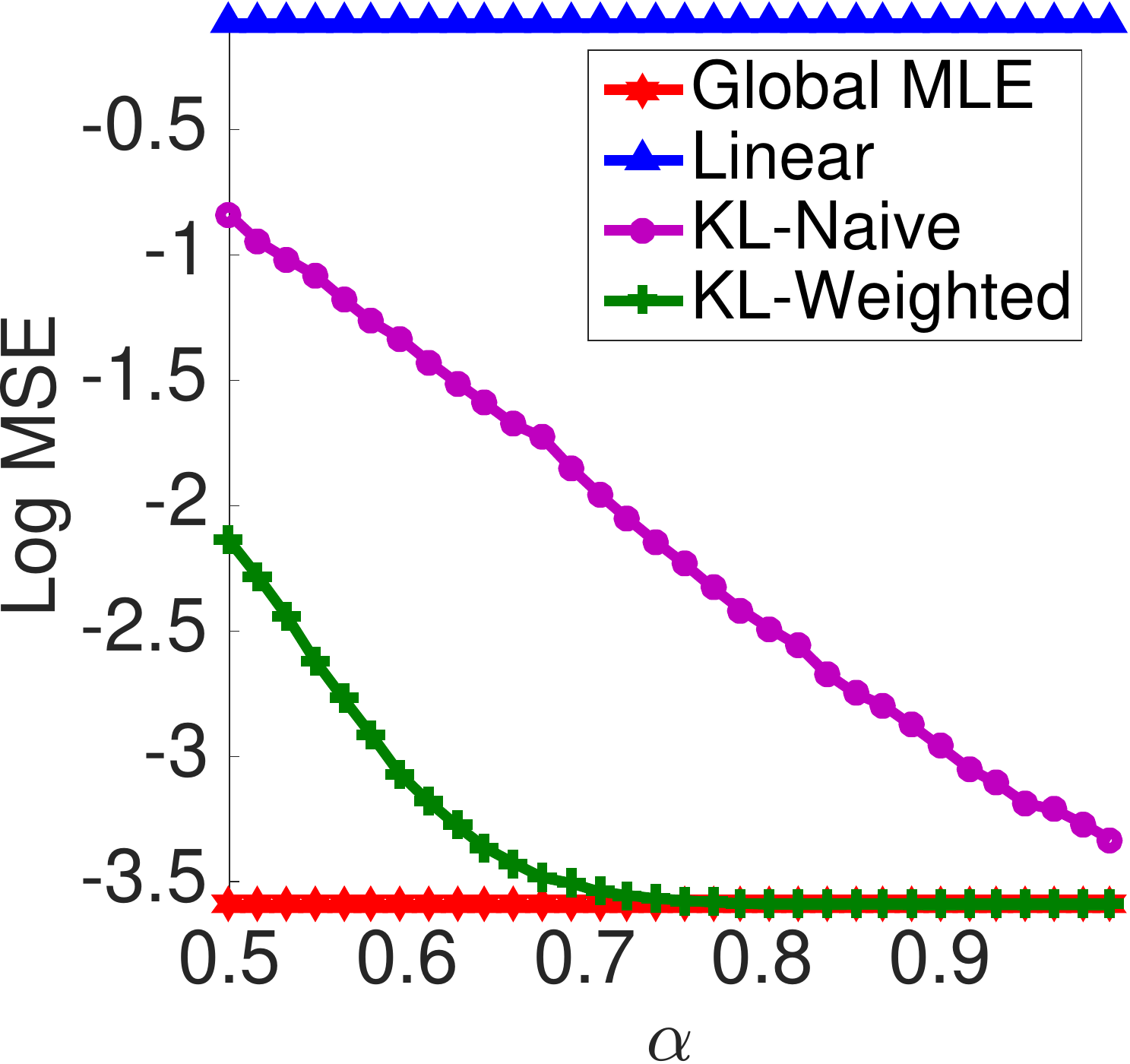} &\hspace{-.8cm}
\includegraphics[width=0.23\textwidth]{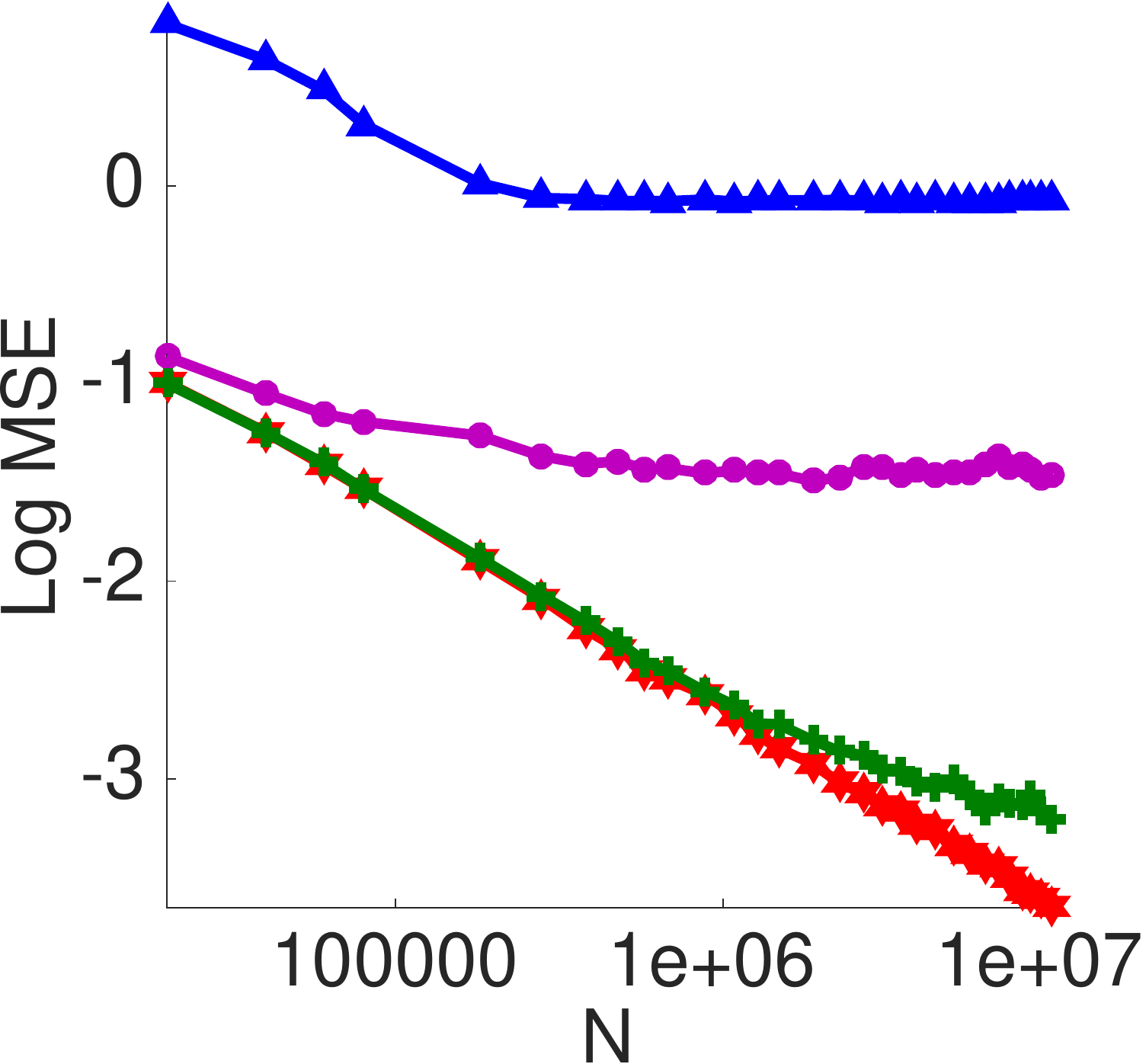} \\
{\small(a) Fix $N$ and $n$} &
{\small(b) Fix $N$ and $n_{\mathrm{tot}}$} &
{\small(c) Fix $N$, $n=(\frac Nd)^{\alpha}$ and $d$} &
 {\small(d) Fix $n$ and $d$}
 \end{tabular}
\caption[Further verification experiments of proposed methods on PPCA with simulated data]{Further experiments on PPCA with simulated data.
(a) varying $n$ with fixed $N=5\times 10^7$. (b) varying $d$ with $N=5\times10^7$, $n_{\mathrm{tot}} = n \times d = 3\times 10^5$.
(c) varying $\alpha$ with  $n=(\frac Nd)^{\alpha}$, $N=10^7$ and $d$. (d) varying $N$ with $n=10^3$ and $d = 20$.
The dimension of data $\boldsymbol{x}$ is 5 and the dimension of latent variables $\boldsymbol{t}$ is 4.}
\label{fig:ppcamore}
\end{centering}
\end{figure}

In Figure~\ref{fig:ppcamore}(a),
we increase the number $d$ of local machines,
while using a fix $n$ and a very large $N$,
and find that both $\vv{\hat\theta}_{\mathrm{KL}}$ and $\vv{\hat\theta}_{\mathrm{KL}-W}$ scales as $O(d^{-1})$ as expected.
Note that since the total observation data size $N$ is fixed, the number of data in each local machine is $(N/d)$ and it decreases as we increase $d$.
It is interesting to see that the performance of the KL-based methods actually increases with more partitions;
this is, of course, with a cost of increasing the total bootstrap sample size $d n$ as $d$ increases.
Figure~\ref{fig:ppcamore}(b) considers a different setting,
in which we increase $d$ when fixing the total observation data size $N$, and the total bootstrap sample size $n_{\mathrm{tot}}=n\times d$.
According to \eqref{globalmse} and  \eqref{globalklw}, the MSEs of
$\vv{\hat\theta}_{\mathrm{KL}}$ and $\vv{\hat\theta}_{\mathrm{KL}-W}$ should be about $O(n_{\mathrm{tot}}^{-1})$ and $O(d n_{\mathrm{tot}}^{-2})$ respectively when $N$ is very large, and this is consistent with the results in Figure~\ref{fig:ppcamore}(b).
It is interesting to note that the MSE of $\vv{\hat\theta}_{\mathrm{KL}}$ is independent with $d$ while that of $\vv{\hat\theta}_{\mathrm{KL}-W}$ increases linearly with $d$.
This is not conflict with the fact that $\vv{\hat\theta}_{\mathrm{KL}-W}$ is better than $\vv{\hat\theta}_{\mathrm{KL}}$,  since we always have $d \leq n_{\mathrm{tot}}$.

Figure~\ref{fig:ppcamore}(c) shows the result when we set $n = (N/d)^{\alpha}$ and vary $\alpha$,
where we find that $\vv{\hat\theta}_{\mathrm{KL}-W}$ quickly converges to the global MLE as $\alpha$ increases, while the KL-naive estimator $\vv{\hat\theta}_{\mathrm{KL}}$ converges significantly slower. 
Figure~\ref{fig:ppcamore}(d) demonstrates the case when we increase $N$ while fix $d$ and $n$,
where we see our KL-weighted estimator $\vv{\hat\theta}_{\mathrm{KL}-W}$ matches closely with $N$, except when $N$ is very large in which case the $O((dn^2)^{-1})$ term starts to dominate, while KL-naive is much worse.
We also find the linear averaging estimator performs poorly, and does not scale with $O(N^{-1})$ as the theoretical rate claims;
this is due to unidentifiable orthonormal transform in the PPCA model that we test on.

\subsection{Gaussian Mixture with Unknown Number of Components}
\begin{figure}[h]
\begin{centering}
\begin{tabular}{ccc}
\includegraphics[height=0.24\textwidth]{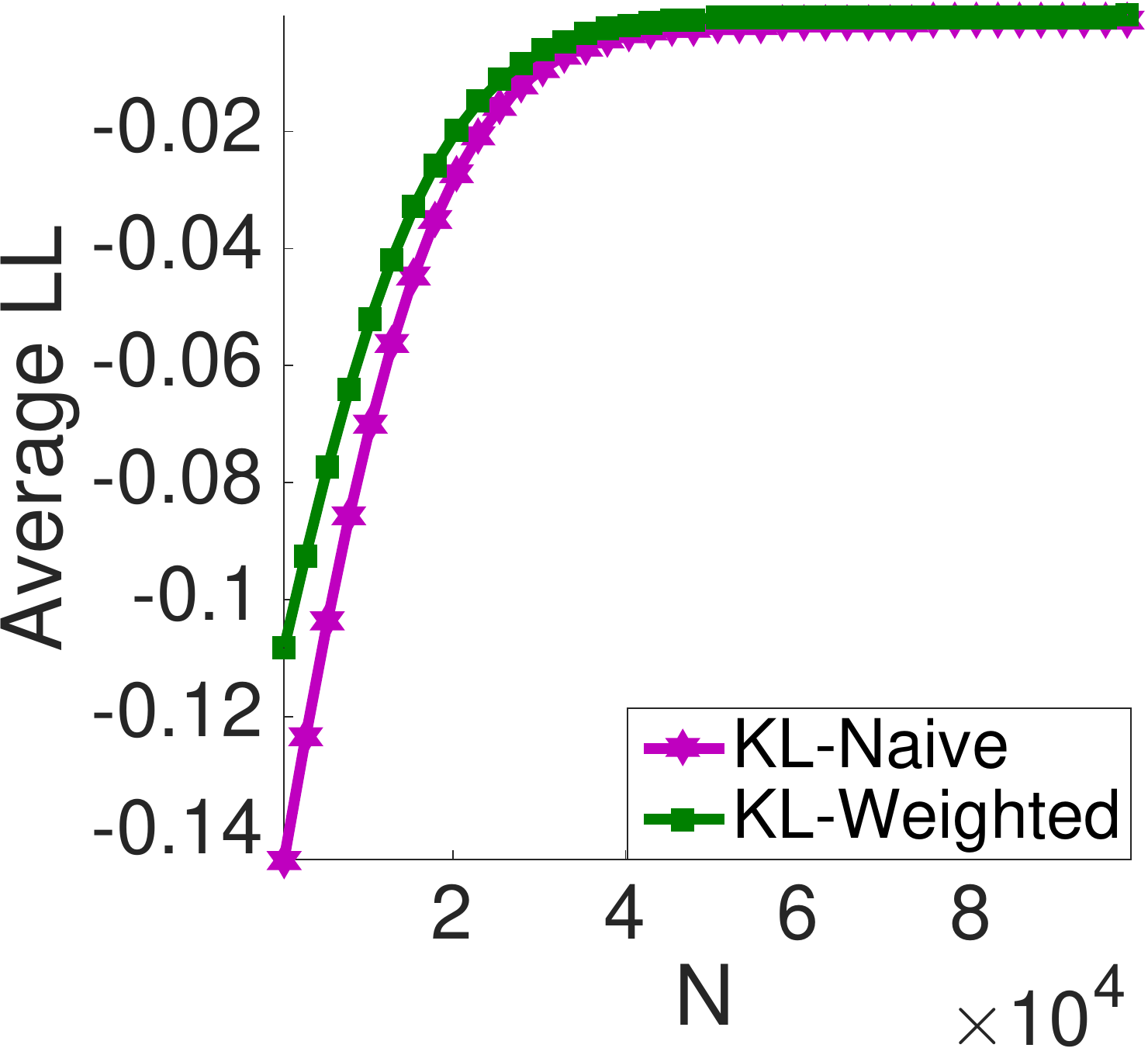} &
\includegraphics[height=0.24\textwidth]{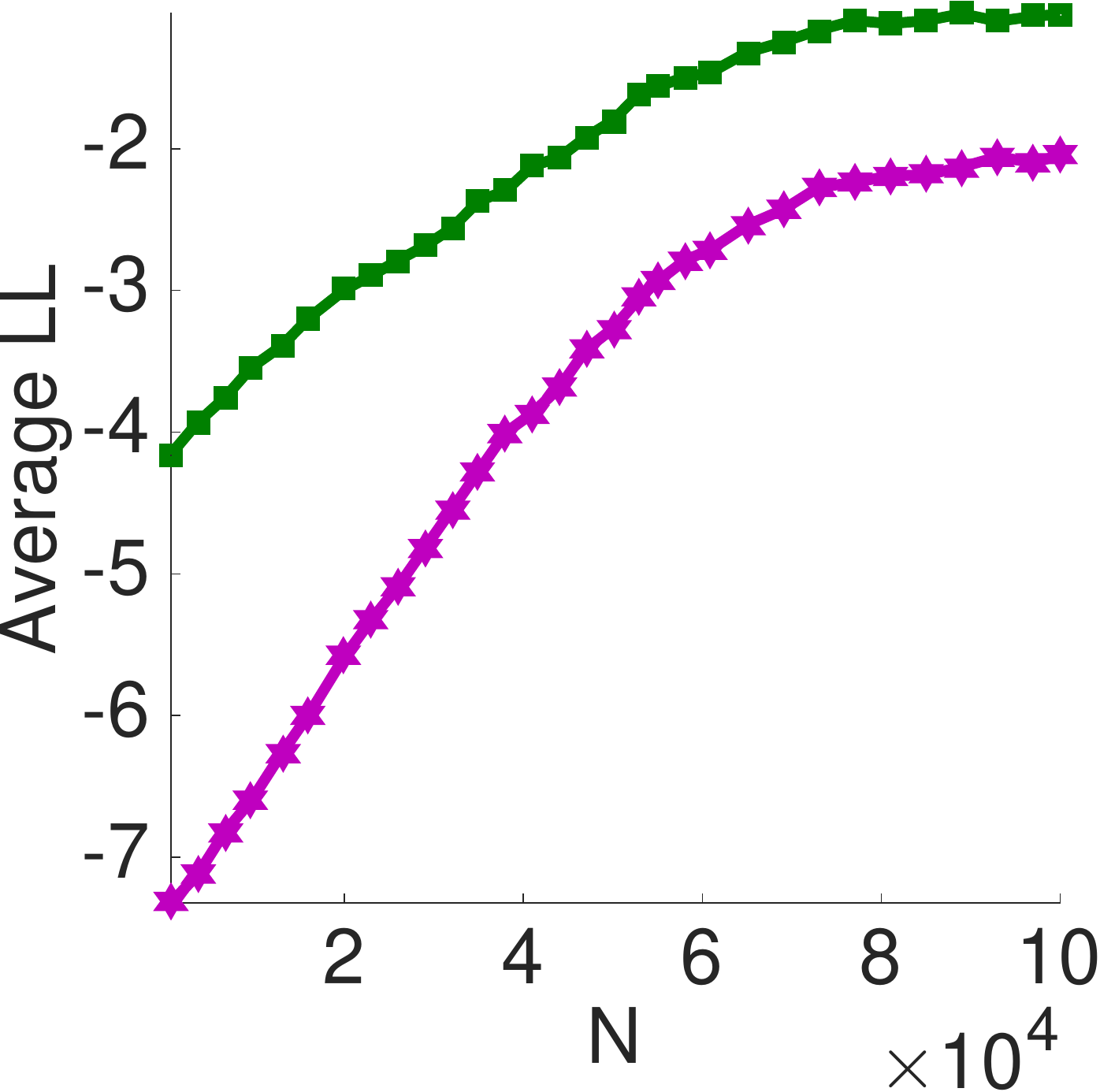} &
\includegraphics[height=0.24\textwidth]{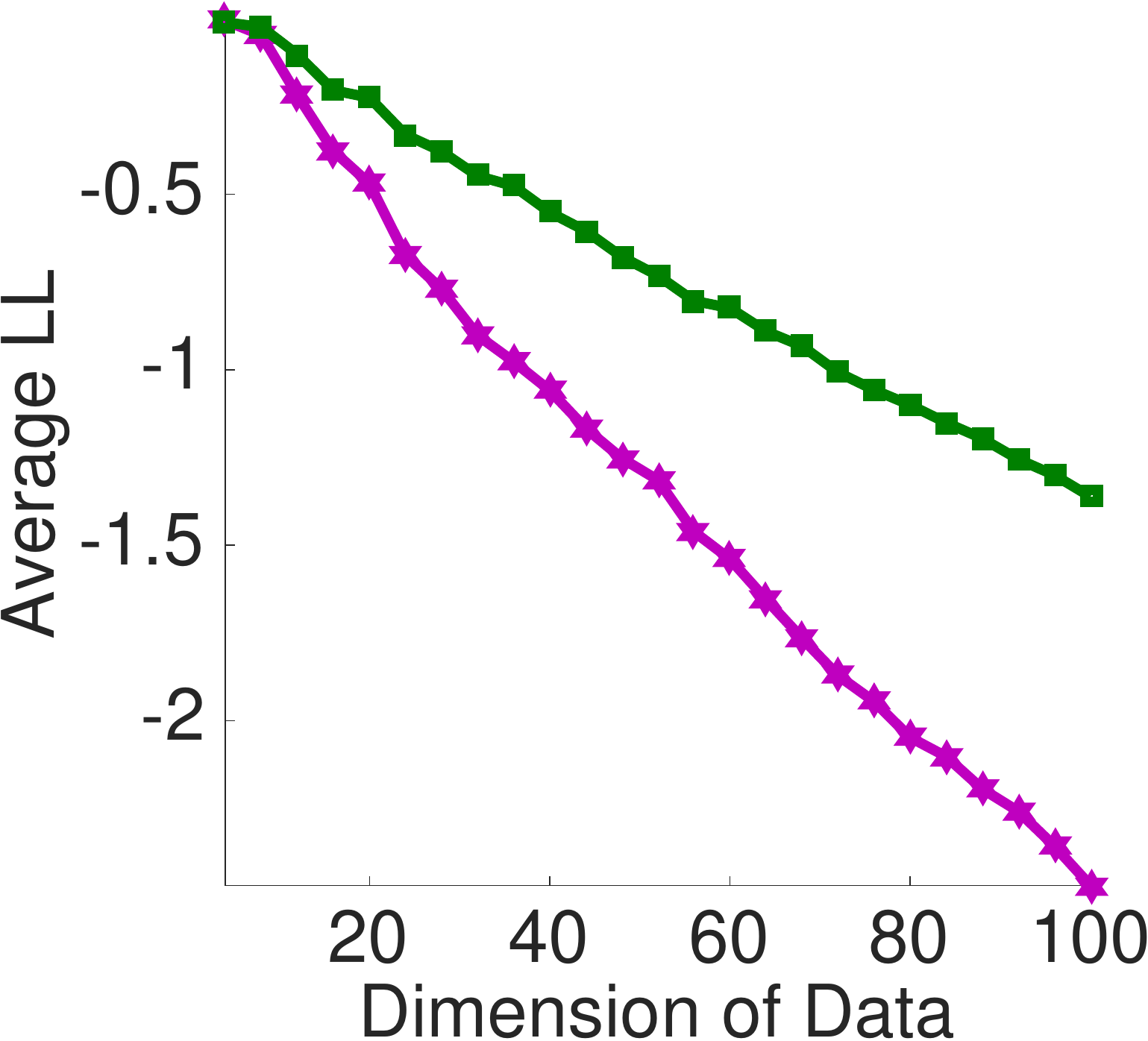} \\
{\small (a)  Dimension of $\vv x$ = 3 } &
{\small (b)  Dimension of $\vv x$ = 80  } &
{\small (c) varying the dimension of $\vv x$}
\end{tabular}
\caption[Performance of $\KL$-averaging methods on GMM with unknown number of mixture components]{GMM with the number of mixture components estimated by Bayesian information criterion. We set $n=600$
and the true number of mixtures to be 10 in all the cases.
(a)-(b) vary the total data size $N$ when the dimension of $\vv x$ is 3 and 80, respectively.
(c) varies the dimension of the data with fixed $N=10^5$.
The y-axis is the testing $\log$ likelihood compared with that of global MLE.
}
\label{fig:fig3}
\end{centering}
\end{figure}

We further apply our methods to a more challenging setting for
distributed learning of GMM when the number of mixture components is unknown.
In this case, we first learn each local model with EM and decide its number of components using BIC selection.
Both linear averaging and KL-control $\vv{\hat\theta}_{\KL-C}$ are not applicable in this setting, and and we only test KL-naive $\vv{\hat\theta}_{\mathrm{KL}}$ and KL-weighted $\vv{\hat\theta}_{\mathrm{KL}-W}$.
Since the MSE is also not computable due to the different dimensions, we evaluate
$\vv{\hat\theta}_{\mathrm{KL}}$ and
$\vv{\hat\theta}_{\mathrm{KL}-W}$
using the log-likelihood on a hold-out testing dataset as shown in Figure~\ref{fig:fig3}.
We can see that $\vv{\hat\theta}_{\mathrm{KL}-W}$ generally outperforms
$\vv{\hat\theta}_{\mathrm{KL}}$ as we expect, and the relative improvement increases significantly as the dimension of the observation data $\vv x$ increases.
This suggests that our variance reduction technique works very efficiently in high dimension problems.

\subsection{Results on Real World Datasets}
\begin{figure}[h]
\begin{centering}
\begin{tabular}{ccc}
\hspace{-.5cm}
\includegraphics[height=0.28\textwidth]{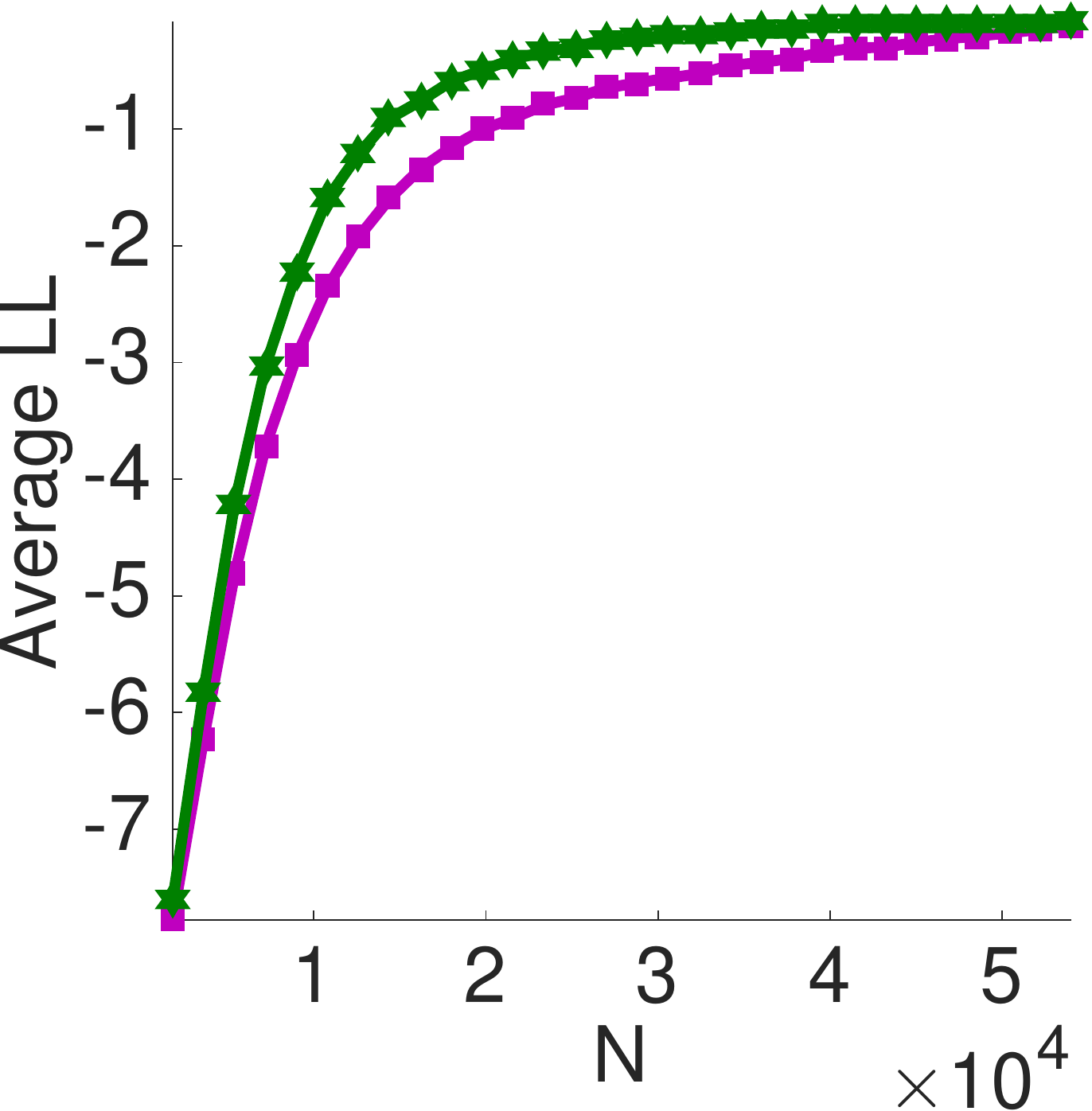} & \hspace{-.8cm}
\includegraphics[height=0.28\textwidth]{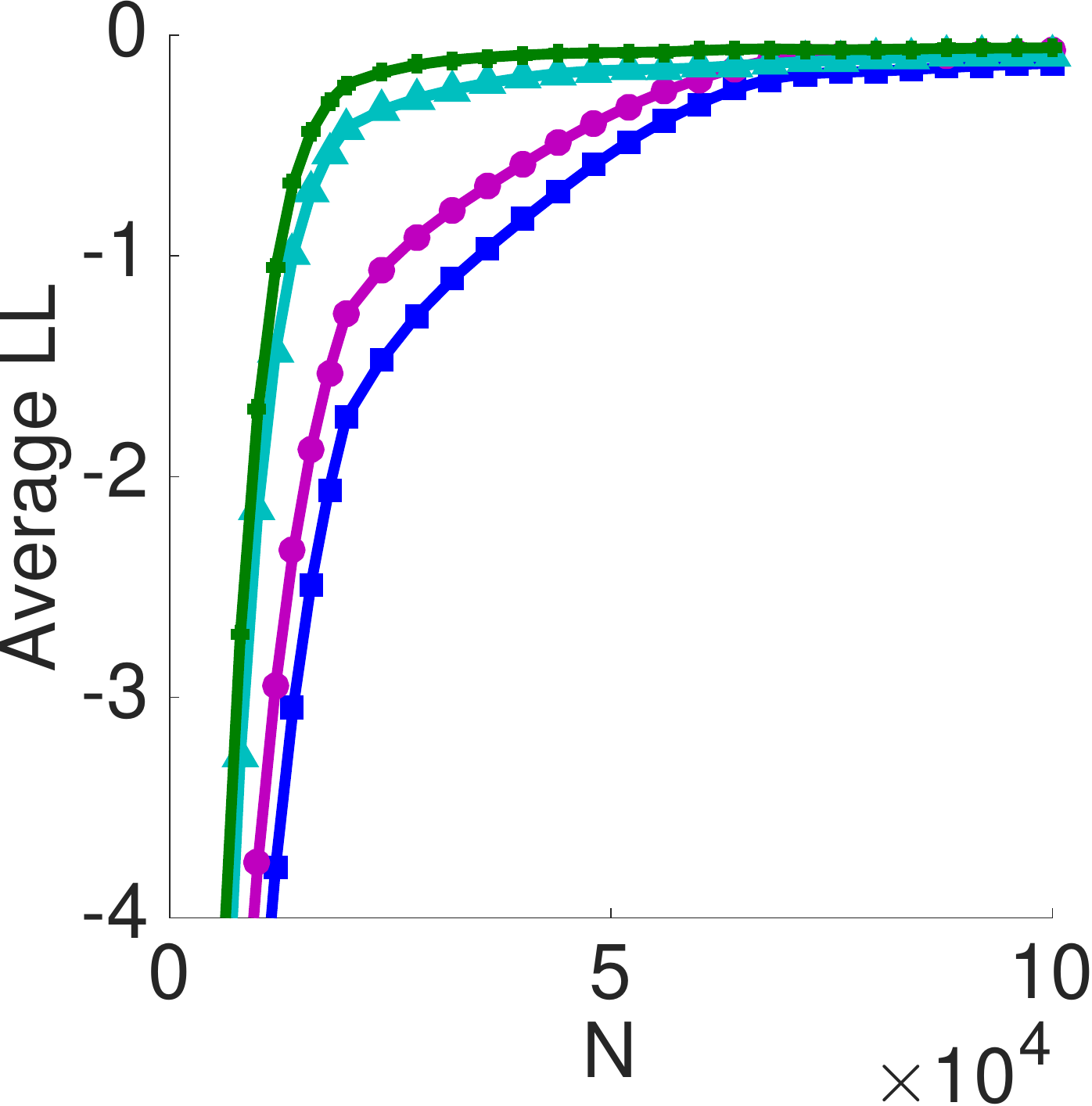} & \hspace{-.5cm}
\includegraphics[height=0.28\textwidth]{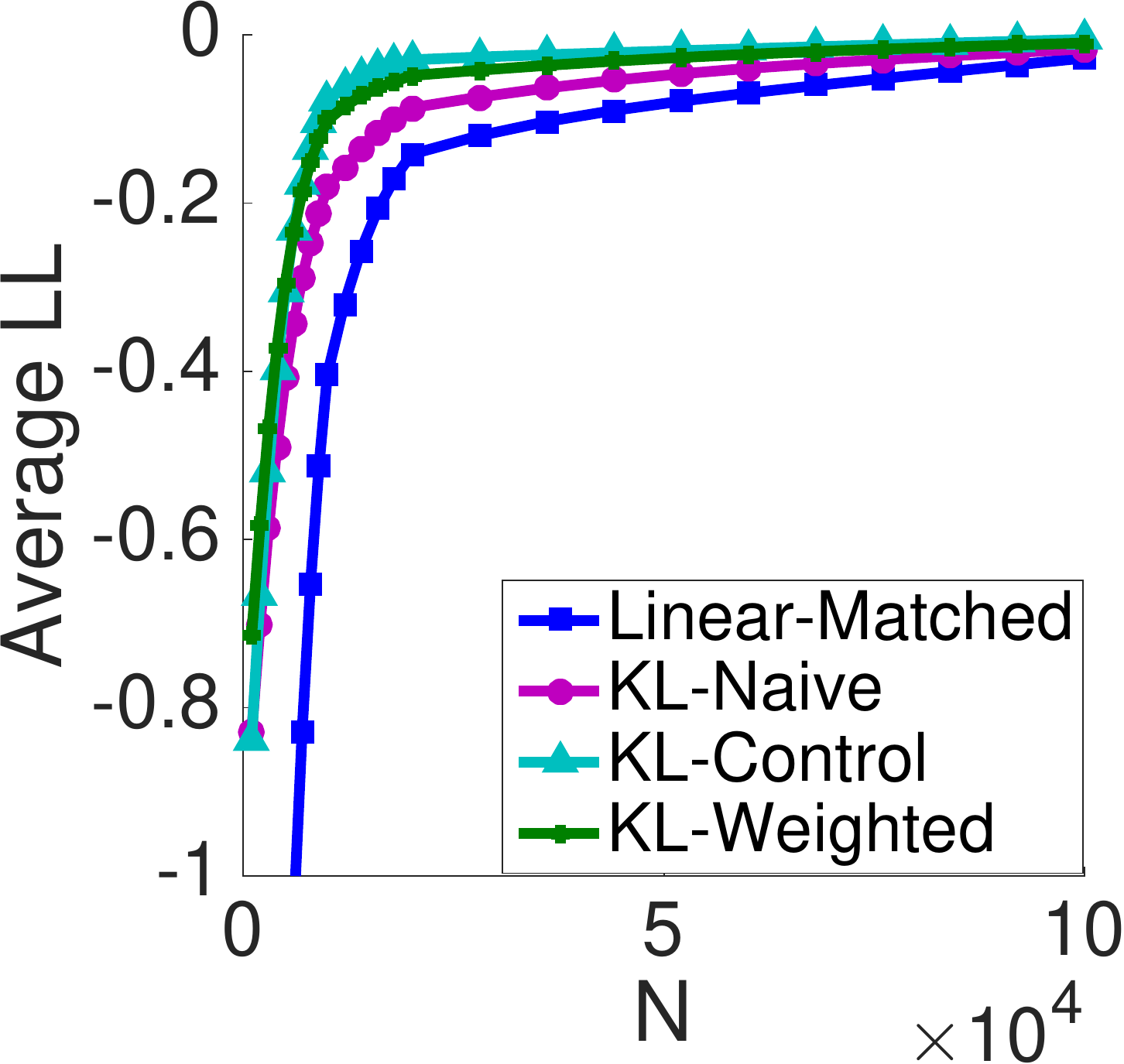} \\
{\small (a) Mixture of PPCA, SensIT Vehicle} &
{\small (b) GMM, Covertype} &
{\small (c) GMM, Epsilon}
\end{tabular}
\caption[Performace of our proposed $\KL$-averaging method with different models on real datasets]{Testing $\log$ likelihood (compared with that of global MLE) on real world datasets.
(a) Learning Mixture of PPCA on SensIT Vehicle. (b)-(c) Learning GMM on Covertype and Epsilon.
The number of local machines is 10 in all the cases,
and the number of mixture components are taken to be the number of labels in the datasets. 
The dimension of latent variables in (a) is 90.
For Epsilon dataset, the PCA is first applied and the top 100 principal components are chosen.
Linear-matched and $\KL$-Control are not applicable on Mixture of PPCA and are not shown on (a).
}
\label{fig:fig4}
\end{centering}
\end{figure}

Finally, we apply our methods to several real world datasets, including
the SensIT Vehicle dataset on which mixture of PPCA is tested, and
the Covertype and Epsilon datasets on which GMM is tested.
From Figure~\ref{fig:fig4}, we can see that our KL-Weight and KL-Control (when it is applicable) again
perform the best. The (matched) linear averaging performs poorly on GMM (Figure~\ref{fig:fig4}(b)-(c)), while is not applicable on mixture of PPCA.


\section{Summary}
This chapter addresses the model aggregation problem in distributed, or privacy-preserving learning. We consider the one-shot communication by sending the probabilistic models estimated from different local repositories $\{p(\boldsymbol{x}|\boldsymbol{\hat{\theta}}_k)\}_{k=1}^d$ to a fusion center. The simple way is to linearly average the parameters of the local models, $\boldsymbol{\hat{\theta}}=\frac{1}{d}\sum_{j=1}^d \boldsymbol{\hat{\theta}}_j$. Although nearly asymptotically optimal error rates can be achieved, this linear-averaging method tends to degenerate in practical scenarios for models with non-convex log-likelihood or non-identifiable parameters (such as latent variable models and neural models), and is not applicable at all for models with non-additive parameters (e.g., when the parameters have discrete or categorical values, the number of parameters in local models are different, or the parameter dimensions of the local models are different). To overcome the limitations of the parameter-averaging based method, we instead find a joint model $p(\boldsymbol{x}|\boldsymbol{\hat{\theta}})$ which has a minimal sum of the $\KL$ divergence between each local model and the joint model, i.e., $\boldsymbol{\hat{\theta}}=\argmax_{\vthe} \sum_{j=1}^d \KL(p(\boldsymbol{x}\mid \boldsymbol{\hat{\theta}}_k)|| p(\vx|\vthe)).$ The similar objective has been used in optimizing the approximate distribution $q_{\vthe}(\vx)$ in variational inference, $\min_{\vthe}\KL(q_{\vthe}(\vx)|| p(\vx)).$ 
As the closed form of the integration induced from the $\KL$ divergence is unavailable in most cases~(for most choices of $p(\boldsymbol{x}\mid \boldsymbol{\hat{\theta}}_k)$, the integration cannot be analytically evaluated), we need to sample bootstrap data $\{\vv{\widetilde{x}}_j^k\}_{j=1}^n$ from each local model $p(\boldsymbol{x}\mid \boldsymbol{\hat{\theta}}_k)$ to do naive Monte Carlo estimation of the integration. Unfortunately, this bootstrapped procedure induces large variance (the mean square error rate between the naive estimator and the ground truth value is $O(N^{-1}+(dn)^{-1})$), when $n$ is typically assumed to be small, which limits its application. In this chapter, we propose two effective variance reduction techniques to learn a joint model $p(\boldsymbol{x}|\boldsymbol{\hat{\theta}}),$ including a KL-weighted estimator that is both statistically efficient and widely applicable for even challenging practical scenarios.
Theoretical analysis is provided to understand the statistical property of our proposed methods. The two estimators proposed in this chapter are unbiased and has MSE rate $O(N^{-1}+(d^{-1}n^{-2}),$ where $N$ is the size of the original dataset and $n$ is the size of the bootstrapped sample for each local model $p(\boldsymbol{x}\mid \boldsymbol{\hat{\theta}}_k)$. We perform experiments on three different models, PPCA, mixture PPCA and GMM. A sufficient empirical experiments for the PPCA and GMM models on simulated datasets exactly verify the theoretical correctness of our proposed two estimators under different choices of $N$, $n$ and $d$. The empirical results on various real datasets demonstrate the practical power of our proposed $\KL$-averaging method when applied to latent variable model and multi-modal models with non-convex nature. Future directions include extending our methods to discriminant learning tasks, as well as the more challenging deep generative networks
on which the exact MLE is not computationally tractable, and surrogate likelihood methods with stochastic gradient descent are needed. 
We note that the same KL-averaging problem also appears in the ``knowledge distillation" problem in Bayesian deep neural networks \citep{korattikara2015bayesian}, and it seems that our technique can be applied straightforwardly. Our $\KL$ averaging method can also be directly applied to optimize the approximate distribution in variational inference. 
 
\chapter{Conclusions and Future Works}
In this thesis, we propose several approximate inference algorithms, which can be effectively applied to continuous-valued distributions and discrete-valued distributions. Traditional MCMC-based methods, which usually runs very long Markov chains to approximate the target distributions, have theoretical guarantee of the convergence of the chain to the target distribution but are very slow. Variational inference methods approximate the target distribution $p(\vx)$ with an approximate distribution $q_{\phi}(\vx)$ and optimize $\phi$, which are much faster but the predefined parametric distribution $q_{\phi}(\vx)$ tends to give a poor approximation as $p(\vx)$ might not be from the predefined parametric family. SVGD~\citep{liu2016stein} is an alternative
framework that integrates both the advantages of particle-based methods and variational algorithms. In this thesis, we propose several new methods in the framework of SVGD, where the original SVGD cannot be readily applied. The whole thesis can be summarized into three main pipelines. Our recently proposed approximate inference framework provide a powerful tool and new direction to perform faster and more accurate inference. While several new questions deserve to be investigated, which might ultimately improve current performance and further achieve the state-of-the-art results in approximate inference. We will discuss them in detail in each pipeline. 

\paragraph{Approximate Inference on Continuous-valued Distribution}
In the following, we first emphasize our contributions of approximate inference algorithms on continuous-valued distributions. Specifically, we propose a nonparametric adaptive importance sampling algorithm by decoupling the iteratively updated particles of SVGD $\{\vx_i^\ell\}_{i=1}^n$ into two sets: \emph{leader particles} $\vx_A^\ell = \{\vx_i^\ell  \colon  i\in A\}$ and \emph{follower particles}
$\vx_B^\ell = \{\vx_i^\ell  \colon  i\in B\}$, with $B =  \{1,\ldots, n\} \setminus A$. The leader particles is applied to construct the transform and the follower particles are updated by the constructed transform, $\vv x_{i}^{\ell+1} \gets \vv x_i^\ell + \epsilon \ff_{\ell+1}(\vv x_i^\ell), ~ \forall i \in A\cup B,$ where $\ff_{\ell+1}(\cdot)$ is constructed by only  
using particles in set $A,$
\begin{align}\notag
 \!\!\!\! \!\!\!\!
\ff_{\ell+1}(\cdot) =  \frac{1}{|A|}\sum_{j\in A} [\nabla \log p(\vv x_j^\ell) k(\vv x_j^\ell,  \cdot) + \nabla_{\vv x_j^\ell} k(\vv x_j^\ell, \cdot)].
\end{align}

With such a transform, the distribution of the updated particles $\{\vv x_j^\ell\}$ in $\vx_B$ satisfies \begin{align}\label{intro:equ:qt}
q_{\ell} =  (\T_{\ell} \circ \cdots \circ \T_{1})\sharp q_0, \quad \ell=1, \ldots, K,
\end{align}
where the importance proposal $q_\ell$ forms increasingly better approximation of the target $p$ as $\ell$ increases. Conditional on $\vx_A^\ell,$ particles in $\vx_B^\ell$ are i.i.d. and hence can provide an unbiased estimation of the integral $\E_{\vx\sim q_\ell}[\frac{p(\vx)}{q_\ell(\vx)}h(\vx)]$ for any function $h(\vx).$ Our importance proposal is not restricted to the predefined distributional family as traditional adaptive importance sampling methods do. The $\KL$ divergence between the updated proposal $q_\ell$ and the target distribution is also maximally decreased in a functional space, which inherits from the theory of SVGD~\citep{liu2017stein}. We apply our proposed algorithm to evaluate the normalization constant of various probability models including restricted Boltzmann machine and deep generative model to demonstrate the effectiveness of our proposed algorithm, where the original SVGD cannot be applied in such tasks. We propose a novel sampling algorithm for continuous-valued target distribution when the gradient information of the target distribution is unavailable. 
iteratively updated by 
$\vx_i \gets \vx_i  + \frac{\epsilon}{n} \Delta \vx_i$, where 
\begin{align}
\label{intro:gf:upd}
 \Delta \vx_i \propto
\sum_{j=1}^n 
\! w(\vx_j) \big[\nabla \log \rho(\vx_j) k( \vx_j, \vx_i) + \nabla_{\vx_j} k(\vx_j, \vx_i) \big], 
\end{align}
which replaces the true gradient $\nabla \log p(
\vx)$ with a surrogate gradient $\nabla\log \rho(\vx)$ of an arbitrary auxiliary distribution $\rho(\vx)$, and then uses an importance weight $w(\vx_j):=\rho(\vx_j)/p(\vx_j)$ to correct the bias introduced by the surrogate $\rho(\vx)$.
Perhaps surprisingly, 
we show that the new update can be derived as a standard SVGD update by using an importance weighted kernel $w(\vx)k(\vx,\vx')w(\vx')$, and hence immediately inherits the theoretical proprieties of SVGD; for example, particles updated by \eqref{intro:gf:upd} can be viewed as a gradient flow of KL divergence similar to the original SVGD \citep{liu2017stein}. Empirical experiments demonstrate that our proposed gradient-free SVGD significantly outperforms gradient-free Markov Chain Monte Carlo sampling baselines on various probability models with intractable normalization constant and unavailable gradient information of the target distribution. 

\paragraph{Approximate Inference on Discrte-valued Distribution}
In the second part of the thesis, we propose two approximate inference algorithms on discrete-valued distributions. We propose a new algorithm to sample from the discrete-valued distributions. Our proposed algorithm is based on the fact that the discrete-valued distributions can be bijectively mapped to the piecewise continuous-valued distributions. Since the piecewise continuous-valued distributions are non-differentiable, gradient-based sampling algorithms cannot be applied in this setting. Our proposed sample-efficient GF-SVGD is a natural choice. To construct effective surrogate distributions $\rho(\vx)$ in GF-SVGD, we propose a simple transformation, the inverse of dimension-wise Gaussian c.d.f., to transform the piecewise continuous-valued distributions to a simple form of continuous distributions. With such a straightforward transform, the effective surrogate distribution $\rho(\vx)$ in GF-SVGD is natural to construct. Empirical experiments on large-scale discrete graphical models demonstrate the effectiveness of our proposed algorithm. 

\paragraph{Principled Ensemble Method to Learn Binarized Neural Networks} As a direct application from sampling on discrete distributions, we propose a principled ensemble method to train the binarized neural networks~(BNN). We train an ensemble of $n$ neural networks (NN) with the same architecture ($n\ge 2$). Let $\vw_i^b$ be the binary weight of model $i$, for $i=1,\cdots, n$, and $p_*(\vw_i^b;D)$ be the target probability model with softmax layer as last layer given the data $D$. Learning the target probability model is framed as drawing $n$ samples $\{\vw_i^b\}_{i=1}^n$ to approximate the posterior distribution $p_*(\vw^b; D)$. Let $p_0(w)$ be the base function, which is the product of the p.d.f. of the standard Gaussian distribution over the dimension $d.$ The distribution of $\vw$ has the form $p_c(\vw;D)\propto p_*(\sign(\vw);D)p_0(\vw)$ with weight $\vw$ and the $\sign$ function is applied to each dimension of $\vw$. The surrogate probability model $\rho(\vw;D)$ is easily constructed by $\wt{p}(\sigma(\vw);D)p_0(\vw)$. Here $\wt{p}(\sigma(\vw);D)$ is a differentiable approximation of $p_*(\sign(\vw);D).$  Then an ensemble of models $\{\vw_i\}_{i=1}^n$ is updated by GF-SVGD as follows, 
$\vw_i \leftarrow \vw_i+\frac{\epsilon_{i}}{\Omega}\Delta \vw_i$, $\forall i=1,\cdots, n,$
\begin{equation}\label{bnn:update}
 \Delta \vw_i \!\! \leftarrow \!\! \! \sum_{j=1}^n \! \gamma_j [\nabla_{\vw}\log \rho(\vw_j;\!D_i)k(\vw_j\!,\!\vw_i)
             +\!\nabla_{\vw_j} k(\vw_j\!,\!\vw_i)]   
\end{equation}
where $D_i$ is batch data $i$ and $\mu_j =\rho(\vw_j; D_i)/p_c(\vw_j;D_i)$, $H(t) \overset{\mathrm{def}}{=}\sum_{j=1}^n \mathbb{I}(\mu_j\ge t)/n$, $\gamma_j= (H(\vw_j))^{-1}$ and $\Omega=\sum_{j=1}^n \gamma_j$.

We compare our ensemble algorithm with typical ensemble method using bagging and AdaBoost (BENN, \citet{zhu2018binary}), BNN \citep{hubara2016binarized} and BNN+\citep{darabi2018bnn+}. Both BNN and BNN+ are trained on a single model with same network structure. Our proposed ensemble algorithm achieves the highest accuracy among all three ensemble by using AlexNet \citep{krizhevsky2012imagenet} on CIFAR-10 dataset. This is because our ensemble model is sufficiently interactive at each iteration during training and our models $\{\vw_i\}$ in principle are approximating the posterior distribution $p(\vw;D).$ Our ensemble algorithm is welled justified from the Bayesian perspective. Our ensemble method of learning BNN provides a new way to train BNN. Future research includes applying our ensemble method to training BNN with larger networks such as VGG net and  larger dataset such as ImageNet dataset.
\paragraph{Goodness-of-fit Test on Discrete-valued Distributions} We propose a new goodness-of-fit testing method on discrete distributions, which evaluates whether a set of data $\{\vz_i\}_{i=1}^n$ match the proposed distribution $p_*(\vz)$. Our algorithm is motivated from the goodness-of-fit test method for continuous-valued distributions~\citep{liu2016kernelized}.
To leverage the gradient-free KSD to perform the goodness-of-fit test, we first transform the data $\{\vz_i\}_{i=1}^n$ and the candidate distribution $p_*(\vz)$ to the corresponding continuous-valued data and distributions using the transformation constructed in discrete distributional sampling aforementioned. Our method performs better and more robust than maximum mean discrepancy and discrete KSD methods under different setting on various discrete models. 

Besides leveraging the computational progress of Stein's method to propose a new framework to perform approximate inference, it is also possible to incorporate some nice properties in Stein's method into traditional inference algorithms to improve their performance. In the following, we will discuss how to leverage Stein's identity to black-box variational inference and adaptive importance sampling to improve the performance of the algorithms in these domains.

\paragraph{Improve Black-Box Variational Inference by Stein Control Variates}
We have introduced variational inference (VI) algorithms in Section~\ref{cha:def:vi}. When the common divergence $\KL$ is chosen as the metric, the goal of the VI algorithms is to minimize 
\begin{equation}
\label{chap:sumry:viobj}
\mathcal{L}(\phi)=\E_{q_{\phi}(\vx)}[\log q_{\phi}(\vx) - \log p(\vx)],
\end{equation}
where $p(\vx)$ is the target distribution and $q_{\phi}(\vx)$ is the approximate distribution parameterized by $\phi.$ In order to maximize \eqref{chap:sumry:viobj}, samples $\{\vx_i\}$ from $q_{\phi}(\vx)$ have to be drawn to estimate \eqref{chap:sumry:viobj}. However, in many applications, the Monte Carlo estimation of \eqref{chap:sumry:viobj} tends to have large variance. In order to have a good estimation of $\mathcal{L}(\phi),$ a large number of samples $\{\vx_i\}$ have to draw, which is impractical when the evaluation of $p(\vx)$ is expensive. Therefore, reducing the variance of the Monte Carlo estimation is critical. In the following, we will discuss how to adopt Stein control variates~\citep{liu2017action} to reduce the variance. For any function $f(\vx),$ let $J(\phi) = E_{q_{\phi}(\vx)}[f(\vx)].$

\begin{equation}
\nabla_{\phi} J(\phi) = E_{q_{\phi}(\vx)}[\nabla_{\phi} \log q_{\phi}(\vx) f(\vx)]
\end{equation}
By the reparameterization trick, $\vx =h(\epsilon, \phi),$ we also have
\begin{equation}
\nabla_{\phi} J(\phi) = E_{q(\epsilon)} [\nabla_{\vx} f(\vx)\nabla_{\phi} h(\epsilon, \phi)],
\end{equation}
Therefore, we have the following identity, motivated from~\citep{liu2017action},
\begin{equation}
E_{q_{\phi}(\vx)}[\nabla_{\phi} \log q_{\phi}(\vx) f(\vx)] - E_{q(\epsilon)} [\nabla_{\vx} f(\vx)\nabla_{\phi} h(\epsilon, \phi)] =0.
\end{equation}
We can propose the following variance reduced form,
 \begin{equation}
 \begin{aligned}
\nabla_{\phi} \mathcal{L}(\phi)=& \E_{q_{\phi}(\vx)}[ \nabla_{\phi}\log q_{\phi}(\vx) (\log q_{\phi}(\vx))-\log p(\vx))] \\
&  + \lambda\big[\E_{q_{\phi}(\vx)}[\nabla_{\phi} \log q_{\phi}(\vx) f(\vx)] -  E_{q(\epsilon)} [\nabla_{\vx} f(\vx)\nabla_{\phi} h(\epsilon, \phi)]\big],
\end{aligned}
\end{equation}
where $\lambda$ can be chosen similarly as the score function method~\eqref{intro:control}. $f(\vx)$ can be chosen arbitrarily. In practice, it is possible to use neural network to parametrize $f_{\gamma}(\vx)$ and optimize $\gamma$ to fit the current samples $\{\vx_i\}$ for minimal variance.

\paragraph{Reduce the variance of the objective in adaptive importance sampling by Stein Control Variates}
In adaptive importance sampling, one popular way is to find the importance proposal $q_{\phi}$ such that the variance of the interested estimation is minimized~\citep{cappe2008adaptive, ryu2014adaptive,  cotter2015parallel},
\begin{equation}
\label{app:imp:opt}
\min_{\phi} \mathcal{L}(\phi)=\min_{\phi} \mathrm{Var}(\frac{p(\vx)}{q_{\phi}(\vx)}f(\vx)) = \int_{\vx} \frac{p^2(\vx)}{q_{\phi}(\vx)}f^2(\vx)d\vx-\mathrm{Constant}.     
\end{equation} 
In order to minimize the objective~\eqref{app:imp:opt}, we need to draw samples $\{\vx_i\}$ from $q_{\phi}(\vx),$ and use Monte Carlo estimation of ~\eqref{app:imp:opt}, which induces large variance when $q_{\phi}$ is different from $p(\vx),$
\begin{equation}
\label{imp:opt:est}
\nabla_{\phi} \mathcal{L}(\phi) = - \E_{q_{\phi}}[\frac{p^2(\vx)}{q_{\phi}^2(\vx)}f^2(\vx) \nabla_{\phi} \log q_{\phi}(\vx)].       
\end{equation}
In order to reduce the variance from the Monte Carlo estimation of \eqref{imp:opt:est}, we have discussed one simple way from the score function method, $\E_{\vx \sim q_{\phi}(\vx)}[\nabla_{\phi} \log q_{\phi}(\vx)]=0,$

\begin{equation}
\label{imp:opt:cont}
\nabla_{\phi} \mathcal{L}(\phi) = - \E_{q_{\phi}}[\frac{p^2(\vx)}{q_{\phi}^2(\vx)}f^2(\vx)  \nabla_{\phi} \log q_{\phi}(\vx)]+\lambda\E_{q_{\phi}}[\nabla_{\phi} \log q_{\phi}(\vx)],   \end{equation}
where the optimal $\lambda$ has closed form,
\begin{equation}
\label{app:imp::opt:coeff}
\lambda = \mathrm{Var}(\nabla_{\phi}\log q_{\phi}(\vx))^{-1}\mathrm{Cov}[\frac{p^2(\vx)}{q_{\phi}^2(\vx)}f^2(\vx)\nabla_{\phi} \log q_{\phi}(\vx), \nabla_{\phi} \log q_{\phi}(\vx)], 
\end{equation}
and can be empirically estimated by samples $\{\vx_i\}_{i=1}^n$ from $q_{\phi}(\vx).$

Based on previous introduction of Stein control variates, we can use the following more efficient variance reduction trick,
 \begin{equation}
 \label{imp:opt:stein}
 \begin{aligned}
\nabla_{\phi} \mathcal{L}(\phi)=& \E_{q_{\phi}(\vx)}[ \frac{p^2(\vx)}{q_{\phi}^2(\vx)}f^2(\vx)  \nabla_{\phi} \log q_{\phi}(\vx)] \\
&  + \lambda\big[\E_{q_{\phi}(\vx)}[\nabla_{\phi} \log q_{\phi}(\vx) f(\vx)] -  E_{q(\epsilon)} [\nabla_{\vx} f(\vx)\nabla_{\phi} h(\epsilon, \phi)]\big],
\end{aligned}
\end{equation}
where $\lambda$ can be chosen similarly as the score function method~\eqref{app:imp::opt:coeff} and $f(\vx)$ can be chosen arbitrarily, for example, parameterized by the neural network  $f_{\gamma}(\vx)$, and optimizing $\gamma$ to fit the current samples $\{\vx_i\}$ from the importance proposal $q_{\phi}(\vx)$ for minimal variance. When the evaluation of the target $p(\vx)$ is expensive, the low variance after introducing Stein control variates enables us to have less samples to estimate \eqref{imp:opt:stein}.

\backmatter
\bibliographystyle{note}
\bibliography{note}
\chapter{Appendices}
\section{Proofs of Theorems in Chapter~\ref{chap:is}}
\label{append:is}

In this section, we provide some theoretical investigation of our proposed algorithm. The analysis of our Stein adaptive importance sampling is based on the theoretical results of SVGD~\citep{liu2017stein}. Firstly, we provide some analysis on the convergence rate of our importance proposal $q_K(\vx)$ to the target distribution $p(\vx)$. Secondly, we establish the convergence property of our algorithm w.r.t. the number of particles.

In the following, we provide some analysis on the convergence property of our importance proposal $q_K(\vx)$ to the target distribution $p(\vx).$ If we take $\epsilon$ in the transformation to be infinitesimal, 
\begin{align}
\vx_i^{\ell+1}  \gets \vx^{\ell}+\epsilon \ff_{\ell+1}(\vx^\ell),
\end{align}
where $\ff_{\ell+1}(\vx)$ is defined in \eqref{is:transf},  the evolution equation of the random variable $\bd{x}^t$ reduces to a partial differential equation(PDE),
\begin{equation}
\label{part}
\frac{d\bd{x}^t}{dt}=\mathbb{E}_{\bd{x}\sim{q_t}(\vx)}[\nabla_{\bd{x}} \log p(\bd{x})k(\bd{x},\vx^t)+\nabla_{\bd{x}} k(\bd{x},\vx^t)].
\end{equation}
Here we use $\bd{x}^t$ to denote the evolved particle at current time $t$ with density function $q_t.$  PDE\eqref{part} captures one type of Vlasov process for interacting particles system~\citep{braun1977vlasov}. Based on the continuous-time Vlasov process, the convergence rate of $q_K$ to $p$ can be more conveniently illustrated.

\begin{thm}
\label{app:is:thm2}
Suppose random variable $\bd{x}^t$ is the solution of PDE \eqref{part},
then the probability density function of $\bd{x}^t$, denoted by $q_t$, satisfies the following PDE,
\begin{equation}
\label{diffode}
\frac{\partial q_t}{\partial t}=-\mathrm{div}(q_t\mathbb{E}_{\bd{x}\sim{q}}[\nabla_{\bd{x}} \log p(\bd{x})k(\bd{x},\bd{z}^t)+\nabla_{\bd{x}} k(\bd{x},\bd{z}^t)]).
\end{equation}
where $\mathrm{div}$ denotes the divergence of a vector.
\end{thm}

The proof of proposition~\ref{pro2} is similar to the proofs of proposition 1.1 in~\citet{jourdain1998propagation} and lemma 1 on the appendix of \citet{dai2019opaque}. 
Proposition~\ref{pro2} characterizes the evolution of the density function $q_t(\bd{x}^t)$ when the random variable $\bd{x}^t$ is evolved by ~\eqref{part}. The continuous system captured by~\eqref{part} and ~\eqref{diffode} is a type of Vlasov process which has wide applications in physics, biology and many other areas~\citep[e.g.,][]{braun1977vlasov}.

{\bf Proof:} Denote $A(X^t,t)=\int_\Omega q(\vx,t)[\nabla_\vx \log p(x)K(\vx,X^t)+\nabla_x K(\vx,X^t)]d
\vx,$ to prove equation (\ref{diffode}), we just need to show for any test function $\psi(\vx,t)\in C^{2,1}_0$ ($C^{2,1}_0$ means the set of functions which are second-order differential in $\vx$ and first-order differential in $t$ and take zeros when $\vx\in \partial\Omega$), we have
\begin{equation}
\int (\frac{\partial \rho(\vx,t)}{\partial t}+\mathrm{div}(\rho(\vx,t)A(\vx,t)))\psi(\vx,t)d\vx=0.
\end{equation}
Let $\mathfrak{F}_t=\sigma(X^s: s\le t),$ and define un-normalized conditional probability $p_{(t)}(\psi_t)=\mathrm{E}[\psi(X^t,t)\mid \mathfrak{F}_t].$
By Ito's formula,
\begin{equation}
\label{ito}
d\psi(X^t,t)=\nabla_\vx\psi\cdot dX^t+\frac{\partial \psi}{\partial t}dt=[-\nabla_\vx\psi\cdot A(X^t,t)+\frac{\partial \psi}{\partial t}]dt.
\end{equation}
As $\rho(\vx,t)$ is the probability density function of $X^t$, by the definition of conditional probability, it satisfies
 \begin{equation}
 \label{formu}
p_{(t)}(\psi_t)=\int \rho(\vx,t)\psi(\vx,t)dx.
 \end{equation}

According to the formula (\ref{ito}), we have the following identity,
\begin{equation}
\label{inte}
\psi(X^t,t)=\psi(X^0,0)+\int_0^t[-\nabla_\vx\psi\cdot A(X^s,s)+\frac{\partial \psi}{\partial t}]ds.
\end{equation}
Based on (\ref{inte}), we have
\begin{equation}
\label{equa}
\mathrm{E}[\psi(X^t,t)\mid \mathfrak{F}_t]=\mathrm{E}[\psi(X^0,0)\mid \mathfrak{F}_t]+\mathrm{E}[\int_0^t(-\nabla_x\psi\cdot A(X^s,s)+\frac{\partial \psi}{\partial t})ds\mid \mathfrak{F}_t].
\end{equation}
By the definition of condition probability and Fubini's theorem, and based on the equality (\ref{formu}), (\ref{equa}), we have
\begin{equation}
\label{fina}
\int \rho(\vx,t)\psi(\vx,t)dx=\int \rho(\vx,0)\psi(\vx,0)dx+\int \int_0^t\rho(\vx,s)[-\nabla_x\psi\cdot A(x,s)+\frac{\partial \psi}{\partial t}]dsd\vx.
\end{equation}
We observe the following formula, $$\nabla_\vx\psi\cdot (\rho(\vx,s)A(\vx,s))=\nabla_\vx\cdot(\psi A(\vx,s)\rho(x,s))-\psi\nabla_
\vx\cdot (\rho(\vx,s)A(\vx,s)).$$ Since $\psi(\vx,s)\in C^{2,1}_0$, then we have $\int_\Omega \nabla_{\vx}\cdot(\psi\rho(\vx,s) A(X^s,s))d\vx=0$ for any $s$. It is easy to verify that
\begin{equation*}
\begin{split}
\int_\Omega\int_0^t \frac{\partial(\rho\psi)}{\partial t}dsd\vx &=\int_\Omega\int_0^t (\psi\frac{\partial\rho}{\partial t}+\rho\frac{\partial\psi}{\partial t})dsdx \\
&=\int_\Omega \rho(\vx,t)\psi(\vx,t)d\vx-\int_\Omega \rho(\vx,0)\psi(\vx,0)d\vx.
\end{split}
\end{equation*}
Equation (\ref{fina}) can be rewritten in the following,
\begin{equation}
\int_0^t\int_\Omega [\frac{\partial \rho }{\partial t}+\nabla\cdot(\rho(\vx,s)A(\vx,s))]\psi(x,s)d\vx ds=0.
\end{equation}
Take derivative w.r.t. $t$, we have
\begin{equation}
\label{final}
\int_\Omega [\frac{\partial \rho }{\partial t}+\nabla\cdot(\rho(\vx,t)A(x,t))]\psi(\vx,s) d\vx=0.
\end{equation}
Since equation (\ref{final}) holds for any test function $\psi(\vx,t)\in C^{2,1}_0$, then we can get $\frac{\partial \rho }{\partial t}=-\nabla\cdot(\rho(\vx,t)A(\vx,t)).$ The proof is complete. $\square$ 

Theorem~\ref{is:thm2} builds a general connection between the evolution of random variable and the evolution of its density function. Theorem~\ref{thm2} helps us establish one importance property of $q_t$ in our algorithm, provided in the following algorithm.

One nice property of algorithm~\ref{VarIS:algo} is that the KL divergence between the iterative distribution $q_\ell$ and $p$ is monotonically decreasing. This property can be more easily understood by considering our iterative system in continuous evolution time as shown in \citet{liu2017stein}. 

\begin{lem}
\label{lemma1}
The evolution equation of the density function $q_t$ satisfies the following PDE,
\begin{equation}
\label{klksd}
\frac{d\mathrm{KL}(q_t\mid\mid p)}{dt}=-\mathbb{D}(q_t,p)^2,
\end{equation}
where $\mathbb{D}(q_t,p)$ is the square of the KSD between density functions $q_t$ and the target density $p$.
\end{lem}
Lemma~\ref{klksd} indicates that the KL divergence between the iteratively transformed distribution and the target distribution $p$ is monotone decreasing w.r.t. time. Equation ~\eqref{klksd} indicates that the KL divergence between the iterative distribution $q_t$ and $p$ is monotonically decreasing with a rate of $\mathbb{D}(q_t~||~ p)^2$. 
If the relationship $ \mathrm{KL}(q_t\mid\mid p) \le \frac{1}{\gamma} \mathbb{D}(q_t,p)$ can be established, $0<\gamma<\infty$, we have the following convergence of $q_t,$
\begin{equation}
\label{converg}
\KL(q_t, p)\le C \exp(-\gamma t).
\end{equation}
Equation (\ref{converg}) indicates that the KL divergence between the evolved density $q_t$ and the target density $p$ has exponential decay. Although it is unclear whether $\gamma$ satisfies $0<\gamma<\infty$ in general cases, numerical experiments on Gaussian mixture models(GMM) have demonstrated that we have $\KL(q_t, p)\le C \exp(-\gamma t)$ in this case when the initial distribution $q_0$ is Gaussian distribution.

The convergence rate w.r.t. the particles size can be more easily illustrated in terms of the empirical measures induced by the evolved particles $\{\bd{z}^j_i\}_{i=1}^N$ and $\{\bd{y}^j_i\}_{i=1}^M$. The empirical measures of these two sets of particles are defined as
\begin{equation*}
\begin{split}
&\widetilde{\mu}^j_N(d\bd{x})=\frac1N\sum_{i=1}^N\delta(\bd{x}-\bd{z}^j_i)d\bd{x},\\
& \hat{\mu}^j_M(d\bd{x})=\frac{1}{M}\sum_{i=1}^M\delta(\bd{x}-\bd{y}^j_i)d\bd{x},
\end{split}
\end{equation*}
where $\delta$ is the dirac function. Denote $\mu_{\infty}^j$ as the exact probability measure with density function $q_j$, which is the density of $\bd{z}^j$ defined by equation \eqref{transform} and \eqref{stein}. We define the bounded Lipschitz of function $f$ as
\begin{equation*}
\|f\|_{\mathrm{BL}}= \max\{\sup_{\bd{x}}|f(\bd{x})|, \sup_{\bd{x}\neq \bd{y}} \frac{|f(\bd{x})-f(\bd{y})|}{\|\bd{x}-\bd{y}\|_2}\}.
\end{equation*}
For vector-valued $\bd{f}=[f_1,\cdots, f_d],$ $\|\bd{f}\|_{\mathrm{BL}}^2=\sum_i|f_i\|_{\mathrm{BL}}^2.$ Denote $$\bd{g}(\bd{x}, \bd{y})=\nabla_{\bd{x}} \log p(\bd{x})k(\bd{x},\bd{y})+\nabla_{\bd{x}} k(\bd{x},\bd{y}).$$ We assume $\|\bd{g}(\bd{x}, \bd{y})\|_{\mathrm{BL}}<\infty.$ We define the Lipschitz metric between measures $\mu$ and $\nu$ as follows,
$$\mathrm{BL}(\mu, \nu)=\sup_f{\mathbb{E}_\mu f-\mathbb{E}_\nu f, \quad \textit{s.t.}\quad \|f\|_{\mathrm{BL}}\le 1}.$$
 Similar to the theoretical result of $\widetilde{\mu}^j_N$, we have the following result for $\hat{\mu}^j_M.$ With mild conditions, for all bounded $h$, the theoretical result of SVGD indicates that $\mathbb{E}_{\widetilde{\mu}^j_N}[h]\rightarrow \mathbb{E}_{\mu^j_{\infty}}[h],$ which means that empirical measure $\widetilde{\mu}^j_N$ weakly converges to $\mu^j_{\infty}.$ 

\begin{thm}
\label{concen}
Suppose $\{\bd{z}^0_i\}_{i=1}^N$ and $\{\bd{y}^0_i\}_{i=1}^M$ are drawn from the distribution with probability measure $\mu_{\infty}^0$ and density function $q_0$, where $q_0$ is log-concave, such as Gaussian distribution. Assume
\begin{equation*}
\lim_{N\rightarrow\infty}\mathrm{BL}(\widetilde{\mu}^0_N, \mu_{\infty}^0)=0, \quad \lim_{M\rightarrow\infty}\mathrm{BL}(\hat{\mu}^0_M, \mu_{\infty}^0)=0,
\end{equation*}
then for $j=1,2,\cdots, K$ and bounded continuous function $h$, we have
$$\sqrt{M}(\mathbb{E}_{\hat{\mu}^j_M}[h]-\mathbb{E}_{\mu^j_{\infty}}[h])\rightarrow \mathcal{N}(0, \sigma^2),$$
where $\mathcal{N}(0, \sigma^2)$ is normal distribution with variance $\sigma^2$.
\end{thm}

{\bf Proof}: Let $\widetilde{\mu}^j_N(d\bd{x})$ be the empirical measure of $\{\bd{x}^j_i\}_{i=1}^N$ and $\hat{\mu}^j_M(d\bd{x})$ be the empirical measure of $\{\bd{y}^j_i\}_{i=1}^M.$ We use $\mu_{\infty}^j$ the exact probability measure of $\bd{x}^j$ defined in equation \eqref{transform} and \eqref{stein}. We define the Lipschitz metric between two probability measures  as
\begin{equation*}
\mathrm{BL}(\mu, \nu)=\sup_f{\mathbb{E}_\mu f-\mathbb{E}_\nu f, \quad \textit{s.t.}\quad \|f\|_{\mathrm{BL}}\le 1},
\end{equation*}
\begin{equation*}
\textit{where } \|f\|_{\mathrm{BL}}= \max\{\sup_{\bd{x}}f(\bd{x}), \sup_{\bd{x}\neq \bd{y}} \frac{|f(\bd{x})-f(\bd{y})|}{\|\bd{x}-\bd{y}\|_2}\}.
\end{equation*}
 
Based on the following triangle inequality
\begin{equation*}
\|\bd{T}_{\mu, p}\mu - \bd{T}_{\widetilde{\mu}, p}\hat{\mu}\|_{\mathrm{BL}}\le \|\bd{T}_{\mu, p}\mu - \bd{T}_{\widetilde{\mu}, p}\widetilde{\mu}\|_{\mathrm{BL}}+\|\bd{T}_{\widetilde{\mu}, p}\widetilde{\mu} - \bd{T}_{\widetilde{\mu}, p}\hat{\mu}\|_{\mathrm{BL}},
\end{equation*}
Since we know $\mathrm{BL}(\widetilde{\mu}^0_N, \hat{\mu}^0_M)\rightarrow 0,$ then it is easy to derive $\mathrm{BL}(\hat{\mu}^1_M, \mu_{\infty}^1)\rightarrow 0.$
Similarly, $\mathrm{BL}(\hat{\mu}^j_N, \mu_{\infty}^j)\rightarrow 0$ can be proved inductively.

Theorem~\ref{concen} indicates particles $\{\bd{y}^j_i\}_{i=1}^M$ with empirical measure $\hat{\mu}^j_M$ and empirical density $\{\hat{q}_j(\bd{y}^j_i)\}_{i=1}^M$ satisfy the concentration property~\citep{spohn2012large}. The convergence rate of $\{\bd{y}^j_i\}_{i=1}^M$ is $O(1/\sqrt{M}).$

\section{Proofs of Theorems in Chapter~\ref{chap:gf}}
\label{append:gf}
Before proving our main theorem, we define some preliminary notations. We always assume $\vx=[x_1,\cdots, x_d]^\top \in \R^d$.
Given a positive definite kernel $k(\vx,\vx')$, there exists a unique reproducing kernel Hilbert space (RKHS) $\H$,
formed by the closure of functions of form $f(\vx) = \sum_{i} a_i k(\vx,\vx_i)$ where $a_i \in \RR$, equipped with inner product
$\la f, ~ g\ra_{\H_0} = \sum_{ij}a_i k(\vx_i, \vx_j) b_j$ for $g(\vx) = \sum_j b_j k(\vx, \vx_j)$. 
Denote by $\H^d = \H \times \cdots \times \H$ the vector-valued function space formed by $\vv f = [f_1, \ldots, f_d]^\top$, where $f_i \in \H$, $i=1,\ldots, d$, equipped with inner product $\la \vv f, ~ \vv g\ra_{\H^d}=\sum_{l=1}^d \la f_l, ~ g_l\ra_{\H}, $ for $\vv g =[g_1, \ldots, g_d]^\top.$ 
Equivalently, $\H$ is the closure of functions of form  $\vv f(\vx) = \sum_{i} \vv a_i k(\vx,\vx_i)$ where $\vv a_i \in \RR^d $ 
with inner product $\la \vv f, ~ \vv g\ra_{\H^d} = \sum_{ij}\vv a_i^\top \vv b_j k(\vx_i, \vx_j)$ for $\vv g(\vx) = \sum_{i} \vv b_i k(\vx,\vx_i)$. 
See e.g.,~\citet{berlinet2011reproducing} for more background on RKHS.

In the following section, we derive a key observation from the importance-weighted Stein's identity and KSD, which can be used to develop our gradient-free SVGD. We also provide one theorem to develop the gradient-free form of the gradient-free KSD which can be used to propose gradient-free black-box importance sampling and the goodness-of-fit on discrete distributions.

\begin{thm}\label{app:pro:wphi}
Let $p(\vx)$, $\prop(\vx)$ be  positive differentiable densities and $w(\vx) = {\prop(\vx)}/{p(\vx)}$. We have 
\begin{align}\label{equ:ws}
 w(\vx)\steinbxtransp\ff(\vx)  = \steinpxtransp \big(w(\vx)\ff (\vx) \big).
\end{align}
Therefore,
$\S_{\F, \prop}(q~||~p)$ in \eqref{equ:grds}
is equivalent to 
\begin{align}
 \S_{\F, \prop}(q~||~p)  
 &  = \max_{\ff \in \F}\big\{ \E_{\vx\sim q} [\steinpxtransp \big( w(\vx)\ff(\vx)\big)]\big\} \label{sbf} \\
 & = \max_{\ff \in w\F} \big \{\E_{\vx\sim q}[\steinpxtransp \ff(\vx)]     \big \}  \label{sbf2}\\
 & = \S_{w\F}(q~||~p). \notag
\end{align}
\end{thm}
{\bf Proof:} By definition, $w(\vx)=\rho(\vx)/p(\vx)$, $\nabla_{\vx}w(\vx)=w(\vx)\bd{s}_\rho(\vx)-w(\vx)\bd{s}_p(\vx)$,
\begin{align}
\steinpxtransp(w(\vx)\ff(\vx)) & = w(\vx) \bd{s}_p(\vx)^\top \ff(\vx) + \nabla_{\vx}^\top (w(\vx)\ff(\vx)) \notag\\
& = w(\vx) \bd{s}_p(\vx)^\top \ff(\vx) + \nabla_{\vx}w(\vx)^\top \ff(\vx) + w(\vx)\nabla_{\vx}^\top \ff(\vx) \notag \\
& = w(\vx) \bd{s}_\rho(\vx)^\top \ff(\vx) + w(\vx)\nabla_{\vx}^\top \ff(\vx) =w(\vx)\steinbxtransp \ff(\vx). \notag    
\end{align}
Therefore, we have 
\begin{align}
 \S_{\F, \prop}(q~||~p)  
 &  = \max_{\ff \in \F}\big\{ \E_{\vx\sim q} [\steinpxtransp \big( w(\vx)\ff(\vx)\big)]\big\} \label{app:sbf1} \\
 & = \max_{\ff \in w\F} \big \{\E_{\vx\sim q}[\steinpxtransp \ff(\vx)]]     \big \}  \label{app:sbf2}\\
 & = \S_{w\F}(q~||~p). \notag
\end{align}

\begin{thm}
When $\Hd$ is an RKHS with kernel $k(\vx,\vx')$, the optimal solution of \eqref{sbf} is ${\ff}^*/||{\ff}^*||_\Hd,$ where 
\begin{align}
{\ff}^*(\cdot) 
& =\E_{\vx\sim q}[\steinpx(w(\vx)k(\vx, \cdot))] \label{newvel}
\\
& =\E_{\vx\sim q}[w(\vx) \steinbx k(\vx, \cdot)], \label{newvel2}
\end{align}
where the Stein operator $\steinbx$ is applied to variable $\vx$, $\steinbx k(\vx, \cdot)=\nabla_\vx \log \rho(\vx)k(\vx, \cdot)+\nabla_{\vx}k(\vx, \cdot).$
 Correspondingly, the optimal decrease rate of KL divergence in \eqref{gradfreeKLmin} equals the square of $\S_{\F, \rho}(q~||~p)$, which equals 
\begin{equation}
\label{newksd}
 \S_{\F, \prop}(q ~||~ p) = (\E_{\vx, \vx'\sim q}[w(\vx)w(\vx')\kappa_\prop(\vx,\vx')])^{\frac12},
\end{equation}
where $\kappa_\prop(\vx,\vx') = (\newsteinbx)^\top (\steinbx k(\vx,\vx'))$ and $\newsteinbx$ is the Stein operator applied on variable $\vx'$.
\label{theom}
\end{thm}
{\bf Proof:} When $\H$ is an RKHS with kernel $k(\vx,\vx')$, then $w\H$ is also an 
RKHS, with an ``importance weighted kernel'' 
\begin{align}\label{app:newkernel}
\tilde k(\vx,\vx') = w(\vx)w(\vx')k(\vx,\vx').
\end{align}
Following Lemma 3.2 in \citet{liu2016stein}, the optimal solution of the optimization problem  \eqref{app:sbf2} is,
\begin{align*}
w(\cdot){\ff}^*(\cdot) 
& = \E_{\vx\sim q}[\bd{s}_p(\vx) w(\vx)k(\vx, \cdot) w(\cdot) +\nabla_\vx (w(\vx)k(\vx, \cdot)w(\cdot))] \label{newvel}
\\
& = w(\cdot) \E_{\vx\sim q}[w(\vx) \steinbx k(\vx, \cdot)]. 
\end{align*}
This gives 
$$\ff^*(\cdot)= \E_{\vx\sim q}[w(\vx) \steinbx k(\vx, \cdot)].$$ 
Following 
Theorem 3.6 \citep{liu2016kernelized}, we can show that 
\begin{equation}
\label{app:newksd}
 \S_{\F, \prop}(q ~||~ p) = (\E_{\vx, \vx'\sim q}[\tilde{\kappa}_p (\vx,\vx')])^{\frac12},
\end{equation}
where 
$$
\tilde{\kappa}_p (\vx,\vx')
=(\stein_p')^\top(\stein_p\tilde k(\vx,\vx')).
$$
and $\stein_p$ and $\stein_p'$ denote the Stein operator applied on variable $\vx$ and $\vx'$, respectively. 
Applying Theorem~\ref{pro:wphi}, we have 
\begin{align*}
    \tilde{\kappa}_p (\vx,\vx')
& =(\stein_p')^\top\left (\stein_p (w(\vx)  w(\vx')k(\vx,\vx'))\right ) \\
& = (\stein_p')^\top(
 w(\vx)  \stein_\rho \left ( w(\vx')k(\vx,\vx'))\right ) \\
 & = (\stein_p')^\top(
 w(\vx') w(\vx)  \stein_\rho \left ( k(\vx,\vx'))\right ) \\
 &=  w(\vx') w(\vx)  (\stein_\rho')^\top(
 \stein_\rho \left ( k(\vx,\vx'))\right ) \\
 & =  w(\vx') w(\vx)  \kappa_\rho(\vx, \vx'),
\end{align*}
where we recall that $\kappa_\rho(\vx, \vx') = (\stein_\rho')^\top 
 \left (\stein_\rho k(\vx,\vx') \right)$. 
 
\begin{align}\label{tmp:newksd}
\tilde{\kappa}_p(\bd{x},  \bd{x}') 
& = \bd{s}_{p}(\bd{x})^\top \tilde{k}(\bd{x},\bd{x}')\bd{s}_{p}(\bd{x}')+\bd{s}_{p}(\bd{x})^\top \nabla_{\bd{x}'}\tilde{k}(\bd{x},\bd{x}') \\ \notag
& +\bd{s}_{p}(\bd{x}')^\top \nabla_{\bd{x}}\tilde{k}(\bd{x},\bd{x}')+\nabla_{\bd{x}}^\top(\nabla_{\bd{x}'}\tilde{k}(\bd{x}, \bd{x}')). 
\end{align}
Note that $\bd{s}_p(\vx)=\bd{s}_{\rho}(\vx) - \bd{s}_{w}(\vx)$ and $\bd{s}_w(\vx)=\nabla_{\vx} w(\vx)/w(\vx).$ The second term in RHS of \eqref{tmp:newksd} is  
\begin{equation}
\label{q1}
\bd{s}_p(\bd{x})^\top w(\vx) k(\vx, \vx')w(\vx') \bd{s}_p(\vx') =(\bd{s}_{b}(\vx) - \bd{s}_{w}(\vx))^\top w(\vx) k(\vx, \vx')w(\vx')  (\bd{s}_{b}(\vx) - \bd{s}_{w}(\vx)),
\end{equation}
\begin{align}
\label{q2}
\bd{s}_p(\bd{x})^\top \nabla_{\vx'}( w(\vx)k(\vx, \vx')w(\vx')) & =w(\vx)w(\vx') [(\bd{s}_{\rho}(\vx) - \bd{s}_{w}(\vx))^\top \nabla_{\vx'} k(\vx, \vx') \\ \notag
& + (\bd{s}_{\rho}(\vx) - \bd{s}_{w}(\vx))^\top k(\vx, \vx')\bd{s}_w(\vx')];
\end{align}
and the third term in RHS \eqref{tmp:newksd} can be derived similarly. The fourth term in RHS \eqref{tmp:newksd} is 
\begin{equation}
\label{q3}
\begin{aligned}
\bd{s}_p(\vx')^\top \nabla_{\bd{x}}(w(\vx)k(\vx, \vx')w(\vx'))&  = (\bd{s}_{b}(\vx') - \bd{s}_{w}(\vx'))^\top w(\vx)w(\vx')\nabla_{\vx} k(\vx, \vx')\\
&+ (\bd{s}_{b}(\vx') - \bd{s}_{w}(\vx'))^\top w(\vx')k(\vx, \vx')\nabla_{\vx} w(\vx),
\end{aligned}
\end{equation}
\begin{align*}
\nabla_{\vx'}^\top (\nabla_{\bd{x}}(w(\vx)k(\vx, \vx')w(\vx'))) & =w(\vx)w(\vx')\big[\nabla_{\vx'}^\top(\nabla_{\bd{x}}(k(\vx, \vx'))+ k(\vx,
\vx')\bd{s}_w(\vx')^\top \bd{s}_w(\vx)  \\
 &~~~~~ +  \nabla_{\vx}  k(\vx, \vx')^\top \bd{s}_w(\vx')  + \nabla_{\vx'}  k(\vx, \vx')^\top\bd{s}_w(\vx)\big]
\end{align*}

Therefore, $\D_{\F, \prop}(q, p)$ in \eqref{app:newksd} equals 
\begin{equation*}
\S_{\F, \prop}(q, p) =(\E_{\vx,\vx'\sim q}[w(\vx)\kappa_{\rho}(\vx, \vx')w(\vx')])^\frac12. 
\end{equation*}
where $\kappa$ is defined in \eqref{tmp:newksd} but with the distribution $\rho(\vx)$ and the kernel $k(\vx, \vx')$. 
\begin{align*}
\label{gfkernel}
\kappa_{\rho} (\bd{x},  \bd{y}) 
= & \bd{s}_{\rho}(\bd{x})^\top k(\bd{x},\bd{y})\bd{s}_{\rho}(\bd{y})+\bd{s}_{\rho}(\bd{x})^\top \nabla_{\bd{y}}k(\bd{x},\bd{y}) \\
& +\bd{s}_{\rho}(\bd{y})^\top \nabla_{\bd{x}} k(\bd{x},\bd{y})+\nabla_{\bd{x}}\cdot(\nabla_{\bd{y}}k(\bd{x}, \bd{y})). \\ 
\end{align*}
This completes the proof. \hfill $\square$

\paragraph{Monotone Decreasing of KL divergence} 
One nice property of the gradient-free SVGD is that the KL divergence between the updated distribution $q_\ell(\vx)$ and the target distribution $p(\vx)$ is monotonically decreasing. This property can be more easily understood by considering our iterative system in continuous evolution time as shown in \citet{liu2017stein}. 
Take the step size $\epsilon$ of the transformation defined in \eqref{update} to be infinitesimal, 
and define the continuous time $t = \epsilon \ell$. Then the evolution equation of random variable $\vx^t$ is governed by the following nonlinear partial differential equation~(PDE), 
\begin{equation}
\label{part}
\frac{d\bd{x}^t}{dt}=\mathbb{E}_{\bd{x}\sim{q_t}}[w(\vx)(\bd{s}_{\rho}(\bd{x})k(\bd{x},\bd{x}^t)+\nabla_{\bd{x}} k(\bd{x},\bd{x}^t))],
\end{equation}
where $t$ is the current evolution time and $q_t$ is the density function of $\vx^t.$ The current evolution time $t= \epsilon \ell$ when $\epsilon$ is small and $\ell$ is the current iteration. We have the following proposition: 
\begin{pro}
\label{gfpro3}
Suppose random variable $\bd{x}^t$ is governed by PDE \eqref{part}, then its density $q_t(\bd{x})$ is characterized by
\begin{equation}
\label{diffodegf}
\frac{\partial q_t(\vx^t)}{\partial t}=-\mathrm{div}(q_t(\vx^t)\mathbb{E}_{\bd{x}\sim{q_t}}[w(\vx)(\bd{s}_{\rho}(\bd{x}) k(\bd{x},\bd{x}^t)+\nabla_{\bd{x}} k(\bd{x},\bd{x}^t))]),
\end{equation}
where $\mathrm{div}(\bd{f})=\trace(\nabla \vv f) = \sum_{i=0}^d \partial f_i(\bd{x})/\partial x_i$, and  $\bd{f}=[f_1,\ldots, f_d]^\top.$ And the derivative of the $\KL$ divergence between the iterative distribution $q_t(\vx)$ and the target $p(\vx)$ satisfies that
\begin{equation}
\frac{d\KL(q_t, p)}{dt} = -  \E_{\vx, \vx'\sim q}[w(\vx)\kappa_\prop(\vx,\vx')w(\vx')] \le 0,
\end{equation}
where $\kappa_\prop(\vx,\vx')$ can be derived as
\begin{equation}
\label{chap:gf:ksd}
\begin{aligned}
\kappa_{\rho} (\bd{x},  \bd{y}) = & (\newsteinbx)^\top (\steinbx k(\vx,\vx'))= \bd{s}_{\rho}(\bd{x})^\top k(\bd{x},\bd{y})\bd{s}_{\rho}(\bd{y}) \\ & +\bd{s}_{\rho}(\bd{x})^\top \nabla_{\bd{y}}k(\bd{x},\bd{y}) 
+\bd{s}_{\rho}(\bd{y})^\top \nabla_{\bd{x}} k(\bd{x},\bd{y})+\nabla_{\bd{x}}\cdot(\nabla_{\bd{y}}k(\bd{x}, \bd{y})).
\end{aligned}
\end{equation}
\end{pro}
Proof: Based on the proof of Theorem~\ref{app:is:thm2}, it is similar to derive the result.

It is interesting to observe that replacing the kernel $k(\vx, \vx')$ in the original SVGD with a new kernel,
\begin{equation}
\wt{k}(\vx,\vx') = \frac{\rho(\vx)}{p(\vx)} k(\vx, \vx')\frac{\rho(\vx')}{p(\vx')},     
\end{equation}
Proposition~\ref{gfpro3} is straightforward to derive from the derviation in SVGD~\citep{liu2017stein}.

In the following, we replace the kernel $k(\bd{x},\bd{y})$ with the kernel $w(\vx)k(\bd{x}, \bd{y})w(\bd{y})$ in RKHS $\mathcal{H}_d$ in KSD~\citep{liu2016kernelized}, we can straightforwardly derive the gradient-free KSD.
\begin{thm}
\label{app:gf-ksd}
Replace the kernel $k(\bd{x},\bd{y})$ with the kernel $\widetilde{k}(\bd{x}, \bd{y})=w(\vx)k(\bd{x}, \bd{y})w(\bd{y})$ in RKHS $\mathcal{H}_d,$ the KSD can be rewritten as follows,  
\begin{equation}
\label{app:bbis:ksd}
\wt{\mathbb{S}}(q, p) =\E_{\bd{x},\bd{y}\sim q}[\wt{\kappa}_p(\bd{x}, \bd{y})]\ge 0,
\end{equation}
where $\wt{\kappa}_{p}(\bd{x}, \bd{y})$ satisfies $
\wt{\kappa}_{p}(\bd{x}, \bd{y}) =w(\bd{x})\kappa_{\rho} (\bd{x},  \bd{y})w(\bd{y}),$ \\
\begin{equation*}
\begin{aligned}
\kappa_{\rho} (\bd{x},  \bd{y}) = & \bd{s}_{\rho}(\bd{x})^\top k(\bd{x},\bd{y})\bd{s}_{\rho}(\bd{y}) +\bd{s}_{\rho}(\bd{x})^\top \nabla_{\bd{y}}k(\bd{x},\bd{y}) \\
&+\bd{s}_{\rho}(\bd{y})^\top \nabla_{\bd{x}} k(\bd{x},\bd{y})+\nabla_{\bd{x}}\cdot(\nabla_{\bd{y}}k(\bd{x}, \bd{y})),
\end{aligned}
\end{equation*}
which does not require the gradient of the target distribution $p(\vx).$
\end{thm}

{\bf Proof:} We provide another derivation of the gradient-free KSD. Just need to replacing $k(\bd{x}, \bd{y})$ with $\wt{k}(\bd{x}, \bd{y})$ in KSD~\eqref{intro:ksd}\citep{liu2016kernelized},
\begin{align}
\wt{\kappa}_p (\bd{x},  \bd{y}) &
= \bd{s}_p(\bd{x})^\top \widetilde{k}(\bd{x},\bd{y})\bd{s}_p(\bd{y}) +\bd{s}_p(\bd{x})^\top \nabla_{\bd{y}}\widetilde{k}(\bd{x},\bd{y}) \\
&+\bd{s}_p(\bd{y})^\top \nabla_{\bd{x}} \widetilde{k}(\bd{x},\bd{y})+\nabla_{\bd{y}}\cdot(\nabla_{\bd{x}}\widetilde{k}(\bd{x}, \bd{y})).
\end{align}

\begin{equation}
\nabla_{\bd{x}} \widetilde{k}(\bd{x}, \bd{y}) = \frac{\nabla_{\bd{x}}k(\bd{x}, \bd{y})}{\ell(\bd{x})\ell(\bd{y})} - \frac{k(\bd{x}, \bd{y}) \nabla_{\bd{x}}\log\ell(\bd{x})}{\ell(\bd{x})\ell(\bd{y})}
\end{equation}

\begin{equation}
\nabla_{\bd{y}} \widetilde{k}(\bd{x}, \bd{y}) = \frac{\nabla_{\bd{y}}k(\bd{x}, \bd{y})}{\ell(\bd{x})\ell(\bd{y})} - \frac{k(\bd{x}, \bd{y}) \nabla_{\bd{y}}\log\ell(\bd{y})}{\ell(\bd{x})\ell(\bd{y})}
\end{equation}

With simple calculation, we can get the following equations,
\begin{equation}
\label{gf:d1}
\begin{aligned}
\nabla_{\bd{y}}\cdot(\nabla_{\bd{x}} \widetilde{k}(\bd{x}, \bd{y})) & = \frac{\nabla_{\bd{y}}\cdot(\nabla_{\bd{x}}k(\bd{x}, \bd{y}))}{\ell(\bd{x})\ell(\bd{y})} -\frac{\nabla_{\bd{x}}k(\bd{x}, \bd{y})\cdot \nabla_{\bd{y}}\log\ell(\bd{y})}{\ell(\bd{x})\ell(\bd{y})}\\
& - \frac{\nabla_{\bd{y}} k(\bd{x}, \bd{y})\cdot \nabla_{\bd{x}}\log\ell(\bd{x})}{\ell(\bd{x})\ell(\bd{y})} + \frac{k(\bd{x}, \bd{y}) \nabla_{\bd{x}}\log\ell(\bd{x})\cdot \nabla_{\bd{y}}\log\ell(\bd{y})}{\ell(\bd{x})\ell(\bd{y})}
\end{aligned}
\end{equation}

\begin{equation}
\label{gf:d2}
\begin{aligned}
\bd{s}_p(\bd{x})^\top \wt{k}(\bd{x},\bd{y})\bd{s}_p(\bd{y}) & =\frac{(\bd{s}_{p_0}(\bd{x})+\bd{s}_{\ell}(\bd{x}))^\T k(\bd{x}, \bd{y})(\bd{s}_{p_0}(\bd{y})+\bd{s}_{\ell}(\bd{y}))}{\ell(\bd{x})\ell(\bd{y})}\\
& = \frac{\bd{s}_{p_0}(\bd{x})^\top k(\bd{x}, \bd{y})\bd{s}_{p_0}(\bd{y})+  \bd{s}_{p_0}(\bd{x})^\top k(\bd{x}, \bd{y})\bd{s}_{\ell}(\bd{y})}{\ell(\bd{x})\ell(\bd{y})} \\ 
& +\frac{\bd{s}_{\ell}(\bd{x})^\top k(\bd{x}, \bd{y})\bd{s}_{p_0}(\bd{y})+  \bd{s}_{\ell}(\bd{x})^\top k(\bd{x}, \bd{y})\bd{s}_{\ell}(\bd{y})}{\ell(\bd{x})\ell(\bd{y})}
\end{aligned}
\end{equation}

\begin{equation}
\label{gf:d3}
\begin{aligned}
& \bd{s}_p(\bd{x})^\top \nabla_{\bd{y}}\widetilde{k}(\bd{x},\bd{y})=( \bd{s}_{p_0}(\bd{x})+\bd{s}_{\ell}(\bd{x}))^\top[\frac{\nabla_{\bd{y}}k(\bd{x}, \bd{y})}{\ell(\bd{x})\ell(\bd{y})} - \frac{k(\bd{x}, \bd{y}) \bd{s}_\ell(\bd{y})}{\ell(\bd{x})\ell(\bd{y})}] \\
& =\frac{( \bd{s}_{p_0}(\bd{x})+\bd{s}_{\ell}(\bd{x}))^\top\nabla_{\bd{y}}k(\bd{x}, \bd{y})}{\ell(\bd{x})\ell(\bd{y})} - \frac{( \bd{s}_{p_0}(\bd{x})+\bd{s}_{\ell}(\bd{x}))^\top k(\bd{x}, \bd{y}) \bd{s}_\ell(\bd{y})}{\ell(\bd{x})\ell(\bd{y})}
\end{aligned}
\end{equation}

\begin{equation}
\label{gf:d4}
\begin{aligned}
&\bd{s}_p(\bd{y})^\top \nabla_{\bd{x}} \widetilde{k}(\bd{x},\bd{y}) =(\bd{s}_{p_0}(\bd{y})+\bd{s}_{\ell}(\bd{y}))^\top [\frac{\nabla_{\bd{x}}k(\bd{x}, \bd{y})}{\ell(\bd{x})\ell(\bd{y})} - \frac{k(\bd{x}, \bd{y}) \bd{s}_\ell(\bd{x})}{\ell(\bd{x})\ell(\bd{y})}]\\
& =  (\bd{s}_{p_0}(\bd{y})+\bd{s}_{\ell}(\bd{y}))^\top\frac{\nabla_{\bd{x}}k(\bd{x}, \bd{y})}{\ell(\bd{x})\ell(\bd{y})} - \frac{(\bd{s}_{p_0}(\bd{y})+\bd{s}_{\ell}(\bd{y}))^\top k(\bd{x}, \bd{y}) \bd{s}_\ell(\bd{x})}{\ell(\bd{x})\ell(\bd{y})}
\end{aligned}
\end{equation}

\begin{equation}
\label{gf:d5}
\begin{aligned}
\nabla_{\bd{y}}\cdot(\nabla_{\bd{x}} \widetilde{k}(\bd{x}, \bd{y})) & = \frac{\nabla_{\bd{y}}\cdot(\nabla_{\bd{x}}k(\bd{x}, \bd{y}))}{\ell(\bd{x})\ell(\bd{y})} -\frac{\nabla_{\bd{x}}k(\bd{x}, \bd{y})\cdot \bd{s}_\ell(\bd{y})}{\ell(\bd{x})\ell(\bd{y})}\\
& - \frac{\nabla_{\bd{y}} k(\bd{x}, \bd{y})\cdot \bd{s}_\ell(\bd{x})}{\ell(\bd{x})\ell(\bd{y})} + \frac{\bd{s}_\ell(\bd{x})^\top k(\bd{x}, \bd{y})  \bd{s}_\ell(\bd{y})}{\ell(\bd{x})\ell(\bd{y})}
\end{aligned}
\end{equation}

Combining equations~\eqref{gf:d1}, \eqref{gf:d2}, \eqref{gf:d3}, \eqref{gf:d4}, \eqref{gf:d5}, we get the form of the gradient-free KSD defined in \eqref{app:bbis:ksd}. The gradient-free KSD in \eqref{app:bbis:ksd} leverages the gradient information of a surrogate distribution $\rho(\vx)$ and corrects its bias with a form of importance weights.

\section{Detail of Network Architecture in Chapter \ref{chap:disc}}
\label{append:disc}
We use the same AlexNet architecture as \citet{zhu2018binary}, where the specific hyper-parameters are provided in the following table~\ref{tab:alexnet}.
\begin{table}[htb]
    \centering
    \begin{tabular}{|c|c|c|} \hline
    Layer & Type & Parameters \\\hline
     1    & Conv & Depth: 96, K: $11\times 11$, S: 4, P:0 \\
     2  &  Relu &  - \\
     3 & MaxPool & K: $3\times 3$, S: 2 \\
     4 & BatchNorm & -  \\
     5 & Conv & Depth: 256, K: $5\times 5$, S: 1, P:1 \\
    6  &  Relu &  - \\
      7 & MaxPool & K: $3\times 3$, S: 2 \\
     8 & BatchNorm & -  \\
     9 & Conv & Depth: 384, K: $3\times 3$, S: 1, P:1 \\
     10  &  Relu &  - \\
     11 & Conv & Depth: 384, K: $3\times 3$, S: 1, P:1 \\
    12  &  Relu &  - \\
    13 & Conv & Depth: 256, K: $3\times 3$, S: 1, P:1 \\
    14  &  Relu &  - \\
    15 & MaxPool & K: $3\times 3$, S: 2 \\
    16  &Dropout & $p=0.5$\\
    17  & FC & Width=4096\\
     18  &  Relu &  - \\
    19  &Dropout & $p=0.5$\\
    20  & FC & Width=4096\\
    21  &  Relu &  - \\
    22 & FC & Width=10\\\hline
\end{tabular}
\caption{Detailed architecture of AlexNet. "K" denotes the kernel size; "S" denotes the stride; "P" denotes the padding. \label{tab:alexnet}}
\end{table}

\section{Proofs of Lemmas in Chapter~\ref{chap:gof}}
\label{append:gof}
The square of the gradient-free KSD between $q(\vx)$ and $p(\vx)$ is 
\begin{equation}
\label{gof:ksd}
\wt{\mathcal{S}}(q, p) = \E_{\vx, \vx'\sim q}[w(\vx)k_{\rho}(\vx, \vx') w(\vx')],  
\end{equation}
where $w(\vx)=\rho(\vx)/p(\vx)$ and  $\kappa_{\rho}(\bd{x},  \bd{x}')$ is defined as,
\begin{align}
\label{imp:kernel}
\!\! \kappa_{\rho}(\bd{x},  \bd{x}')\!\! & = \!\! \bd{s}_{\rho}(\bd{x})^\top k(\bd{x},\bd{x}')\bd{s}_{\rho}(\bd{x}')
+\bd{s}_{\rho}(\bd{x})^\top \nabla_{\bd{x}'}k(\bd{x},\bd{x}') \\ \notag
 & \!\! +\bd{s}_{\rho}(\bd{x}')^\top \nabla_{\bd{x}}  k(\bd{x},\bd{x}')\!\! +\!\!\nabla_{\bd{x}}\!\cdot\!(\nabla_{\bd{x}'}k(\bd{x}, \bd{x}')),
\end{align}
$\bd{s}_{\rho}(\bd{x})$ is the score function of the surrogate distribution $\rho(\vx).$
With $\{\vx_i\}_{i=1}^n$ from $q(\vx)$, the GF-KSD between $q(\vx)$ and $p(\vx)$ can be estimated by the U-statistics,
\begin{equation}
\label{emp:ksd}
\hat{\mathbb{S}}(q, p) =\frac{1}{(n-1)n}\sum_{1\le i\neq j\le n} w(\vx_i)\kappa_{\rho}(\bd{x}_i, \vx_j)w(\vx_j) 
\end{equation}
\begin{lem}
Let $k(\vx,\vx')$ be a positive definite kernel. Suppose $\|p(\vx)(s_{q}(\vx)-s_{\rho}(\vx)) \|_2^2<\infty,$ and 
$\wt{\mathcal{S}}(q, p) = \E_{\vx, \vx'\sim q}[w(\vx)k_{\rho}(\vx, \vx') w(\vx')]< \infty,$ we have:
\begin{enumerate}
\item If $q\neq p,$ then $\sqrt{n}(\hat{\mathbb{S}}(q, p)- \mathbb{S}(q, p) )\rightarrow \mathcal{N}(0, \sigma_u^2)$ in distribution with the variance $\sigma_u^2=\mathrm{var}_{\vx\sim q}(\E_{\vx'\sim} [w(\vx)k_{\rho}(\vx, \vx') w(\vx')]),$ and $\sigma_u^2\neq 0.$
\item If $q=p,$ then $\sigma_u^2\neq 0.$ And we have
$$n\hat{\mathbb{S}}(q, p)\rightarrow \sum_{j=1}^\infty c_j(Z_j^2-1), ~ \text{in distribution,} $$
where $\{Z_j\}$ are i.i.d. standard Gaussian random variable, and $\{c_j\}$ are the eigenvalues of kernel $w(\vx)k_{\rho}(\vx, \vx') w(\vx')$ under distribution $q.$
\end{enumerate}
\end{lem}
{\bf Proof}: Based on the standard asymptotic results of U-statistics~\citep{serfling2009approximation} and the proof of Theorem 4.1 in ~\citet{liu2016kernelized}, it is straightforward to derive the two results above. 

{\bf Bootstrap Sample} The asymptotic distribution of $\hat{\mathbb{S}}(q_c, p_c)$ under the null hypothesis cannot be evaluated. In order to perform goodness-of-fit test, we draw random multinomial weights $u_1, \cdots, u_n\sim \mathrm{Multi}(n;1/n,\cdots,1/n),$ and calculate
\begin{equation}
\label{boot:ksd}
\hat{\mathbb{S}}^*(q_c, p_c) =\sum_{i\neq j } (u_i\!-\!\frac1n)w(\vx_i) \kappa_{\rho}(\bd{x}_i, \vx_j)w(\vx_j) (u_j-\frac1n).     
\end{equation}
We repeat this process by $m$ times and calculate the critical values of the test by taking the $(1-\alpha)$-th quantile, denoted by $\gamma_{1-\alpha}$, of the bootstrapped statistics $\{\hat{\mathbb{S}}_i^*(q_c, p_c)\}.$ 
\begin{pro}
\label{gof:pro}
Suppose the conditions in \ref{gof:lem} hold. For any fixed $q_c\neq p_c,$ the limiting power of the test that rejects the null hypothesis $q_c \neq p_c$ when $\hat{\mathbb{S}}^*(q_c, p_c)\ge\gamma_{1-\alpha}$
is one, which means the test is consistent in power against any fixed $q_c\neq p_c.$
\end{pro}
The proof is similar to the procedure in Proposition~\citet{liu2016kernelized}.  The Proposition~\ref{gof:pro} theoretically justifies the correctness of our proposed goodness-of-fit testing algorithm.
\section{Proofs of Theorems in Chapter~\ref{chap:boot}}
\label{append:boot}
In this section, we study the statistical efficiency of the estimators $\vv {\hat\theta}_{\KL-C}$ and $\vv {\hat\theta}_{\KL-W}$ proposed in Chapter~\ref{chap:boot}. First, we study the asymptotic property of the KL-naive estimator $\vv {\hat\theta}_{\KL}$, and prove Theorem~\ref{boot:thm1}. Then we analyze the asymptotic property of the KL-Control estimator $\vv {\hat\theta}_{\KL-C}.$ Finally, we study the asymptotic property of the KL-Weighted estimator $\vv {\hat\theta}_{\KL-W}.$


\paragraph{Notations and Assumptions}
To simplify the notations for the proofs in the following, we define the following notations.
\begin{equation}
\label{notation}
\begin{aligned}
&s(\boldsymbol{x};\boldsymbol{\theta})=\log p(\boldsymbol{x}\mid\boldsymbol{\theta});\quad \text{\.{s}}(\boldsymbol{x};\boldsymbol{\theta})=\frac{\partial \log p(\boldsymbol{x}\mid\boldsymbol{\theta})}{\partial \boldsymbol{\theta}};\quad \text{\"{s}}(\boldsymbol{x};\boldsymbol{\theta})=\frac{\partial^2 \log p(\boldsymbol{x}\mid\boldsymbol{\theta})}{\partial \boldsymbol{\theta}^2};\\
&I(\boldsymbol{\theta})=\mathbb{E}(\text{\"{s}}(x,\boldsymbol{\theta})); \quad I(\boldsymbol{\hat{\theta}}_k,\boldsymbol{\theta}_{\KL}^*)=\mathbb{E} (\text{\"{s}}(\boldsymbol{x},\boldsymbol{\theta}_{\KL}^*)\mid \boldsymbol{\hat{\theta}}_k).
\end{aligned}
\end{equation}
The theoretical results are based on the following assumptions.
\begin{ass}
\label{assump}
 1. $\log p(\boldsymbol{x}\mid\boldsymbol{\theta}),$
 $\frac{\partial\log p(\boldsymbol{x}\mid\boldsymbol{\theta})}{\partial\boldsymbol{\theta}},$
 and $\frac{\partial^2\log p(\boldsymbol{x}\mid\boldsymbol{\theta})}{\partial\boldsymbol{\theta}\partial\boldsymbol{\theta}^\top}$
 are continuous for $\forall \boldsymbol{x}\in\mathcal{X}$ and $\forall \boldsymbol{\theta}\in\Theta;$ 2. $\frac{\partial^2\log p(\boldsymbol{x}\mid\boldsymbol{\theta})}{\partial\boldsymbol{\theta}\partial\boldsymbol{\theta}^\top}$ is positive definite and $C_1\leq \|\frac{\partial^2\log p(\boldsymbol{x}\mid\boldsymbol{\theta})}{\partial\boldsymbol{\theta}\partial\boldsymbol{\theta}^\top}\|\leq C_2$ in a neighbor of $\boldsymbol{\theta}^*$ for $\forall x\in\mathcal{X}$, and $C_1$, $C_2$ are some positive constants.
\end{ass}
We start with investigating the theoretical property of $\boldsymbol{\hat{\theta}}_{\KL}$. 
\begin{lem}
 Based on Assumption $\ref{assump}$, as $n\to\infty,$ we have $\mathbb{E}(\boldsymbol{\hat{\theta}}_{\KL}-\boldsymbol{\theta}_{\KL}^*)=o((dn)^{-1}).$  Further, in terms of estimating the true parameter, we have
\begin{equation}
\label{globalklc}
\mathbb{E}\|\boldsymbol{\hat{\theta}}_{\KL}-\boldsymbol{\theta}^*\|^2=O(N^{-1}+(dn)^{-1}).
\end{equation}
\end{lem}
{\bf Proof:} Based on Equation (\ref{KLdivmax}) and (\ref{KLdivmaxapprox}), we know
\begin{equation}
\label{KLdiffen}
\sum_{k=1}^d\frac{1}{n}\sum_{j=1}^n \text{\.{s}}(\boldsymbol{\widetilde{x}}_j^k;\boldsymbol{\hat{\theta}}_{\KL})- \sum_{k=1}^d \int p(x|\boldsymbol{\hat{\theta}}_k)\text{\.{s}}(\boldsymbol{x};\boldsymbol{\theta}_{\KL}^*) d\boldsymbol{x}=0.
\end{equation}
By the law of large numbers, we can rewrite Equation (\ref{KLdiffen}) as
\begin{equation}
\label{KLdiffer}
\sum_{k=1}^d \int p(\boldsymbol{x}| \boldsymbol{\hat{\theta}}_k)\text{\.{s}}(\boldsymbol{x}; \boldsymbol{\hat{\theta}}_{\KL})d\boldsymbol{x}-\sum_{k=1}^d \int p(x| \boldsymbol{\hat{\theta}}_k)\text{\.{s}}(\boldsymbol{x};\boldsymbol{\theta}_{\KL}^*)d\boldsymbol{x}=o_p(\frac1n).
\end{equation}
We also observe that $\text{\.{s}}(\boldsymbol{x}; \boldsymbol{\hat{\theta}}_{\KL})-\text{\.{s}}(\boldsymbol{x};\boldsymbol{\theta}_{\KL}^*)= \big[ \int_0^1 \text{\"{s}}(\boldsymbol{x};\boldsymbol{\theta}_{\KL}^*+t(\boldsymbol{\hat{\theta}}_{\KL}-\boldsymbol{\theta}_{\KL}^*))dt \big] ~ (\boldsymbol{\theta}_{\KL}^*-\boldsymbol{\hat{\theta}}_{\KL}).$
Therefore,  Equation (\ref{KLdiffer}) can be written as
\begin{equation}
\label{KLdifferen}
\bigg [\sum_{k=1}^d \int p(x| \boldsymbol{\hat{\theta}}_k)\int_0^1 \text{\"{s}}(\boldsymbol{x};\boldsymbol{\theta}_{\KL}^*+t(\boldsymbol{\hat{\theta}}_{\KL}-\boldsymbol{\theta}_{\KL}^*)) dt d\boldsymbol{x} \bigg] ~ (\boldsymbol{\theta}_{\KL}^*-\boldsymbol{\hat{\theta}}_{\KL}) = o_p(\frac1n).
\end{equation}
Under our Assumption \ref{assump}, the Fish Information matrix $I(\boldsymbol{\theta})$ is positive definite in a neighborhood of $\boldsymbol{\theta}^*,$ then we can find constant $C_1$, $C_2$ such that $C_1\leq\|\int p(x| \boldsymbol{\hat{\theta}}_k)\int_0^1 \text{\"{s}}(\boldsymbol{x};\boldsymbol{\theta}_{\KL}^*+t(\boldsymbol{\hat{\theta}}_{\KL}-\boldsymbol{\theta}_{\KL}^*)) dtd\boldsymbol{x}\|\leq C_2$.  Therefore, we can get $\mathbb{E}(\boldsymbol{\hat{\theta}}_{\KL}-\boldsymbol{\theta}_{\KL}^*)=o((dn)^{-1}).$ $\square$

 The following theorem provides the MSE between $\boldsymbol{\hat{\theta}}_{\KL}$ and $\boldsymbol{\theta}_{\KL}^*$ and that between $\boldsymbol{\hat{\theta}}_{\KL}$ and $\boldsymbol{\theta}^*$.
\begin{thm}
Based on Assumption $\ref{assump}$, as $n\to\infty$, $\mathbb{E}\|\boldsymbol{\hat{\theta}}_{\KL}-\boldsymbol{\theta}_{\KL}^*\|^2=O(\frac{1}{nd}).$ Further, in terms of estimating the true parameter, we have
\begin{equation}
\mathbb{E}\|\boldsymbol{\hat{\theta}}_{\KL}-\boldsymbol{\theta}^*\|^2=O(N^{-1}+(dn)^{-1}).
\end{equation}
\end{thm}
{\bf Proof:} According to the Equation (\ref{KLdivmaxapprox}),
\begin{equation}
\label{unn}
\boldsymbol{\hat{\theta}}_{\KL}=\argmax_{\boldsymbol{\theta}\in\Theta}\sum_{k=1}^d\frac{1}{n}\sum_{j=1}^n s(\boldsymbol{\widetilde{x}}_j^k;\boldsymbol{\theta}).
\end{equation}
Then the first order derivative of Equation (\ref{unn}) with respect to $\boldsymbol{\theta}$ at $\boldsymbol{\theta}=\boldsymbol{\hat{\theta}}_{\KL}$ is zero,
\begin{equation}
\label{taylorapp}
\sum_{k=1}^d\frac{1}{n}\sum_{j=1}^n \text{\.{s}}(\boldsymbol{\widetilde{x}}_j^k;\boldsymbol{\hat{\theta}}_{\KL})=0.
\end{equation}
By Taylor expansion of Equation (\ref{taylorapp}), we get
$$\sum_{k=1}^d\frac{1}{n}\sum_{j=1}^n (\text{\.{s}}(\boldsymbol{\widetilde{x}}_j^k;\boldsymbol{\theta}_{\KL}^*)+\text{\"{s}}(\boldsymbol{\widetilde{x}}_j^k;\boldsymbol{\hat{\theta}}_{\KL})(\boldsymbol{\hat{\theta}}_{\KL}-\boldsymbol{\theta}_{\KL}^*))+o_p(\boldsymbol{\hat{\theta}}_{\KL}-\boldsymbol{\theta}_{\KL}^*)=0.$$
By the law of large numbers, $\frac{1}{n}\sum_{j=1}^n\text{\"{s}}(\boldsymbol{\widetilde{x}}_j^k;\boldsymbol{\hat{\theta}}_{\KL}^*)=I(\boldsymbol{\hat{\theta}}_k,\boldsymbol{\theta}_{\KL}^*)+o_p(\frac{1}{n}).$ 
Under our Assumption \ref{assump},  $I(\boldsymbol{\theta})$ is positive definite in a neighborhood of $\boldsymbol{\theta}^*.$ Since $\hat{\boldsymbol{\theta}}_k$ are in the neighborhood of $\boldsymbol{\theta}^*$, $I(\boldsymbol{\hat{\theta}}_k, \boldsymbol{\theta}_{\KL}^*)$ is positive definite, for $k=1\in [d].$ Then we have
\begin{equation}
\label{varianceKL}
\boldsymbol{\hat{\theta}}_{\KL}-\boldsymbol{\theta}_{\KL}^*=(\sum_{k=1}^dI(\boldsymbol{\hat{\theta}}_k,\boldsymbol{\theta}_{\KL}^*))^{-1}\sum_{k=1}^d\frac{1}{n}\sum_{j=1}^n \text{\.{s}}(\boldsymbol{\widetilde{x}}_j^k;\boldsymbol{\theta}_{\KL}^*)+o_p(\frac{1}{n})=0.
\end{equation}
By the central limit theorem, $\frac{1}{\sqrt{n}}\sum_{j=1}^n \text{\.{s}}(\boldsymbol{\widetilde{x}}_j^k;\boldsymbol{\theta}_{\KL}^*)$ converges to a normal distribution. By some simple calculation, we have
\begin{equation}
\label{cova}
\mathrm{Cov}(\boldsymbol{\hat{\theta}}_{\KL}-\boldsymbol{\theta}_{\KL}^*,\boldsymbol{\hat{\theta}}_{\KL}-\boldsymbol{\theta}_{\KL}^*)=\frac{1}{n}(\sum_{k=1}^dI(\boldsymbol{\hat{\theta}}_k,\boldsymbol{\theta}_{\KL}^*))^{-1}\sum_{k=1}^d \mathrm{Var}(\text{\.{s}}(\boldsymbol{x};\boldsymbol{\theta}_{\KL}^*)\mid\boldsymbol{\hat{\theta}}_k)(\sum_{k=1}^dI(\boldsymbol{\hat{\theta}}_k,\boldsymbol{\theta}_{\KL}^*))^{-1}.
\end{equation}
According to our Assumption \ref{assump}, we already know $I(\boldsymbol{\hat{\theta}}_k,\boldsymbol{\theta}_{\KL}^*)$ is positive definite, $C_1\le \|I(\boldsymbol{\hat{\theta}}_k,\boldsymbol{\theta}_{\KL}^*)\|\le C_2$. We have $(\sum_{k=1}^dI(\boldsymbol{\hat{\theta}}_k,\boldsymbol{\theta}_{\KL}^*))^{-1}=O(\frac{1}{d})$ and $\sum_{k=1}^d \mathrm{Var}(\text{\.{s}}(\boldsymbol{x};\boldsymbol{\theta}_{\KL}^*)\mid\boldsymbol{\hat{\theta}}_k)=O(d).$
Therefore,  $\mathbb{E}\|\boldsymbol{\hat{\theta}}_{\KL}-\boldsymbol{\theta}_{\KL}^*\|^2=\mathrm{trace}(\mathrm{Cov}(\boldsymbol{\hat{\theta}}_{\KL}-\boldsymbol{\theta}_{\KL}^*,\boldsymbol{\hat{\theta}}_{\KL}-\boldsymbol{\theta}_{\KL}^*))=O(\frac{1}{nd}).$ Because the MSE between the exact $\KL$ estimator $\boldsymbol{\theta}_{\KL}^*$ and the true parameter $\boldsymbol{\theta}^*$ is
$O(N^{-1})$
as shown in \citet{liu2014distributed}, the MSE between $\boldsymbol{\hat{\theta}}_{\KL}$ and the true parameter $\boldsymbol{\theta}^*$ is
\begin{equation*}
\mathbb{E}\|\boldsymbol{\hat{\theta}}_{\KL}-\boldsymbol{\theta}^*\|^2\approx\mathbb{E}\|\boldsymbol{\hat{\theta}}_{\KL}-\boldsymbol{\theta}^*_{\KL}\|^2+\mathbb{E}\|\boldsymbol{\theta}_{\KL}^*-\boldsymbol{\theta}^*\|^2=O(N^{-1}+(dn)^{-1}).
\end{equation*}
We complete the proof of this theorem. $\square$
\paragraph{Theoretical Result on $\mathrm{\KL-C}$ Estimator $\boldsymbol{\hat{\theta}}_{\mathrm{\KL-C}}$}
In this section, we analyze the MSE of our proposed estimator $\boldsymbol{\hat{\theta}}_{\KL-C}$ and prove Theorem~\ref{Control}.
\begin{thm}
Under Assumptions \ref{assump}, we have
$$\text{as }~ n\to \infty,\quad n\mathbb{E}\|\boldsymbol{\hat{\theta}}_{\KL-C}-\boldsymbol{\theta}_{\KL}^*\|^2 < n\mathbb{E}\|\boldsymbol{\hat{\theta}}_{\KL}-\boldsymbol{\theta}_{\KL}^*\|^2.$$
\end{thm}
Since $\widetilde{\boldsymbol{\theta}}_k$ is the MLE of data $\{\boldsymbol{\widetilde{x}}_j^k\}_{j=1}^n$, then we have
\begin{equation}
\label{MLE}
(\boldsymbol{\widetilde{\theta}}_k-\boldsymbol{\hat{\theta}}_k)= -I(\boldsymbol{\hat{\theta}}_k)^{-1}\frac{1}{n}\sum_{j=1}^n\dot{s}(\boldsymbol{\widetilde{x}}_j^k;\boldsymbol{\hat{\theta}}_k)+o_p(\frac{1}{n}).
\end{equation}
Then $\mathbb{E}(\boldsymbol{\widetilde{\theta}}_k-\boldsymbol{\hat{\theta}}_k)=o(\frac{1}{n}).$ According to Theorem (\ref{boot:thm1}), when $\boldsymbol{\mathfrak{B}}_k$ is a constant matrix, for $k\in[d],$
$$\mathbb{E}(\boldsymbol{\hat{\theta}}_{\KL-C}-\boldsymbol{\theta}_{\KL}^*)=\mathbb{E}(\boldsymbol{\hat{\theta}}_{\KL}-\boldsymbol{\theta}_{\KL}^*)+\sum_{k=1}^d \boldsymbol{\mathfrak{B}}_k\mathbb{E}(\boldsymbol{\widetilde{\theta}}_k-\boldsymbol{\hat{\theta}}_k)=o(\frac{1}{n}).$$

Notice that $\frac{1}{n}\sum_{j=1}^n\text{\.{s}}(\boldsymbol{\widetilde{x}}_j^r;\boldsymbol{\hat{\theta}}_{r})$ and $\frac{1}{n}\sum_{j=1}^n\text{\.{s}}(\boldsymbol{\widetilde{x}}_j^t;\boldsymbol{\hat{\theta}}_{t})$ are independent when $r\neq t.$
According to Equation (\ref{varianceKL}), we know $\sum_{k=1}^d\frac1n\sum_{j=1}^n \text{\.{s}}(\boldsymbol{\widetilde{x}}_j^k;\boldsymbol{\theta}_{\KL}^*)$ and $\frac{1}{n}\sum_{j=1}^n\dot{s}(\boldsymbol{\widetilde{x}}_j^k;\boldsymbol{\hat{\theta}}_k)$ are correlated to each other for $k\in [d],$
\begin{equation*}
\begin{split}
&\mathrm{Cov}((\boldsymbol{\hat{\theta}}_{\KL-C}-\boldsymbol{\theta}_{\KL}^*),(\boldsymbol{\hat{\theta}}_{\KL-C}-\boldsymbol{\theta}_{\KL}^*))=\mathrm{Cov}(\boldsymbol{\hat{\theta}}_{\KL}-\boldsymbol{\theta}_{\KL}^*,\boldsymbol{\hat{\theta}}_{\KL}-\boldsymbol{\theta}_{\KL}^*)\\
&+2\sum_{k=1}^d\boldsymbol{\mathfrak{B}}_k
\mathrm{Cov}(\boldsymbol{\hat{\theta}}_{\KL}-\boldsymbol{\theta}_{\KL},\boldsymbol{\widetilde{\theta}}_k-\hat{\boldsymbol{\theta}}_{k})^T+
\sum_{k=1}^d\boldsymbol{\mathfrak{B}}_k\mathrm{Cov}((\boldsymbol{\widetilde{\theta}}_k-\boldsymbol{\hat{\theta}}_{k}),(\boldsymbol{\widetilde{\theta}}_k-\boldsymbol{\hat{\theta}}_{k}))\boldsymbol{\mathfrak{B}}_k^T.
\end{split}
\end{equation*}
When $\mathfrak{\boldsymbol{B}}_k=-(\mathrm{Cov}(\boldsymbol{\widetilde{\theta}}_k-\boldsymbol{\hat{\theta}}_{k},\boldsymbol{\widetilde{\theta}}_k-\boldsymbol{\hat{\theta}}_{k}))^{-1}\mathrm{Cov}(\boldsymbol{\hat{\theta}}_{\KL}-
\boldsymbol{\theta}_{\KL}^*,\boldsymbol{\widetilde{\theta}}_k-\hat{\boldsymbol{\theta}}_{k}),$
we have
\begin{equation}
\label{controleff}
\begin{split}
&\mathrm{Cov}(\boldsymbol{\hat{\theta}}_{\KL-C}-\boldsymbol{\theta}_{KL}^*,\boldsymbol{\hat{\theta}}_{\KL-C}-\boldsymbol{\theta}_{\KL}^*)=\mathrm{Cov}(\boldsymbol{\hat{\theta}}_{\KL}-\boldsymbol{\theta}_{\KL}^*,\boldsymbol{\hat{\theta}}_{\KL}-\boldsymbol{\theta}_{\KL}^*)-\\
&\sum_{k=1}^d\mathrm{Cov}(\boldsymbol{\widetilde{\theta}}_k-\boldsymbol{\hat{\theta}}_{k},\boldsymbol{\widetilde{\theta}}_k-\boldsymbol{\hat{\theta}}_{k})^{-1}\mathrm{Cov}(\boldsymbol{\hat{\theta}}_{KL}-
\boldsymbol{\theta}_{\KL}^*,\boldsymbol{\widetilde{\theta}}_k-\hat{\boldsymbol{\theta}}_{k})\mathrm{Cov}(\boldsymbol{\hat{\theta}}_{\KL}-\boldsymbol{\theta}_{\KL}^*,\boldsymbol{\widetilde{\theta}}_k-
\boldsymbol{\hat{\theta}}_{k})^T.
\end{split}
\end{equation}
We know $\mathbb{E}\|\boldsymbol{\hat{\theta}}_{\KL-C}-\boldsymbol{\theta}_{\KL}^*\|^2=\mathrm{trace}(\mathrm{Cov}(\boldsymbol{\hat{\theta}}_{\KL-C}-\boldsymbol{\theta}_{\KL}^*,\boldsymbol{\hat{\theta}}_{\KL-C}-\boldsymbol{\theta}_{\KL}^*))$, $ \mathbb{E}\|\boldsymbol{\hat{\theta}}_{\KL}-\boldsymbol{\theta}_{\KL}^*\|^2=\mathrm{trace}(\mathrm{Cov}(\boldsymbol{\hat{\theta}}_{\KL}-\boldsymbol{\theta}_{\KL}^*,\boldsymbol{\hat{\theta}}_{\KL}-\boldsymbol{\theta}_{\KL}^*)).$ The second term of Equation (\ref{controleff}) is a positive definite matrix, therefore we have
$n\mathbb{E}\|\boldsymbol{\hat{\theta}}_{\KL-C}-\boldsymbol{\theta}_{\KL}^*\|^2< n\mathbb{E}\|\boldsymbol{\hat{\theta}}_{\KL}-\boldsymbol{\theta}_{\KL}^*\|^2$ as $n\to\infty.$
We complete the proof of this theorem. $\square$
\begin{thm}
Under Assumption \ref{assump}, when $N> n\times d$, we have $E\|\boldsymbol{\hat{\theta}}_{\KL-C}- \boldsymbol{\theta}_{\KL}^*\|^2=O(\frac{1}{dn^2})$ as $n\to\infty.$ Further, in terms of estimating the true parameter, we have
\begin{equation*}
\label{globalklc}
\mathbb{E}\|\boldsymbol{\hat{\theta}}_{\KL-C}-\boldsymbol{\theta}^*\|^2=O(N^{-1}+(dn^{2})^{-1}).
\end{equation*}
\end{thm}
From Equation (\ref{KLdivmaxapprox}), we know
\begin{equation}
\label{MLEequation}
\sum_{k=1}^d\frac1n\sum_{j=1}^n\frac{\partial \log p(\boldsymbol{\widetilde{x}}_j^k| \boldsymbol{\hat{\theta}}_{\KL})}{\partial \boldsymbol{\theta}}=0.
\end{equation}
By Taylor expansion, Equation (\ref{MLEequation}) can be rewritten as
\begin{equation}
\label{MLEapproximate}
\sum_{k=1}^d[\frac1n\sum_{j=1}^n\text{\.{s}}(\boldsymbol{\widetilde{x}}_j^k;\boldsymbol{\hat{\theta}}_{k})+ \text{\"{s}}(\boldsymbol{\widetilde{x}}_j^k;\hat{\boldsymbol{\theta}}_{k})(\boldsymbol{\hat{\theta}}_{\KL}-\boldsymbol{\hat{\theta}}_{k}))+O_p(\|\boldsymbol{\hat{\theta}}_{\KL}-\boldsymbol{\hat{\theta}}_{k}\|^2)]=0.
\end{equation}
$\|\boldsymbol{\hat{\theta}}_{\KL}-\boldsymbol{\hat{\theta}}_{k}\|^2\leq \|\boldsymbol{\hat{\theta}}_{\KL}-\boldsymbol{\theta}_{\KL}^*\|^2+\|\boldsymbol{\theta}_{\KL}^*-\boldsymbol{\hat{\theta}}_{k}\|^2$. As we know from \citet{liu2014distributed}, we have
\begin{equation}
\label{liuproof}
\|\boldsymbol{\theta}_{\KL}^*-\boldsymbol{\hat{\theta}}_{k}\|^2\le\|\boldsymbol{\theta}_{\KL}^*- \boldsymbol{\theta}^*\|^2+\|\boldsymbol{\theta}^*- \boldsymbol{\hat{\theta}}_k\|^2=O_p(\frac{d}{N}),
\end{equation}
When $N> n\times d$, we have $\|\boldsymbol{\hat{\theta}}_{\KL}-\boldsymbol{\hat{\theta}}_{k}\|^2=O_p(\frac{1}{nd})$. And it is also easy to derive
\begin{equation}
\label{asym}
\boldsymbol{\hat{\theta}}_{\KL}-\boldsymbol{\hat{\theta}}_{k}=\boldsymbol{\hat{\theta}}_{\KL}-\boldsymbol{\theta}_{\KL}^*+\boldsymbol{\theta}_{\KL}^*-\boldsymbol{\theta}^*+\boldsymbol{\theta}^*-\boldsymbol{\hat{\theta}}_{k}=
o_p(\frac{1}{N})+o_p(\frac{1}{N})+o_p(\frac{d}{N})=o_p(\frac{1}{nd}+\frac{d}{N}),
\end{equation}
where $\boldsymbol{\theta}_{KL}^*-\boldsymbol{\theta}^*=o_p(\frac1N)$ has been proved in Liu and Ihler's paper(2014).
According to the law of large numbers, $\frac1n\sum_{j=1}^n\text{\"{s}}(\boldsymbol{\widetilde{x}}_j^k;\boldsymbol{\hat{\theta}}_{k})= I(\boldsymbol{\hat{\theta}}_k)+o_p(\frac{1}{n})$, then we have
\begin{equation}
\label{MLEcalculate}
(\boldsymbol{\hat{\theta}}_{\KL}-\boldsymbol{\theta}_{\KL}^*)=-(\sum_{k=1}^dI(\boldsymbol{\hat{\theta}}_k))^{-1}\sum_{k=1}^d\frac{1}{n}\sum_{j=1}^n\text{\.{s}}(\boldsymbol{\widetilde{x}}_j^k;\boldsymbol{\hat{\theta}}_{k})+
O_p(\frac{1}{nd}).
\end{equation}
Notie that $\frac{1}{n}\sum_{j=1}^n\text{\.{s}}(\boldsymbol{\widetilde{x}}_j^r;\boldsymbol{\hat{\theta}}_{r})$ and $\frac{1}{n}\sum_{j=1}^n\text{\.{s}}(\boldsymbol{\widetilde{x}}_j^t;\boldsymbol{\hat{\theta}}_{t})$ are independent when $r\neq t.$
Therefore from (\ref{MLE}) and (\ref{MLEcalculate}), the covariance matrix of $n(\boldsymbol{\hat{\theta}}_{\KL}-\boldsymbol{\theta}_{\KL}^*)$ and $n(\widetilde{\boldsymbol{\theta}}_k-\boldsymbol{\hat{\theta}}_k)$ is
$$
\mathrm{Cov}(n(\boldsymbol{\hat{\theta}}_{\KL}-\boldsymbol{\theta}_{\KL}^*),n(\boldsymbol{\widetilde{\theta}}_k-\boldsymbol{\hat{\theta}}_{k}))=n(\sum_{k=1}^dI(\boldsymbol{\hat{\theta}}_k))^{-1}+(\sum_{k=1}^dI(\boldsymbol{\hat{\theta}}_k))^{-1}O(1),
$$
for $k\in [d].$ According to Assumption \ref{assump}, we know $\sum_{k=1}^dI(\boldsymbol{\hat{\theta}}_k)=O(d)$. Then we will have 
\begin{equation}
\label{covariance}
\mathrm{Cov}(n(\boldsymbol{\hat{\theta}}_{\KL}-\boldsymbol{\theta}_{\KL}^*),n(\boldsymbol{\widetilde{\theta}}_k-\boldsymbol{\hat{\theta}}_{k}))=n(\sum_{k=1}^dI(\boldsymbol{\hat{\theta}}_k))^{-1}+O(\frac1d), ~~\text{for}~~k\in[d].
\end{equation}

According to Theorem \ref{boot:thm1} and Equation (\ref{cova}), by the law of large numbers, it is easy to derive
$$\mathrm{Cov}(n(\boldsymbol{\hat{\theta}}_{\KL}-\boldsymbol{\theta}_{KL}^*),n(\boldsymbol{\hat{\theta}}_{\KL}-\boldsymbol{\theta}_{\KL}^*))=n(\sum_{k=1}^dI(\boldsymbol{\hat{\theta}}_k))^{-1}+o(1).$$
\begin{equation}
\label{controlcov}
\begin{split}
&\mathrm{Cov}(n(\boldsymbol{\hat{\theta}}_{\KL-C}-\boldsymbol{\theta}_{\KL}^*),n(\boldsymbol{\hat{\theta}}_{\KL-C}-\boldsymbol{\theta}_{\KL}^*))=\mathrm{Cov}(n(\boldsymbol{\hat{\theta}}_{\KL}-\boldsymbol{\theta}_{\KL}^*),n(\boldsymbol{\hat{\theta}}_{\KL}-\boldsymbol{\theta}_{\KL}^*)\\
&+2\sum_{k=1}^d\boldsymbol{\mathfrak{B}}_k
\mathrm{Cov}(n(\boldsymbol{\hat{\theta}}_{\KL}-\boldsymbol{\theta}_{\KL}^*),n(\boldsymbol{\widetilde{\theta}}_k-\boldsymbol{\hat{\theta}}_{k}))^\top+
\sum_{k=1}^d\boldsymbol{\mathfrak{B}}_k\mathrm{Cov}(n(\boldsymbol{\widetilde{\theta}}_k-\boldsymbol{\hat{\theta}}_{k}),n(\boldsymbol{\widetilde{\theta}}_k-\boldsymbol{\hat{\theta}}_{k}))\boldsymbol{\mathfrak{B}}_k^T,
\end{split}
\end{equation}
where $\boldsymbol{\mathfrak{B}}_k$ is defined in (\ref{scorecoeff}),
\begin{equation*}
\boldsymbol{\mathfrak{B}}_k=-(\sum_{k=1}^dI(\boldsymbol{\hat{\theta}}_k))^{-1}I(\boldsymbol{\hat{\theta}}_k), \quad k\in [d].
\end{equation*}

According to Equation (\ref{MLE}), we know $\mathrm{Cov}(n(\boldsymbol{\widetilde{\theta}}_k-\boldsymbol{\hat{\theta}}_{k}),n(\boldsymbol{\widetilde{\theta}}_k-\boldsymbol{\hat{\theta}}_{k}))=n(I(\vv {\hat\theta}_k))^{-1}+o(1).$ By some simple calculation, we know that $n^2\mathrm{Cov}(\boldsymbol{\hat{\theta}}_{\KL-C}-\boldsymbol{\theta}_{\KL}^*,\boldsymbol{\hat{\theta}}_{\KL-C}-\boldsymbol{\theta}_{\KL}^*)=O(\frac{1}{d}).$  Therefore, under the Assumption \ref{assump}, when $N> n\times d,$  we get the following result,
$$
\mathbb{E}\|\boldsymbol{\hat{\theta}}_{\KL-C}- \boldsymbol{\theta}_{\KL}^*\|^2=\mathrm{trace}(\mathrm{Cov}(\boldsymbol{\hat{\theta}}_{\KL-C}-\boldsymbol{\theta}_{\KL}^*,\boldsymbol{\hat{\theta}}_{\KL-C}-\boldsymbol{\theta}_{\KL}^*))=O(\frac{1}{dn^2}).$$
We know $\mathbb{E}\|\boldsymbol{\theta}_{\KL}^*-\vv \theta^*\|^2=O(N^{-1})$ from \citet{liu2014distributed}. Then we have
$$\mathbb{E}\|\boldsymbol{\hat{\theta}}_{\KL-C}-\boldsymbol{\theta}^*\|^2\approx\mathbb{E}\|\boldsymbol{\hat{\theta}}_{\KL-C}-\boldsymbol{\theta}^*_{\KL}\|^2+\mathbb{E}\|\boldsymbol{\theta}_{\KL}^*-\boldsymbol{\theta}^*\|^2=O(N^{-1}+(dn^2)^{-1}).$$
The proof of this theorem is complete. $\square$
\paragraph{Theoretical Result on $\mathrm{\KL-W}$ Estimator $\boldsymbol{\hat{\theta}}_{\mathrm{\KL-W}}$}
In this section, we analyze the asymptotic property of $\boldsymbol{\hat{\theta}}_{\mathrm{\KL-W}}$ and prove Theorem~\ref{boot:thm3}. We show the MSE between $\boldsymbol{\hat{\theta}}_{\mathrm{\KL-W}}$ and $\boldsymbol{\theta}_{\KL}^*$ is much smaller than the MSE between the $\KL$-naive estimator $\boldsymbol{\hat{\theta}}_{\KL}$ and $\boldsymbol{\theta}_{\KL}^*.$
\begin{lem}
Under Assumption \ref{assump}, as $n\to \infty$, $\widetilde{\eta}(\boldsymbol{\theta})$ is a more accurate estimator of $\eta(\boldsymbol{\theta})$ than $\hat{\eta}(\boldsymbol{\theta})$, i.e.,
\begin{equation}
n\mathrm{Var}(\widetilde{\eta}(\boldsymbol{\theta}))\leq n\mathrm{Var}(\hat{\eta}(\boldsymbol{\theta})), \quad \text{for any } \boldsymbol{\theta}\in\Theta.
\end{equation}
\end{lem}
By Taylor expansion,
\begin{equation}
\frac{p(\boldsymbol{x}| \boldsymbol{\hat{\theta}}_k)}{p(\boldsymbol{x}|\boldsymbol{ \widetilde{\theta}}_k)}=1+(\log p(\boldsymbol{x}|\boldsymbol{\hat{\theta}}_k)-\log p(\boldsymbol{x}| \boldsymbol{\widetilde{\theta}}_k))+O_p(\|\widetilde{\boldsymbol{\theta}}_k-\boldsymbol{\hat{\theta}}_k\|^2),
\end{equation}
we will have
\begin{equation*}
\widetilde{\eta}(\boldsymbol{\theta})=\sum_{k=1}^d[\frac1n\sum_{j=1}^n(1+(s(\boldsymbol{\widetilde{x}}_j^k; \boldsymbol{\hat{\theta}}_k)-s(\boldsymbol{\widetilde{x}}_j^k; \boldsymbol{\widetilde{\theta}}_k)))s(\boldsymbol{\widetilde{x}}_j^k;\boldsymbol{\theta})+O_p(\|\boldsymbol{\widetilde{\theta}}_k-\boldsymbol{\hat{\theta}}_k\|^2)],
\end{equation*}
Since $s(\boldsymbol{x}; \boldsymbol{\hat{\theta}}_k)-s(\boldsymbol{x}; \widetilde{\boldsymbol{\theta}}_k)=\text{\.{s}}(\boldsymbol{x}; \boldsymbol{\hat{\theta}}_k)(\boldsymbol{\hat{\theta}}_k-\widetilde{\boldsymbol{\theta}}_k),$ according to equation (\ref{MLE}), we have
\begin{equation*}
\widetilde{\eta}(\boldsymbol{\theta})=\hat{\eta}(\boldsymbol{\theta})-\sum_{k=1}^d\frac1n\sum_{j=1}^ns(\boldsymbol{\widetilde{x}}_j^k;\boldsymbol{\theta})\text{\.{s}}(\boldsymbol{\widetilde{x}}_j^k; \boldsymbol{\hat{\theta}}_k)(\boldsymbol{\widetilde{\theta}}_k-\boldsymbol{\hat{\theta}}^k)+O_p(\|\boldsymbol{\widetilde{\theta}}_k-\boldsymbol{\hat{\theta}}_k\|^2),
\end{equation*}
Then according to equation (\ref{MLE}), we have
\begin{equation*}
\hat{\eta}(\boldsymbol{\theta})=\widetilde{\eta}(\boldsymbol{\theta})-\sum_{k=1}^d\mathbb{E}(s(\boldsymbol{\widetilde{x}}_j^k;\boldsymbol{\theta})\text{\.{s}}(\boldsymbol{\widetilde{x}}_j^k; \boldsymbol{\hat{\theta}}_k)\mid \boldsymbol{\hat{\theta}}_k))I(\boldsymbol{\hat{\theta}}_k)^{-1}\frac{1}{n}\sum_{j=1}^n\dot{s}(\boldsymbol{\widetilde{x}}_j^k;\boldsymbol{\hat{\theta}}_k)+O_p(\frac{d}{n}),
\end{equation*}
Denote $\hat{\xi}(\boldsymbol{\theta})=-\sum_{k=1}^d\mathbb{E}(s(\boldsymbol{\widetilde{x}}_j^k;\boldsymbol{\theta})\text{\.{s}}(\boldsymbol{\widetilde{x}}_j^k; \boldsymbol{\hat{\theta}}_k)\mid\boldsymbol{\hat{\theta}}_k))I(\boldsymbol{\hat{\theta}}_k)^{-1}\frac{1}{n}\sum_{j=1}^n\dot{s}(\boldsymbol{x}_j^k;\boldsymbol{\hat{\theta}}_k)$. According to Henmi et al. (2007), $\hat{\xi}(\boldsymbol{\theta})$ is the orthogonal projection of $\hat{\eta}(\boldsymbol{\theta})$ onto the linear space spanned by the score vector component for each $\boldsymbol{\hat{\theta}}_k$, where $k\in [d]$. Then we will have
$\mathrm{Var}(\hat{\eta}(\boldsymbol{\theta}))=\mathrm{Var}(\widetilde{\eta}(\boldsymbol{\theta}))+\mathrm{Var}(\hat{\xi}(\boldsymbol{\theta})).$ Therefore, $n\mathrm{Var}(\widetilde{\eta}(\boldsymbol{\theta}))\leq n\mathrm{Var}(\hat{\eta}(\boldsymbol{\theta})).$
\begin{thm}
Under the Assumption \ref{assump}, for any $\{\boldsymbol{\hat{\theta}}_{k}\}$, we have that $$\text{as } n\to\infty,\quad n\mathbb{E}\|\boldsymbol{\hat{\theta}}_{\KL-W}-\boldsymbol{\theta}_{\KL}^*\|^2\leq n\mathbb{E}\|\boldsymbol{\hat{\theta}}_{\KL}-\boldsymbol{\theta}_{\KL}^*\|^2.$$
\end{thm}
{\bf Proof:} From Equation (\ref{KLweigthed}), we know
$$\sum_{k=1}^d\frac1n\sum_{j=1}^n\frac{p(\boldsymbol{\widetilde{x}}_j^k| \boldsymbol{\hat{\theta}}_k)}{p(\boldsymbol{\widetilde{x}}_j^k| \boldsymbol{\widetilde{\theta}}_k)}\text{\.{s}}(\boldsymbol{\widetilde{x}}_j^k;\boldsymbol{\hat{\theta}}_{\KL-W})=0.$$
Since $\frac{p(\boldsymbol{x}| \boldsymbol{\hat{\theta}}_k)}{p(\boldsymbol{x}| \boldsymbol{\widetilde{\theta}}_k)}=\exp\{\log p(\boldsymbol{x}| \boldsymbol{\hat{\theta}}_k)-\log p(\boldsymbol{x}| \boldsymbol{\widetilde{\theta}}_k)\}=1+(\log p(\boldsymbol{x}| \boldsymbol{\hat{\theta}}_k)-\log p(\boldsymbol{x}| \boldsymbol{\widetilde{\theta}}_k))+O_p(\|\boldsymbol{\widetilde{\theta}}_k-\boldsymbol{\hat{\theta}}_k\|^2),$  we have
\begin{equation}
\label{KLapprox}
\sum_{k=1}^d\frac1n\sum_{j=1}^n\text{\.{s}}(\boldsymbol{x}_j^k;\boldsymbol{\hat{\theta}}_{\KL-W})-\sum_{k=1}^d[\frac1n\sum_{j=1}^n\text{\.{s}}(\boldsymbol{x}_j^k;\boldsymbol{\hat{\theta}}_{\KL-W})
\text{\.{s}}(\boldsymbol{x}_j^k;\boldsymbol{\hat{\theta}}_{k})^T(\boldsymbol{\widetilde{\theta}}_k-\boldsymbol{\hat{\theta}}_k)+O_p(\|\boldsymbol{\widetilde{\theta}}_k-\boldsymbol{\hat{\theta}}_k\|^2)]=0.
\end{equation}
From the asymptotic property of MLE, we know
$\mathbb{E}\|\boldsymbol{\widetilde{\theta}}_k-\boldsymbol{\hat{\theta}}_k\|^2=\frac1n\mathrm{trace}(I(\boldsymbol{\hat{\theta}}_k)).$
Therefore, we know $\|\boldsymbol{\widetilde{\theta}}_k-\boldsymbol{\hat{\theta}}_k\|^2=O_p(\frac{1}{n})$ and $\sum_{k=1}^d \|\boldsymbol{\widetilde{\theta}}_k-\boldsymbol{\hat{\theta}}_k\|^2=O_p(\frac{d}{n}).$

Similar to the derivation of equation (\ref{varianceKL}), according to equation (\ref{MLE}), we have the following equation,
\begin{equation*}
\begin{split}
\boldsymbol{\hat{\theta}}_{\KL-W}-&\boldsymbol{\theta}_{\KL}^*=(\sum_{k=1}^dI(\boldsymbol{\hat{\theta}}_k,\boldsymbol{\theta}_{\KL}^*))^{-1}\sum_{k=1}^d\frac{1}{n}\sum_{j=1}^n \text{\.{s}}(\boldsymbol{\widetilde{x}}_j^k;\boldsymbol{\theta}_{\KL}^*)-\\
&(\sum_{k=1}^dI(\boldsymbol{\hat{\theta}}_k,\boldsymbol{\theta}_{\KL}^*))^{-1}\sum_{k=1}^d\mathbb{E}(\text{\.{s}}(\boldsymbol{\widetilde{x}}_j^k;\boldsymbol{\hat{\theta}}_{\KL-W})^T\text{\.{s}}(\boldsymbol{\widetilde{x}}_j^k;\boldsymbol{\hat{\theta}}_{k})
\mid\boldsymbol{\hat{\theta}}_k)\frac{1}{n}\sum_{j=1}^n\dot{s}(\boldsymbol{\widetilde{x}}_j^k;\boldsymbol{\hat{\theta}}_k)=O_p(\frac{d}{n}).
\end{split}
\end{equation*}
Then we have,
\begin{equation*}
\begin{aligned}
\boldsymbol{\hat{\theta}}_{\KL}-&\boldsymbol{\theta}_{\KL}^*=\boldsymbol{\hat{\theta}}_{\KL-W}-\boldsymbol{\theta}_{\KL}^*\\
&-(\sum_{k=1}^dI(\boldsymbol{\hat{\theta}}_k,\boldsymbol{\theta}_{\KL}^*))^{-1}\sum_{k=1}^d\mathbb{E}(\text{\.{s}}(\boldsymbol{\widetilde{x}}_j^k;\boldsymbol{\hat{\theta}}_{\KL-W})^T\text{\.{s}}(\boldsymbol{\widetilde{x}}_j^k;\boldsymbol{\hat{\theta}}_{k})
\mid\boldsymbol{\hat{\theta}}_k)\frac{1}{n}\sum_{j=1}^n\dot{s}(\boldsymbol{\widetilde{x}}_j^k;\boldsymbol{\hat{\theta}}_k)=O_p(\frac{d}{n}).
\end{aligned}
\end{equation*}
According to Henmi et al.(2007), we know the second term of above equation is the orthogonal projection of $(\boldsymbol{\hat{\theta}}_{\KL}-\boldsymbol{\theta}_{\KL}^*)$ onto the linear space spanned by the score component for each $\boldsymbol{\hat{\theta}}_k$, for $k\in [d].$
Then
$$n\mathbb{E}\|\boldsymbol{\hat{\theta}}_{\KL-W}-\boldsymbol{\theta}_{\KL}^*\|^2\leq n\mathbb{E}\|\boldsymbol{\hat{\theta}}_{\KL}-\boldsymbol{\theta}_{KL}^*\|^2.$$
We complete the proof of this theorem. $\square$
\begin{thm}
Under the Assumptions \ref{assump}, when $N> n\times d$, $\mathbb{E}\|\boldsymbol{\hat{\theta}}_{\KL-W}- \boldsymbol{\theta}_{\KL}^*\|^2=O(\frac{1}{dn^2}).$
Further, its MSE for estimating the true parameter $\vv{\theta}^*$ is
\begin{align*}
\label{globalklw}
\mathbb{E}\|\boldsymbol{\hat{\theta}}_{\KL-W}-\boldsymbol{\theta}^*\|^2
=O(N^{-1}+(dn^2)^{-1}).
\end{align*}
\end{thm}


According to Equation (\ref{KLapprox}),
\begin{equation*}
\sum_{k=1}^d\frac1n\sum_{j=1}^n\text{\.{s}}(\boldsymbol{\widetilde{x}}_j^k;\boldsymbol{\hat{\theta}}_{\KL-W})-\sum_{k=1}^d\frac1n\sum_{j=1}^n\text{\.{s}}(\boldsymbol{\widetilde{x}}_j^k;\boldsymbol{\hat{\theta}}_{\KL-W})
\text{\.{s}}(\boldsymbol{\widetilde{x}}_j^k;\boldsymbol{\hat{\theta}}_{k})^T(\boldsymbol{\widetilde{\theta}}_k-\boldsymbol{\hat{\theta}}_k)=O_p(\frac{d}{n}).
\end{equation*}

Approximating the first term of the above equation by Taylor expansion, we will get
\begin{equation}
\label{Taylorapprox}
\begin{aligned}
\sum_{k=1}^d\frac{1}{n}\sum_{j=1}^n\text{\.{s}}(\boldsymbol{\widetilde{x}}_j^k;\boldsymbol{\hat{\theta}}_{\KL-W})&=\sum_{k=1}^d[\frac{1}{n}\sum_{j=1}^n\text{\.{s}}(\boldsymbol{\widetilde{x}}_j^k;\boldsymbol{\hat{\theta}}_{k})\\
&+\sum_{k=1}^d\frac{1}{n}\sum_{j=1}^n\text{\"{s}}(\boldsymbol{\widetilde{x}}_j^k;\boldsymbol{\hat{\theta}}_{k})(\boldsymbol{\hat{\theta}}_{\KL-W}-\boldsymbol{\hat{\theta}}_k)+O_p(\|\boldsymbol{\hat{\theta}}_{\KL-W}-\boldsymbol{\hat{\theta}}_k\|^2)].
\end{aligned}
\end{equation}
Since $\|\boldsymbol{\hat{\theta}}_{\KL-W}-\boldsymbol{\hat{\theta}}_k\|^2\leq \|\boldsymbol{\hat{\theta}}_{\KL-W}-\boldsymbol{\theta}_{\KL}^*\|^2+\|\boldsymbol{\theta}_{\KL}^*-\boldsymbol{\hat{\theta}}_k\|^2$, according to equation (\ref{liuproof}), then $\|\boldsymbol{\hat{\theta}}_{\KL-W}-\boldsymbol{\hat{\theta}}_k\|^2=O_p(\|\boldsymbol{\hat{\theta}}_{\KL-W}-\boldsymbol{\theta}_{\KL}^*\|^2+\frac{d}{N}).$
We can easily derive $\text{\.{s}}(\boldsymbol{\widetilde{x}}_j^k;\boldsymbol{\hat{\theta}}_{\KL-W})=\text{\.{s}}(\boldsymbol{\widetilde{x}}_j^k;\boldsymbol{\hat{\theta}}_{k})+O_p(\boldsymbol{\hat{\theta}}_{\KL-W}-\boldsymbol{\hat{\theta}}_{k})$ for $k\in[d].$ When $N > n\times d$, we will have
\begin{equation}
\label{WeigthedFinal}
\begin{split}
\sum_{k=1}^d\frac{1}{n}\sum_{j=1}^n&\text{\.{s}}(\boldsymbol{\widetilde{x}}_j^k;\boldsymbol{\hat{\theta}}_{k})+\sum_{k=1}^d\frac{1}{n}\sum_{j=1}^n\text{\"{s}}(\boldsymbol{\widetilde{x}}_j^k;\boldsymbol{\hat{\theta}}_{k}) (\boldsymbol{\hat{\theta}}_{\KL-W}-\boldsymbol{\hat{\theta}}_k)\\
&-\sum_{k}\frac{1}{n}\sum_{j=1}^n\text{\.{s}}(\boldsymbol{x}_j^k;\boldsymbol{\hat{\theta}}_{k})\text{\.{s}}(\boldsymbol{\widetilde{x}}_j^k;\boldsymbol{\hat{\theta}}_{k})^T
(\boldsymbol{\widetilde{\theta}}_k-\boldsymbol{\hat{\theta}}_k)+O_p(\|\boldsymbol{\hat{\theta}}_{\KL-W}-\boldsymbol{\theta}_{\KL}^*\|^2)=O(\frac{d}{n}).
\end{split}
\end{equation}
$\frac{1}{n}\sum_{j=1}^n\text{\"{s}}(\boldsymbol{\widetilde{x}}_j^k;\boldsymbol{\hat{\theta}}_{k})= I(\boldsymbol{\hat{\theta}}_{k})+o_p(\frac{1}{n})$ and we also know that $\frac{1}{n}\sum_{j=1}^n\text{\.{s}}(\boldsymbol{\widetilde{x}}_j^k;\boldsymbol{\hat{\theta}}_{k})\text{\.{s}}(\boldsymbol{\widetilde{x}}_j^k;\boldsymbol{\hat{\theta}}_{k})^T=I(\boldsymbol{\hat{\theta}}_{k})+o_p(1).$ From (\ref{asym}), we know  $\boldsymbol{\theta}_{\KL}^*-\boldsymbol{\hat{\theta}}_{k}=o_p(\frac{d}{N})=o_p(\frac{1}{n}).$ When $N>n\times d,$ we have
\begin{equation}
\label{WeigthedFinal}
\begin{split}
\sum_{k=1}^d\frac{1}{n}\sum_{j=1}^n\text{\.{s}}(\boldsymbol{\widetilde{x}}_j^k;\boldsymbol{\hat{\theta}}_{k})+&\sum_{k=1}^d I(\boldsymbol{\hat{\theta}}_{k})(\boldsymbol{\hat{\theta}}_{\KL-W}-\boldsymbol{\theta}_{\KL}^*)\\
&+\sum_{k=1}^d\frac{1}{n}I(\boldsymbol{\hat{\theta}}_{k})
(\boldsymbol{\widetilde{\theta}}_k-\boldsymbol{\hat{\theta}}_k))
+O_p(\|\boldsymbol{\hat{\theta}}_{\KL-W}-\boldsymbol{\theta}_{\KL}^*\|^2)=O(\frac{d}{n}).
\end{split}
\end{equation}
Based on the Equation (\ref{MLE}), the first term and the third term of Equation (\ref{WeigthedFinal}) are cancelled. By some simple calculation, we will get
\begin{equation}
\label{Weightedorder}
n^2(\boldsymbol{\hat{\theta}}_{\KL-W}-\boldsymbol{\theta}_{\KL}^*)^T(\sum_{k=1}^dI(\boldsymbol{\hat{\theta}}_{k}))(\sum_{k=1}^d I(\boldsymbol{\hat{\theta}}_{k}))(\boldsymbol{\hat{\theta}}_{\KL-W}-\boldsymbol{\theta}_{\KL}^*)=O_p(d).
\end{equation}
This indicates, $\mathrm{Cov}(n(\sum_{k=1}^dI(\boldsymbol{\hat{\theta}}_{k}))(\boldsymbol{\hat{\theta}}_{\KL-W}-\boldsymbol{\theta}_{\KL}^*),n(\sum_{k=1}^dI(\boldsymbol{\hat{\theta}}_{k}))(\boldsymbol{\hat{\theta}}_{\KL-W}-\boldsymbol{\theta}_{\KL}^*))=O(d)$ as $n\to\infty.$ We know $n^2\mathbb{E}\|\boldsymbol{\hat{\theta}}_{\KL-W}- \boldsymbol{\theta}_{\KL}^*\|^2=\mathrm{trace}(\mathrm{Cov}(n(\boldsymbol{\hat{\theta}}_{\KL-W}-\boldsymbol{\theta}_{\KL}^*),n(\boldsymbol{\hat{\theta}}_{\KL-W}-\boldsymbol{\theta}_{\KL}^*))$. According to Assumption \ref{assump}, $I(\boldsymbol{\hat{\theta}}_{k})$ is positive definite and then $\mathrm{trace}(\sum_{k=1}^dI(\boldsymbol{\hat{\theta}}_{k}))=O(d).$ Therefore, we have
$$\mathbb{E}\|\boldsymbol{\hat{\theta}}_{\KL-W}- \boldsymbol{\theta}_{\KL}^*\|^2=O(\frac{d}{d^2n^2})=O(\frac{1}{dn^2}).$$
We know $\mathbb{E}\|\boldsymbol{\theta}_{\KL}^*-\vv \theta^*\|^2=O(N^{-1})$ from \citet{liu2014distributed}. Then we have
$$\mathbb{E}\|\boldsymbol{\hat{\theta}}_{\KL-W}-\boldsymbol{\theta}^*\|^2\approx\mathbb{E}\|\boldsymbol{\hat{\theta}}_{\KL-W}-\boldsymbol{\theta}^*_{\KL}\|^2+\mathbb{E}\|\boldsymbol{\theta}_{\KL}^*-\boldsymbol{\theta}^*\|^2=O(N^{-1}+(dn^2)^{-1}).$$
The proof of this theorem is complete. $\square$

\end{document}